%% file: main.tex
\documentclass[nohyperref]{article}

\newif\ifexport
\exporttrue
\newif\ifarxiv

\PassOptionsToPackage{sort}{natbib}
\input{math_commands}

\usepackage{microtype}
\usepackage{graphicx}
\usepackage{caption}
\usepackage{subcaption}
\usepackage{booktabs} %
\usepackage{paralist} %
\usepackage{enumitem}
\usepackage{multirow}
\usepackage{afterpage}

\usepackage{tikz}
\usetikzlibrary{bayesnet}

\usepackage{hyperref}
\usepackage{xcolor, colortbl}
\usepackage[capitalize,noabbrev]{cleveref}
\usepackage{arydshln}

\newcommand{\OurApproach}{CITRIS}
\newcommand{\iVAEAdapt}{iVAE$^{*}$}
\newcommand{\eg}{\emph{e.g.}}
\newcommand{\ie}{\emph{i.e.}}
\newcommand{\highlight}[1]{\textbf{#1}}
\newcommand{\obs}{h}
\newcommand{\boxcolorone}[0]{blue}
\newcommand{\boxcolortwo}[0]{orange}
\newcommand{\disentanglement}{\delta}
\newcommand{\disentanglementClass}{\Delta}
\usepackage{comment}

\definecolor{instantcolor}{HTML}{117733} %
\definecolor{tempcolor}{HTML}{004488}
\definecolor{intvcolor}{HTML}{c83c04}
\definecolor{confcolor}{HTML}{886600}
\definecolor{obscolor}{HTML}{7a1f5c}

\definecolor{dark-green}{HTML}{00ba16}
\ifexport
    
    \newcommand{\TODO}[1]{}
    \newcommand{\phillip}[1]{}
    \newcommand{\taco}[1]{}
    \newcommand{\stratis}[1]{}
    \newcommand{\yuki}[1]{}
    \newcommand{\sara}[1]{}
    \newcommand{\sindy}[1]{}
\else
    
    \newcommand{\TODO}[1]{\textcolor{red}{[\textbf{TODO: #1}]}}
    \newcommand{\phillip}[1]{{\color{blue}\textsf{[Phillip: {#1}]}}}
    \newcommand{\taco}[1]{{\color{olive}\textsf{[Taco: {#1}]}}}
    \newcommand{\stratis}[1]{{\color{magenta}\textsf{[Stratis: {#1}]}}}
    \newcommand{\yuki}[1]{{\color{dark-green}\textsf{[Yuki: {#1}]}}}
    \newcommand{\sara}[1]{{\color{cyan}\textsf{[Sara: {#1}]}}}
    \newcommand{\sindy}[1]{{\color{violet}\textsf{[Sindy: {#1}]}}}
\fi

\usepackage[accepted]{icml2022}
\icmltitlerunning{CITRIS: Causal Identifiability from Temporal Intervened Sequences}

\usepackage{titletoc}
\renewcommand{\paragraph}[1]{\textbf{#1}}
\begin{document}

\twocolumn[
\icmltitle{CITRIS: Causal Identifiability from Temporal Intervened Sequences}

\icmlsetsymbol{equal}{*}

\begin{icmlauthorlist}
\icmlauthor{Phillip Lippe}{quva,uva}
\icmlauthor{Sara Magliacane}{uva,ibm}
\icmlauthor{Sindy L\"owe}{uva,buva}
\icmlauthor{Yuki M. Asano}{quva,uva}
\icmlauthor{Taco Cohen}{qualcomm}
\icmlauthor{Efstratios Gavves}{quva,uva}
\end{icmlauthorlist}

\icmlaffiliation{quva}{QUVA Lab, University of Amsterdam, Amsterdam, The Netherlands}
\icmlaffiliation{buva}{UvA-Bosch Delta Lab, University of Amsterdam, Amsterdam, The Netherlands}
\icmlaffiliation{uva}{Institute of Informatics, University of Amsterdam, Amsterdam, The Netherlands}
\icmlaffiliation{qualcomm}{Qualcomm AI Research, Amsterdam, The Netherlands. Qualcomm AI Research is an initiative of Qualcomm Technologies, Inc.}
\icmlaffiliation{ibm}{MIT-IBM Watson AI Lab}

\icmlcorrespondingauthor{Phillip Lippe}{p.lippe@uva.nl}

\icmlkeywords{Causal Representation Learning; Identifiability; Normalizing Flows; Variational Autoencoders; Generalization}

\vskip 0.3in
]

\printAffiliationsAndNotice{}  %

\begin{abstract}
\input{sections/0_abstract}
\end{abstract}

\input{sections/1_introduction}
\input{sections/2_preliminaries}
\input{sections/3.1_theory}
\input{sections/3.2_method}
\input{sections/4_related_work}
\input{sections/5_experiments}
\input{sections/6_conclusion}

\input{sections/7_acknowledgements}

\newpage
\bibliography{references_nourl}
\bibliographystyle{icml2022}

\newpage
\appendix
\onecolumn

\newcommand{\apphref}[3][black]{\hyperref[#2]{\color{#1}{#3}}}
\textbf{\LARGE Appendix}
\vskip 8mm
\startcontents[sections]\vbox{\sc\Large Table of Contents}\vspace{4mm}\hrule height .5pt\vspace{4mm}
\printcontents[sections]{l}{1}{\setcounter{tocdepth}{2}}
\phantom{}\hspace{3mm}\textbf{\cref{sec:appendix_reproducibility} - Reproducibility Statement}~\hfill~\pageref{sec:appendix_reproducibility}\\[10pt]
\phantom{}\hspace{3mm}\textbf{\cref{sec:appendix_proofs} - Proofs}~\hfill~\pageref{sec:appendix_proofs}\\
\phantom{}\hspace{8mm}\apphref{sec:appendix_proofs_preliminaries}{B.1. Preliminaries}~\dotfill~\pageref{sec:appendix_proofs_preliminaries}\\
\phantom{}\hspace{8mm}\apphref{sec:appendix_proofs_outline}{B.2. Proof Outline}~\dotfill~\pageref{sec:appendix_proofs_outline}\\
\phantom{}\hspace{8mm}\apphref{sec:appendix_proofs_step_1}{B.3. Step 1: True Disentanglement $\disentanglement^*$ is one of the Global Maxima}~\dotfill~\pageref{sec:appendix_proofs_step_1}\\
\phantom{}\hspace{8mm}\apphref{sec:appendix_proofs_step_2}{B.4. Step 2: Characterizing the Disentanglement Class $\disentanglementClass$}~\dotfill~\pageref{sec:appendix_proofs_step_2}\\
\phantom{}\hspace{8mm}\apphref{sec:appendix_proofs_step_3}{B.5. Step 3: Deriving the Final Theorem}~\dotfill~\pageref{sec:appendix_proofs_step_3}\\
\phantom{}\hspace{8mm}\apphref{sec:appendix_proofs_step_3}{B.6. Non-Identifiability without Interventions}~\dotfill~\pageref{sec:appendix_proofs_proposition}\\[10pt]
\phantom{}\hspace{3mm}\textbf{\cref{sec:appendix_experimental_details} - Experimental Details}~\hfill~\pageref{sec:appendix_experimental_details}\\
\phantom{}\hspace{8mm}\apphref{sec:appendix_experimental_details_causal3d_dataset}{C.1. Temporal Causal3DIdent Dataset}~\dotfill~\pageref{sec:appendix_experimental_details_causal3d_dataset}\\
\phantom{}\hspace{8mm}\apphref{sec:appendix_experimental_details_interventional_pong}{C.2. Interventional Pong}~\dotfill~\pageref{sec:appendix_experimental_details_interventional_pong}\\
\phantom{}\hspace{8mm}\apphref{sec:appendix_experimental_details_experimental_design}{C.3. Experimental Design}~\dotfill~\pageref{sec:appendix_experimental_details_experimental_design}\\
\phantom{}\hspace{8mm}\apphref{sec:appendix_experimental_design_hyperparameters}{C.4. Hyperparameters}~\dotfill~\pageref{sec:appendix_experimental_design_hyperparameters}\\[10pt]
\phantom{}\hspace{3mm}\textbf{\cref{sec:appendix_additional_experiments} - Additional Experiments and Results}~\hfill~\pageref{sec:appendix_additional_experiments}\\
\phantom{}\hspace{8mm}\apphref{sec:appendix_additional_experiments_causal3d}{D.1. Temporal Causal3DIdent}~\dotfill~\pageref{sec:appendix_additional_experiments_causal3d}\\
\phantom{}\hspace{8mm}\apphref{sec:appendix_additional_experiments_pong}{D.2. Interventional Pong}~\dotfill~\pageref{sec:appendix_additional_experiments_pong}\\
\phantom{}\hspace{8mm}\apphref{sec:appendix_additional_experiments_ball_in_box}{D.3. Ball-in-Boxes Example}~\dotfill~\pageref{sec:appendix_additional_experiments_ball_in_box}\\

\vskip 4mm
\hrule height .5pt
\vskip 10mm
\normalsize

\input{sections/appendix_sections/0_statements}
\input{sections/appendix_sections/1_proofs}
\newpage
\input{sections/appendix_sections/2_experimental_details}
\newpage
\input{sections/appendix_sections/3_additional_experiments}

\end{document}

%% file: math_commands.tex
\usepackage{amsthm,amsmath,amssymb,amsfonts,bm,bbm,stmaryrd}

\def\eqref#1{equation~\ref{#1}}

\def\1{\bm{1}}

\DeclareMathAlphabet{\mathsfit}{\encodingdefault}{\sfdefault}{m}{sl}
\SetMathAlphabet{\mathsfit}{bold}{\encodingdefault}{\sfdefault}{bx}{n}

\newcommand{\E}{\mathbb{E}}

\newcommand{\sigmoid}{\sigma}

\newcommand{\range}[2]{\llbracket#1..#2\rrbracket}

\newcommand\independent{\protect\mathpalette{\protect\independenT}{\perp}}
\def\independenT#1#2{\mathrel{\rlap{$#1#2$}\mkern2mu{#1#2}}}

\newtheorem{theorem}{Theorem}[section]

\newtheorem{lemma}[theorem]{Lemma}
\newtheorem{proposition}[theorem]{Proposition}
\newtheorem{definition}[theorem]{Definition}

\newcommand{\sindep}{s^{\mathrm{inv}}}
\newcommand{\sdep}{s^{\mathrm{var}}}
\newcommand{\sdepStar}{s_i^{\mathrm{var}^*}}
\newcommand{\sindepStar}{s_i^{\mathrm{inv}^*}}
\newcommand{\noisevarcap}{E}
\newcommand{\noisevarlow}{\epsilon}
\newcommand{\zpsi}[1]{z_{\psi_{#1}}}
\newcommand{\zpsistar}[1]{z_{\psi^*_{#1}}}

\newcommand{\pa}[1]{\text{pa}(#1)} %
\newcommand{\paG}[1]{\text{pa}_G(#1)}
\newcommand{\paGG}[1]{\text{pa}_{G'}(#1)}

%% file: sections/0_abstract.tex
Understanding the latent causal factors of a dynamical system from visual observations is considered a crucial step towards agents reasoning in complex environments. In this paper, we propose CITRIS, a variational autoencoder framework that learns causal representations from temporal sequences of images in which underlying causal factors have possibly been intervened upon. In contrast to the recent literature, CITRIS exploits temporality and observing intervention targets to identify scalar \textit{and} multidimensional causal factors, such as 3D rotation angles. Furthermore, by introducing a normalizing flow, CITRIS can be easily extended to leverage and disentangle representations obtained by already pretrained autoencoders. Extending previous results on scalar causal factors, we prove identifiability in a more general setting, in which only some components of a causal factor are affected by interventions. In experiments on 3D rendered image sequences, CITRIS outperforms previous methods on recovering the underlying causal variables. Moreover, using pretrained autoencoders, CITRIS can even generalize to unseen instantiations of causal factors, opening future research areas in sim-to-real generalization for causal representation learning. 

%% file: sections/1_introduction.tex
\section{Introduction}
\label{sec:introduction}

\begin{figure}
    \centering
    \includegraphics[width=0.98\columnwidth]{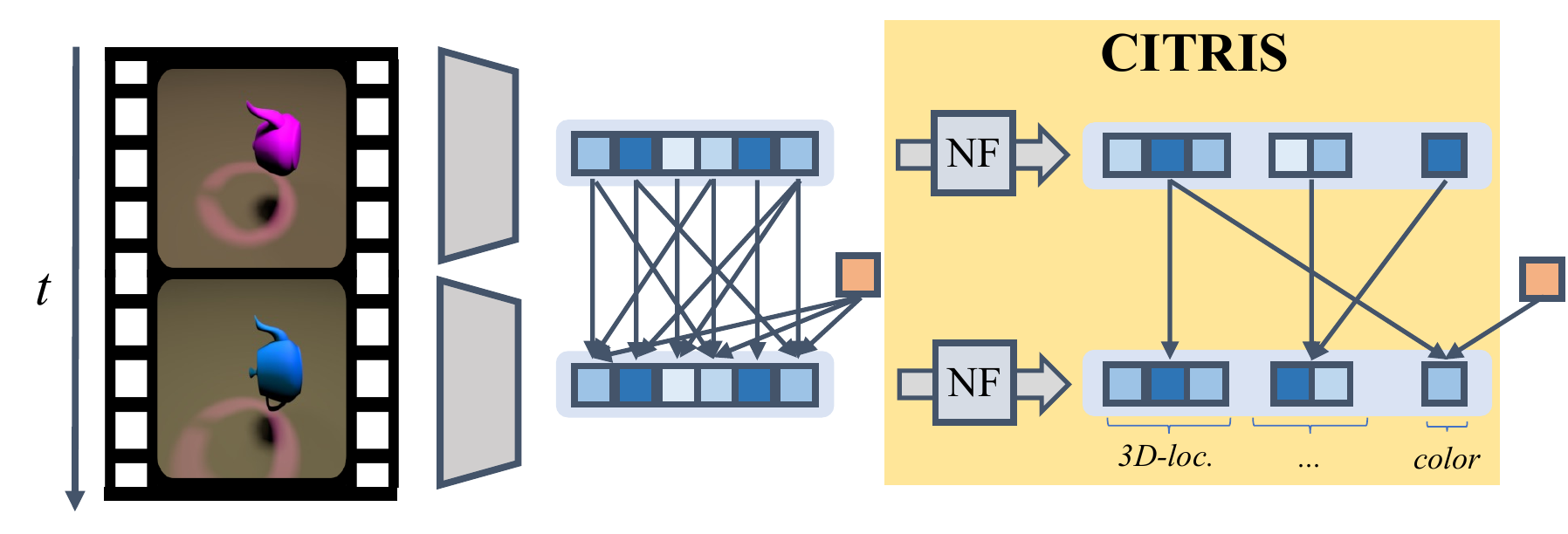}
    \caption{From a sequence of images with interventions (orange), \OurApproach{} learns a normalizing flow, which maps a pretrained autoencoder's entangled latent representation to a causal representation. In this, multiple latent dimensions (blue blocks) can represent a single causal factor, where causal relations exist over time.
    }
    \label{fig:introduction_overall_setup}
\end{figure}

Causal representation learning \cite{schoelkopf2021towards, locatello2020weakly, yang2021causalvae, khemakhem2020variational, lachapelle2021disentanglement} focuses on learning representations of causal factors from high-dimensional observations, such as images.
Commonly, in learning causal representations, causal factors are assumed to be scalars.
As settings become more complex and high-dimensional, however, so do the causal dynamics, where estimating every scalar causal variable becomes impractical.
Consider for instance a set of objects interacting in a three-dimensional space.
For each object, we can describe its position by three variables $x,y,z$ and its rotation in multiple angles.
However, it is often more natural and sufficient to consider those factors as two \emph{multidimensional} variables, \ie{} position and rotation, especially when the definition of the default axes is ambiguous.
Such multidimensional causal factors are reminiscent of macrovariables explored before in causality \cite{chalpuka2016unsupervised, chalpuka2016multilevel, chalpuka2015visual, hoel2013quantifying, holtgen2021encoding}, however, not in causal representation learning yet.

Hence, different from previous causal representation learning approaches, we consider causal factors as potentially multidimensional vectors in this paper.
This requires a representation model to learn a latent space of disentangled factors of variation, where latent variables of the same causal factor are grouped together.
To identify these multidimensional causal variables, we use sequences of observations where interventions with known targets may have been performed at any time step.
This setup resembles a reinforcement learning environment or an interactive real-world setting, with an agent performing actions over time representing interventions on multidimensional causal factors.
However, instead of learning a policy to optimize a reward function, we aim at using this setting to identify the causal factors.
We assume we can observe the intervention targets in our data, which occurs for instance when we have access to a set of observation and action trajectories.
We refer to this setup as TempoRal Intervened Sequences (TRIS).

Under this setting and common assumptions like an invertible observation function and a stationary causal process, we can identify the \textit{minimal causal variables}, which only model the information of a causal factor that is strictly affected by a provided intervention.
With this, we identify the factors which can be directly and independently influenced by, \eg{}, the action of an agent, and thus of the most practical relevance.
Meanwhile, all information that cannot be assigned to a causal factor with certainty and hence not directly influenced, is collected in a separate group of latent variables.
As a practical implementation of this, we propose \OurApproach{} for Causal Identifiability from TempoRal Intervened Sequences.
\OurApproach{} is a variational autoencoder that learns an assignment of latent variables to causal factors, and promotes disentanglement by conditioning each latent's prior distribution only on its respective intervention target.
In experiments on realistically rendered video datasets, \OurApproach{} is able to find and disentangle the causal factors with high accuracy.
Moreover, we extend \OurApproach{} to pretrained autoencoders.
By using a normalizing flow \cite{rezende2015variational}, \OurApproach{} learns a mapping from the entangled autoencoder representation to a disentangled causal representation, see \cref{fig:introduction_overall_setup}.
We empirically show that the normalizing flow can even generalize its disentanglement to unseen instantiations of causal factors, holding promise for future work on generalization of causal representations.

Our contributions are summarized as follows:
\begin{compactitem}
    \setlength\itemsep{1mm}
    \item We show that multidimensional causal factors can be identified from temporal sequences with interventions up to their minimal causal variables.
    \item We propose \OurApproach{}, a VAE architecture for disentangling causal factors in latent space based on this setup.
    \item Finally, we extend \OurApproach{} to pretrained autoencoders by learning a map from an entangled to a causally disentangled latent space using normalizing flows.
\end{compactitem}

%% file: sections/2_preliminaries.tex
\section{Preliminaries and Causal Assumptions}
\label{sec:preliminaries}

We assume that the underlying latent causal process is a dynamic Bayesian network (DBN) \cite{DBN, Murphy_DBN} $G$ over a set of $K$ causal variables.
In the corresponding graph $G=(V,E)$, each node $i\in V$ is associated with a causal variable $C_i$, which can be scalar or vector valued. 
Each edge $(i,j)\in E$ represents a causal relation from $C_i$ to $C_j$: $C_i\to C_j$, where $C_i$ is a \emph{parent} of $C_j$ and $\paG{C_i}$ are all parents of $C_i$ in $G$.
Further, we assume that the DBN is first-order Markov, stationary, and without instantaneous effects. 
This means that in $G$ each causal factor $C_i$ is instantiated at each time step $t$, denoted by $C_i^t$, and its causal parents can only be causal factors at time $t-1$, denoted as $C_j^{t-1}$, including its own previous value $C_i^{t-1}$. 
In other words, for $t\in\range{1}{T}$ and for each causal factor $i\in\range{1}{K}$ we can model $C_i^t = f_i(\paG{C_i^t}, \epsilon_i)$, where $\paG{C_i^t} \subseteq \{ C_1^{t-1}, \dots, C^{t-1}_K \}$.
We denote the set of all causal variables at time $t$ as $C^t=(C^t_1, \dots, C^t_K)$, where $C^t$ inherits all edges from its components $C_i$ for $i\in\range{1}{K}$ without introducing cycles.
In this setting the structure of the graph is time-invariant, \ie,  $\paG{C_i^t} = \paG{C_i^{1}}$ for any $t\in\range{1}{T}$. We also assume all $\epsilon_i$ for $i\in\range{1}{K}$ are mutually independent noises.

We use a binary intervention vector $I^t \in \{0,1\}^{K}$ to indicate that a variable $C^t_i$ in $G$ is intervened upon if and only if $I^t_i=1$. 
We consider that the intervention vector components $I^t_i$ might be confounded by another $I^t_j, i \neq j$, and represent these dependencies with an unobserved regime variable $R^t$ \cite{didelez2006direct, mooij2020joint}.
With this, we construct an augmented DAG $G'=(V', E')$, where $V'=\{\{C^t_i\}_{i=1}^K \cup \{I^t_i\}_{i=1}^K \cup R^t \}_{t=1}^T$ and $E'= \{ \{\paG{C^t_i} \to C^t_i\}_{i=1}^K \cup \{I^t_i \to C^t_i\}_{i=1}^K \cup \{R^t\to I^t_i\}_{i=1}^K \}_{t=1}^T$. 
We say that a distribution $p$ is \emph{Markov} w.r.t. the augmented DAG $G'$ if it factors as $p(V') = \prod_{j \in V'} p(V_j \mid \paGG{V_j})$, where $V_j$ includes the causal factors $C^t_i$, the intervention vector components $I^t_i$, and the regime $R^t$. 
Moreover, we say that $p$ is \emph{faithful} to a causal graph $G'$, if there are no additional conditional independences to the d-separations one can read from the graph $G'$.
The augmented graph $G'$ can model interventions with an arbitrary number of targets, including observational data. 
In this paper we will consider \emph{soft} interventions \cite{eberhardt2007causation}, in which the conditional distribution changes, \ie, $p(C^t_i|\paG{C^t_i}, I^t_i=1) \not = p(C^t_i|\paG{C^t_i}, I^t_i=0)$, which include as a special case \emph{perfect} interventions $\text{do}(C_i=c_i)$ \cite{pearl2009causality}.

%% file: sections/3.1_theory.tex
\section{Identifiability of Minimal Causal Variables}
\label{sec:identifiability}
We first describe our setting, TempoRal Intervened Sequences (TRIS). 
In this setting, we show that identifying the underlying causal factors is not always possible, especially when considering multidimensional causal factors.
Therefore, we define the concept of \emph{minimal causal variable}, which represent the manipulable part of each causal factor.
Finally, we show under which conditions we can recover the minimal causal variables in TRIS.

\subsection{TempoRal Intervened Sequences (TRIS)}
\label{sec:identifiability_setup_assumptions}

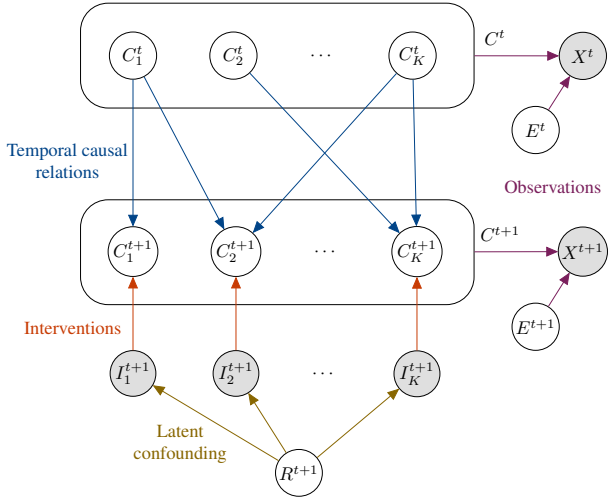
\begin{figure}[t!]
    \centering
    \resizebox{\linewidth}{!}{
    \input{figures/causal_graph}
    }
    \caption{An example causal graph in TRIS, with observed variables shown in gray and latent variables in white. A latent causal factor $C^{t+1}_i$ has as parents a subset of the causal factors at the \textcolor{tempcolor}{previous time step} $C_1^{t}, \dots, C_{K}^{t}$, and its \textcolor{intvcolor}{intervention target} $I^{t+1}_i$. All causal variables $C^{t+1}$ and noise $\noisevarcap^{t+1}$ cause the \textcolor{obscolor}{observation} $X^{t+1}$. $R^{t+1}$ is a \textcolor{confcolor}{latent confounder} between the intervention targets.}
    \label{fig:theory_causal_graph}
    \vspace{-2mm}
\end{figure}

In TRIS, we consider data generated by an underlying latent temporal causal process with $K$ \emph{causal factors} $(C^t_1,C^t_2,...,C^t_K)_{t=1}^T$. 
At each time step $t$, we observe a high-dimensional observation $X^T$ representing a noisy, entangled view of all causal factors.
The following paragraphs describe the details of this setup, visualized in \cref{fig:theory_causal_graph}.

\paragraph{Multidimensional Causal Factors:}  
As opposed to most work on causal representation learning, which considers causal factors to be one-dimensional \cite{klindt2021towards,lachapelle2021disentanglement,khemakhem2020variational}, we allow them to be potentially multidimensional, \ie, $C_i \in \mathcal{D}_i^{M_i}$ with $M_i \geq 1$ and in practice we let $\mathcal{D}_i$ be $\mathbb{R}$ for continuous variables (\eg, spatial position), $\mathbb{Z}$ for discrete variables (\eg, the score of a player) or mixed.
This allows modeling different levels of causal variables (\eg{} a 2D-position encoded in a single factor with two dimensions instead of two different causal factors). 
We define the causal factor space as $\mathcal{C} = \mathcal{D}_1^{M_1}\times\mathcal{D}_2^{M_2}\times...\times\mathcal{D}_K^{M_K}$.

\paragraph{Observation Function:} 
We define the observation function $\obs(C^t_1,C^t_2,...,C^t_K,\noisevarcap^t_{o})=X^t$, where $\noisevarcap^t_{o}$ represents any noise independent of the causal factors that influence the observations, and $\obs: \mathcal{C} \times \mathcal{E} \rightarrow \mathcal{X}$ is a function from the causal factor space $\mathcal{C}$ and the space of the noise variables $\mathcal{E}$ to the observation space $\mathcal{X}$. 
We assume that $\obs$ is bijective, implying that the joint dimensionality of the noise and causal model is limited to the image size.
This allows us to identify each causal factor uniquely from observations by learning an approximation of $f$, while disregarding irrelevant features in the observation space.

\paragraph{Availability of Intervention Targets:}
Crucially, we assume that in each time-step some causal factors might (or might not) have been intervened upon and that we have access to the corresponding intervention targets, but not the intervention values.
We denote these intervention targets by the binary vector $I^{t}\in \{0,1\}^{K}$ where $I^{t}_i=1$ refers to an intervention on the causal variable $C_i^{t}$.

\subsection{Necessary Condition for Disentanglement in TRIS}
\begin{figure}[t!]
    \centering
    \includegraphics[width=0.55\columnwidth]{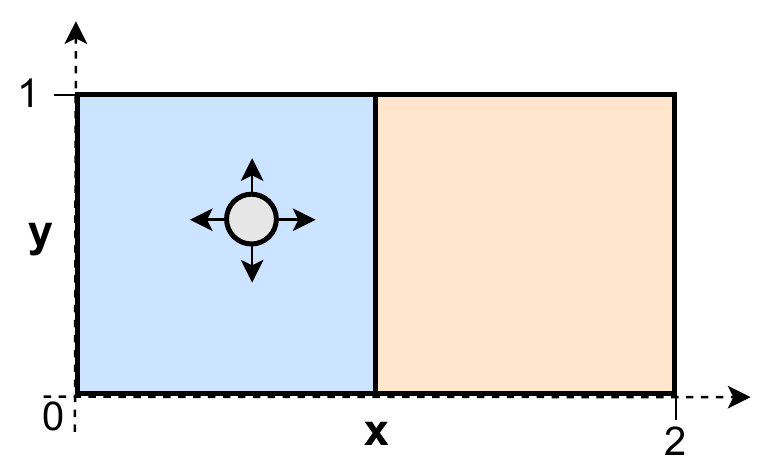}
    \caption{Example causal process of a ball with a two-dimensional position $x,y$. The ball can only swap between the two boxes (\boxcolorone{} and \boxcolortwo{}) through an intervention, which does not influence the relative position within a box.}
    \label{fig:method_intervention_dependent_mechanism}
\end{figure}

In TRIS, we generally cannot disentangle two causal factors if they are always intervened upon jointly, or, on the contrary, if they are never intervened upon.
\begin{proposition}
    \label{prop:method_tris_nonidentifiability}
    In TRIS, if two causal factors $C_i$ and $C_j$ have only been jointly intervened on or not at all, then there exists a causal graph in which $C_i$ and $C_j$ cannot be uniquely identified from observations $X$ and intervention targets $I$.
\end{proposition}
We provide an example of such a graph in \cref{fig:method_intervention_dependent_mechanism}, where a ball can move in two dimensions, $x$ and $y$.
If both $x$ and $y$ follow a Gaussian distribution with stationary variances over time, then any two orthogonal axes can describe the distribution equally well \cite{hyvaerinen2017nonlinear, belouchrani1997blind}, making it impossible to uniquely identify them without interventions.
Similarly, if we only observe joint interventions on $x,y$ together, we cannot identify them either due to the same reasoning. 
We include the proof for this proposition in \cref{sec:appendix_proofs_proposition}.

Additionally, in TRIS where the latent causal factors may correspond to multidimensional vectors, we cannot even completely reconstruct said factors, when by the nature of the system the provided interventions leave some of the causal factor's dimensions unaffected. 
In the next section, we will instead introduce the concept of minimal causal variables to characterize what we can identify instead.

\subsection{Minimal Causal Variables}
\label{sec:identifiability_minimal_causal_mechanisms}

To visualize this scenario, consider again the example in \cref{fig:method_intervention_dependent_mechanism} with a ball in one of two boxes.
Over time, the ball can move freely within the box it is currently in, but it can only jump into another box if there is an intervention. 
The intervention moves the ball to the other box, but keeps the relative position of the ball within the box intact. 
While one could define this process by a single causal variable $x$ over time, it can also be described by two causal variables: the relative position within the box $x'$ and the current box $b$.
Since only $b$ is affected by the intervention, and we consider causal factors to potentially be multidimensional, we could not identify which causal factor $x'$ belongs to.

We formalize this intuition as follows.
Suppose for each causal factor $C_i \in \mathcal{D}^{M_i}$, there exists an invertible map $s_i : \mathcal{D}_i^{M_i} \rightarrow \mathcal{D}_i^{\textup{var}} \times \mathcal{D}_i^{\textup{inv}}$ that splits the domain $\mathcal{D}^{M_i}$ of $C_i$ into a part that is variant and a part that is invariant under intervention.
We denote the two parts of this map as 
\begin{equation}
    s_i(C^t_i) = (\sdep_i(C^t_i), \sindep_i(C^t_i))
\end{equation}
The split $s$ must be invertible, so that we can map back and forth between $\mathcal{D}_i^{M_i}$ and $\mathcal{D}_i^{\textup{var}} \times \mathcal{D}_i^{\textup{var}}$ without losing information.
Furthermore, to be called a split in this setup, $s$ must satisfy $\sindep_i(C_i^{t}) \independent I_i^{t} \mid \pa{C_i^{t}}$, \ie{}, $\sindep_i(C_i^{t})$ is independent of the intervention variable $I_i^{t}$ given the parents of $C_i^{t}$.
Also, both parts of the split must be conditionally independent, \ie{} 
$\sindep_i(C_i^{t}) \independent \sdep_i(C_i^{t}) \mid \pa{C_i^{t}}, I_i^{t}$.
This means that $\sdep_i(C_i^{t})$ will contain the manipulable, or \emph{variable}, part of $C^t_i$.
In contrast, $\sindep_i(C_i^{t})$ is the \emph{invariable} part of $C_i^t$ which is independent of the intervention.

For any causal variable, there may exist multiple possible splits. There is always at least the trivial split where $\mathcal{D}_i^{\textup{var}} = \mathcal{D}_i^{M_i}$ is the original domain of $C_i$, and $\mathcal{D}_i^{\textup{inv}} = \{0\}$ is the one-element set (no invariant information). 
But not all splits are trivial: 
For the example in \cref{fig:method_intervention_dependent_mechanism}, we can split the causal factor $x$ in $\sdep(x)$ such that the box identifier $b$ is in $\sdep(x)$ and the relative position in the box $x'$ is in $\sindep(x)$.

Intuitively, we want to identify the split where $\sdep_i$ contains \emph{only} the manipulable information:
\begin{definition}
The \emph{minimal causal split} of a variable $C^t_i$ with respect to its intervention variable $I^t_i$ is the split $s_i$ which maximizes the information content $H(\sindep_i(C^{t}_i)|\pa{C^{t}_i})$. Under this split, $\sdep_i(C^{t}_i)$ is defined as the \emph{minimal causal variable} and denoted by $\sdepStar(C^{t}_i)$.
\end{definition}
Here $H$ denotes the entropy in the discrete case, or the limiting density of discrete points (LDDP) \cite{jaynes1957information, jaynes1968prior} which measures information content of a continuous variable and is invariant under the change of variables induced by the map $s$.
Intuitively, this ensures that only the information which truly depends on the intervention is represented in $\sdep_i(C_i)$.
The definition of the minimal causal split depends on the characteristics of the provided intervention.
In our previous example, when an intervention on $x$ would also change the internal box position, the minimal causal variable would contain the full causal factor $x$.
Our goal becomes to identify these minimal causal variables.

Depending on the interventions, the causal graph among the minimal causal graph may differ from the original graph on $C_1,...,C_K$.
If the interventions are soft, the minimal causal variable, $\sdep_i(C_i)$, has as parents the subset of $\pa{C_i}$, whose relation to $C_i$ is influenced by the intervention.
For instance, consider a three-dimensional causal variable, of which each dimension has a different set of parents.
If an intervention only affects the first dimension, $\sdep_i(C_i)$ has the same parents as the first dimension, and $\sindep_i(C_i)$ the same parents as the last two dimensions.
For perfect interventions, the intervention-invariant part, $\sindep_i(C_i)$, has no parents since all temporal dependencies are influenced by the intervention. 
Thus, in this case, the parents of $\sdep_i(C_i)$ are the same as of the true causal variable, $\pa{C_i}$.

\subsection{Learning Minimal Causal Variables}
\label{sec:identifiability_learning_setup}

As a practical example of TRIS, we consider a dataset $\mathcal{D}$ of tuples $\{x^t,x^{t+1},I^{t+1}\}$ where $x^{t},x^{t+1} \in \mathbb{R}^{N}$ represent the observations at time step $t$ and $t+1$ respectively, and $I^{t+1}$ describes the targets of the interventions performed on $C^{t+1}$.
To learn a causal representation, we consider a latent space $\mathcal{Z}$ larger than the latent causal factor space $\mathcal{C}$, i.e. $\mathcal{Z}\subseteq\mathbb{R}^{M}, M \geq \text{dim}(\mathcal{E})+\sum_{i=1}^{K} M_i = \text{dim}(\mathcal{E})+\text{dim}(\mathcal{C}) $. 
In this latent space, we aim to disentangle the causal factors.
Since we may not know the exact dimensions $M_1,...,M_K$, we overestimate its size in the latent space $\mathcal{Z}$.

Our goal is to approximate the inverse of the observation function $\obs$ by learning two components.
First, we learn an invertible mapping from observations to latent space, $g_{\theta}: \mathcal{X}\to \mathcal{Z}$.
Second, we learn an assignment function $\psi: \range{1}{M}\to\range{0}{K}$ that maps each dimension of $\mathcal{Z}$ to a causal factor.
Learning a flexible assignment function $\psi$ allows us to allocate any dimension size per causal factor without knowing the individual causal factor dimensions $M_1,...,M_K$ in advance.
Further, it also benefits the optimization process, since some variables like circular angles or categorical factors with many categories can have simpler distributions when modelled in more dimensions, which we verify empirically in \cref{sec:experiments_causal3dident}.
In addition to the $K$ causal factors, we use $\psi(j)=0,j\in\range{1}{M}$ to indicate that the latent dimension $z_j$ does not belong to any minimal causal variable.
Instead, those dimensions might model $\sindep_i(C_i)$ for some causal factor $C_i$ or the observation noise $\noisevarcap^t_o$.
Finally, we denote the set of latent variables that $\psi$ assigns to the causal factor $C_i$ with $\zpsi{i}=\{z_j|j\in\range{1}{M}, \psi(j)=i\}$.

To enforce a disentanglement of causal factors, we model a prior distribution in latent space, $p_{\phi}(z^{t+1}|z^{t},I^{t+1})$, with $z^{t},z^{t+1}\in\mathcal{Z}$, $z^{t}=g_{\theta}(x^{t}), z^{t+1}=g_{\theta}(x^{t+1})$.
This transition prior enforces a disentanglement by conditioning each latent variable on exactly one of the intervention targets:
\begin{equation} \label{eq:method_prior_distribution}
    p_{\phi}\left(z^{t+1}|z^{t}, I^{t+1}\right) = \prod_{i=0}^{K}p_{\phi}\left(\zpsi{i}^{t+1}|z^{t}, I_{i}^{t+1}\right)
\end{equation}
where $I_0^{t+1}=0$.
Then, the objective of the model is to maximize the likelihood:
\begin{equation}
    \resizebox{0.9\columnwidth}{!}{
    $p_{\phi,\theta}(x^{t+1}|x^{t},I^{t+1})=\left|\frac{\partial g_{\theta}(x^{t+1})}{\partial x^{t+1}}\right|p_{\phi}(z^{t+1}|z^{t}, I^{t+1})$
    }
\end{equation}

Under the assumptions stated in \cref{sec:identifiability_setup_assumptions}, we can prove the following identifiability result for this setup:
\begin{theorem}
    \label{theo:method_intv_over_time}
    Suppose that $\phi^{*}$, $\theta^{*}$ and $\psi^{*}$ are the parameters that, under the constraint of maximizing the likelihood $p_{\phi,\theta}(x^{t+1}|x^{t},I^{t+1})$, maximize the information content of $p_{\phi}(\zpsi{0}^{t+1}|z^{t})$. 
    Then, with sufficient latent dimensions, the model $\phi^{*},\theta^{*},\psi^{*}$ learns a latent structure where $\zpsi{i}^{t+1}$ models the minimal causal variable of $C_i$ if \mbox{$C^{t+1}_i\not\independent I^{t+1}_i | C^{t},I^{t+1}_j$} for any $i\neq j$.
    All remaining information is modeled in $\zpsi{0}$.
\end{theorem}
We provide the proof for this statement in \cref{sec:appendix_proofs}, which relies on $I^{t+1}_i$ not being a deterministic function of any other intervention target.
A sufficient condition for this is the interventions being independent of each other, or single-target interventions with observational data.
Finding the minimal variables intuitively means that the latent variables $\zpsi{i}$ model only the information of $C_i$ which strictly depends on the intervention target $I_{i}^{t+1}$, thus defining causal variables by their intervention dependency.
Going back to the example of the $x$ position of the ball in \cref{fig:method_intervention_dependent_mechanism}, we would model the box identifier in $\zpsi{1}$ if $C_1=x$, while the internal box position is modeled in $\zpsi{0}$. We empirically verify the learned split for this example in \cref{sec:appendix_additional_experiments_ball_in_box}.
Nonetheless, one can always ensure that for any definition of the causal process, $\zpsi{i}$ only contains information of causal variable $C_i$ of that process, and no other causal variable.

%% file: figures/causal_graph.tex
\tikz{ %
	\node (Ct) at (2.75,0) [draw,rounded corners=15pt,minimum width=7.5cm,minimum height=2cm] {};
	\node (Ct1) at (2.75,-3.75) [draw,rounded corners=15pt,minimum width=7.5cm,minimum height=2cm] {};
	
	\node[latent, minimum size=.9cm] (C1t) {$C^{t}_1$} ; %
	\node[latent, minimum size=.9cm, right=of C1t] (C2t) {$C^{t}_2$} ; %
	\node[const, right=of C2t] (dotdott) {$\cdots$} ; %
	\node[latent, minimum size=.9cm, right=of C2t, xshift=1.5cm] (CKt) {$C^{t}_{K}$} ; %
	
	\node[latent, below=of C1t, yshift=-1.8cm] (C1t1) {$C^{t+1}_1$} ; %
	\node[latent, right=of C1t1] (C2t1) {$C^{t+1}_2$} ; %
	\node[const, right=of C2t1] (dotdott1) {$\cdots$} ; %
	\node[latent, right=of dotdott1] (CKt1) {$C^{t+1}_{K}$} ; %
	\node[obs, right=of CKt1, xshift=1.2cm, minimum size=.9cm] (Xt1) {$X^{t+1}$} ; %
	\node[latent, below=of Xt1, xshift=-.9cm, yshift=.5cm, minimum size=.9cm] (Eot1) {$\noisevarcap^{t+1}$} ; %
	\node[obs, above=of Xt1, yshift=1.8cm, minimum size=.9cm] (Xt) {$X^t$} ; %
	\node[latent, below=of Xt, xshift=-.9cm, yshift=.5cm, minimum size=.9cm] (Eot) {$\noisevarcap^t$} ; %
	
	\node[obs, below=of C1t1, yshift=-.4cm] (I1t1) {$I^{t+1}_1$} ; %
	\node[obs, below=of C2t1, yshift=-.4cm] (I2t1) {$I^{t+1}_2$} ; %
	\node[const, right=of I2t1] (dotdotIt1) {$\cdots$} ; %
	\node[obs, below=of CKt1, yshift=-.4cm] (IKt1) {$I^{t+1}_{K}$} ; %
	\node[latent, below=of I2t1, xshift=1.2cm] (Rt1) {$R^{t+1}$} ; %
	
	\node[align=center,tempcolor] at (-1.25,-2.0) {Temporal causal\\relations};
	\node[align=center,intvcolor] at (-1.15,-5.2) {Interventions};
	\node[align=center,confcolor] at (.9,-7.4) {Latent\\confounding};
	\node[align=center,obscolor] at (8,-2.5) {Observations};
	\node[] at (7,-3.4) {$C^{t+1}$};
	\node[] at (6.9,0.35) {$C^{t}$};
	
	\edge[obscolor]{Ct}{Xt} ;
	\edge[obscolor]{Eot}{Xt} ;
	\edge[obscolor]{Ct1}{Xt1} ;
	\edge[obscolor]{Eot1}{Xt1} ;
	
	\edge[tempcolor]{C1t}{C1t1} ;
	\edge[tempcolor]{C1t}{C2t1} ;
	\edge[tempcolor]{C2t}{CKt1} ;
	\edge[tempcolor]{CKt}{CKt1} ;
	\edge[tempcolor]{CKt}{C2t1} ;
	
	\edge[intvcolor]{I1t1}{C1t1} ;
	\edge[intvcolor]{I2t1}{C2t1} ;
	\edge[intvcolor]{IKt1}{CKt1} ;
	
	\edge[confcolor]{Rt1}{I1t1} ;
	\edge[confcolor]{Rt1}{I2t1} ;
	\edge[confcolor]{Rt1}{IKt1} ;
}

%% file: sections/3.2_method.tex
\section{Causal Identifiability from Temporal Intervened Sequences}
\label{sec:method}
\label{sec:method_practical_implementation}
To identify causal factors from high-dimensional temporal observations with interventions, we propose \OurApproach{} (Causal Identifiability from Temporal Intervened Sequences).
Below, we discuss its architecture and variants.

\begin{figure*}[t!]
    \centering
    \begin{tabular}{c:c}
        \includegraphics[width=0.95\columnwidth]{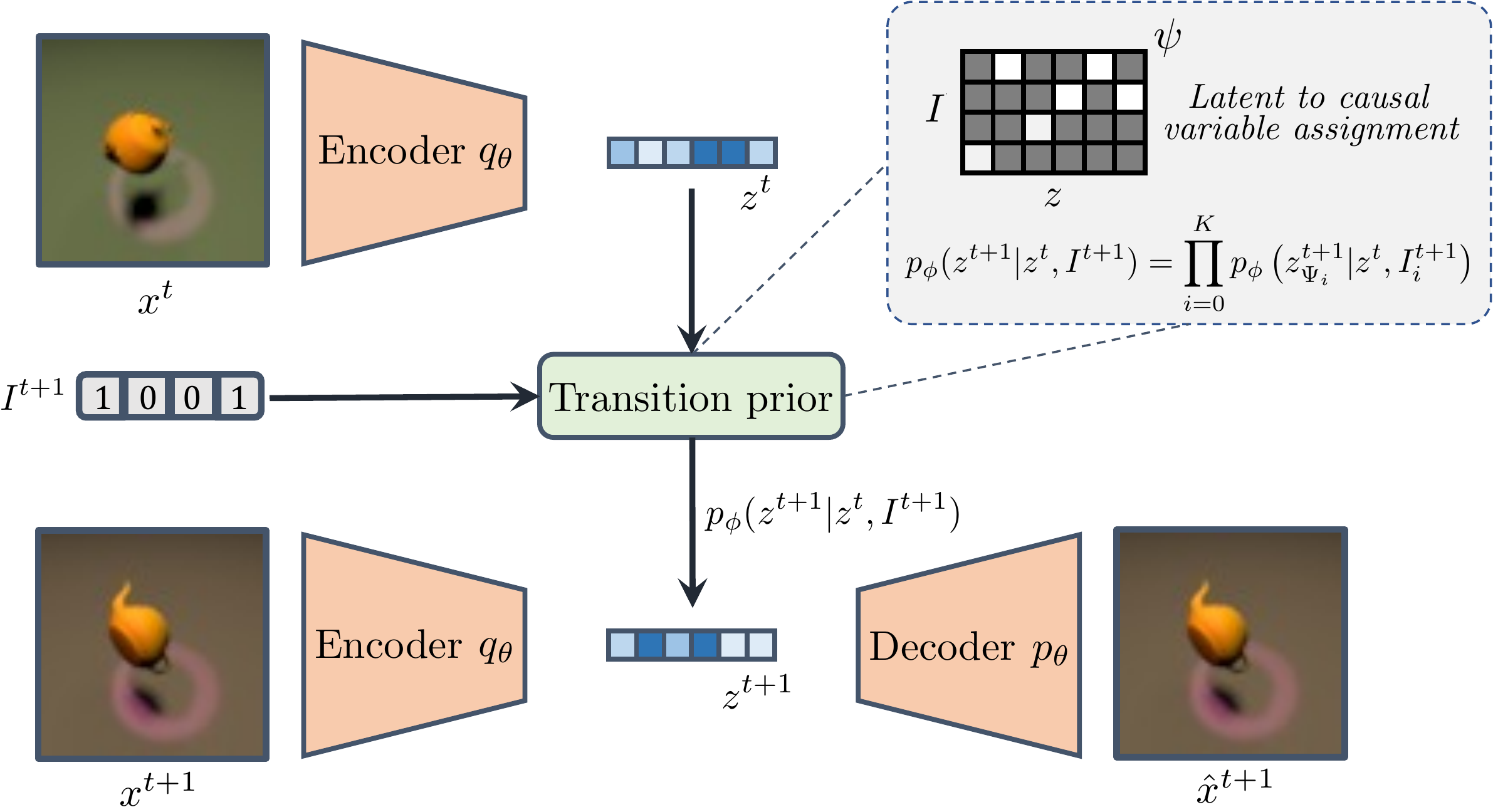}\hspace{3mm} & 
        \hspace{3mm}\includegraphics[width=\columnwidth]{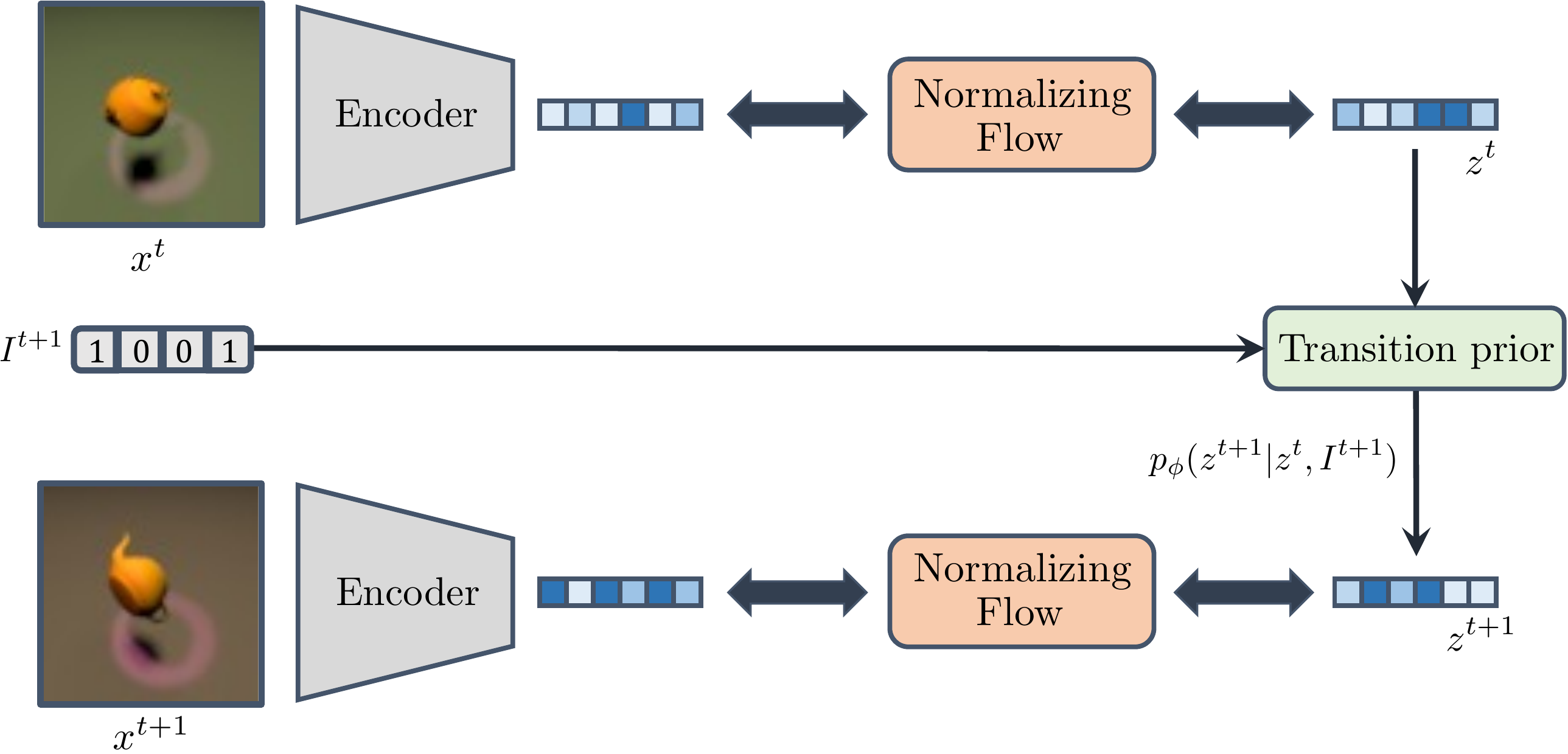} \\
    \end{tabular}
    \caption{Comparing the VAE and AE+NF setup of \OurApproach{}. \textbf{Left}: In the VAE, the encoder and decoder provide an approximate invertible mapping. The transition prior promotes disentanglement in the latent space by conditioning each latent variable on only one intervention target. \textbf{Right}: The Normalizing Flow setup uses a pretrained autoencoder, which remains frozen during training. The flow learns to map the autoencoder latents to a new space, that promotes the disentanglement similar to the VAE. The decoder is not needed during training.}
    \label{fig:method_vae_architecture}
    \label{fig:method_nf_architecture}
\end{figure*}

\subsection{Variational Autoencoder Setup}
\label{sec:method_variational_autoencoder_setup}

Inspired by several previous works \cite{locatello2020weakly, higgins2017beta,traeuble2021disentangled}, we implement the framework of \cref{sec:identifiability_learning_setup} by learning a variational autoencoder (VAE) \cite{kingma2014auto}, visualized in \cref{fig:method_vae_architecture}. 
The encoder $q_{\theta}$ and decoder $p_{\theta}$ approximate the invertible mapping $g_{\theta}$ from observations to latent space, and $p_{\phi}\left(z^{t+1}|z^{t},I^{t+1}\right)$ is the transition prior on the latent variables. 
In this VAE setup, the objective of the model becomes the Evidence Lower Bound (ELBO):
\begin{equation}
    \label{eq:method_elbo_loss_function}
    \resizebox{.9\hsize}{!}{$
    \begin{aligned}
        & \mathcal{L}_{\text{ELBO}} = -\E_{z^{t+1}}\left[\log p_{\theta}\left(x^{t+1}|z^{t+1}\right)\right]+\\
        & \E_{z^{t},\psi}\left[\sum_{i=0}^{K} D_{\mathrm{KL}}\left(q_{\theta}(\zpsi{i}^{t+1}|x^{t+1})||p_{\phi}(\zpsi{i}^{t+1}|z^{t}, I_{i}^{t+1})\right)\right]
    \end{aligned}$
    }
\end{equation}
The KL divergence uses the prior definition of \cref{eq:method_prior_distribution}. This ensures that, conditioned on the previous time step and the interventions, the different blocks of latent variables are independent.
Thereby, the assignment function of latent to causal variables, $\psi$, is learned via a Gumbel-Softmax distribution \cite{jang2017categorical} per latent variable. 
Hence, during training, we sample a latent-to-causal variable assignment from these distributions, while for inference, we can use the argmax to obtain a unique assignment.
To encourage information independent of any intervention to be modeled in $\zpsi{0}$, we weight the KL divergence of $\zpsi{0}$ with $1-\lambda$, where $\lambda>0$ is a hyperparameter (usually $\lambda=0.01$).

The prior $p_{\phi}$ for each set of latents $\zpsi{i}^{t+1}$ is implemented by an autoregressive model.
For each set of latents $\zpsi{i}^{t+1}$, the model takes $z^{t}$, $I_i^{t+1}$ and $z^{t+1}$ as input, where we sample from $\psi$ and mask the dimensions of $z^{t+1}$ for which $\psi(j)\neq i$.
From this input, the model predicts one Gaussian per latent variable.
The autoregressive nature of the prior allows complex distributions over the multiple latent dimensions, while still being independent across causal variables. %

\paragraph{Target Classifier}
\label{sec:method_target_classifier}
Since a causal variable $C_i$ is independent of any other target variable when conditioned on $I_i$, we can use those independence relations to guide the disentanglement in latent space.
For this, we propose a target classifier, which is an additional small network trained to predict the intervention targets from the latent variables over time, \ie{} modeling $p(I^{t+1}|z^{t},\zpsi{i}^{t+1})$ for $i\in\range{0}{K}$.
To guide the disentanglement, we take the gradients of the same objective with respect to $z^{t},\zpsi{i}^{t+1}$ and $\psi$, but only for $I^{t+1}_i$.
In other words, we explicitly train $\zpsi{i}^{t+1}$ having a higher mutual information with the intervention target $I^{t+1}_i$ of its intended causal variable $C_i$.
For all other targets $I_j^{t+1}$, we change the label to the conditional distribution $p(I_j|I_i)$, since $I^{t+1}_j$ should not be fully identifiable by any other set of latents except $\zpsi{j}^{t+1}$. 
This way, the target classifier ensures that information about all causal variables is encoded in the latent space, while also guiding the disentanglement in latent space.
We provide further details in \cref{sec:appendix_experimental_details_target_classifier}, and empirically verify its benefits in \cref{sec:experiments_causal3dident}.

\subsection{Using Pretrained Autoencoders}
\label{sec:method_normalizing_flow}

Despite the improved optimization process by the addition of the target classifiers, VAEs can still struggle to model high-dimensional complex images, especially when small details in the image are relevant. 
To overcome this issue, we propose an adaptation of \OurApproach{} to pretrained autoencoders. 
In this setting, the invertible map $g_{\theta}$ is implemented by a deep autoencoder, which is trained to encode and decode the high-dimensional observations to low-dimensional feature vectors, independently of any disentanglement. 
During training, we add small Gaussian noise to the latents to prevent the latent distribution from collapsing to single delta peaks. 
However, we do not enforce a certain prior like in a VAE, hence allowing complex marginal distributions. 

In a second step, after the autoencoder converged, we freeze its parameters and learn a normalizing flow \cite{rezende2015variational} that maps the entangled latent representation to a disentangled version.
The invertibility of the normalizing flow ensures that no information is lost when mapping from the entangled to the disentangled latent space, and thus we can use the pretrained decoder to reconstruct the observations without requiring any fine-tuning.
Compared to the VAE setup in \cref{sec:method_variational_autoencoder_setup}, we replace the encoder by a successive application of the frozen encoder and a normalizing flow on the encoded latents, shown in \cref{fig:method_nf_architecture}.
Besides that, we deploy the same setup, using both the transition prior structure and target classifier. 

Compared to the VAE setup, this approach has three main benefits. 
First, the autoencoder can be trained on observational data alone, potentially reducing the amount of interventional data required.
Second, learning a separate autoencoder provides an opportunity for generalizing causal factors beyond the known dataset. 
For instance, one could train an autoencoder on two datasets, where only one has interventions, \eg{} synthetic and real-world data.
Then, since the autoencoder uses a joint latent space for both datasets, training the normalizing flow on only the dataset with interventions can lead to a disentanglement function that generalizes to the purely observational dataset. 
We verify the viability of this approach in a restricted setting in \cref{sec:experiments_causal3dident_multishape}, which opens up great potential to practical applications.
Finally, the setup is easier to optimize since the autoencoder can compress the information in an almost unrestricted latent space, while the normalizing flow solely focuses on disentanglement.

%% file: sections/4_related_work.tex
\section{Related Work}
\label{sec:related_work}

Identifying independent factors of variations from data is a well-studied field in machine learning \cite{reed2014learning, higgins2017beta, kumar2018variational, locatello2019challenging, locatello2020disentangling, klindt2021towards, locatello2020weakly}.
One of the first lines of work is Independent Component Analysis (ICA) \cite{comon1994independent, hyvaerinen2001independent}.
ICA tries to recover independent latent variables that were transformed by some invertible transformation.
Although not generally possible in the non-linear case \cite{hyvarinen1999nonlinear}, ICA was recently extended to this setting by exploiting auxiliary variables under which the latents become conditionally mutually independent \cite{hyvarinen2016unsupervised, hyvaerinen2019nonlinear}. 
Several follow-up works extended this work to deep learning architectures like VAEs \cite{khemakhem2020variational, sorrenson2020disentanglement, khemakhem2020ice, zimmermann2021contrastive}.
Recent works draw a connection between causality and ICA \cite{gresele2021independent, monti2019causal}.
In particular, \citet{lachapelle2021disentanglement, yao2021learning} discuss identifiability from temporal sequences and bring it into context to causality.
While \citet{lachapelle2021disentanglement} can model interventions as external actions, \citet{yao2021learning} can model soft interventions through their non-stationary noise. On the other hand, they do not exploit the knowledge of the intervention targets as we do and therefore require additional assumptions in terms of sufficient variation. 
Moreover, both of these works require scalar causal variables, while \OurApproach{} generalizes to multidimensional causal factors.

A second, related line of work is causal representation learning \cite{schoelkopf2021towards}, which aims at discovering causal structures and variables from data.
For instance, \citet{locatello2020weakly} showed that one can identify independent latent causal factors from pairs of observations that only differ in $k$ causal factors.
\citet{yang2021causalvae} propose a VAE that integrates a structural causal model in its prior, but requires the true causal variables as labels during training.
\citet{vonkuegelgen2021self} demonstrated that common contrastive learning methods can block-identify the causal variables that remain unchanged under augmentations.
\OurApproach{} similarly identifies multidimensional causal factors as a block, and, furthermore, disentangles individual causal factors by grouping the latent space.

%% file: sections/5_experiments.tex
\section{Experiments}
\label{sec:experiments}

We evaluate \OurApproach{} on two video datasets, and compare it to common disentanglement methods.
We include further details for reproducibility in \cref{sec:appendix_experimental_details}, and make our code publicly available.\footnote{\url{https://github.com/phlippe/CITRIS}}

\subsection{Experimental Setup}
\label{sec:experiments_setup}
\label{sec:experiments_setup_dataset}

\begin{figure}[t!]
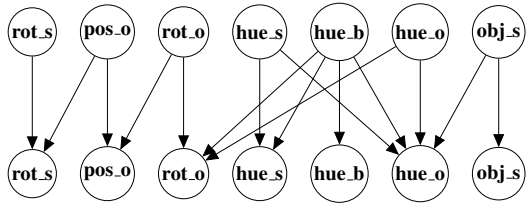

    \centering
    \resizebox{0.85\columnwidth}{!}{
        \tikz{ %
    		\node[latent] (pos) {\textbf{pos\_o}} ; %
    		\node[latent, right=of pos, xshift=-.6cm] (rot) {\textbf{rot\_o}} ; %
    		\node[latent, left=of pos, xshift=.6cm] (rots) {\textbf{rot\_s}} ; %
    		\node[latent, right=of rot, xshift=-.6cm] (hues) {\textbf{hue\_s}} ; %
    		\node[latent, right=of hues, xshift=-.6cm] (hueb) {\textbf{hue\_b}} ; %
    		\node[latent, right=of hueb, xshift=-.6cm] (hueo) {\textbf{hue\_o}} ; %
    		\node[latent, right=of hueo, xshift=-.6cm] (objs) {\textbf{obj\_s}} ; %
    		
    		\node[latent, below=of pos, yshift=-.5cm] (pos1) {\textbf{pos\_o}} ; %
    		\node[latent, right=of pos1, xshift=-.6cm] (rot1) {\textbf{rot\_o}} ; %
    		\node[latent, left=of pos1, xshift=.6cm] (rots1) {\textbf{rot\_s}} ; %
    		\node[latent, right=of rot1, xshift=-.6cm] (hues1) {\textbf{hue\_s}} ; %
    		\node[latent, right=of hues1, xshift=-.6cm] (hueb1) {\textbf{hue\_b}} ; %
    		\node[latent, right=of hueb1, xshift=-.6cm] (hueo1) {\textbf{hue\_o}} ; %
    		\node[latent, right=of hueo1, xshift=-.6cm] (objs1) {\textbf{obj\_s}} ; %
    		
    		\edge{pos}{pos1} ;
    		\edge{rot}{rot1} ;
    		\edge{rots}{rots1} ;
    		\edge{hues}{hues1} ;
    		\edge{hueb}{hueb1} ;
    		\edge{hueo}{hueo1} ;
    		\edge{objs}{objs1} ;
    		
    		\edge{rot}{pos1} ;
    		\edge{hueb}{rot1} ;
    		\edge{hueo}{rot1} ;
    		\edge{hueb}{hues1} ;
    		\edge{objs}{hueo1} ;
    		\edge{hues}{hueo1} ;
    		\edge{hueb}{hueo1} ;
    		\edge{pos}{rots1} ;
    	}
    }
    \caption{The temporal causal graph among the 7 dimensions of variation. The graph structure covers a wide range of scenarios, such as confounders, chains, and up to 4 parents per variable.}
    \label{fig:experiments_dataset_causal3d}
    \label{fig:experiments_dataset_causal3d_examples}
    \label{fig:experiments_dataset_causal3d_causal_graph}
\end{figure}

\textbf{Temporal Causal3DIdent} We evaluate \OurApproach{} on an adapted, temporal version of the Causal3DIdent identifiability benchmark \cite{zimmermann2021contrastive, vonkuegelgen2021self}. 
The dataset consists of 3D renderings of objects, for which we consider seven different causal factors: the object position as multidimensional vector $[x,y,z]\in[-2,2]^3$; the object rotation with two dimensions $[\alpha,\beta]\in [0,2\pi)^2$; the hue of the object, background and spotlight in $[0,2\pi)$; the spotlight's rotation in $[0,2\pi)$; and the object shape (categorical with seven values). 
The relations among those variables are shown in \cref{fig:experiments_dataset_causal3d_causal_graph}. 
Each continuous variable follows a Gaussian distribution over time, where the mean is a (non-linear) function of the parents.
Overall, this causes strong dependencies among the variables. 
We perform perfect interventions with $I^{t}_i\sim\text{Bernoulli}(0.1)$, ensuring the minimal causal variables to be the true factors.
Experiments with confounded interventions are included in \cref{sec:appendix_additional_experiments_causal3d} with similar results.
Further details on this dataset are summarized in Appendix~\ref{sec:appendix_experimental_details_causal3d_dataset}. 

\begin{figure}[t!]
    \centering
    \begin{tabular}{cccc}
        \hspace{-1mm}\includegraphics[width=0.1\textwidth]{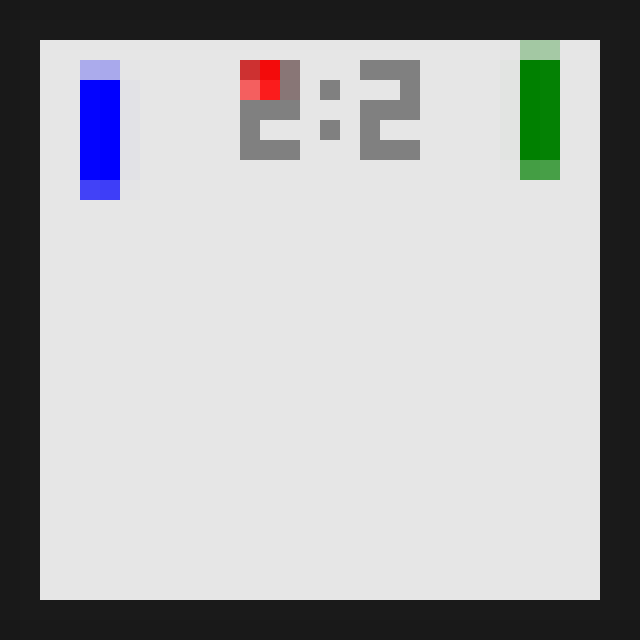} & 
        \includegraphics[width=0.1\textwidth]{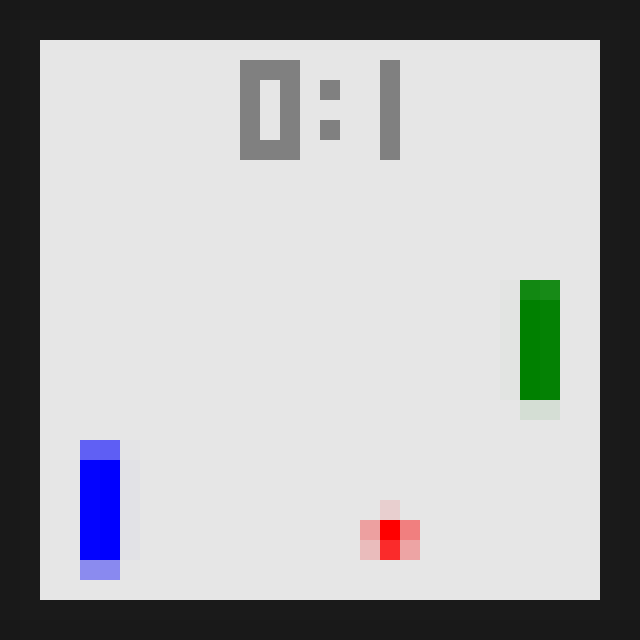} & 
        \includegraphics[width=0.1\textwidth]{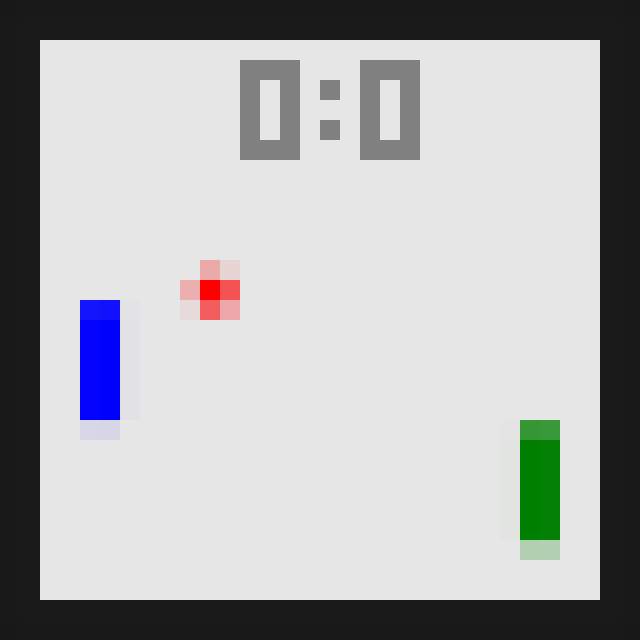} & 
        \includegraphics[width=0.1\textwidth]{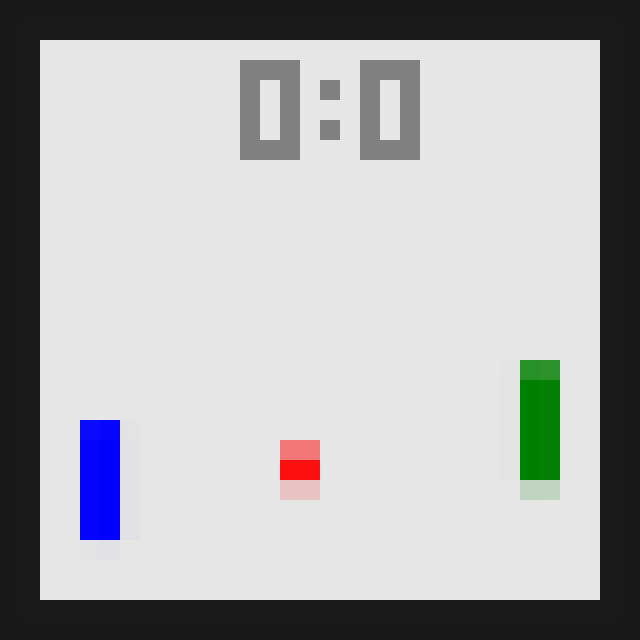} \\
    \end{tabular}
    \caption{Examples of Interventional Pong. The score represents a causal factor on which we do not have any interventions. The velocity of the ball is encoded in a 4th channel, not visualized here.}
    \label{fig:experiments_dataset_pong_examples}
\end{figure}

\textbf{Interventional Pong} As a second benchmark, we adapt the popular Atari game Pong \cite{bellemare2013arcade}, in which we define 5 causal factors: the $x,y$ position of the ball, the $y$-positions of the two paddles, and the velocity direction of the ball (angle in $[0,2\pi)$). 
Further, we add the score as a sixth causal factor, which is part of the game's dynamics.
However, we do not provide any interventions on it.
Hence, in \OurApproach{}, this information should be modeled in $\zpsi{0}$.
The temporal dependencies follow the dynamics of Pong, where the ball can collide with the paddles and walls, and the two paddles move towards the ball to hit it.
We provide single-target interventions for the variables.
For the paddles, the imperfect interventions override the policy and take a random action (up/down), while keeping a smooth motion.
For the causal factors of the ball, we uniformly sample a new value when intervened, causing a possible jump.
Example frames of the environment are shown in \cref{fig:experiments_dataset_pong_examples}. 

\paragraph{Baselines}
\label{sec:experiments_setup_baselines}
We compare \OurApproach{} to SlowVAE \cite{klindt2021towards}, a state-of-the-art disentanglement method for temporal sequences.
Notably, SlowVAE assumes that the factors of variation are independent.
As we show, this assumption, which often cannot be met in more complex settings like Temporal Causal3DIdent, has detrimental effects on modelling, underlining the importance of modelling the correlations between causal variables.
The second baseline we consider is iVAE \cite{khemakhem2020variational}, a VAE model which conditions its components on additional observed variables $u$.
In our setup, $u$ corresponds to the previous time step observation $x^{t}$ and the intervention targets $I^{t+1}$.
As we aim to find a mapping from image to a causal space, which is independent of those factors, we must adapt iVAE to only condition its prior on $u$.
We refer to this model variant as \iVAEAdapt{}.
With this change, the main difference between \iVAEAdapt{} and \OurApproach{} becomes the structure of the prior $p(z_{t+1} | z_t, I_{t+1})$, showcasing the significance of \OurApproach{}'s identifiability guarantees in this setup.
Hyperparameter details are listed in \cref{sec:appendix_experimental_details}, and the code is provided in the supplementary material.

\newcolumntype{C}{c<{\hspace{4mm}}}
\newcolumntype{L}{c>{\hspace{2mm}}}
\begin{table*}[t!]
    \centering
    \caption{Results on the Temporal-Causal3DIdent dataset. All triplet distances have a distance at random of $1$ and optimum of $0$. \textit{Oracle} represents the accuracy of a CNN trained supervised to predict the causal factors from images, hence constituting a lower error limit. In the correlation metrics, \textit{diag} refers to the average score of the predicted causal factor to its true value (optimal 1), and \textit{sep} for the average of the maximum correlation per predicted causal variable besides its true factor (optimal 0). Standard deviations over 3 seeds shown in \cref{sec:appendix_experimental_details_causal3d_dataset}. \OurApproach{} is able to disentangle the causal factors well, and \OurApproach{}-NF even accurately models all 7 shapes. }
    \label{tab:experiments_results_teapot}
    \label{tab:experiments_results_all_shapes}
    \resizebox{\textwidth}{!}{
    \begin{tabular}{lccccccccccCcccc}
        \toprule
        & \multicolumn{11}{c}{\textbf{Triplet evaluation distances} $\downarrow$} & \multicolumn{4}{c}{\textbf{Correlation metrics}}\\\cmidrule(r{4mm}){2-12}\cmidrule{13-16}
        & \texttt{pos\_x} & \texttt{pos\_y} & \texttt{pos\_z} & \texttt{rot\_}$\alpha$ & \texttt{rot\_}$\beta$ & \texttt{rot\_s} & \texttt{hue\_s} & \texttt{hue\_b} & \texttt{hue\_o} & \texttt{obj\_s} & Mean & $R^2$ diag $\uparrow$ & $R^2$ sep $\downarrow$ & Spearman diag $\uparrow$ & Spearman sep $\downarrow$\\ 
        \midrule
        \multicolumn{16}{c}{\cellcolor{gray!25}\textbf{Temporal-Causal3DIdent Teapot}}\\
        \textbf{Oracle} & 0.02 & 0.02 & 0.02 & 0.02 & 0.03 & 0.01 & 0.02 & 0.01 & 0.02 & - & 0.02 & - & - & - & - \\
        \midrule
        \textbf{SlowVAE} & 0.13 & 0.10 & 0.12 & 0.50 & 0.59 & 0.22 & 0.64 & 0.21 & 0.17 & - & 0.30 & 0.65 & 0.20 & 0.62 & 0.27\\
        \textbf{\iVAEAdapt{}} - 9dim & 0.11 & 0.09 & 0.12 & 0.70 & 0.76 & 0.06 & 0.67 & 0.02 & 0.12 & - & 0.30 & 0.65 & 0.11 & 0.65 & 0.13\\
        \textbf{\iVAEAdapt{}} - 32dim & \highlight{0.04} & \highlight{0.03} & \highlight{0.04} & 0.25 & 0.31 & \highlight{0.03} & 0.58 & 0.02 & 0.05 & - & 0.15 & 0.78 & 0.21 & 0.77 & 0.17\\
        \midrule
        \textbf{\OurApproach-VAE} & 0.05 & 0.04 & 0.05 & 0.10 & 0.20 & \highlight{0.03} & 0.08 & 0.02 & 0.05 & - & 0.07 & 0.96 & 0.02 & 0.95 & \highlight{0.04} \\
        - No target classifier & 0.05 & 0.04 & 0.05 & 0.62 & 0.66 & 0.19 & \highlight{0.04} & 0.02 & 0.17 & - & 0.20 & 0.79 & 0.15 & 0.76 & 0.12 \\
        \textbf{\OurApproach-NF} & \highlight{0.04} & \highlight{0.03} & \highlight{0.04} & \highlight{0.06} & \highlight{0.10} & \highlight{0.03} & \highlight{0.04} & \highlight{0.01} & \highlight{0.04} & - & \highlight{0.04} & \highlight{0.98} & \highlight{0.01} & \highlight{0.97} & 0.05 \\
        \midrule\midrule
        \multicolumn{16}{c}{\cellcolor{gray!25}\textbf{Temporal-Causal3DIdent 7-shapes}}\\
        \textbf{Oracle} & 0.08 & 0.06 & 0.08 & 0.06 & 0.09 & 0.04 & 0.04 & 0.01 & 0.04 & 0.00 & 0.05 & - & - & - & - \\
        \midrule
        \textbf{SlowVAE} & 0.44 & 0.25 & 0.41 & 0.69 & 0.75 & 0.25 & 0.57 & 0.10 & 0.14 & 0.37 & 0.40 & 0.61 & 0.23 & 0.59 & 0.27 \\
        \textbf{\iVAEAdapt{}} & 0.26 & 0.23 & 0.34 & 0.58 & 0.65 & 0.10 & 0.31 & 0.02 & 0.09 & 0.14 & 0.27 & 0.80 & 0.29 & 0.77 & 0.28\\
        \midrule
        \textbf{\OurApproach-VAE} & 0.15 & 0.13 & 0.23 & 0.54 & 0.71 & 0.07 & 0.05 & 0.02 & 0.06 & 0.18 & 0.21 & 0.89 & 0.10 & 0.88 & 0.12\\
        \textbf{\OurApproach-NF} & \highlight{0.12} & \highlight{0.08} & \highlight{0.11} & \highlight{0.09} & \highlight{0.14} & \highlight{0.05} & \highlight{0.05} & \highlight{0.02} & \highlight{0.06} & \highlight{0.00} & \highlight{0.07} & \highlight{0.98} & \highlight{0.04} & \highlight{0.97} & \highlight{0.08} \\
        \bottomrule
    \end{tabular}
    }
\end{table*}

\paragraph{Correlation Metrics}
\label{sec:experiments_setup_evaluation_metrics} 
Following common practice, we report the correlation of the learned latent variables to the ground truth causal factors. 
Since in our setup, multiple latent variables can jointly describe a single causal variable, we first learn a mapping between such, \eg{}, with an MLP.
For \OurApproach{}, we apply one MLP per set of latent variables that are assigned to the same causal factor by $\psi$. 
The MLP is then trained to predict \emph{all} causal factors per set of latents, on which we measure the correlation.
Thereby, no gradients are propagated through the model.
\iVAEAdapt{} and SlowVAE do not learn an assignment of latent to causal factors.
As an alternative, we assign each latent dimension to the causal factor it has the highest correlation with.
Although this gives the baselines a considerable advantage, it shows whether \OurApproach{} can improve upon the baselines beyond finding a good latent to causal factor assignment.
We report both the $R^2$ coefficient of determination \cite{wright1921correlation} and the Spearman's rank correlation coefficient \cite{spearman1904proof} of the predicted values against the ground truth causal factors.
However, in our datasets, the causal factors are correlated themselves. This makes it difficult to spot spurious correlations between latents and causal factors.
To overcome this issue, we measure the correlations on a test dataset for which we sample the causal factors independently.

\begin{figure}[t!]
    \centering
    \scriptsize
    \begin{tabular}{cccc}
        \hspace{-1mm}\includegraphics[width=0.1\textwidth]{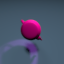} & 
        \includegraphics[width=0.1\textwidth]{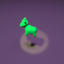} & 
        \includegraphics[width=0.1\textwidth]{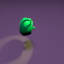} & 
        \includegraphics[width=0.1\textwidth]{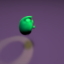} \\
        Image 1 & Image 2 & Ground Truth & Prediction \\[7pt]
        \hspace{-1mm}\includegraphics[width=0.1\textwidth]{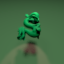} & 
        \includegraphics[width=0.1\textwidth]{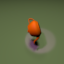} & 
        \includegraphics[width=0.1\textwidth]{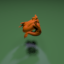} & 
        \includegraphics[width=0.1\textwidth]{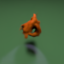} \\
        Image 1 & Image 2 & Ground Truth & Prediction \\[7pt]
        \hspace{-1mm}\includegraphics[width=0.1\textwidth]{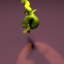} & 
        \includegraphics[width=0.1\textwidth]{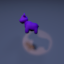} & 
        \includegraphics[width=0.1\textwidth]{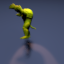} & 
        \includegraphics[width=0.1\textwidth]{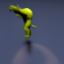} \\
        Image 1 & Image 2 & Ground Truth & Prediction \\
    \end{tabular}
    \caption{Example triplet evaluations on the Temporal-Causal3DIdent. In the first row, the ground truth combines the spotlight rotation and object shape from Image 1 with all other causal factors from Image 2. The prediction is generated by encoding the two images, and performing the same combination of causal factors in latent space. The shown prediction by \OurApproach{}-NF only slightly differs in the object rotation. For details on the other predictions, see \cref{sec:appendix_additional_experiments_causal3d}.}
    \label{fig:experiments_triplet_visualizations}
\end{figure}

\textbf{Triplet Evaluation} 
To also evaluate the decoding part of the model, we propose another parameter-free evaluation, \textit{triplet evaluation}, to reveal complex dependencies between latent variables.
For this, we create triplets of images: the first two are randomly sampled test images, while the third one is created based on a random combination of causal factors of the first two images. 
For example, in \cref{fig:experiments_triplet_visualizations}, we take the spotlight rotation and object shape from image 1, and all other causal factors from image 2.
For evaluation, we then encode the two test images independently, perform the combination of ground-truth causal factors as done for the third image in latent space, and use the decoder to generate a new image, which ideally resembles the ground truth third image.
Since the reconstruction error is not descriptive of the errors being made, \eg{} the rotation in \cref{fig:experiments_triplet_visualizations}, we train an additional CNN in a supervised manner that maps images to the causal factors. 
With this model, we can extract the causal factors from the generated image, and report the average distance of these to the ground truth causal factors.

\subsection{Temporal Causal3DIdent Experiments}
\label{sec:experiments_causal3dident}

\paragraph{Teapot Experiments} 
The Causal3DIdent benchmark constitutes a challenging dataset due to its various interactions of causal factors in the high-dimensional observational space.
To slightly simplify the problem, especially for learning a VAE on the high-dimensional observational space, we first show experiments on the teapot shape only, Temporal-Causal3DIdent Teapot.
All results are summarized in \cref{tab:experiments_results_teapot}.
Both the VAE and Normalizing Flow version of \OurApproach{} considerably outperformed the two baselines and were able to achieve an average $R^{2}$ and Spearman correlation above 0.9, while keeping the correlation between factors low. 
Moreover, \OurApproach{}-NF achieves close-to optimal scores on the triplet evaluation, and especially outperforms the VAE-based approaches in modeling the rotations.
This is because the autoencoder is able to reconstruct images with negligible error, while the VAE optimization involves balancing the small difference in output space with its KL divergence regularization in latent space.
Further, an ablation study on training \OurApproach{}-VAE without the target classifier verifies that the additional guidance can considerably help in optimization.
Finally, in \cref{sec:appendix_additional_experiments_causal_graph}, we show that the original causal graph can be found from \OurApproach{}'s latents.

In contrast, the SlowVAE entangles the causal factors due to their strong correlation over time.
The \iVAEAdapt{} disentangles the position and the rotation of the spotlight well.
However, the hue of the spotlight was highly entangled since its appearance differs with different background and object colors.
We further report the performance of an \iVAEAdapt{} with equal number of latent dimensions as causal factors.
It is apparent that the smaller latent space is particularly insufficient for modeling the rotation angles, showing the need for higher dimensional latent spaces, not only in \OurApproach{}.

\begin{table}[t!]
    \centering
    \caption{Results on the Temporal-Causal3DIdent dataset with \OurApproach{}-NF trained on 5 object shapes. The same metrics as in \cref{tab:experiments_results_teapot} are reported, with ``Spr'' being the Spearman correlation score. The full table is in \cref{sec:appendix_additional_experiments_causal3d}. For 2 unseen shapes, \OurApproach{}-NF can yet disentangle their causal factors well.}
    \label{tab:experiments_results_causal3d_generalization}
    \resizebox{\columnwidth}{!}{%
    \begin{tabular}{lccccc}
        \toprule
        \textbf{\OurApproach-NF} & \textbf{Triplets} $\downarrow$ & \textbf{$R^2$ diag $\uparrow$} & \textbf{$R^2$ sep $\downarrow$} & \textbf{Spr diag $\uparrow$} & \textbf{Spr sep $\downarrow$}\\
        \midrule
        5 seen shapes & 0.09 & 0.98 & 0.05 & 0.97 & 0.10\\
        2 unseen shapes & 0.23 & 0.94 & 0.15 & 0.93 & 0.19\\
        \bottomrule
    \end{tabular}%
    }
\end{table}

\paragraph{7-shapes Experiments} We apply all models on the Temporal Causal3DIdent dataset with all seven shapes (\cref{tab:experiments_results_all_shapes}).
The models have to align the rotation axes and central points of all shapes, which caused higher triplet distances for rotations across models.
While \OurApproach{}-VAE still outperformed the other baselines, \OurApproach{}-NF significantly improved upon that, still maintaining a disentanglement similar to the single shape experiments.
This underlines the optimization benefits of using pretrained autoencoders for disentanglement learning on complex, high-dimensional observations.

\paragraph{Generalization of Causal Representations}\label{sec:experiments_causal3dident_multishape} 
Since the autoencoder is trained on observational data, we evaluate whether its causal representation can generalize to new, unseen settings.
For this, we reuse the same autoencoder as before, but train the Normalizing Flow on an interventional dataset which excludes any observations from two shapes (Head and Cow).
Afterwards, we test its zero-shot generalization to the two unseen shapes (see \cref{tab:experiments_results_causal3d_generalization}). 
Note that optimal performance cannot be achieved here, since the central point and default rotation of an object cannot be generalized to other objects.
Nonetheless, the results in \cref{tab:experiments_results_causal3d_generalization} indicate a strong disentanglement among factors, with slight decreases in position and rotation due to the forementioned limitations.
This shows that the learned disentanglement function can indeed generalize to unseen instantiations of causal factors, promising potential for future work on generalizing causal representations to unseen settings with \OurApproach{}.

\begin{table}[t!]
    \centering
    \caption{Results on the Interventional Pong dataset. The same metrics as in \cref{tab:experiments_results_teapot} are reported, with ``Spr'' being the Spearman correlation score. The full table is in \cref{sec:appendix_additional_experiments_pong}. The results how that \OurApproach{} can handle imperfect interventions.}
    \label{tab:experiments_results_pong}
    \resizebox{\columnwidth}{!}{%
    \begin{tabular}{lccccc}
        \toprule
        & \textbf{Triplets} $\downarrow$ & \textbf{$R^2$ diag $\uparrow$} & \textbf{$R^2$ sep $\downarrow$} & \textbf{Spr diag $\uparrow$} & \textbf{Spr sep $\downarrow$}\\
        \midrule
        \textbf{SlowVAE} & 0.34 & 0.61 & 0.17 & 0.66 & 0.23\\
        \textbf{\iVAEAdapt{}} & 0.09 & 0.91 & 0.04 & 0.92 & 0.06\\
        \midrule
        \textbf{\OurApproach-VAE} & 0.03 & 0.99 & \highlight{0.01} & 0.99 & \highlight{0.05} \\
        \textbf{\OurApproach-NF} & \highlight{0.02} & \highlight{1.00} & 0.04 & \highlight{1.00} & 0.10\\
        \bottomrule
    \end{tabular}%
    }
\end{table}

\subsection{Interventional Pong}
\label{sec:experiments_interventional_pong}

Finally, we report the disentanglement results of \OurApproach{} on the Interventional Pong dataset in \cref{tab:experiments_results_pong}.
The challenge of this dataset is its imperfect and correlated interventions, and the score being an unintervened causal variable.
Still, \OurApproach{} is able to disentangle the factors and also assign the score variables to its correct set of latents, \ie{} $\zpsi{0}$, showing that it can indeed handle imperfect interventions.
Although \OurApproach{}-NF obtains a higher correlation among learned latents, it again did not affect its triplet generations.
SlowVAE entangles the two paddles, since they are strongly correlated by following the same movement policy and having imperfect interventions.
\iVAEAdapt{} showed unstable behavior over seeds, and entangled the ball position and velocity.

%% file: sections/6_conclusion.tex
\section{Conclusion}
\label{sec:conclusion}

We propose \OurApproach{}, a VAE framework for learning causal representations.
\OurApproach{} identifies the minimal causal variables of a dynamical system from temporal, intervened sequences.
Furthermore, by using normalizing flows, \OurApproach{} learns to disentangle the representation of pretrained autoencoders. 
In experiments, \OurApproach{} reliably recovered the causal factors of 3D rendered images.
Moreover, we empirically showed that \OurApproach{} can generalize to unseen instantiations of causal factors.
This promises great potential for future work on simulation-to-real generalization research for causal representation learning.
As future work, \OurApproach{} can be extended to an active learning setup, allowing for more data-efficient causal identifiability methods in practice.
Further, future work could consider the setup of robotic simulators \cite{szot2021habitat, makoviychuk2021isaac}, where interventions are available through sequences of actions.

%% file: sections/7_acknowledgements.tex
\section*{Acknowledgements}
We thank Johann Brehmer and Pim de Haan for valuable discussions throughout the project.
We also thank SURFsara for the support in using the Lisa Compute Cluster.
This work is financially supported by Qualcomm Technologies Inc., the University of Amsterdam and the allowance Top consortia for Knowledge and Innovation (TKIs) from the Netherlands Ministry of Economic Affairs and Climate Policy.

%% file: sections/appendix_sections/0_statements.tex
\section{Reproducibility Statement}
\label{sec:appendix_reproducibility}

To ensure reproducibility, we publish the code for all models used in this paper at \url{https://github.com/phlippe/CITRIS}.
Further, we include the code for generating the Interventional Pong dataset and the sequences of the causal factors in the Temporal-Causal3DIdent dataset.
The complete datasets of this paper are released with corresponding licenses, and links to those datasets are available in the code repository.
Moreover, we give a detailed overview of the datasets and more visual examples in \cref{sec:appendix_experimental_details_causal3d_dataset} and \cref{sec:appendix_experimental_details_interventional_pong}.

Further, for all experiments of \cref{sec:experiments}, we have included a detailed overview of the hyperparameters in \ref{sec:appendix_experimental_design_hyperparameters} and additional implementation details of the evaluation metrics and model architecture components in \cref{sec:appendix_experimental_details_experimental_design}.
All experiments have been repeated with 3 seeds to obtain stable, reproducible results.
We provide an overview of the standard deviations, as well as additional results in \cref{sec:appendix_additional_experiments}.

Finally, all experiments in this paper were performed on a single NVIDIA TitanRTX GPU with a 6-core CPU.
The overall computation time of all experiments together in this paper correspond to approximately $80$ GPU days (excluding hyperparameter search and trials during the research).

%% file: sections/appendix_sections/1_proofs.tex
\section{Proofs}
\label{sec:appendix_proofs}

The following section contains the proof for \cref{theo:method_intv_over_time}.
We first give an overview of the notation and additional preliminary discussions in \cref{sec:appendix_proofs_preliminaries}.
Then, we give an outline of the proof.
The remaining sections provide the details of the proof.

Additionally, we provide the proof and more details on Proposition~\ref{prop:method_tris_nonidentifiability} in \cref{sec:appendix_proofs_proposition}.

\subsection{Preliminaries}
\label{sec:appendix_proofs_preliminaries}
\subsubsection{Summary of Notation}

We summarize the notation, which is the same as used for the main paper, as follows:
\begin{compactitem}
    \item We assume $K$ causal factors $C_1, \dots, C_K$ such that $C_i \in \mathcal{D}_i^{M_i}$ with $M_i \geq 1$; 
    \item We can group all causal factors in a single variable $C= (C_1, \dots, C_K) \in \mathcal{C}$, where $\mathcal{C}$ is the causal factor space $\mathcal{C} = \mathcal{D}_1^{M_1}\times\mathcal{D}_2^{M_2}\times...\times\mathcal{D}_K^{M_K}$;
    \item The data is generated by a latent Dynamic Bayesian network with variables $(C^t_1,C^t_2,...,C^t_K)_{t=1}^T$;
    \item We assume to know at each time step the binary intervention vector $I^{t}\in \{0,1\}^{K+1}$ where $I^{t}_i=1$ refers to an intervention on the causal factor $C_i^{t+1}$. As a special case $I^t_0=0$ for all $t$;
    \item For each causal factor $C_i$, there exists a split $\sdep_i(C_i),\sindep_i(C_i)$ such that $\sdep_i(C_i)$ represents the variable/manipulable part of $C_i$, while $\sindep_i(C_i)$ represents the invariable part of $C_i$;
    \item The minimal causal split is defined as the one which only contains the intervention-dependent information in $\sdep_i(C_i)$, and everything else in $\sindep_i(C_i)$. This split is denoted by $\sdepStar(C_i)$ and $\sindepStar(C_i)$
    \item At each timestep we can access observations $x^t, x^{t+1}\in \mathcal{X} \subseteq \mathbb{R}^{N}$;
    \item $\mathcal{E}$ is the space of the noise variables that affect the observation without changing the encoding of the causal factors. For example, this could be random color shifts in Pong, or brightness shifts in Causal3D, since no causal factor is encoded in color and brightness in these setups respectively;
    \item Observation function $\obs: \mathcal{C} \times \mathcal{E} \rightarrow \mathcal{X}$, where $\mathcal{E}$ is the space of the noise variables;
    \item Latent vector $z^t \in \mathcal{Z} \subseteq\mathbb{R}^{M}$, where $\mathcal{Z}$ is the latent space of dimension $M \geq \text{dim}(\mathcal{E})+\text{dim}(\mathcal{C})$;
    \item Inverse of the observation function in the latent space $g_{\theta}: \mathcal{X}\to \mathcal{Z}$;
    \item Assignment function from latent dimensions to causal factors $\psi: \range{1}{M}\to\range{0}{K}$;
    \item Disentanglement function $\disentanglement^*: \mathcal{X} \to \mathcal{\tilde{C}}\times \mathcal{\tilde{E}}$ with $\mathcal{\tilde{C}}=\mathcal{D}^{\tilde{M_1}}\times...\times \mathcal{D}^{\tilde{M_K}}$ and $\tilde{M_i}$ being the number of latent dimensions assigned to causal factor $C_i$ by $\psi^*$. We denote the output of $\disentanglement^*$ for an observation $X$ as $\disentanglement^*(X)=(\tilde{C}_1,\tilde{C}_2,...,\tilde{\noisevarcap})$. Then, $\disentanglement^*$ is a disentanglement function if there exist a set of deterministic functions $h_0,h_1,...,h_K$ for which, for any $X=\obs(C,\noisevarcap)$, $h_i(\tilde{C}_i)=C_i$ for all $i\in\range{1}{K}$, and $h_0(\tilde{\noisevarcap})=\noisevarcap$.
    \item The representation of $\disentanglement^{*}$ in terms of the learnable function is denoted by $g_{\theta}^*$ and $\psi^{*}$;
    \item Latent variables assigned to each causal factor $C_i$ by $\psi$ are denoted as $\zpsi{i}=\{z_j|j\in\range{1}{M}, \psi(j)=i\} = \{g_\theta(x^t)_j |j\in\range{1}{M}, \psi(j)=i\}$;
    \item The remaining latent variables that are not assigned to any causal factor are denoted as $\zpsi{0}$;
    \item The goal is to learn for each $C_i$: $ p_{\phi}\left(\zpsi{i}^{t+1}|z^{t}, I_{i}^{t+1}\right) \approx     p\left(\sdep_i(C^{t+1}_i)|C^{t},I^{t+1}_i\right)$; 
\end{compactitem}

\subsubsection{Limiting Density of Discrete Points}

In this section, we give a short overview on the difference between differential entropy and the limiting density of discrete points approach, introduced by \citet{jaynes1957information, jaynes1968prior}.
Differential entropy on a continuous random variable $X$ with a distribution $p(X)$ is defined as:
\begin{equation}
    H(X)=-\int p(X)\log p(X) dx
\end{equation}
While for discrete variables, entropy has the intuitive explanation of an uncertainty measure, or the 'information' of a variable, one cannot draw the same relation so easily for continuous variables.
This is because differential entropy lacks properties that would be necessary for that.
For one, the entropy can become negative.
Secondly, and most importantly for the use case in this paper, it is not invariant under invertible transformations, \ie{} a change of variables.
For the example of the random variable $X$, the entropy of $H(X)$ does not necessarily equal to $H(aX)$ where $a$ is a constant factor, \eg{} $a=2$.
Thus, it becomes difficult to use differential entropy as a measure of information content of a continuous variable, like in the discrete case.

One approach that was proposed to overcome these issues is the limiting density of discrete points (LDDP) \citet{jaynes1957information, jaynes1968prior}.
It adjusts the definition of differential entropy by introducing an \emph{invariant measure} $m(X)$, which can be seen as a reference distribution we measure the entropy of $p(X)$ to.
Intuitively, the LDDP adjustment is derived from arguing that the continuous entropy should be derived by taking the limit of increasingly dense discrete distributions.
In the limit of infinitely many discrete points, one arrives at the entropy for continuous functions, which becomes:
\begin{equation}
    H(X) = -\int p(X)\log \frac{p(X)}{m(X)}dx
\end{equation}
Note that in some formulations, a constant $\log N$ is added to this equation, where $N$ is the number of discrete points considered which goes against infinity in the limit.
Since for this paper, we only require to compare two entropy values with each other and do not require the entropy to take a specific value, we can neglect this constant.

One crucial property of LDDP, which we use in the following proof, is that the entropy stays invariant under a change of variable.
This is achieved by transforming the invariant measure $m(X)$ by the exact same invertible transformation as done for $p(X)$.
Therefore, when coming back to the example of scaling $X$ by a constant factor, both $p(X)$ and $m(X)$ change in the same way, resulting in $H(X)=H(aX)$.

\subsection{Proof Outline}
\label{sec:appendix_proofs_outline}

The goal of this section is to proof \cref{theo:method_intv_over_time}: the global optimum of \OurApproach{} will find the minimal causal variables.
We will take the following steps in the proof:
\begin{enumerate}
    \item (\cref{sec:appendix_proofs_step_1}) Firstly, we show that the function $\disentanglement^{*}$ that disentangles the true latent variables $C_1,...,C_K$ and assigns them to the corresponding sets $\zpsi{1},...,\zpsi{K}$ constitutes a global, but not necessarily unique, optimum for maximizing the likelihood of \cref{eq:method_prior_distribution}. 
    \item (\cref{sec:appendix_proofs_step_2}) Next, we characterize the class of disentanglement functions $\disentanglementClass^{*}$ which all represent a global maximum of the likelihood, \ie{} get the same score as the true disentanglement. In particular, we show that in all optimal disentanglement functions, each assignment set $\zpsi{i}$ contains the variable part of the causal factor $\sdep_i(C_i)$, but that it might contain also the invariable parts of any other causal factor, thus creating multiple optimal solutions. 
    We do this in two sub-steps:
    \begin{enumerate}
        \item First, we assume that all intervention targets are independent, \ie{} $I^{t+1}_i\independent I^{t+1}_j|C^{t}$ for any $i\neq j$.
        \item Secondly, we extend it to a wider group of intervention settings where interventions might be confounded, and show that all of them fall in the same class $\disentanglementClass^{*}$.
    \end{enumerate} 
    \item (\cref{sec:appendix_proofs_step_3}) Finally, we derive \cref{theo:method_intv_over_time} by showing that the function $\hat{\disentanglement}\in \disentanglementClass^{*}$, which maximizes the entropy of $\zpsi{0}$, identifies the minimal causal mechanisms, which intuitively represent the parts of the causal factors that are affected by the available interventions.
\end{enumerate}

\begin{figure*}[t!]
    \centering
    \resizebox{0.6\textwidth}{!}{
        \input{figures/causal_graph}
    }
    \caption{An example temporal causal graph in TRIS, with observed variables shown in gray and latent variables in white. A latent causal factor $C^{t+1}_i$ has as parents a subset of the causal factors at the \textcolor{tempcolor}{previous time step} $C_1^{t}, \dots, C_{K}^{t}$, and its \textcolor{intvcolor}{intervention target} $I^{t+1}_i$. All causal variables $C^{t+1}$ and the noise $\noisevarcap^{t+1}$ cause the \textcolor{obscolor}{observation} $X^{t+1}$. $R^{t+1}$ is a \textcolor{confcolor}{latent confounder} between the intervention targets, but is not strictly necessary (\ie{} $R^{t+1}$ may also be constant). The goal is to identify the causal factors $C_1,...,C_K$.}
    \label{fig:appendix_general_causal_graph}
\end{figure*}
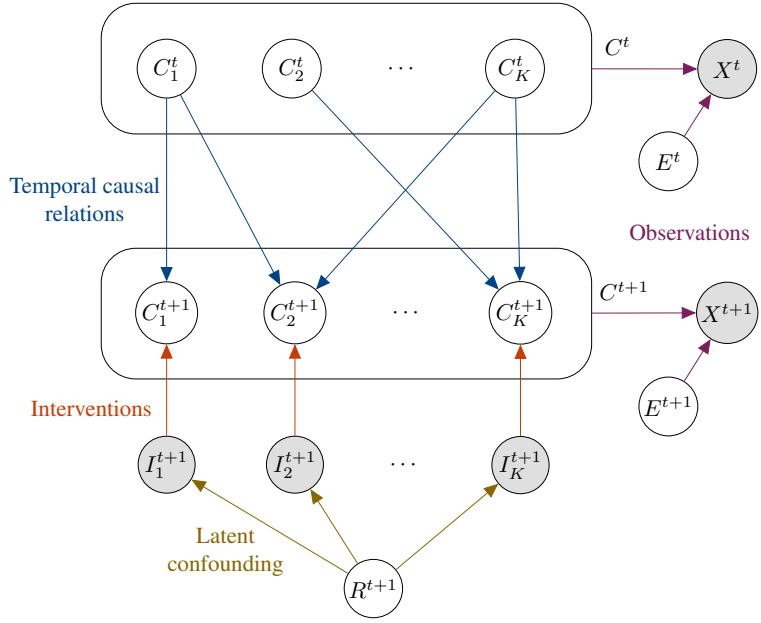

Along the way, we will make use of \cref{fig:appendix_general_causal_graph} summarizing the temporal causal graph.
For the remainder of the proof, we assume that the prior $p_{\phi}\left(z^{t+1}|z^{t}, I^{t+1}\right)$ and the invertible map $g_{\theta}$ are sufficiently complex to approximate any possible function and distribution one might encounter. 
To simplify the exposition, we also assume that the latent dimension size is unlimited, \ie{} $M=\infty$, so there are no limitations on how many latent variables $\zpsi{i}$ can be used to represent a causal factor $C_i$.
In practice, however, this is not a limiting factor as long as we can overestimate the dimensions of the causal factors and noise variables.

Throughout the proof, we will use $C^{t}$ to refer to the set of all causal factors at time step $t$, \ie{} $C^{t}=\{C_1^t,...,C_K^t\}$.
Similarly, we define $I^{t+1}=\{I^{t+1}_1,...,I^{t+1}_K\}$.

\subsection{Step 1: True Disentanglement $\disentanglement^*$ is one of the Global Maxima of the Conditional Likelihood $p(X^{t+1}|X^{t}, I^{t+1})$}
\label{sec:appendix_proofs_step_1}

We start by proving the following Lemma:
\begin{lemma}
The true disentanglement function $\disentanglement^*$ that correctly disentangles the true causal factors $C^{t+1}_1,...,C^{t+1}_K$ from observations $X^t, X^{t+1}$ using the true $\psi^*$ assignment function on the true latent variables $Z^{t+1}$ is one of the global maxima of the likelihood of $p(X^{t+1}|X^{t}, I^{t+1})$. 
\end{lemma}

We are interested in optimizing $p(X^{t+1}|X^{t}, I^{t+1})$. 
We can first try to simplify this equation with the knowledge of the causal graph in \cref{fig:appendix_general_causal_graph}, \ie{} using the true underlying generative model, since we aim to show that learning the causal factors and aligning them correspondingly in the prior of \cref{eq:method_prior_distribution} represents a global optimum of maximizing $p(X^{t+1}|X^{t}, I^{t+1})$. 
Using the conditional independence relations of the graph in \cref{fig:appendix_general_causal_graph}, we write the joint distribution of all the variables in the true generative model as:
\begin{equation}
    p(X^{t}, X^{t+1}, C^{t}, C^{t+1}, I^{t+1})=  p(X^{t+1}|C^{t+1}) \cdot  \left[\prod_{i=1}^{K} p(C_i^{t+1}|C^{t}, I_i^{t+1})\right] \cdot p(X^{t}|C^{t}) \cdot p(C^t) \cdot p(I^{t+1})
\end{equation}
We can now condition on $X^t$ and $I^{t+1}$, marginalize out $C^t$ and $C^{t+1}$ and write the conditional likelihood as:
\begin{equation}
    p(X^{t+1}|X^{t}, I^{t+1})=\int_{C^{t+1}}\int_{C^{t}} 
    p(X^{t+1}|C^{t+1})  \cdot \left[\prod_{i=1}^{K} p(C_i^{t+1}|C^{t}, I_i^{t+1})\right]  \cdot 
    p(C^{t}|X^t) dC^{t} dC^{t+1}
\end{equation}
In our assumptions of \cref{sec:identifiability_setup_assumptions}, we have defined the observation function $\obs$ to be bijective, meaning that there exists an inverse $f$ that can identify the causal factors $C^{t}$ and noise variable $\noisevarcap^t_o$ from $X^{t}$. 
Thus, we can write $p(C^{t}|X^{t}) = \delta_{f(X^{t})=C^{t}}$, where $\delta$ is a Dirac delta.
Since the noise on the observations, $\noisevarcap^t_o$, is said to be independent of $X^{t+1}$ and $C^{t+1}$, we can remove it from being in the conditioning set.
This leads us to:
\begin{equation}
    p(X^{t+1}|X^{t}, I^{t+1})=\int_{C^{t+1}} \left[\prod_{i=1}^{K} p(C_i^{t+1}|C^{t}, I_i^{t+1})\right] \cdot p(X^{t+1}|C^{t+1}) dC^{t+1}
\end{equation}
Since we have assumed $\obs$ to be bijective, we know that for each $X^{t+1}$, there exist only one combination of $C^{t+1}$ and $\noisevarcap^{t+1}_o$.
Thus, by using the change of variables formula, we can rewrite the equation above by:
\begin{equation} \label{eq:appendix_proof_step_1_maxlike_true_gen}
    p(X^{t+1}|X^{t}, I^{t+1})=\left|J_{\obs}\right|^{-1}\cdot\left[\prod_{i=1}^{K} p(C_i^{t+1}|C^{t}, I_i^{t+1})\right] \cdot p(\noisevarcap^{t+1}_o)
\end{equation}
where $J_{\obs}=\frac{\partial \obs(C^{t+1},\noisevarcap^{t+1}_o)}{\partial C^{t+1}\partial\noisevarcap^{t+1}_o}$ denotes the Jacobian of the bijective/invertible observation function $\obs$.
\cref{eq:appendix_proof_step_1_maxlike_true_gen} constitutes a global optimum of the maximum likelihood, since it represents the true underlying dynamics.

We relate this conditional likelihood to the prior setup of \OurApproach{}. We show that assigning $C^{t+1}_i$ to $\zpsi{i}^{t+1}$, \ie, learning the true assignment function $\psi^*$, provides us with the same maximum likelihood solution as in \cref{eq:appendix_proof_step_1_maxlike_true_gen}.
We have defined our objective in \cref{sec:identifiability} in \cref{eq:method_prior_distribution} as:
\begin{equation}
    p_{\phi}\left(z^{t+1}|z^{t}, I^{t+1}\right) = \prod_{i=0}^{K}p_{\phi}\left(\zpsi{i}^{t+1}|z^{t}, I_{i}^{t+1}\right) 
\end{equation}
Since we know that $g_{\theta}^*$ is an invertible function between $\mathcal{X}$ and $\mathcal{Z}$, we know that $z^{t}$ must include all information of $X^{t}$.
Thus, we can also replace it with $z^{t}=[C^{t},\noisevarcap_o^{t}]$, giving us:
\begin{equation}
    \label{eq:appendix_proof_step_1_prior_ct}
    p_{\phi}\left(z^{t+1}|C^{t}, \noisevarcap_o^{t}, I^{t+1}\right) = \prod_{i=0}^{K}p_{\phi}\left(\zpsi{i}^{t+1}|C^{t}, \noisevarcap_o^{t}, I_{i}^{t+1}\right)
\end{equation}
The optimal assignment function $\psi^*$ assigns sufficient dimensions to each causal factor $C_1,...,C_K$.
Since $\mathcal{Z}$ can have a larger space than $\mathcal{E}\times\mathcal{C}$, but $\mathcal{E}\times\mathcal{C}$ is sufficient to describe $\mathcal{X}$, we know that the remaining dimensions of $\mathcal{Z}$ do not contain any information.
Thus, the assignment function $\psi^*$ can map them to any causal factor without a change in distribution.
Using this assignment function, we now consider $\zpsistar{i}^{t+1}=C^{t+1}_i$ for $i=1,...,K$.
Then, \cref{eq:appendix_proof_step_1_prior_ct} becomes:
\begin{equation}
    \label{eq:appendix_proof_step_1_prior_ct_plugged_in_2} 
    p_{\phi}\left(z^{t+1}|C^{t}, \noisevarcap_o^{t}, I^{t+1}\right) = \left[\prod_{i=1}^{K}p_{\phi}\left(\zpsistar{i}^{t+1}=C^{t+1}_i|C^{t},  I_{i}^{t+1}\right)\right]\cdot p(\zpsistar{0}^{t+1}|C^{t}, \noisevarcap_o^{t})
\end{equation}
where we remove $\noisevarcap^t_o$ from the conditioning set for the causal factors, since know that $C^{t+1}$ and $\noisevarcap_o^{t+1}$ is independent of $\noisevarcap_o^t$. We further simplify by noting that $\zpsistar{0}^{t+1} = \noisevarcap_o^{t+1}$ is independent of any other factor.
\begin{equation}
    \label{eq:appendix_proof_step_1_prior_ct_plugged_in_3} 
    p_{\phi}\left(z^{t+1}|C^{t}, \noisevarcap_o^{t}, I^{t+1}\right) = \left[\prod_{i=1}^{K}p_{\phi}\left(\zpsistar{i}^{t+1}=C^{t+1}_i|C^{t},  I_{i}^{t+1}\right)\right]\cdot p(\zpsistar{0}^{t+1} = \noisevarcap_o^{t+1})
\end{equation}
Finally, by using $g_{\theta}^*$, we can replace the distribution on $z^{t+1}$ by a distribution on $X^{t+1}$ by the change of variables formula: 
\begin{equation}
    \label{eq:appendix_proof_step_1_prior_ct_plugged_in_4} 
    p_{\phi}\left(X^{t+1}|C^{t}, \noisevarcap_o^{t}, I^{t+1}\right) = \left|\frac{\partial g_{\theta}^*(z^{t+1})}{\partial z^{t+1}}\right|\cdot \left[\prod_{i=1}^{K}p_{\phi}\left(\zpsistar{i}^{t+1}=C^{t+1}_i|C^{t},  I_{i}^{t+1}\right)\right]\cdot p(\zpsistar{0}^{t+1} = \noisevarcap_o^{t+1})
\end{equation}

Thereby, it is apparent that $g_{\theta}^*$ is equal to $\obs^{-1}$, since both are identical invertible functions between the same spaces ($\mathcal{Z}$ becomes $\mathcal{E}\times\mathcal{C}$ here).
Hence, \cref{eq:appendix_proof_step_1_prior_ct_plugged_in_4} represents the exact same distribution as \cref{eq:appendix_proof_step_1_maxlike_true_gen}. 
Therefore, we have shown that the function $\disentanglement^{*}$ that disentangles the true latent variables $C_1,...,C_K$ and assigns them to the corresponding sets $\zpsi{1},...,\zpsi{K}$ constitutes a global, but not necessarily unique, optimum for maximizing the likelihood of \cref{eq:method_prior_distribution}.

Note that while assigning $C^{t+1}_i$ to $\zpsi{i}^{t+1}$ provides us with the same maximum likelihood solution as in \cref{eq:appendix_proof_step_1_maxlike_true_gen}, this is not the only possible representation.
Additional possible representation will be discussed in Step 2.

\subsection{Step 2: Characterizing the Disentanglement Class $\disentanglementClass$}
\label{sec:appendix_proofs_step_2}

Showing that the correct disentanglement constitutes a global optimum is not sufficient for showing that a model trained on solving the maximum likelihood solution converges to it, since there might potentially be multiple global optima.
Hence, this section discusses the class of causal representation functions $\disentanglement\in\disentanglementClass^{*}$ which can achieve the same maximum likelihood optimum as the true causal factor disentanglement discussed in \cref{sec:appendix_proofs_step_1}.
For this, we first need to distinguish between the \emph{variable} and \emph{invariable} information of a causal variable $C_i$, which is introduced in \cref{sec:appendix_proof_intervention_independence}.
Next, we will discuss the causal representation function class $\disentanglementClass$ for the setting where interventions are independent, \ie{} $I^{t+1}_i\independent I^{t+1}_j|C^{t}$ for any $i\neq j$, and finally extend it to confounded interventions.

\subsubsection{Intervention-Independent Variables}
\label{sec:appendix_proof_intervention_independence}

\begin{figure*}[t!]
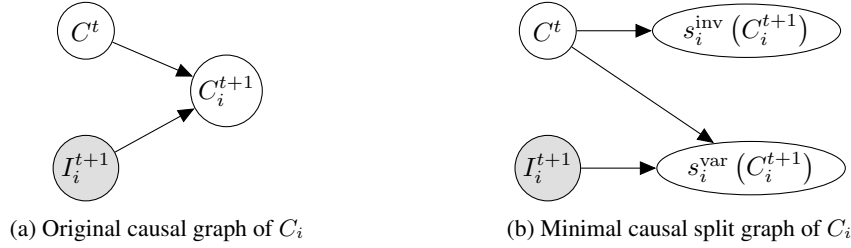

    \centering
    \begin{subfigure}[b]{0.4\textwidth}
        \centering
        \tikz{ %
    		\node[latent] (Ct) {\hspace{1mm}$C^{t}$\hspace{1mm}\phantom{}} ; %
    		\node[obs, below=of Ct] (Iit1) {$I^{t+1}_i$} ; %
    		\node[latent, right=of Ct, yshift=-.8cm] (Cit1) {$C^{t+1}_i$} ; %
    		
    		\edge{Ct}{Cit1} ;
    		\edge{Iit1}{Cit1} ;
    	}
    	\caption{Original causal graph of $C_i$}
    \end{subfigure}
    \begin{subfigure}[b]{0.4\textwidth}
        \centering
        \tikz{ %
    		\node[latent] (Ct) {\hspace{1mm}$C^{t}$\hspace{1mm}\phantom{}} ; %
    		\node[obs, below=of Ct] (Iit1) {$I^{t+1}_i$} ; %
    		\node[latent, right=of Iit1, ellipse] (Cit1sdep) {$\sdep_i\left(C^{t+1}_i\right)$} ; %
    		\node[latent, right=of Ct, ellipse] (Cit1sindep) {$\sindep_i\left(C^{t+1}_i\right)$} ; %
    		
    		\edge{Ct}{Cit1sdep} ;
    		\edge{Iit1}{Cit1sdep} ;
    		\edge{Ct}{Cit1sindep} ;
    	}
    	\caption{Minimal causal split graph of $C_i$}
    \end{subfigure}
    \caption{Splitting the causal variable $C_i$ in its minimal causal split. (a) In the original causal graph, $C_i^{t+1}$ has $C^t$ (or an eventual subset of it) and $I_i^{t+1}$ as its parents. (b) In the minimal causal split, only the variable part $\sdep_i(C_i^{t+1})$ depends on the intervention. The invariable part, $\sindep_i(C_i^{t+1})$, is independent of $I_i^{t+1}$, hence giving us an additional conditional independence. Note that $\sdep_i(C_i^{t+1})$ and $\sindep_i(C_i^{t+1})$ are conditionally independent.}
    \label{fig:appendix_minimal_mechanism_causal_graph}
\end{figure*}

Interventions allow us to identify a causal variable by seeing the caused change in its conditional distribution.
However, especially when talking about multidimensional causal variables, one might have interventions that only affect a subset of the actual causal variable dynamics, while the rest remains independent of the intervention.
As we will see later, this can have an influence on the identifiability result, making the found causal factors intervention-dependent.

We start by considering a single causal factor $C_i \in \mathcal{D}_i^{M_i}$ in the setup of \cref{fig:appendix_general_causal_graph} under our previously discussed assumptions.
Suppose for each causal factor $C_i \in \mathcal{D}^{M_i}$, there exists an invertible map $s_i : \mathcal{D}_i^{M_i} \rightarrow \mathcal{D}_i^{\textup{var}} \times \mathcal{D}_i^{\textup{inv}}$ that splits the domain $\mathcal{D}^{M_i}$ of $C_i$ into a part that is invariant and a part that is variant under intervention.
We denote the two parts of this map as 
\begin{equation}
    s_i(C^t_i) = (\sdep_i(C^t_i), \sindep_i(C^t_i)) 
\end{equation}
The split $s$ must be invertible, so that we can map back and forth between $\mathcal{D}_i^{M_i}$ and $\mathcal{D}_i^{\textup{var}} \times \mathcal{D}_i^{\textup{var}}$ without losing information.
Furthermore, to be called a split, $s$ must satisfy $\sindep_i(C_i^{t}) \independent I_i^{t} \mid \pa{C_i^{t}}$, \ie, $\sindep_i(C_i^{t})$ is independent of the intervention variable $I_i^{t}$ given the parents of $C_i^{t}$.
Further, both parts of the split must be conditionally independent, \ie{} 
$\sindep_i(C_i^{t}) \independent \sdep_i(C_i^{t}) \mid \pa{C_i^{t}}, I_i^{t}$.
Hence, we can write their distributions as:
\begin{equation}
    \label{eqn:appendix_intervention_indep_split_def}
    p\left(s_i(C^{t+1}_i)|C^{t},I^{t+1}_i\right) = p\left(\sdep_i(C^{t+1}_i)|C^{t},I^{t+1}_i\right) \cdot  p\left(\sindep_i(C^{t+1}_i)|C^{t}\right)
\end{equation}
This means that $\sdep_i(C_i^{t})$ will contain the manipulable, or \emph{variable}, part of $C^t_i$.
In contrast, $\sindep_i(C_i^{t})$ is the \emph{invariable} part of $C_i^t$ which is independent of the intervention.
This relation is visualized in \cref{fig:appendix_minimal_mechanism_causal_graph}.

For any causal variable, there may exist multiple possible splits, but there is always at least the trivial split where $\mathcal{D}_i^{\textup{var}} = \mathcal{D}_i^{M_i}$ is the original domain of $C_i$, and $\mathcal{D}_i^{\textup{inv}} = \{0\}$ is the one-element set (no invariant information). 
However, there might also exist splits in which $\sindep_i(C^{t+1}_i)\neq\emptyset$.
For instance, in a multidimensional causal variable $\hat{C}\in\mathbb{R}^{3}$, if an intervention only affects the first two dimensions while the last one remains unaffected, we obtain the split $\sdep([\hat{C}_1, \hat{C}_2]),\sindep(\hat{C}_3)$.
Nonetheless, this can even happen for scalar variables, since we do not constraint the possible distributions of $C_i$.
We give an example for such a case below.

\begin{figure}[t!]
    \centering
    \includegraphics[width=0.3\textwidth]{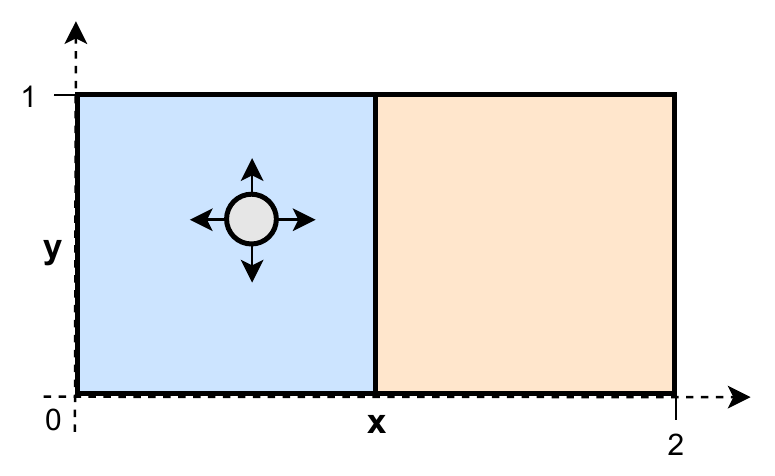}
    \caption{Example for splitting a causal variable into an intervention-dependent and -independent part. The two ground truth causal variables are the $x$ and $y$ positions of the ball. See \cref{sec:appendix_proof_intervention_independence} for details.}
    \label{fig:appendix_proof_intervention_independent_example}
\end{figure}

\paragraph{Example 1}
Consider the scenario in \cref{fig:appendix_proof_intervention_independent_example} where we have a ball with its two positional dimensions $x$ and $y$ as its causal factors.
For now, we only focus on its $x$ position (in the remainder of the section, $x^{t}$ refers to the position of the ball on the $x$-axis, not the full observation $X^{t}$ which we denote by a capital letter).
Over time, the ball moves within one of the two boxes, but cannot jump in between boxes.  
An example of such a conditional could be:
\begin{equation}
    p(x^{t+1}|x^{t},I_x^{t+1}=0) = \begin{cases}
        \min(\max(x^{t} + \noisevarlow, 0), 1) & \text{if }x^{t} < 1\\
        \min(\max(x^{t} + \noisevarlow, 1), 2) & \text{otherwise}\\
    \end{cases}
\end{equation}
with $\noisevarlow\sim\mathcal{N}(0, 0.1)$.
Intuitively, the ball therefore moves randomly around its previous position, while being bounded by the box it is in.
Due to its modular conditional distribution, we can rewrite the causal variable $x$ and its distribution in terms of two different variables: its position within its current box, $u\in [0,1]$, and a binary variable indicating in which box the ball is, $b\in{0,1}$ (left/\boxcolorone{} vs right/\boxcolortwo{} in \cref{fig:appendix_proof_intervention_independent_example}).
Then, its conditional distribution becomes:
\begin{equation}
    p(x^{t+1}|x^{t},I_x^{t}) = p\left(b^{t+1}|x^{t},I^{t+1}_x\right) \cdot p\left(u^{t+1}|x^{t},I^{t+1}_x\right)
\end{equation}
Now, suppose that an intervention $I^{t+1}_x$ changes the box the ball is in, while the relative position keeps evolving as it would under no intervention, \ie{} still depending on its parents.
Then, we can yet write its conditional distribution as:
\begin{equation}
    p(x^{t+1}|x^{t},I_x^{t}) = p\left(b^{t+1}|x^{t},I^{t+1}_x\right) \cdot p\left(u^{t+1}|x^{t}\right)
\end{equation}
Using the notation above, we therefore can define the split $\sdep(x)=b,\sindep(x)=u$, where $b$ depends on the intervention, while $u$ does not.
Note that $\sdep(x)=x,\sindep(x)=\emptyset$ is yet another valid split in this case.

\paragraph{Example 2}
Consider the same example as before, however, now with a different intervention setup.
Suppose that an intervention $I^{t+1}_x$ constitutes a perfect intervention on $x$, under which $x^{t+1}\independent x^{t}|I^{t+1}_x=1$.
Then, the previous split $\sdep(x)=b,\sindep(x)=u$ is not valid anymore, if the intervention target $I^{t+1}_x$ cannot be deterministically deduced from $x^{t}$, since an intervention changes the distribution of the relative position $u^{t+1}$.
Hence, the only valid split is $\sdep(x)=x,\sindep(x)=\emptyset$.
This shows that the possible space of such splits depends on the available interventions.

\paragraph{Minimal causal variables}
In the examples above, one can see that for certain situations, a causal variable can have multiple valid splits $\sdep_i(C_i),\sindep_i(C_i)$ since intervention-independent information can be modeled in either $\sdep_i(C_i)$ or $\sindep_i(C_i)$.
The split that will be the most relevant for the identifiability discussion here is the one that assigns only the intervention-dependent information to $\sdep_i(C_i)$, and the rest to $\sindep_i(C_i)$.
We define this as follows:
\begin{definition}
The \emph{minimal causal split} of a variable $C^t_i$ with respect to its intervention variable $I^t_i$ is the split $s_i$ which maximizes the entropy of $H(\sindep_i(C^{t}_i)|\pa{C^{t}_i})$. Under this split, $\sdep_i(C^{t}_i)$ is defined as the \emph{minimal causal variable} and denoted by $\sdepStar(C^{t}_i)$.
\end{definition}
Additionally, we also define:
\begin{definition}
The \underline{minimal causal mechanism} of a variable $C_i$ with respect to its intervention $I_i$ is defined as the conditional distribution $p\left(\sdepStar(C^{t}_i)|\pa{C^{t+1}_i},I^{t+1}_i\right)$.
\end{definition}
We refer to $p\left(\sdepStar(C^{t}_i)|C^{t},I^{t+1}_i\right)$ as \textit{minimal} causal mechanism, since it is the distribution for which as little as possible information depends on $I^{t+1}_i$.
Hence, the definition of this mechanism depends on the characteristics of the provided intervention.
As we will see later, while we cannot guarantee to find the full causal mechanism, we can yet identify the minimal causal mechanism.

The existence of a split where $\sindep(C_i)\neq\emptyset$ for any causal factor $C_i$ creates additional, possible solutions that obtain the same maximum likelihood as the true split.
This is because $\sindep(C_i)$ is independent of the intervention target $I_i$, allowing it to be modeled by any set $\zpsi{j}$ without losing information.
The following subsections further characterize the space of new solutions with the existence of such splits.

\subsubsection{Independent Interventions}
\label{sec:appendix_proofs_step_2_independent_interventions}

In this section, we show by which class of disentanglement functions $\disentanglementClass$ a maximum likelihood solution of the generative model can be found.

However, to simplify the first steps, we assume that all intervention targets are independent of each other given the causal factors of the previous time step, \ie{} $I^{t+1}_i\independent I^{t+1}_j|C^{t}$ for any $i\neq j \in 1, \dots, K$. By construction, we also assume $I^{t+1}_0 = 0$ for all $t$.
We will extend it afterwards in \cref{sec:appendix_proofs_step_2_confounded_interventions}.

\paragraph{Solutions for $\sindep_i(C_i)=\emptyset$ for all $i\in\range{1}{K}$} As a first step, we assume that for all causal factors $C_1,...,C_K$, there does not exist any minimal causal mechanism split besides $\sindep_i(C_i)$ being the empty set.
Therefore, all of $C_i$ is dependent on $I^{t+1}_i$.
For this case, consider an arbitrary partition of $C_i$, $s^{0}(C_i),s^{1}(C_i)$, with the same invertibility constraints as $\sdep,\sindep$ and conditionally independence between $s^{0}(C_i),s^{1}(C_i)$, but with a non-empty invariable part, \ie, $s^{1}(C_i)\neq\emptyset$.
For this partition, the conditional entropy of $s^{1}(C_i)$ given $C^t$ must be strictly lower when conditioning also on $I^{t+1}_i$:
\begin{equation}
    H\left(s^{1}(C^{t+1}_i)|C^{t},I^{t+1}_i\right) < H\left(s^{1}(C^{t+1}_i)|C^{t}\right)
\end{equation}
If the conditional entropy was equal, then $s^1(C^{t+1}_i)$ would be independent of $I^{t+1}_i$ given $C^{t}$, which is only true for $\sindep_i(C_i)$. Since we assume $\sindep_i(C_i)$ is empty, while $s^{1}(C^{t+1}_i)$ is not, this can never happen. 
Thus, to model a causal factor $C_i$ where $\sindep_i(C_i)=\emptyset$, a maximum likelihood solution can only be found if all of $\sdep_i(C_i)$ is conditioned on $I^{t+1}$.
Further, since in this setting $C_j^{t+1} \independent C_i^{t+1} | C^t,I^{t+1}$ for all $i,j \in \range{1}{K}, i \neq j$, there cannot exist any split across multiple causal factors that violate the entropy inequality above for $I^{t+1}_i$ and $I^{t+1}_j$ while still modeling the true conditional distributions.

Similarly, in this setting under the specified assumptions, the information of $I^{t+1}_i$ cannot be determined by any other target variable $I^{t+1}_j,i\neq j$, since otherwise, the targets would not be independent.
Hence, we can write the following entropy inequality for any $i\neq j$:
\begin{equation}
     H\left(C^{t+1}_i|C^{t},I^{t+1}_i\right) < H\left(C^{t+1}_i|C^{t}\right) = H\left(C^{t+1}_i|C^{t},I^{t+1}_j\right) 
\end{equation}
Therefore, one can only achieve the maximum likelihood solution (\ie, the minimum entropy solution) if all information of $C^{t+1}_i$ is conditioned on $I^{t+1}_i$ for $i=1,...,K$ in addition to $C^t$.
This implies that each factor $p_\phi(\zpsi{i}^{t+1} | z^t, I^{t+1}_i)$ described in \cref{eq:method_prior_distribution} will have $p(C^{t+1}_i | z^t, I^{t+1}_i)$, and therefore $\zpsi{i}^{t+1} = C^{t+1}_i$ as its maximum likelihood solution.

Nonetheless, this excludes $\zpsi{0}$, \ie{} the factors independent of any intervention, in this case being the noise $\noisevarcap_o^{t+1}$.
Since it is independent of any interventions, any distribution of $\noisevarcap_o^{t+1}$ across the different causal factor sets $\zpsi{0},...,\zpsi{K}$ will achieve the same likelihood score, as long as the part of $\noisevarcap_o^{t+1}$ across factors is independent.
Hence, in conclusion for this scenario, we can guarantee that $\zpsi{i}$ will model all information of $C_i$ and no other causal factor $C_j, i\neq j$, but can contain additional information from $\noisevarcap^{t+1}$.

\paragraph{Solutions with invariable parts} Next, we consider the scenario where there exists a split with $\sindep_i(C_i)\neq\emptyset$ for some causal variables in $i\in\range{1}{K}$.
For this case, we can write the maximum likelihood solution of \cref{eq:appendix_proof_step_1_maxlike_true_gen} as:
\begin{align}
    p(X^{t+1}|X^{t}, I^{t+1}) & = \left|\frac{\partial g_{\theta}^*(z^{t+1})}{\partial z^{t+1}}\right|\cdot\left[\prod_{i=1}^{K} p(C_i^{t+1}|I_i^{t+1}, C^{t})\right] \cdot p(\noisevarcap^{t+1})\\
    & = \left|\frac{\partial g_{\theta}^*(z^{t+1})}{\partial z^{t+1}}\right|\cdot\left[\prod_{i=1}^{K} p(\sdep_i(C_i^{t+1})|I_i^{t+1}, C^{t})\right] \cdot \left[\prod_{i=1}^{K} p(\sindep_i(C_i^{t+1})|C^{t})\right] \cdot p(\noisevarcap^{t+1})
\end{align}
This equation shows that one can assign $\sindep_1(C_1),...,\sindep_K(C_K)$ to any latent variable set $\zpsi{0},...,\zpsi{K}$ or split it across, while achieving the same optimal likelihood, since they are independent of any intervention target.
The remaining information in $\sindep_1(C_1),...,\sindep_K(C_K)$ thereby acts the same way as the noise variable $\noisevarcap^{t+1}$.
Thus, there exist multiple maximum likelihood solutions with different splits of information to causal factors.

However, on the other hand, the solution space is yet restricted by the assignment of $\sdep_i(C_i)$.
In particular, if $\sdep_i(C_i)$ cannot be split further into an invariable/intervention-independent part, we can rely on the same results from the previous setting, when considering $\sdep_i(C_i)$ as new causal variables.
In case there exist another split of $C_i$ which would add more information to $\sindep_i(C_i)$, this part could not be guaranteed to be matched to the causal factor $C_i$ due to its independence.
Hence, in conclusion here, we can guarantee that $\zpsi{i}$ will model all information of $\sdep(C_i)$ and no other causal factor $\sdep(C_j), i\neq j$, if there does not exist another split of $\sdep(C_i)$.
The additional information of $\noisevarcap^{t+1}$ as well as $\sindep(C_j)$ can be assigned to any causal variable.
In the third step of the proof (\cref{sec:appendix_proofs_step_3}), we discuss how one can yet obtain a unique solution.

\subsubsection{Confounded Interventions}
\label{sec:appendix_proofs_step_2_confounded_interventions}

In the previous discussion, we have used the assumption that interventions are independent of each other: $I^{t+1}_i\independent I^{t+1}_j|C^{t}$.
This assumption was required for showing that conditioning information of $C^{t+1}$ on any other target will lead to the same entropy as having it without a target, \ie{} $H\left(C^{t+1}_i|C^{t},I^{t+1}_j\right) = H\left(C^{t+1}_i|C^{t}\right)$.
In this section, however, we consider a wider range of interventions.
Specifically, we assume that the intervention targets $I^{t+1}_1,...,I^{t+1}_K$ are confounded by some unobserved variable $R^{t+1}$ besides $C^{t}$.
This allows the modeling of, for example, single-target interventions or groups of interventions, \eg{} $I^{t+1}\in\{[0,0,0], [1,1,0], [0,1,1]\}$ for a three-variable case.
Under such a setup, the entropy equation from before, \ie{} $H\left(C^{t+1}_i|C^{t},I^{t+1}_j\right) \neq H\left(C^{t+1}_i|C^{t}\right)$, is not valid anymore since $I^{t+1}_j$ and $I^{t+1}_i$ are not necessarily independent anymore and hence $C^{t+1}_i\not\independent I^{t+1}_j$ can occur for some $i,j\in\range{1}{K}, i\neq j$.

Despite that, a causal factor $C^{t+1}_i$ is still independent of any other target $I^{t+1}_j,i\neq j$, as long as it is conditioned on its true target and previous time step: $C^{t+1}_i\independent I^{t+1}_j|C^{t},I^{t+1}_i$.
This is because $C^{t}$ and $I^{t+1}_i$ are all the parents of $C^{t+1}_i$, as shown in the causal graph of \cref{fig:appendix_general_causal_graph}. 
Further, suppose that there exist information of $C^{t+1}_i$ which is statistically independent of the intervention $I^{t+1}_i$, \ie $\sindep_i(C_i)\neq\emptyset$.
Then, this will also be independent of any other intervention target $I^{t+1}_j$, since $\sindep_i(C_i)\independent I^{t+1}_i|C^{t}$, and all paths from $C_i$ to $I^{t+1}_j$ include $I^{t+1}_i$.
Hence, our discussion of the intervention-independent parts follow the same logic as in \cref{sec:appendix_proofs_step_2_independent_interventions}, and we are left with showing that $\sdep_i(C_i)$ is modeled by $\zpsi{i}$ in any maximum likelihood solution.

For this, we consider a pair of variables $C_i,C_j$, for which $I^{t+1}_i\not\independent I^{t+1}_j|C^{t}$, and show under which circumstances we can guarantee that no information of $\sdep_i(C_i)$ will be modeled in $\zpsi{j}$.
It is sufficient to limit the discussion to pairs of variables, since one latent variable can be only assigned to a single causal variable, hence to the actual one it belongs to, $C_i$, or any other variable $C_j$ here.
A crucial insight to the discussion will be that the influence of $I^{t+1}_j$ to $C^{t+1}_i$ solely relies on $I^{t+1}_i$ being correlated to both variables.
Further, one requirement is that no additional conditional independence relations exist, such as $\sdep_i(C_i)\independent I^{t+1}_i|I^{t+1}_j,C^{t}$, which is covered by our faithfulness assumption. 
Now, under this setup, we consider three cases: 
\begin{enumerate}
    \item for every time step $t$, the two variables $C_i,C_j$ have always been intervened on together, \ie{} $I^{t+1}_i=I^{t+1}_j$ for any $t$;
    \item there exist a time step $t$ at which $I^{t+1}_i=0,I^{t+1}_j=1$;
    \item there exist a time step $t$ at which $I^{t+1}_i=1,I^{t+1}_j=0$.
\end{enumerate}
Note that the only excluded case is when for every time step $t$, $I^{t+1}_i=0,I^{t+1}_j=0$.
This case refers to not having observed interventions for any of the two variables, and goes back to \cref{sec:appendix_proofs_step_2_independent_interventions}, where the variable part of $C_i$ is empty, \ie{} $\sdep_i(C_i)=\emptyset$.
Hence, in that case, $\sdep_i(C_i)$ would have no influence on the modeled solution.

In the first case, since the two factors have always been intervened on together, we know that $I^{t+1}_{i}=I^{t+1}_{j}$ for any time step $t$.
Hence, one can assign the information of $\sdep_i(C^{t+1}_i)$, $\sdep_j(C^{t+1}_j)$, or the union of both to either intervention target $I^{t+1}_{i}$ or $I^{t+1}_{j}$, without losing any information.
Moreover, if $\sdep_i(C^{t+1}_i)$ has multiple independent dimensions, \ie{} can be written as a product of multiple, conditionally independent variables, one can even split information of $C_i$ over the two targets.
This shows that in the general case, we cannot disentangle between two variables which have always been intervened on together. 
Similarly, if more than 2 variables have always been intervened on together, we cannot disentangle among all those variables.

For the second case, we can deduce that there must be interventions provided for at least the observational case, \ie{} $I^{t+1}_i=0,I^{t+1}_j=0$, the case where $C_j$ is intervened on but not $C_i$, \ie{} $I^{t+1}_i=0,I^{t+1}_j=1$, and either the joint intervention on both $C_i,C_j$ or only interventions on $C_i$, not $C_j$.
The reason why one of the two latter cases needs to exist is that if it would not be the case, $I^{t+1}_i=0$ would be zero for any $t$. 
In that case, the minimal causal mechanism of $C_i$ uses $\sdep_i(C_i)=\emptyset$, hence making the modeling of $\sdep_i(C_i)$ irrelevant for the maximum likelihood solution.

Thus, from these different intervention settings, it is apparent that there cannot exist a deterministic function $f$ with which we can determine $I^{t+1}_i$ from seeing $I^{t+1}_j$.
If we observe joint interventions on both variables, then for $I^{t+1}_j=1$, both $I^{t+1}_i=0$ and $I^{t+1}_i=1$ can occur.
Similarly, if we observe interventions on $C_i$ when $C_j$ is not intervened on, then both $I^{t+1}_i=0$ and $I^{t+1}_i=1$ can occur for $I^{t+1}_j=0$.
If both joint interventions and single interventions on $C_i$ have been observed, we cannot determine $I^{t+1}_i$ from $I^{t+1}_j$ at either $I^{t+1}_j=0$ or $I^{t+1}_j=1$.
Since $\sdep_i(C^{t+1}_i) \independent I_j^{t+1} | C^t, I^{t+1}_i$ by definition and $\sdep_i(C^{t+1}_i) \not \independent I^{t+1}_i  | C^t, I_j^{t+1}$ (the latter because $I_i^{t+1}$ is not a deterministic function of $I_j^{t+1}$, and therefore the dependence holds), we can write:
\begin{equation}
    \label{eq:appendix_step_2_3_entropy_inequality}
    H\left(\sdep_i(C^{t+1}_i)|C^{t},I^{t+1}_i\right) = H\left(\sdep_i(C^{t+1}_i)|C^{t},I^{t+1}_i, I^{t+1}_j\right) < H\left(\sdep_i(C^{t+1}_i)|C^{t},I^{t+1}_j\right)
\end{equation}
In conclusion, we cannot find the maximum likelihood solution if any information of $C^{t+1}_i$, which depends on $I^{t+1}_i$, is assigned to latent variables $\zpsi{j}$.
Hence, the maximum likelihood solution will strictly model $\sdep_i(C_i)$ in $\zpsi{i}$.

Finally, in the third case, we can take a similar argument as for the second case.
The only difference is that any of the additional intervention cases (joint or single on $C_j$), we have that from $I^{t+1}_j=0$, both $I^{t+1}_i=0$ and $I^{t+1}_i=1$ can occur.
Hence, the inequality in \cref{eq:appendix_step_2_3_entropy_inequality} is still valid, and we cannot replace $I^{t+1}_i$ by $I^{t+1}_j$ for any subset of information of $\sdep_i(C_i)$.
In summary, the maximum likelihood solution will strictly model $\sdep_i(C_i)$ in $\zpsi{i}$ also for this case.

Therefore, we can summarize the results in the following statement.
We can disentangle the intervention-dependent part of any two variables $C_i,C_j$, if there does not exist a deterministic function $f$ for which $I_i^{t}=f(I_j^{t})$ holds for every time step $t$.

\subsection{Step 3: Deriving the Final Theorem}
\label{sec:appendix_proofs_step_3}

Now that we have discussed the class of disentanglement functions $\disentanglementClass^{*}$ with their corresponding solutions, we can take the final step by adding constraints that ensure a unique solution.
In all the settings discussed in \cref{sec:appendix_proofs_step_2_independent_interventions} and \cref{sec:appendix_proofs_step_2_confounded_interventions}, the problem is that intervention-independent information can be represented in any of the latent sets $\zpsi{0},...,\zpsi{K}$ without affecting the optimal likelihood.
However, our main goal in getting a causal representation is that we disentangle information from different causal factors, meaning that we want to guarantee that the latents of $\zpsi{i}$ will only model information of the causal factor $C_i$, and no other causal factor $C_j,i\neq j$. 
Thus, we can do this by collecting all intervention-independent information in $\zpsi{0}$.
In other words, our ideal solution would be to have the latents of $\zpsi{i}$ model $\sdep_i(C_i)$ ($i=1,...,K$), and $\zpsi{0}$ to model $\{\sindep_1(C_1),...,\sindep_K(C_K)\}$, where the split $\sdep_i,\sindep_i$ was chosen to maximize the entropy of $p(\sindep_i(C_i^{t+1})|C^t)$.
To find both the right splits and collecting all intervention-independent information in $\zpsi{0}$, we thus want to find the representation function $\hat{\disentanglement}\in\disentanglementClass^{*}$ which maximizes the entropy of $\zpsi{0}$ while maintaining the optimal likelihood.
If any intervention-independent information would not be modeled in $\zpsi{0}$, it implies that there must exist another solution with greater entropy in $\zpsi{0}$, since all $\sindep_1(C_1),...,\sindep_K(C_K)$ as well as $\sdep_1(C_1),...,\sdep_K(C_K)$ are conditionally independent of each other (\ie{} adding parts to $\zpsi{0}$ cannot reduce the entropy).
Further, since we try to maximize the entropy of $\zpsi{0}$, we find the information splits $\sdep_i,\sindep_i$ that maximize the entropy of its intervention-independent part.
This is the same split as we had defined as minimal causal mechanisms in \cref{sec:appendix_proof_intervention_independence}.
Thus, we can summarize this result as follows:
\begin{theorem}
    \label{theo:appendix_method_intv_over_time}
    Suppose that $\phi^{*}$, $\theta^{*}$ and $\psi^{*}$ are the parameters that, under the constraint of maximizing the likelihood $p_{\phi}(g_{\theta}(x^{t+1})|g_{\theta}(x^{t}), I^{t+1})$, maximize the entropy of $p_{\phi}(\zpsi{0}^{t+1}|z^{t}, I^{t+1})$. 
    Then, with sufficient latent dimensions, the model $\phi^{*},\theta^{*},\psi^{*}$ learns a latent structure where $\zpsi{i}^{t+1}$ models the minimal causal variable of $C_i$ if $C^{t+1}_i\not\independent I^{t+1}_i | C^{t},I^{t+1}_j$ for any $i\neq j$. All remaining information independent of any interventions is modeled in $\zpsi{0}$.
\end{theorem}
The conditional independence $C^{t+1}_i\not\independent I^{t+1}_i | C^{t},I^{t+1}_j$  ensures that there exists no deterministic function $f$ for which $I^{t+1}_i=f(I^{t+1}_j)$.
This also includes when $I^{t+1}_i$ is constant, \ie{}, when $C_i$ is intervened all the time or not at all, since then, $C_i$ becomes independent of $I^{t+1}_i$.

\subsection{Non-Identifiability without Interventions}
\label{sec:appendix_proofs_proposition}

This section discusses Proposition~\ref{prop:method_tris_nonidentifiability} and its accompanying proof.
Proposition~\ref{prop:method_tris_nonidentifiability} states that in TRIS, if two causal factors $C_i$ and $C_j$ have only been jointly intervened on or not at all, then there exists a causal graph in which $C_i$ and $C_j$ cannot be uniquely identified from observations $X$ and intervention targets $I$.
In other words, without interventions, it cannot be guaranteed that for all possible settings, the causal variables can be uniquely identified.
A similar situation occurs in (non-linear) independent component analysis \cite{hyvaerinen2001independent, hyvaerinen2019nonlinear}, where the goal is to disentangle and identify $N$ independent sources/signals from observations in which the signals are entangled.
In \citet[Section 7.5]{hyvaerinen2001independent}, it is shown that if we have two, independent variables $s_1,s_2$ that follow a standard Gaussian distribution $\mathcal{N}(0,1)$, then any mixing of the two variables with an orthogonal matrix $\bm{A}$ results in two variables, $x_1,x_2$ with $\bm{x}=\bm{A}\bm{s}$, which yet again follow a standard Gaussian distribution.
Thus, from the perspective of the probability density function, the two representations cannot be distinguished, and hence no unique solution can be found.
This also implies that simply from samples of this distribution, one cannot uniquely identify the true sources.

In TRIS, we do not have two independent variables, but instead variables that have dependencies over time.
This corresponds to the setting described in \citet[Section 18]{hyvaerinen2001independent}, \citet{belouchrani1997blind}, or \citet{hyvaerinen2017nonlinear}.
Suppose that we have two variables $s^t_1,s^t_2$ at a time step $t$, which follow a Gaussian distribution with constant variance over time: 
\begin{equation}
    s^t_1\sim\mathcal{N}(\mu_{1}(s^{t-1}_1, s^{t-1}_2), \sigma^2), x^t_2\sim\mathcal{N}(\mu_{2}(s^{t-1}_1, s^{t-1}_2), \sigma^2)
\end{equation}
where $\mu_1,\mu_2$ are arbitrary functions on the variables of the previous time step, $s^{t-1}_1, s^{t-1}_2$.
The probability density function of it is:
\begin{equation}
    p(s^t_1, s^t_2|\bm{s}^{t-1})=\frac{1}{\sqrt{\det(2\pi\bm{\Sigma})}}\exp\left(-\frac{1}{2}(\bm{s}^t-\bm{\mu}^t)^T\bm{\Sigma}^{-1}(\bm{s}^t-\bm{\mu}^t)\right)
\end{equation}
where $\bm{\mu}^t$ represents the mean based on $\bm{s}^{t-1}$, and $\bm{\Sigma}=\begin{bmatrix}\sigma^2 & 0\\0 & \sigma^2\end{bmatrix}$.
Now, consider a mixing of the sources by an orthogonal matrix $\bm{A}$, i.e. $\bm{x}^t=\bm{A}\bm{s}^t$.
Then, the probability density function of those are:
\begin{align}
    p(x^t_1, x^t_2|\bm{s}^{t-1}) & =\frac{1}{\sqrt{\det(2\pi\bm{A}^T\bm{\Sigma}\bm{A})}}\exp\left(-\frac{1}{2}(\bm{A}\bm{s}^t-\bm{A}\bm{\mu}^t)^T(\bm{A}^T\bm{\Sigma}\bm{A})^{-1}(\bm{A}\bm{s}^t-\bm{A}\bm{\mu}^t)\right) \\
    & =\frac{1}{\sqrt{2\pi\det(\bm{A}^T)\det(\bm{\Sigma})\det(\bm{A})}}\exp\left(-\frac{1}{2}(\bm{s}^t-\bm{\mu}^t)^T\bm{A}^T\bm{A}\bm{\Sigma}^{-1}\bm{A}^T\bm{A}(\bm{s}^t-\bm{\mu}^t)\right) \\
    & =\frac{1}{\sqrt{2\pi\det(\bm{\Sigma})}}\exp\left(-\frac{1}{2}(\bm{s}^t-\bm{\mu}^t)^T\bm{\Sigma}^{-1}(\bm{s}^t-\bm{\mu}^t)\right) \\
\end{align}
where $\bm{A}^T\bm{A}=\bm{I}$ since $\bm{A}^T=\bm{A}^{-1}$ for rotation matrices, and $\det(\bm{A})=1$.
This shows that the density functions for the rotated and original signal are identical.
Thus, similar to the time-independent case, we cannot uniquely identify the original sources.

Coming back to the setting of TRIS, the two sources $s^{t-1}_1, s^{t-1}_2$ can represent two causal variables, $C_1$ and $C_2$, and the orthogonal matrix $\bm{A}$ represents a valid observation function $h$.
Hence, the previous conclusion also holds here, meaning that there exist a setting in which $C_1$ and $C_2$ cannot be uniquely identified from observational data only.
This concludes the proof for Proposition~\ref{prop:method_tris_nonidentifiability}.

%% file: sections/appendix_sections/2_experimental_details.tex
\section{Experimental Details}
\label{sec:appendix_experimental_details}

In this section, we describe the datasets and hyperparameter used in the experiments of \cref{sec:experiments} in detail.
First, we discuss the two datasets, the Temporal Causal3DIdent dataset (\cref{sec:appendix_experimental_details_causal3d_dataset}) and the Interventional Pong dataset (\cref{sec:appendix_experimental_details_interventional_pong}), including giving more examples and insights in the underlying dynamics.
The datasets will be publicly released upon publication.
In the second part, we discuss the experimental design (\cref{sec:appendix_experimental_details_experimental_design}), which includes a closer discussion of the correlation metrics as well as implementation details of \OurApproach{}.
Finally, we give an overview of the used hyperparameters in \cref{sec:appendix_experimental_design_hyperparameters}.

\subsection{Temporal Causal3DIdent Dataset}
\label{sec:appendix_experimental_details_causal3d_dataset}

\begin{figure}[t!]
    \centering
    \footnotesize
    \begin{tabular}{ccccccc}
        \ifarxiv
            \includegraphics[width=0.1\textwidth]{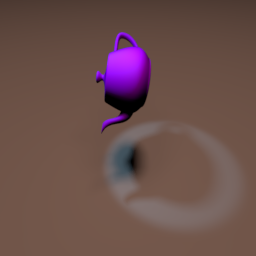} & 
            \includegraphics[width=0.1\textwidth]{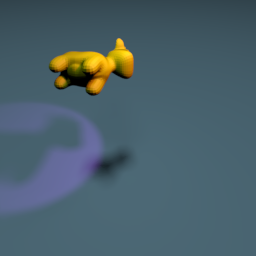} & 
            \includegraphics[width=0.1\textwidth]{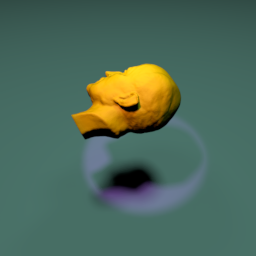} & 
            \includegraphics[width=0.1\textwidth]{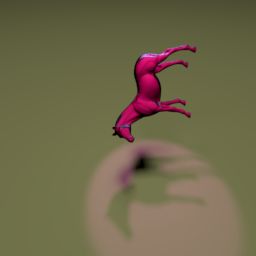} & 
            \includegraphics[width=0.1\textwidth]{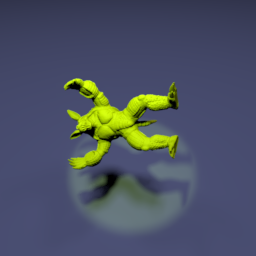} & 
            \includegraphics[width=0.1\textwidth]{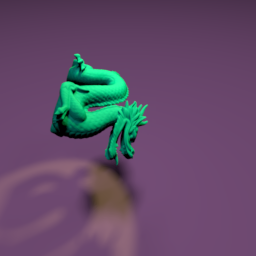} & 
            \includegraphics[width=0.1\textwidth]{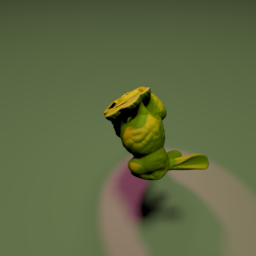} \\
        \else
            \includegraphics[width=0.12\textwidth]{figures/causal3d_dataset/teapot_high_res_01.png} & 
            \includegraphics[width=0.12\textwidth]{figures/causal3d_dataset/cow_high_res_01.png} & 
            \includegraphics[width=0.12\textwidth]{figures/causal3d_dataset/head_high_res_01.png} & 
            \includegraphics[width=0.12\textwidth]{figures/causal3d_dataset/horse_high_res_01.png} & 
            \includegraphics[width=0.12\textwidth]{figures/causal3d_dataset/armadillo_high_res_01.png} & 
            \includegraphics[width=0.12\textwidth]{figures/causal3d_dataset/dragon_high_res_01.png} & 
            \includegraphics[width=0.12\textwidth]{figures/causal3d_dataset/hare_high_res_01.png}
            \\
        \fi
        (a) Teapot & (b) Cow & (c) Head & (d) Horse & (e) Armadillo & (f) Dragon & (g) Hare\\
    \end{tabular}
    \caption{An example image for each object shape in the Temporal-Causal3DIdent dataset.}
    \label{fig:appendix_causal3d_shapes_example}
\end{figure}

The creation of the Temporal Causal3DIdent dataset closely followed the setup of \citet{vonkuegelgen2021self}.
We used the code provided by \citet{zimmermann2021contrastive}\footnote{\url{https://github.com/brendel-group/cl-ica}} to render the images via Blender \cite{blender2021blender}.
We will publish the adapted code for this dataset generation as well as the full datasets used here upon publication with an accompanying license.

\subsubsection{Causal Factor Description}

\begin{figure}
    \centering
    \setlength{\tabcolsep}{2pt}
    \begin{tabular}{cccccccccc}
        \textbf{pos\_x} & \textbf{pos\_y} & \textbf{pos\_z} & \textbf{rot\_$\alpha$} & \textbf{rot\_$\beta$} & \textbf{rot\_s} & \textbf{hue\_s} & \textbf{hue\_b} & \textbf{hue\_o} & \textbf{obj\_s} \\
        \includegraphics[width=0.09\textwidth]{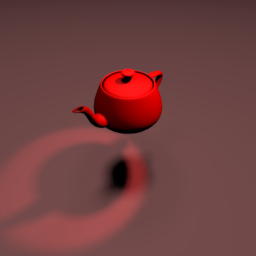} & 
        \includegraphics[width=0.09\textwidth]{figures/causal3d_dataset/example_figs/0000000.png} & 
        \includegraphics[width=0.09\textwidth]{figures/causal3d_dataset/example_figs/0000000.png} & 
        \includegraphics[width=0.09\textwidth]{figures/causal3d_dataset/example_figs/0000000.png} & 
        \includegraphics[width=0.09\textwidth]{figures/causal3d_dataset/example_figs/0000000.png} & 
        \includegraphics[width=0.09\textwidth]{figures/causal3d_dataset/example_figs/0000000.png} & 
        \includegraphics[width=0.09\textwidth]{figures/causal3d_dataset/example_figs/0000000.png} & 
        \includegraphics[width=0.09\textwidth]{figures/causal3d_dataset/example_figs/0000000.png} & 
        \includegraphics[width=0.09\textwidth]{figures/causal3d_dataset/example_figs/0000000.png} & 
        \includegraphics[width=0.09\textwidth]{figures/causal3d_dataset/example_figs/0000000.png} \\
        \includegraphics[width=0.09\textwidth]{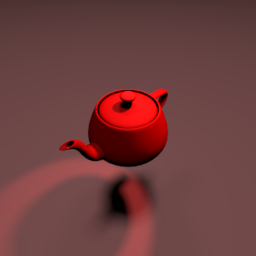} & 
        \includegraphics[width=0.09\textwidth]{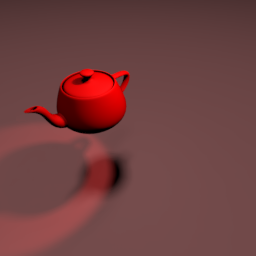} & 
        \includegraphics[width=0.09\textwidth]{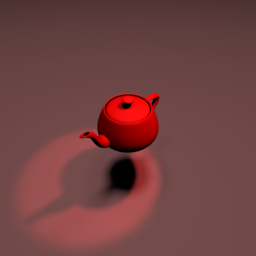} & 
        \includegraphics[width=0.09\textwidth]{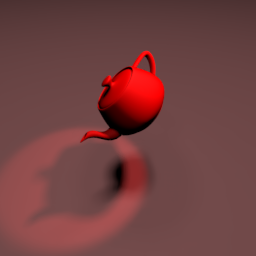} & 
        \includegraphics[width=0.09\textwidth]{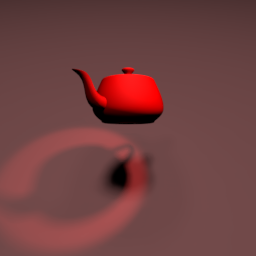} & 
        \includegraphics[width=0.09\textwidth]{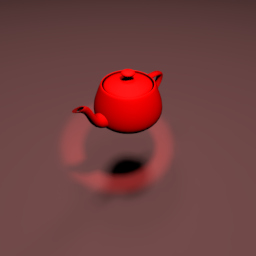} & 
        \includegraphics[width=0.09\textwidth]{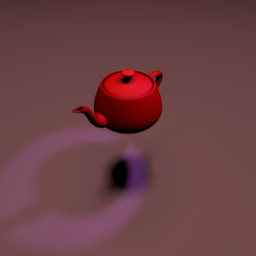} & 
        \includegraphics[width=0.09\textwidth]{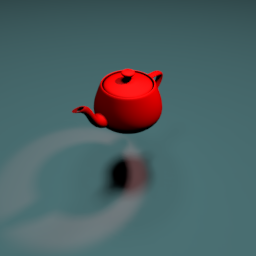} & 
        \includegraphics[width=0.09\textwidth]{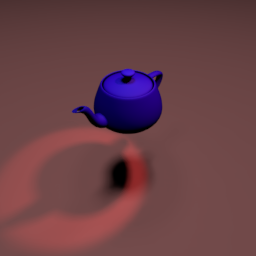} & 
        \includegraphics[width=0.09\textwidth]{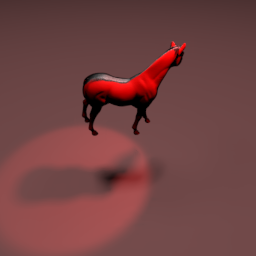} \\
        \includegraphics[width=0.09\textwidth]{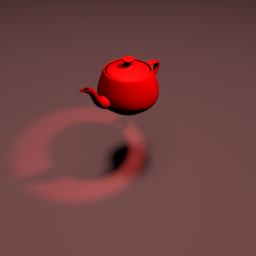} & 
        \includegraphics[width=0.09\textwidth]{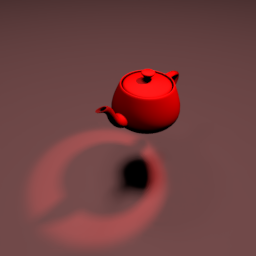} & 
        \includegraphics[width=0.09\textwidth]{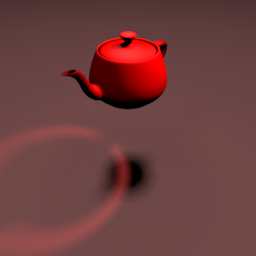} & 
        \includegraphics[width=0.09\textwidth]{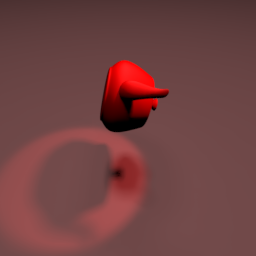} & 
        \includegraphics[width=0.09\textwidth]{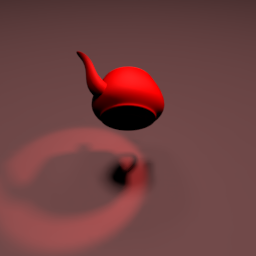} & 
        \includegraphics[width=0.09\textwidth]{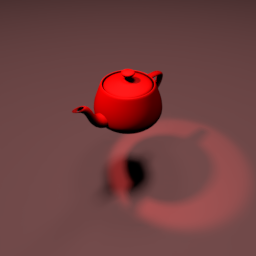} & 
        \includegraphics[width=0.09\textwidth]{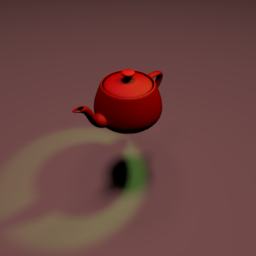} & 
        \includegraphics[width=0.09\textwidth]{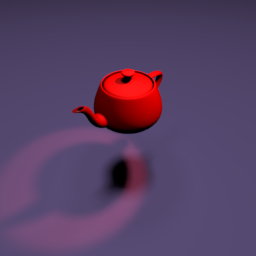} & 
        \includegraphics[width=0.09\textwidth]{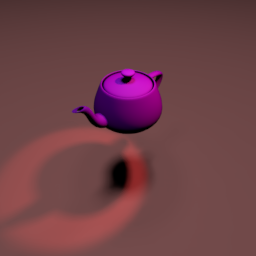} & 
        \includegraphics[width=0.09\textwidth]{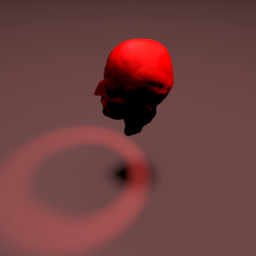} \\
        \includegraphics[width=0.09\textwidth]{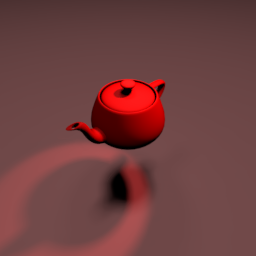} & 
        \includegraphics[width=0.09\textwidth]{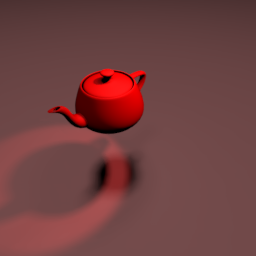} & 
        \includegraphics[width=0.09\textwidth]{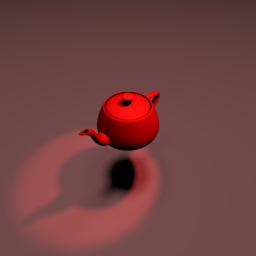} & 
        \includegraphics[width=0.09\textwidth]{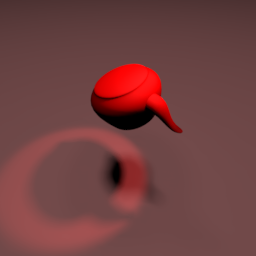} & 
        \includegraphics[width=0.09\textwidth]{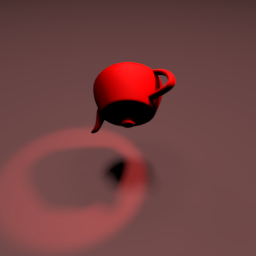} & 
        \includegraphics[width=0.09\textwidth]{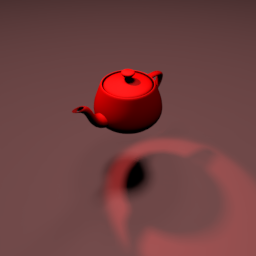} & 
        \includegraphics[width=0.09\textwidth]{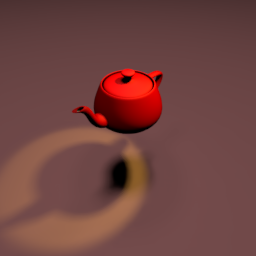} & 
        \includegraphics[width=0.09\textwidth]{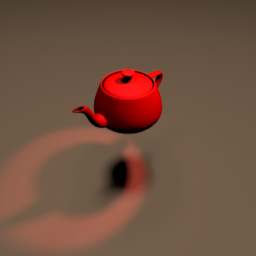} & 
        \includegraphics[width=0.09\textwidth]{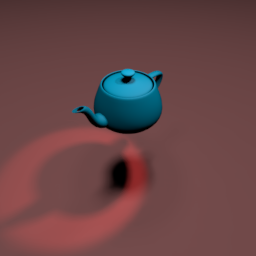} & 
        \includegraphics[width=0.09\textwidth]{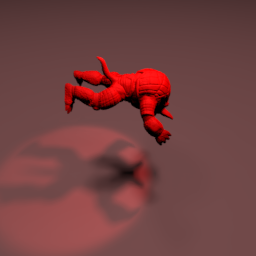} \\
    \end{tabular}
    \caption{Visualizing the different factors of variation in the Temporal Causal3DIdent dataset. Each column represents one dimension of a causal factor, and the different rows show the original image (first row) with only the corresponding causal factor being changed.}
    \label{fig:appendix_causal3d_factors_of_variation}
\end{figure}

To begin with, we give a more detailed description of the 7 causal factors here, and provide examples of varying individual factors in \cref{fig:appendix_causal3d_factors_of_variation}:
\begin{itemize}
    \item The \textbf{object position} (pos\_o) is modeled in 3 dimensions ($x$ - depth dimension, $y$ - horizontal, $z$ - vertical). All values are scaled between -2 as minimum and a maximum of 2, following \citet{zimmermann2021contrastive}. For the $y$ and $z$ dimension, this ensures that the object stays within the frame. For the $x$ dimensions, it ensures that the object does not cover the whole camera image, but also does not become too small in resolution for recognizing rotations and shapes.
    \item The \textbf{object rotation} (rot\_o) is modeled in 2 dimensions ($\alpha$ - roll angle, $\beta$ - pitch angle). All dimensions use circular values of $[0,2\pi)$, meaning that in distributions, we consider values close to $0$ and $2\pi$ as close as well. The rotation is restricted to two dimensions to guarantee that every object rotation has a unique value assignment of the angles. This can be violated when modeling three angles with a value range of $[0,2\pi)$.
    \item The \textbf{spotlight rotation} (rot\_s) is the positioning of the spotlight as an angle. The value range is $[0,2\pi)$, where, similarly as the object rotation, we consider it to be circular. 
    \item The \textbf{spotlight hue} is the color of the spotlight. The value range is, again, $[0,2\pi)$, where $0$ corresponds to red. Note that the color appearance of the spotlight changes with the object and background color, since we only see the combined reflected color.
    \item The \textbf{background hue} (hue\_b) is the color of the background. The value range is $[0,2\pi)$ with the same color spectrum as the spotlight hue.
    \item The \textbf{object hue} (hue\_o) is the color of the object, and follows the same setup as the background hue.
    \item The \textbf{object shape} (obj\_s) is a categorical variable describing the object shape. For the 7-shape dataset version, we consider the same object shapes as \citet{vonkuegelgen2021self}: Cow \cite{keenan2021cow}, Head \cite{rusinkiewicz2021head}, Dragon \cite{curless1996volumetric}, Hare \cite{turk1994zippered}, Armadillo \cite{krishnamurthy1996fitting}, Horse \cite{praun2000lapped}, Teapot \cite{newell1975utilization}. An example image for each of the objects is shown in \cref{fig:appendix_causal3d_shapes_example}.
\end{itemize}

\subsubsection{Dataset Generation}

The datasets are generated by starting at a random sample of all causal factors.
For each next time step, we generate a sample according to the conditional distributions of each causal factor (see \cref{sec:appendix_causal3d_temporal_causal_relations} for details on the distributions).
Additionally, we sample intervention targets $I^{t+1}_1,...,I^{t+1}_7$ for all 7 causal factors.
For the datasets in \cref{sec:experiments}, we sample the targets from $I^{t+1}_i\sim\text{Bernoulli}(0.1)$, and show additionally intervention settings on this dataset in \cref{sec:appendix_additional_experiments_causal3d_confounded}.
For each causal factor for which the intervention target is one, we replace its previously sampled value with a new value randomly sampled from a uniform distribution.
For angles and hues, the distribution is $U(0,2\pi)$, while for the positions, we use $U(-2,2)$.
For the object shape, we uniformly sample one out of the seven shapes.
After performing the interventions, the sampled vector of causal factors is used to generate an image of a resolution of $64\times 64$ using Blender.
Note that for visualization purposes, the depicted images in this section are shown in higher resolution ($256\times 256$).
This makes it easier to recognize the object shapes and their rotations.
However, we use a resolution of $64\times 64$ in experiments to keep the computational cost of the experiments in a reasonable range.

Repeating this generation procedure for several steps results in one long sequence, which we use as a dataset.
For the experiments on Temporal-Causal3DIdent Teapot, we generate a sequence of $150$,$000$ images.
For the experiments on Temporal-Causal3DIdent 7-shapes, the chosen dataset size was $250$,$000$ images.
The large dataset sizes were chosen to prevent any sampling bias and focus the experiments on general identifiability.
We noticed that smaller dataset sizes such as $50$,$000$ images still gave good scores on the correlation metrics, but a decrease in the triplet evaluation was noticeable for most causal factors, especially the position and rotation.

\subsubsection{Temporal Causal Relations}
\label{sec:appendix_causal3d_temporal_causal_relations}
\begin{figure*}[t!]
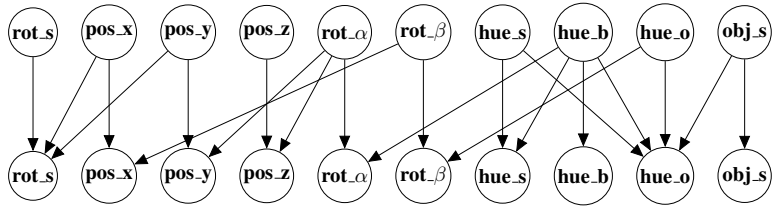

    \centering
    \resizebox{0.6\textwidth}{!}{
        \tikz{ %
    		\node[latent] (posx) {\textbf{pos\_x}} ; %
    		\node[latent, right=of posx, xshift=-.6cm] (posy) {\textbf{pos\_y}} ; %
    		\node[latent, right=of posy, xshift=-.6cm] (posz) {\textbf{pos\_z}} ; %
    		\node[latent, right=of posz, xshift=-.6cm] (rota) {\textbf{rot\_$\alpha$}} ; %
    		\node[latent, right=of rota, xshift=-.6cm] (rotb) {\textbf{rot\_$\beta$}} ; %
    		\node[latent, left=of posx, xshift=.6cm] (rots) {\textbf{rot\_s}} ; %
    		\node[latent, right=of rotb, xshift=-.6cm] (hues) {\textbf{hue\_s}} ; %
    		\node[latent, right=of hues, xshift=-.6cm] (hueb) {\textbf{hue\_b}} ; %
    		\node[latent, right=of hueb, xshift=-.6cm] (hueo) {\textbf{hue\_o}} ; %
    		\node[latent, right=of hueo, xshift=-.6cm] (objs) {\textbf{obj\_s}} ; %
    		
    		\node[latent, below=of posx, yshift=-.5cm] (posx1) {\textbf{pos\_x}} ; %
    		\node[latent, right=of posx1, xshift=-.6cm] (posy1) {\textbf{pos\_y}} ; %
    		\node[latent, right=of posy1, xshift=-.6cm] (posz1) {\textbf{pos\_z}} ; %
    		\node[latent, right=of posz1, xshift=-.6cm] (rota1) {\textbf{rot\_$\alpha$}} ; %
    		\node[latent, right=of rota1, xshift=-.6cm] (rotb1) {\textbf{rot\_$\beta$}} ; %
    		\node[latent, left=of posx1, xshift=.6cm] (rots1) {\textbf{rot\_s}} ; %
    		\node[latent, right=of rotb1, xshift=-.6cm] (hues1) {\textbf{hue\_s}} ; %
    		\node[latent, right=of hues1, xshift=-.6cm] (hueb1) {\textbf{hue\_b}} ; %
    		\node[latent, right=of hueb1, xshift=-.6cm] (hueo1) {\textbf{hue\_o}} ; %
    		\node[latent, right=of hueo1, xshift=-.6cm] (objs1) {\textbf{obj\_s}} ; %
    		
    		\edge{posx}{posx1} ;
    		\edge{posy}{posy1} ;
    		\edge{posz}{posz1} ;
    		\edge{rota}{rota1} ;
    		\edge{rotb}{rotb1} ;
    		\edge{rots}{rots1} ;
    		\edge{hues}{hues1} ;
    		\edge{hueb}{hueb1} ;
    		\edge{hueo}{hueo1} ;
    		\edge{objs}{objs1} ;
    		
    		\edge{rotb}{posx1} ;
    		\edge{rota}{posy1} ;
    		\edge{rota}{posz1} ;
    		\edge{hueb}{rota1} ;
    		\edge{hueo}{rotb1} ;
    		\edge{hueb}{hues1} ;
    		\edge{objs}{hueo1} ;
    		\edge{hues}{hueo1} ;
    		\edge{hueb}{hueo1} ;
    		\edge{posx}{rots1} ;
    		\edge{posy}{rots1} ;
    	}
    }
    \caption{Causal relations between all dimensions of the causal factors of the Temporal Causal3DIdent dataset. The causal graph of \cref{fig:experiments_dataset_causal3d_causal_graph} summarizes pos\_x, pos\_y, pos\_z into pos\_o and rot\_$\alpha$, rot\_$\beta$ into rot\_o. See \cref{sec:appendix_causal3d_temporal_causal_relations} for details on the conditional distributions.}
    \label{fig:appendix_dataset_causal3d_full_causal_graph}
\end{figure*}

\begin{figure}[t!]
    \centering
    \ifarxiv
        \tiny
    \else
        \footnotesize
    \fi
    \begin{tabular}{ccccc}
        \includegraphics[width=0.12\textwidth]{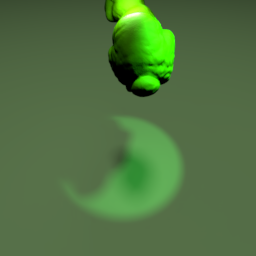} & 
        \includegraphics[width=0.12\textwidth]{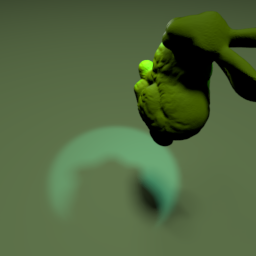} & 
        \includegraphics[width=0.12\textwidth]{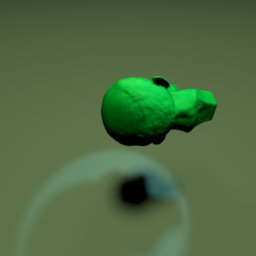} & 
        \includegraphics[width=0.12\textwidth]{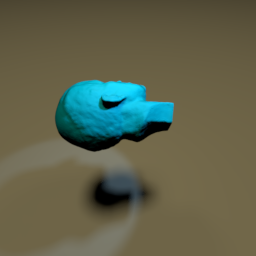} & 
        \includegraphics[width=0.12\textwidth]{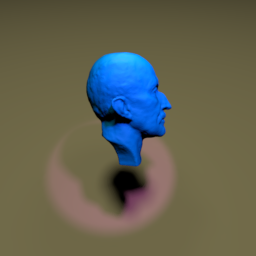}
        \\
        (1) - & (2) rot-$\alpha$ & (3) pos-x & (4) pos-y, hue-b & (5) rot-$\alpha$ \\[7pt]
        \includegraphics[width=0.12\textwidth]{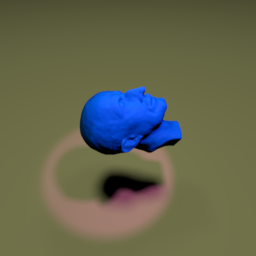} & 
        \includegraphics[width=0.12\textwidth]{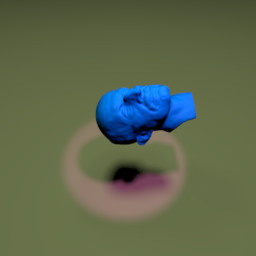} & 
        \includegraphics[width=0.12\textwidth]{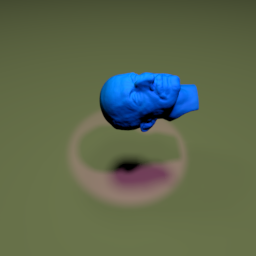} & 
        \includegraphics[width=0.12\textwidth]{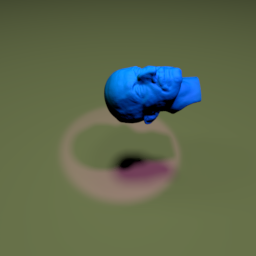} & 
        \includegraphics[width=0.12\textwidth]{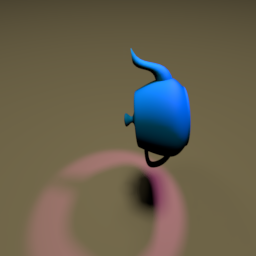}
        \\
        (6) pos-y & (7) - & (8) - & (9) - & (10) hue-b, obj-s \\[7pt]
        \includegraphics[width=0.12\textwidth]{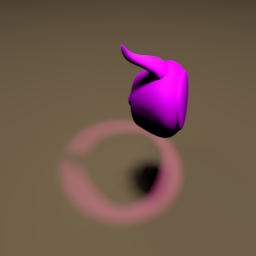} & 
        \includegraphics[width=0.12\textwidth]{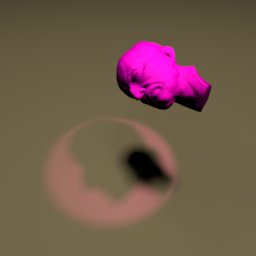} & 
        \includegraphics[width=0.12\textwidth]{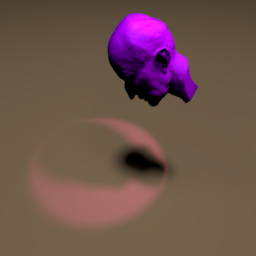} & 
        \includegraphics[width=0.12\textwidth]{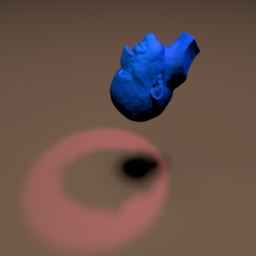} & 
        \includegraphics[width=0.12\textwidth]{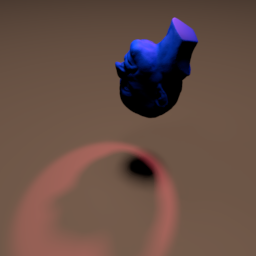}
        \\
        (11) rot-$\beta$ & (12) - & (13) rot-$\beta$ & (14) rot-$\beta$ & (15) - \\[7pt]
        \includegraphics[width=0.12\textwidth]{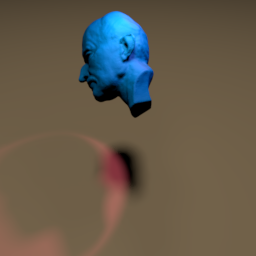} & 
        \includegraphics[width=0.12\textwidth]{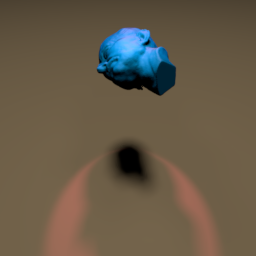} & 
        \includegraphics[width=0.12\textwidth]{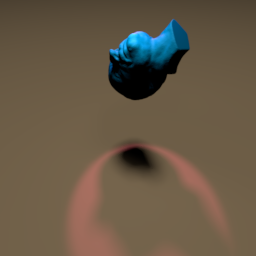} & 
        \includegraphics[width=0.12\textwidth]{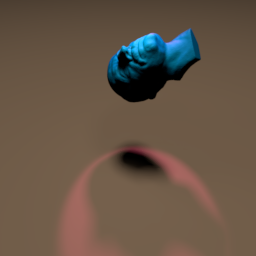} & 
        \includegraphics[width=0.12\textwidth]{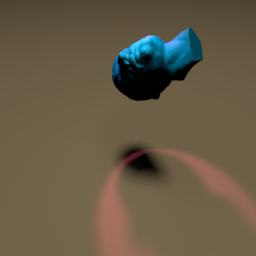}
        \\
        (16) pos-y, rot-$\beta$ & (17) - & (18) - & (19) - & (20) rot-s \\
    \end{tabular}
    \caption{An example sequence with 20 frames in the Temporal Causal3DIdent 7-shapes dataset with higher resolution (from left to right, top to bottom). The causal variables denoted below each image indicate the variables which were intervened on at this time step. For instance, when transitioning from the first to the second image, all variables were sampled according to their temporal dependency except rot-$\alpha$.}
    \label{fig:appendix_causal3d_sequence_example}
\end{figure}

Below, we define the transition functions used in the causal graph of Temporal Causal3DIdent dataset, as shown in \cref{fig:experiments_dataset_causal3d_causal_graph} (see \cref{fig:appendix_dataset_causal3d_full_causal_graph} for the relations on individual causal dimensions).
The chosen function forms are inspired by the ones defined by \citet{vonkuegelgen2021self} for the Causal3DIdent dataset. 

For the position, rotation, and hue values, we sample new values over time with the following functions:
\begin{eqnarray}
    \label{eq:causal3dident_time_scm_start}
    \text{pos\_x}^{t+1} & = & f\left(1.5\cdot\sin(\text{rot\_}\beta^t), \text{pos\_x}^t, \epsilon^t_x\right)\\
    \label{eq:causal3dident_time_scm_pos_y}
    \text{pos\_y}^{t+1} & = & f\left(1.5\cdot\sin(\text{rot\_}\alpha^t), \text{pos\_y}^t, \epsilon^t_y\right)\\
    \text{pos\_z}^{t+1} & = & f\left(1.5\cdot\cos(\text{rot\_}\alpha^t), \text{pos\_z}^t, \epsilon^t_z\right)\\
    \text{rot\_}\alpha^{t+1} & = & f\left(\text{hue\_b}^t, \text{rot\_}\alpha^t, \epsilon^t_{\alpha}\right)\\
    \text{rot\_}\beta^{t+1} & = & f\left(\text{hue\_o}^t, \text{rot\_}\beta^t, \epsilon^t_{\beta}\right)\\
    \text{rot\_s}^{t+1} & = & f\left(\text{atan2}(\text{pos\_x}^t, \text{pos\_y}^t), \text{rot\_s}^t, \epsilon^t_{rs}\right)\\
    \text{hue\_s}^{t+1} & = & f\left(2\pi-\text{hue\_b}^t, \text{hue\_s}^t, \epsilon^t_{hs}\right)\\
    \label{eq:causal3dident_time_scm_end}
    \text{hue\_b}^{t+1} & = & \text{hue\_b}^t + \epsilon^t_{b}
\end{eqnarray}
where $f(a,b,c)=\frac{a-b}{2}+c$, and all $\epsilon$-variables being independent samples from a Gaussian distribution with standard deviation $0.1$ for positions, and $0.15$ for angles and hues.
Intuitively, the function $f$ represents that we create a 'goal' position for each variable based on its parents, and move towards the goal by taking the average between goal and current position, with additive noise.
This gives us a simulation of a moving system, which, however, also permits large changes without interventions.

The function of the object hue depends on the categorical object shape, which is outlined in \cref{tab:appendix_causal3d_object_hue}.
We use the same setup as \citet{vonkuegelgen2021self}, where the hare is trying to blend into the background and spotlight, while the dragon tries to stand out.
The colors of the other objects are spread out across the color ring.

Finally, for the object shape, we use a noisy identity map over time.
With a probability of 5\%, we change the current object shape with a newly sampled one from a uniform distribution.
This introduces additional noise to the object shape besides the interventions.

\begin{table}[t!]
    \centering
    \begin{tabular}{cc}
        \toprule
        Object shape & Object hue goal \\
        \midrule
        Teapot & $0$\\
        Armadillo & $\frac{1}{5}\cdot 2\pi$\\
        Hare & $\text{avg}(\text{hue\_spot}, \text{hue\_back})$\\
        Cow & $\frac{2}{5}\cdot 2\pi$\\
        Dragon & $\pi+\text{avg}(\text{hue\_spot}, \text{hue\_back})$\\
        Head & $\frac{3}{5}\cdot 2\pi$\\
        Horse & $\frac{4}{5}\cdot 2\pi$\\
        \bottomrule
    \end{tabular}
    \caption{The causal relation of the object shape, background hue, and spotlight hue to the object hue goal $g$ with which we determine the next step as $\text{hue\_o}^{t+1} = f\left(g, \text{hue\_o}^t, \epsilon^t_{ho}\right)$. The angle mean is defined as $\text{avg}(\alpha, \beta)=\text{atan2}\left(\frac{\sin(\alpha)+\sin(\beta)}{2}, \frac{\cos(\alpha)+\cos(\beta)}{2}\right)$.}
    \label{tab:appendix_causal3d_object_hue}
\end{table}

\begin{figure}
    \centering
    \begin{subfigure}{0.3\textwidth}
        \includegraphics[width=\textwidth]{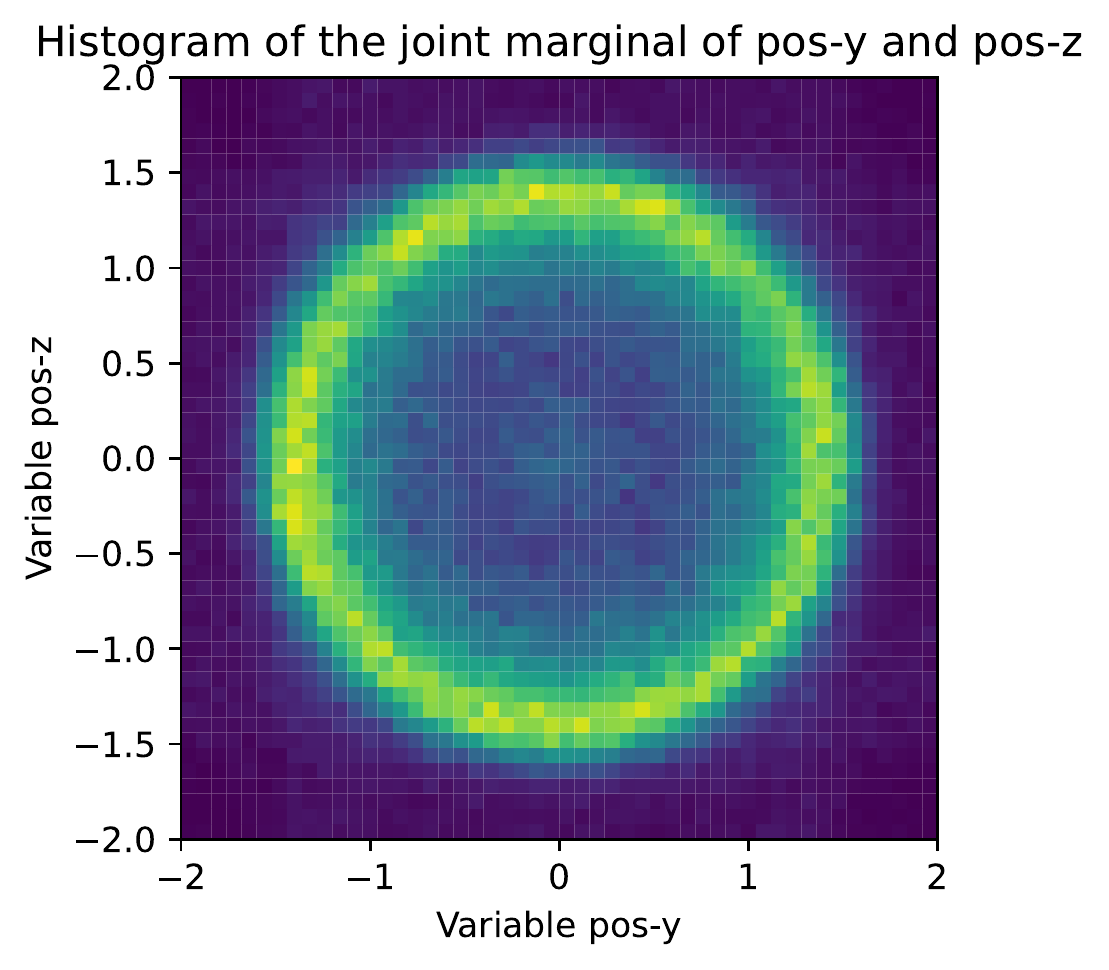}
        \caption{$p(\text{pos\_y},\text{pos\_z})$}
    \end{subfigure}
    \hfill
    \begin{subfigure}{0.3\textwidth}
        \includegraphics[width=\textwidth]{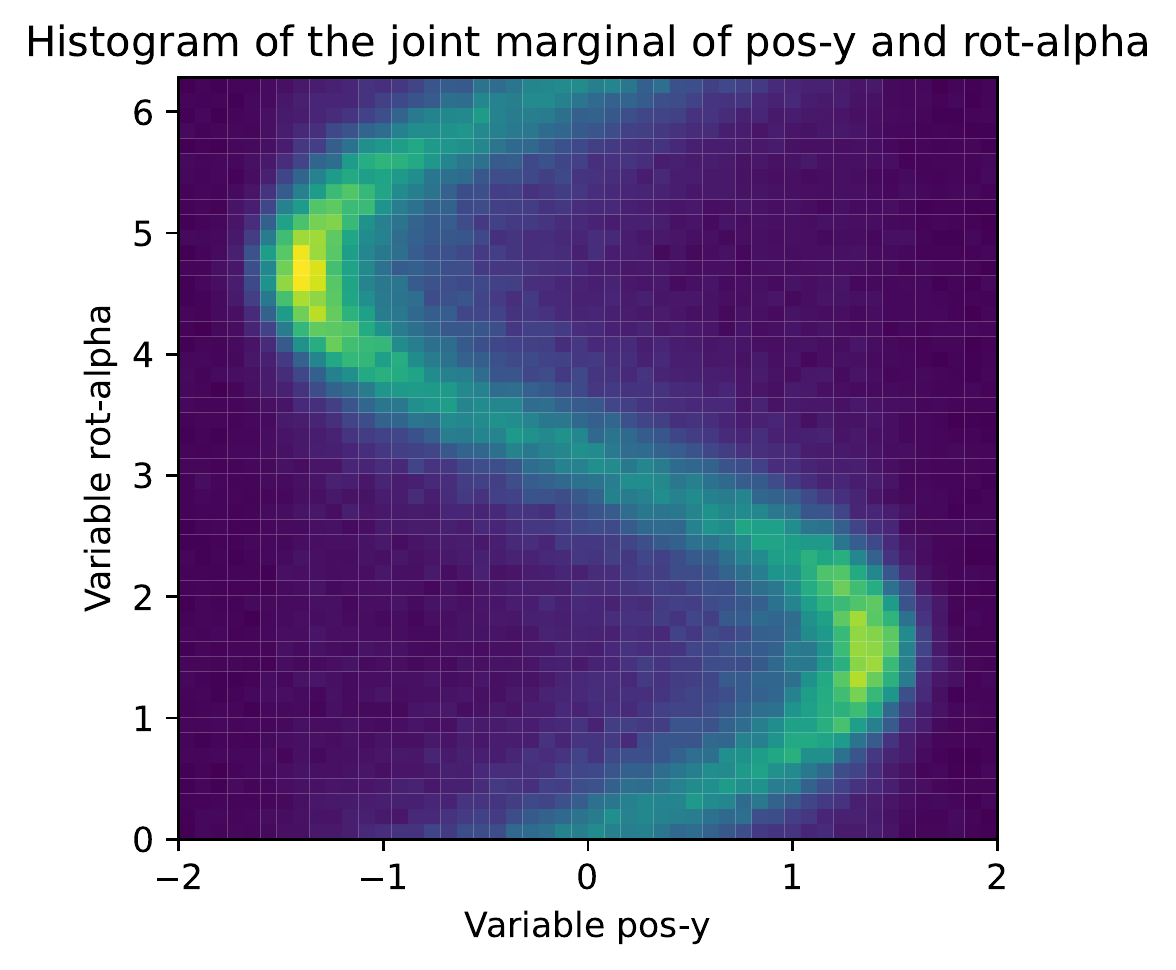}
        \caption{$p(\text{pos\_y},\text{rot\_}\alpha)$}
    \end{subfigure}
    \hfill
    \begin{subfigure}{0.3\textwidth}
        \includegraphics[width=\textwidth]{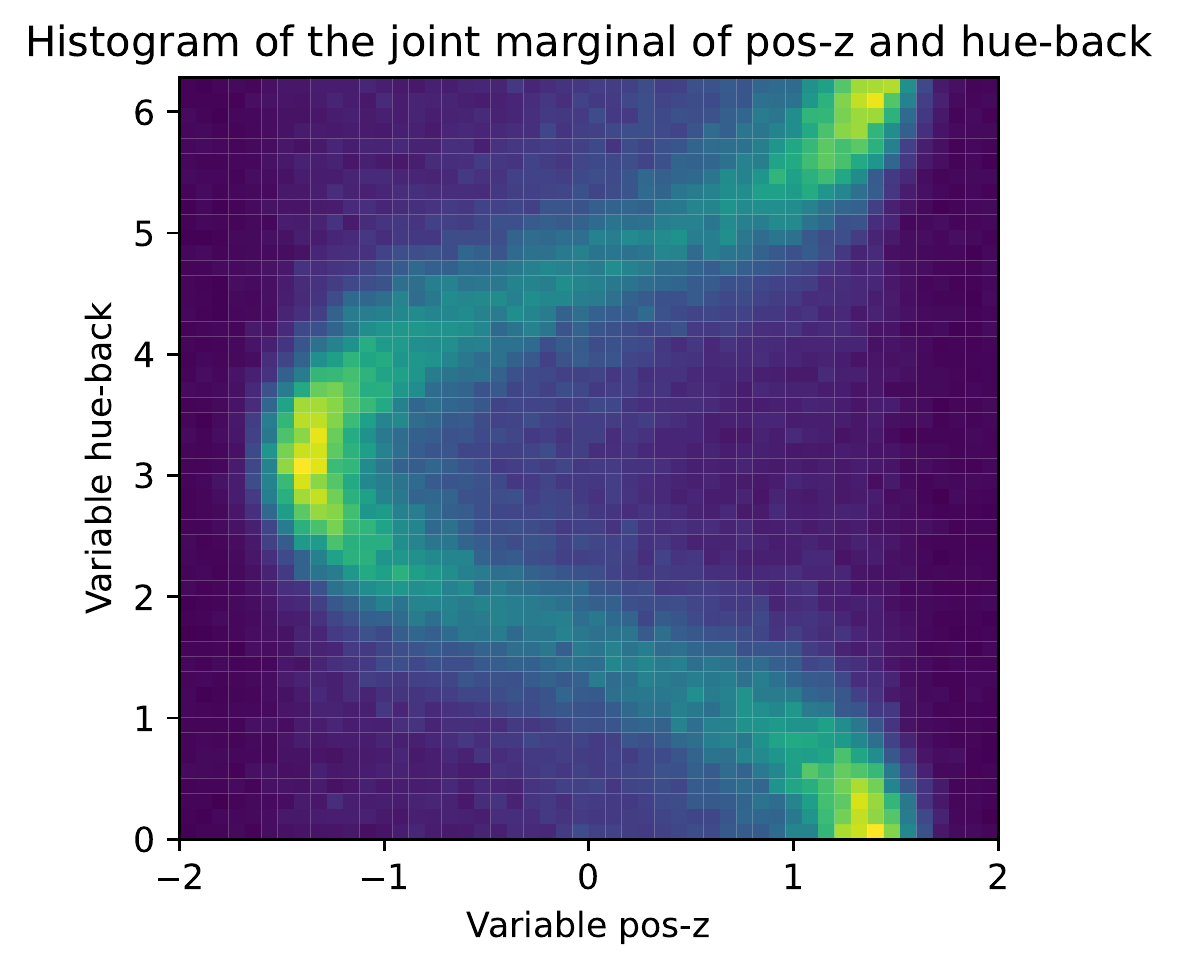}
        \caption{$p(\text{pos\_z},\text{hue\_b})$}
    \end{subfigure}
    \caption{Overview of the joint marginal distributions of selected variables in the Temporal Causal3DIdent dataset, showcasing the correlations among variables. The figures show histograms over the two variables, where yellow indicates a high likelihood/frequency, while dark blue has close to zero samples. (a) The two causal variables pos\_y and pos\_z share a common confounder, $\text{rot\_}\alpha$, which causes them to follow a circle with radius 1.5. (b) $\text{rot\_}\alpha$ is a parent of $\text{pos\_y}$, and one can see that the marginal closely follows its functional form (see \cref{eq:causal3dident_time_scm_pos_y}). (c) The hue of the background is an ancestor of pos\_z, with \text{rot}\_$\alpha$ in between the two variables. Yet, the marginal clearly follows the cosine signal, showing that the correlation goes beyond parents.}
    \label{fig:appendix_causal3d_temporal_dependency_histograms}
\end{figure}

To showcase the dependency among causal variables, we plot the marginal distribution of tuples of variables in \cref{fig:appendix_causal3d_temporal_dependency_histograms}. 
The distributions are plotted based on a histogram of a dataset with 250,000 samples.
Despite the occasional interventions, a clear correlation among variables with a confounder and ancestor-descendant relations can be seen.
Overall, this shows that the chosen functions in Equation~\ref{eq:causal3dident_time_scm_start} to \ref{eq:causal3dident_time_scm_end} introduce strong correlations among variables, which makes disentangling the factors a difficult task. 

\subsection{Interventional Pong}
\label{sec:appendix_experimental_details_interventional_pong}

\begin{figure}
    \centering
    \ifarxiv
        \tiny
    \else
        \footnotesize
    \fi
    \begin{tabular}{cccccccccc}
        \includegraphics[width=0.1\textwidth]{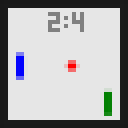} & 
        \includegraphics[width=0.1\textwidth]{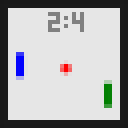} & 
        \includegraphics[width=0.1\textwidth]{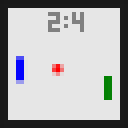} & 
        \includegraphics[width=0.1\textwidth]{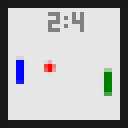} & 
        \includegraphics[width=0.1\textwidth]{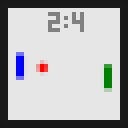} & 
        \includegraphics[width=0.1\textwidth]{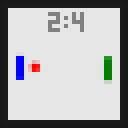} & 
        \includegraphics[width=0.1\textwidth]{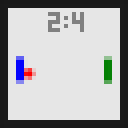} \\
         (1) - & (2) - & (3) - & (4) - & (5) - & (6) ball\_vel\_dir & (7) ball\_vel\_dir \\[7pt]
        \includegraphics[width=0.1\textwidth]{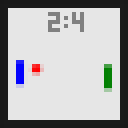} &
        \includegraphics[width=0.1\textwidth]{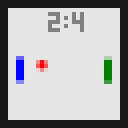} & 
        \includegraphics[width=0.1\textwidth]{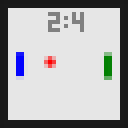} &
        \includegraphics[width=0.1\textwidth]{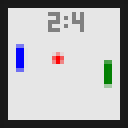} &
        \includegraphics[width=0.1\textwidth]{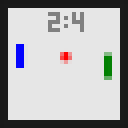} & 
        \includegraphics[width=0.1\textwidth]{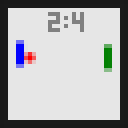} & 
        \includegraphics[width=0.1\textwidth]{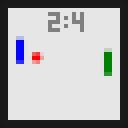}
        \\
        (8) pright\_y & (9) - & (10) - & (11) pright\_y & (12) - & (13) ball\_x & (14) pright\_y \\[7pt] 
        \includegraphics[width=0.1\textwidth]{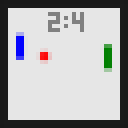} & 
        \includegraphics[width=0.1\textwidth]{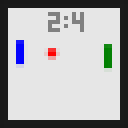} &
        \includegraphics[width=0.1\textwidth]{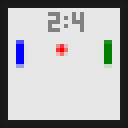} & 
        \includegraphics[width=0.1\textwidth]{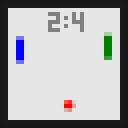} &
        \includegraphics[width=0.1\textwidth]{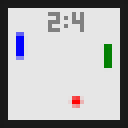} & 
        \includegraphics[width=0.1\textwidth]{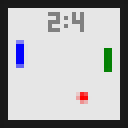}& 
        \includegraphics[width=0.1\textwidth]{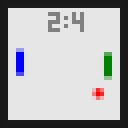}
        \\
        (15) pleft\_y & (16) - & (17) - & (18) ball\_y & (19) pleft\_y & (20) pleft\_y & (21) ball\_x \\
    \end{tabular}
    \caption{An example sequence with 21 frames in the Interventional Pong dataset. The causal variables denoted below each image indicate the variables which were intervened on at this time step. For instance, when transitioning from the first to the second image, all variables were sampled according to their temporal dependency except. The ball velocity direction is encoded in a 4th channel, which is not visualized here. It encodes the velocity by showing the ball moved by the velocity at the current time step.}
    \label{fig:appendix_pong_sequence_example}
\end{figure}

The Interventional Pong dataset models the dynamics of the popular Atari game Pong \cite{bellemare2013arcade}.
We show a sequence of 21 frames in \cref{fig:appendix_pong_sequence_example}.
Thereby, each frame has a resolution of $32\times 32$.

In this environment, we have six underlying causal factors:
\begin{itemize}
    \item The \textbf{ball x-position} (ball\_x) describes the x-position (left to right) of the ball. The value range is limited to $[-1,1]$ where $-1$ corresponds to hitting the left border, and $1$ the right. Under interventions, we sample the x-position uniformly in the space between the two paddles, which corresponds to a value range of approximately $[-0.7,0.7]$.
    \item The \textbf{ball y-position} (ball\_y) describes the y-position (top to bottom) of the ball. The value range is limited to $[-1,1]$ where $-1$ corresponds to hitting the upper border, and $1$ the lower border. Under interventions, we sample the y-position uniformly from $U(-1,1)$.
    \item The \textbf{ball velocity direction} (ball\_vel\_dir) describes the angle of the velocity of the ball. The value range is $[0,2\pi)$ where $0$ corresponds to the ball heading to the right. Under interventions, we sample the velocity direction uniformly from $U(0,2\pi)$.
    \item The \textbf{paddle-left y-position} (pleft\_y) describes the y-position of the left paddle (blue in the figures). It has the same value range as the causal factor ball\_y. Over time, the paddle takes a step towards the ball\_y position of the previous time step. The step size is sampled from a Gaussian mean of $0.05$ and standard deviation of $0.017$. If an intervention is performed on pleft\_y, it flips the sign of the step, \ie{} going up instead of down, with a chance of 50\%. This replaces the policy of following the ball with a random one.
    \item The \textbf{paddle-right y-position} (pright\_y) describes the y-position of the right paddle (green in the figures), and follows the same dynamics as pleft\_y.
    \item The \textbf{score} (score) describes the player's score (gray background numbers). It contains two scores (left player and right player), each with an integer value between $0$ and $4$. When the ball hits one of the two border, the score for the corresponding player is increased. Further, if a player scores their 5th goal, the score is reset to $0,0$. No interventions are provided on this causal factor.
\end{itemize}

\begin{figure}
    \centering
    \footnotesize
    \begin{tabular}{cccccccccc}
        \includegraphics[width=0.1\textwidth]{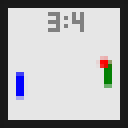} & 
        \includegraphics[width=0.1\textwidth]{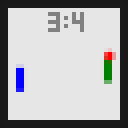} & 
        \includegraphics[width=0.1\textwidth]{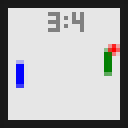} & 
        \includegraphics[width=0.1\textwidth]{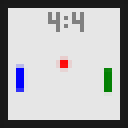} & 
        \includegraphics[width=0.1\textwidth]{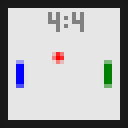} & 
        \includegraphics[width=0.1\textwidth]{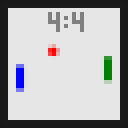} & 
        \includegraphics[width=0.1\textwidth]{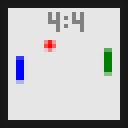} \\
    \end{tabular}
    \caption{An example sequence of 7 frames (left to right) in the Interventional Pong dataset, where the environment is being reset at the fourth time step. The ball is returned to the center position and gets a new velocity direction, as well as the paddles being moved to new random positions. The score of the respective player is increased as well.}
    \label{fig:appendix_pong_sequence_example_reset}
\end{figure}

\begin{figure*}[t!]
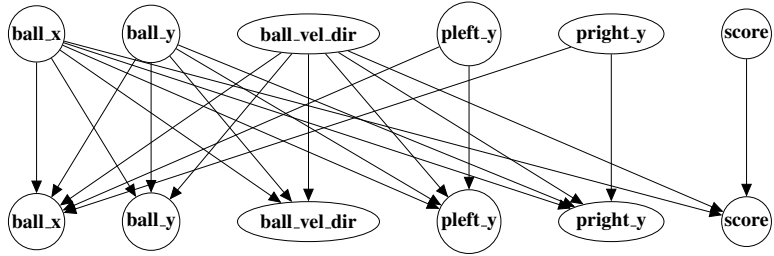

    \centering
    \resizebox{0.6\textwidth}{!}{
        \tikz{ %
    		\node[latent] (posx) {\textbf{ball\_x}} ; %
    		\node[latent, right=of posx] (posy) {\textbf{ball\_y}} ; %
    		\node[latent, right=of posy, ellipse] (vel) {\textbf{ball\_vel\_dir}} ; %
    		\node[latent, right=of vel] (pleft) {\textbf{pleft\_y}} ; %
    		\node[latent, right=of pleft, ellipse] (pright) {\textbf{pright\_y}} ; %
    		\node[latent, right=of pright] (score) {\textbf{score}} ; %
    		
    		\node[latent, below=of posx, yshift=-1.3cm] (posx1) {\textbf{ball\_x}} ; %
    		\node[latent, right=of posx1] (posy1) {\textbf{ball\_y}} ; %
    		\node[latent, right=of posy1, ellipse] (vel1) {\textbf{ball\_vel\_dir}} ; %
    		\node[latent, right=of vel1] (pleft1) {\textbf{pleft\_y}} ; %
    		\node[latent, right=of pleft1, ellipse] (pright1) {\textbf{pright\_y}} ; %
    		\node[latent, right=of pright1] (score1) {\textbf{score}} ; %
    		
    		\edge{posx}{posx1} ;
    		\edge{posy}{posy1} ;
    		\edge{vel}{vel1} ;
    		\edge{pleft}{pleft1} ;
    		\edge{pright}{pright1} ;
    		\edge{score}{score1} ;
    		
    		\edge{posx}{posy1} ;
    		\edge{posx}{vel1} ;
    		\edge{posx}{pleft1} ;
    		\edge{posx}{pright1} ;
    		\edge{posx}{score1} ;
    		\edge{posy}{posx1} ;
    		\edge{posy}{vel1} ;
    		\edge{posy}{pleft1} ;
    		\edge{posy}{pright1} ;
    		\edge{vel}{posx1} ;
    		\edge{vel}{posy1} ;
    		\edge{vel}{pleft1} ;
    		\edge{vel}{pright1} ;
    		\edge{vel}{score1} ;
    		\edge{pleft}{posx1} ;
    		\edge{pright}{posx1} ;
    	}
    }
    \caption{Causal relations between all dimensions of the causal factors of the Interventional Pong dataset. Note that although the graph is very dense, not all relations have an effect on the causal variables at each time step. For instance, the velocity only influences the paddles and the score when the ball x position moved by the velocity ends up in one of the two boundaries.}
    \label{fig:appendix_dataset_pong_full_causal_graph}
\end{figure*}

Given a previous time step, the game engine performs the following steps in order to determine the next step:
\begin{enumerate}
    \item Using the velocity direction at the previous time step, we determine the velocity in $x$ and $y$ direction. We then move the x-position of the ball accordingly. 
    \item If this new ball-x position hits one of the two borders, we increase the score of the corresponding player and reset the game. Resetting the game includes setting the ball to the center of the field, randomly sampling a ball velocity direction, and replacing the position of the paddles with a new random position between $-0.66$ and $0.66$. The latter is done to prevent the paddles of ending up in one of the two corners when the ball is reset. An example of resetting the environment is shown in \cref{fig:appendix_pong_sequence_example_reset}.
    \item If the game was not reset, we do the following steps:
    \begin{itemize}
        \item We first move the two paddles according to their dynamics, as described before, moving towards the ball.
        \item We move the ball y position by the velocity in y direction. If the new ball\_y position collides with the upper or lower boundary, we calculate the new y-position it would have after the collision, and mirror the velocity angle on the x-axis.
        \item If the new ball x and y position collide with one of the two paddle positions in the previous time step, we reflect the ball accordingly and calculate its new x position, as well as mirroring the velocity angle on the y-axis.
        \item Finally, we add Gaussian noise with standard deviation of $0.015$ to the ball x and y position, as well as the velocity direction. The small standard deviation was chosen to keep the dynamics similar to Pong and not divert too much into random movements.
    \end{itemize}
    \item With a chance of about 35\%, we do not perform any intervention. Otherwise, we randomly sample one out of the five causal factors (excluding the score), and perform an intervention as described before.
\end{enumerate}
We provide the full code of the dynamics and dataset generation for this Pong environment in the supplementary.
The causal graph implied by these dynamics are shown in \cref{fig:appendix_dataset_pong_full_causal_graph}.
Similarly to the Temporal Causal3DIdent dataset, we generate a dataset by sampling one long sequence with a dataset size of $100$,$000$.

\subsection{Experimental Design}
\label{sec:appendix_experimental_details_experimental_design}

In this section, we give a more detailed description of elements used in the experiments.
We first discuss the evaluation via the correlation metrics and triplet evaluation, and then describe the implementation details of the target classifier.

\subsubsection{Correlation Metrics}

As described in \cref{sec:experiments_setup_evaluation_metrics}, we evaluate the models by measuring the correlation of the learned latent variables to the ground truth causal factors. 
Typically, in the literature, the latent variables are one-dimensional.
In our setup, however, the latent variables span over multiple dimensions and can describe a causal factor in a non-linear relation.
Thus, plotting the correlation per variable would not reveal the true correlation between the learned, multidimensional latent variables and causal factors.

As a first step, we therefore need to learn a mapping between sets of latent variables and true causal factors. 
We emphasize that during this whole evaluation, the models are frozen and will not be updated.
We only make use of their encoders to map observations to latent vectors.
For \OurApproach{}, we therefore apply one MLP per set of latent variables that, according to the learned assignment function $\psi$, belong to the same causal factor. 
The MLP is then trained to predict \emph{all} causal factors per set of latents, on which we measure the correlation.
To this end, we use a two-layer MLP with a latent dimensionality of 128.
The MLP size was chosen sufficiently large to uncover any possible correlations between the latent and causal factors.
Using larger MLPs did not show any noticeable change in the correlation found, verifying that the MLP is large enough to learn any potential non-linear mapping.

To train the MLP, we split the test dataset into two subsets: one for training the MLP ($40\%$ of the dataset), and one for measuring the correlation metrics on ($60\%$ of the dataset).
The test datasets are generated by randomly sampling sets of causal variables of the respective dataset.
We use a dataset size of $25$,$000$ images to prevent any form of overfitting, and ensure a sufficiently large sample size for the correlation metrics.
However, in our datasets, the causal factors are correlated themselves. 
This means that the optimal disentanglement function would yet measure a correlation from, \eg{}, the latents describing the ball\_y position to the paddles' y positions.
This makes it difficult to spot spurious correlations between latents and causal factors, since with sufficient latent dimensions, a model can describe the same causal factors in multiple sets of latent dimensions.
Since this would not follow the goal of a good disentanglement, we prevent measuring such correlations by perform this evaluation on test datasets without temporal correlations among causal factors.
In other words, we generate samples by sampling each causal factor independently, and mapping those into the observation space.

As a loss function for training the MLP, we use a mean-squared-error loss (MSE) for all continuous values excluding the circular values (angles and hues).
For the circular values between 0 and $2\pi$, using the MSE loss would be disadvantageous since predicting a value of $2\pi-\epsilon$, when the label is $0$, causes a large error although on a unit circle, the two points would be very close.
Hence, as a better alternative, we train the MLP by predicting a vector in two dimensions.
We then use the cosine distance between the predicted vector and the vector of the ground truth angle projected onto the unit circle.
This gives us a loss that respects the circular nature of the causal factor.
Since the loss is independent of the length of the predicted vector, we apply a small regularizer to stabilize the vector length.
Finally, for the categorical causal factors, we use cross entropy as a training loss.

After training the MLPs, we can use them to create predictions for all causal factors for a set of latents.
Then, we determine the correlation metrics between these predictions and the ground truth values for the individual causal factors.
Note that for multidimensional causal factors, we calculate the metrics for each dimension individually, and take the average correlation coefficient over dimensions afterwards.

The $R^2$ correlation coefficient \cite{spearman1904proof} is a metric which compares the prediction of a variable from a conditional input set to the average prediction. For continuous variables, this is:
\begin{equation}
    R^2=1-\frac{\sum_i (x_i - \hat{x}_i)^2}{\sum_i (x_i - \bar{x})^2}
\end{equation}
where $x_i$ is the ground truth value of the data point $i$, $\hat{x}_i$ is the prediction, and $\bar{x}=\sum_i x_i/N$ is the average of the ground truth values.
For angles, we adjust the $R^2$ metric by replacing the difference with the cosine distance, and take the angle mean to calculate $\bar{x}$, since the standard mean of two angles, i.e. $\frac{\alpha+\beta}{2}$, does not take into account the circular nature of the angle values. 
The mean over more than 2 angles is defined as:
\begin{equation}
    \text{avg}(\alpha_1,...,\alpha_n)=\text{atan2}\left(\frac{1}{n}\sum_i \sin(\alpha_i), \frac{1}{n}\sum_i \cos(\alpha_i)\right)
\end{equation}
with $\text{atan2}$ being the 2-argument $\arctan$.
For the categorical variables, we replace $\bar{x}$ with the most frequent category in the dataset, and distance by a simple equal, \ie{}, $\delta_{x_i=\hat{x_i}}$.
While one could also apply more suitable methods for finding correlations on categorical variables, we found this simple adjustment sufficient for the two categorical variables we consider in this paper (the object shape in Temporal Causal3DIdent and the score in Pong).

The Spearman's rank correlation \cite{wright1921correlation} ranks both the prediction and ground truth values, and checks the difference in ranking using the Pearson correlation coefficient.
We can apply it as is for both categorical and continuous values, excluding angles.
For the circular values (angles and hues), we report the average Spearman's rank of their sine and cosine transform, since the original values do not have a unique ranking.

Both metrics result in a correlation matrix, where we have on one dimension the sets of latent variables, and the ground truth causal variables on the other.
We show examples of these matrices in \cref{fig:appendix_additional_experiments_correlation_matrices_causal3d_teapot} and \cref{fig:appendix_additional_experiments_correlation_matrices_causal3d_all7}.
The optimal matrix would contain 1s where the sets of latents was assigned to its respective causal factor by $\psi$ in \OurApproach{}, and zero everywhere else.
To summarize such a matrix, we therefore determine the average of these matching values (usually the diagonal).
Further, to show the entanglement with other causal factors, we determine for each set of latents the causal factor with the highest correlation besides the matching one, and average these correlations across latents.
Since the baselines, SlowVAE and \iVAEAdapt{}, do not learn an assignment function like $\psi$, we first need to determine one before calculating above's metrics.
We do this by first calculating all metrics for all latent dimensions independently, and then assigning a latent variable to the causal factor it has the highest correlation to according to the $R^2$ correlation coefficient.

\subsubsection{Triplet Evaluation}

The triplet evaluation provides a parameteric-free evaluation, which can reveal complex dependencies between latent variables in any disentanglement model.
To repeat the general approach, we create triplets of images: the first two are randomly sampled test images, while the third one is created based on a random combination of causal factors of the first two images. 
The test images are taken from a test dataset which contains a sequence of $10$,$000$, generated in the same fashion as the training dataset.
Hence, it also contains the common correlations between causal factors.
The third image is generated by, \eg{}, taking the spotlight rotation and object shape from image 1, and all other causal factors from image 2.
We overall create a set of $10$,$000$ of such triplets.

For evaluation, we then encode the two test images independently with the model that is supposed to be evaluated.
We then perform the combination of ground-truth causal factors as done for the third image in latent space.
This means that for each latent dimension $z_i$, we pick the value of image 1 if the causal factor $C_j$, where $C_j$ is the factor $z_i$ is assigned to, is taken from image 1 for the ground truth third image.
Otherwise, we pick the value of image 2.
We repeat this procedure for every latent dimension, and then use the decoder of the model to generate a new image.
Note that for the normalizing flow, this also includes inverting the flow before applying the decoder.

Since the reconstruction error is not descriptive of the errors being made, \eg{} the rotation in \cref{fig:experiments_triplet_visualizations}, we train an additional CNN in a supervised manner that maps images to the causal factors. 
This CNN has the same architecture as the VAE encoder (see \cref{sec:appendix_experimental_design_hyperparameters}), and is trained on the same training dataset while using the ground truth causal factors as labels.
We use the same training methods for angles as for the MLPs in the correlation metrics.
With this model, we can extract the causal factors from the generated image, and report the average distance of these to the ground truth causal factors.
However, one common difference between the model's prediction and the ground truth is that the prediction is more blurry, and can cause surprising mispredictions of the CNN when it focused on clear corners in the original dataset.
We found it to be beneficial for both the VAE and the AE+NF approach to train the CausalEncoder on reconstructions from the pretrained autoencoder.
Since the autoencoder is able to reconstruct the images relatively well and only has minor blurring effects compared to the VAE, the CNN still managed to predict most factors very accurately and, for the Pong dataset and Teapot, with no noticeable difference to the original ground truth data.

\subsubsection{Target Classifier Implementation}
\label{sec:appendix_experimental_details_target_classifier}

\begin{figure}[t!]
    \centering
    \includegraphics[width=0.25\textwidth]{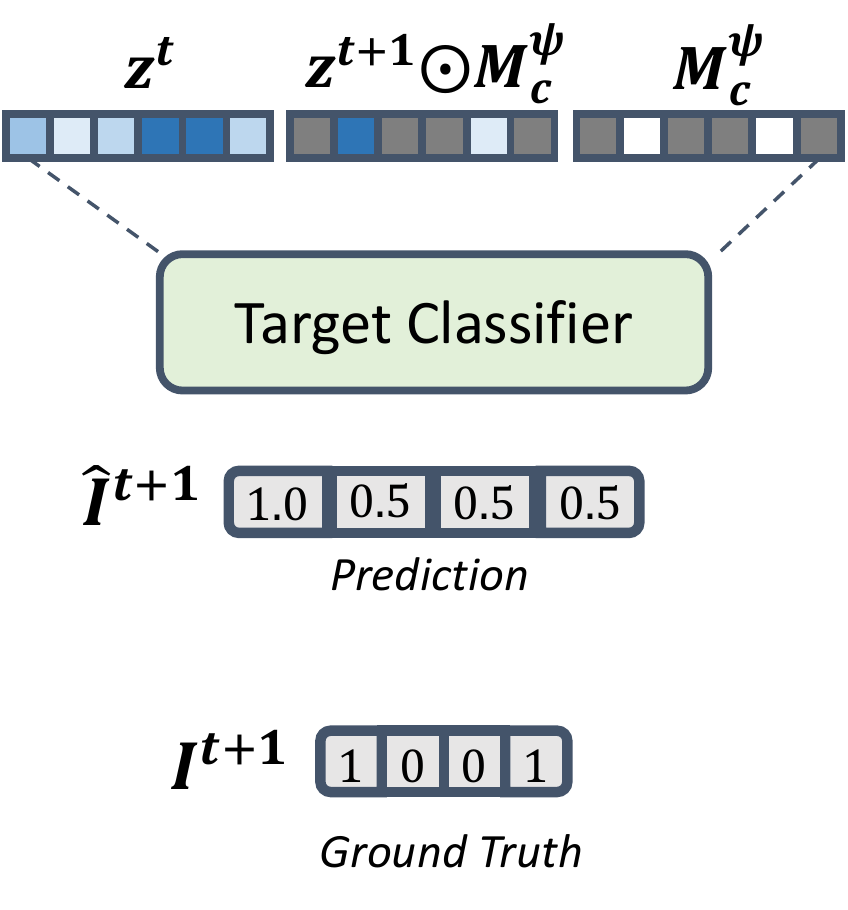}
    \caption{Visualization of the target classifier. From the latents of the first time step, $z^t$, and masked latents of the second time step, $\zpsi{i}^{t+1}$ (here shown in the mask setting), the network tries to predict the intervention targets. If the intervention targets are independent and the model has learned a good disentanglement, only $I_i^{t+1}$ should be predicted well.}
    \label{fig:appendix_target_classifier}
\end{figure}

In this section, we give a detailed description of the target classifier.
The target classifier is an additional small network with the goal of predicting the intervention targets from the latent variables over time. 
While one cannot identify the intervention target 100\% of the time, the distribution change between time steps is yet often detectable.
This way, the target classifier ensures that information about all causal variables is encoded in the latent space, while guiding the disentanglement in latent space and the assignment function $\psi$.

In detail, the classifier takes as input the latents $z^{t}$ and a masked set of $z^{t+1}$ that, according to a sample from $\psi$, are assigned to the causal variable $C_i$. 
We denote this mask with $T^{\psi_i}_j\in \{0,1\}^{M}$, where $T^{\psi_i}_j$ is sampled from the Gumbel Softmax distribution, modeling the assignment of latent dimension $j$ to the causal factor $i$.
Then, the classifier is trained to recover all intervention targets from those latents by optimizing:
\begin{equation}
    \mathcal{L}^{\varphi}_{\text{target}}=-\E_{T,z}\left[\sum_{i=0}^{K}\log p_{\varphi}\left(I^{t+1}|z^{t},z^{t+1}\odot T^{\psi}_i, T^{\psi}_i\right)\right]
\end{equation} 
where $\varphi$ are the parameters of the classifier.
Intuitively, we train the parameters $\varphi$ to predict as many intervention targets from any input as possible, thus revealing any correlations between latents and intervention targets.

When training the classifier for a causal factor $C_i$, we also take gradients with respect to the latent variables $z^{t}$ and $z^{t+1}$ as well as the assignment $\psi$ to optimize the target classification for $I^{t+1}_i$.
However, for all other target variables, $z^{t+1}\odot T^{\psi}_i$ should be uninformative since the intervention target $I_i^{t+1}$ only affects the causal variable $C_i^{t+1}$ while being conditionally independent to all other causal variables. 
Hence, in this case, the latents and $\psi$ are optimized to make the classifier's prediction follow the target's marginal $p(I^{t+1}_j|I^{t+1}_i)$ (or $p(I^{t+1}_j)$ if $I_i^{t+1}\independent I_j^{t+1}$). 
This way, the latents and their assignment to causal variables are trained to only contain information from the causal variables that they are assigned to by $\psi$, and not any others. 
This training objective can be summarized as:
\begin{equation}
    \mathcal{L}^{z,\psi}_{\text{target}}= -\E_{T,z}\Bigg[\sum_{i=0}^{K}\log p_{\varphi}\left(I_i^{t+1}|Tz_i\right)+\sum_{j\neq i}\sum_{l\in\{0,1\}}p\left(I^{t+1}_j=l|I_i^{t+1}\right)\log p_{\varphi}\left(I_j^{t+1}=l|I_i^{t+1},Tz_i\right)\Bigg]
\end{equation} 
where $Tz_i=[z^{t},z^{t+1}\odot T^{\psi}_i, T^{\psi}_i]$.
The second part of the objective is a cross-entropy between the marginal $p(I^{t+1}_j|I^{t+1}_i)$ and the classifier's prediction.
Hence, the full loss for \OurApproach{} is $\mathcal{L}=\mathcal{L}_{\text{ELBO}}+\beta_{\text{class}}\cdot\left(\mathcal{L}^{\varphi}_{\text{target}}+\mathcal{L}^{z,\psi}_{\text{target}}\right)$, where $\beta_{\text{class}}>0$ is a hyperparameter.
In common Deep Learning frameworks like PyTorch \cite{paszke2019pytorch}, this two-folded loss can be implemented by stopping the gradients of $z$ and $\psi$ for $\mathcal{L}^{\phi}_{\text{target}}$, and $\varphi$ for $\mathcal{L}^{z,\psi}_{\text{target}}$.
We empirically verify that this simple addition provides considerable benefits in disentanglement and overall optimization of the VAE (see \cref{sec:experiments_causal3dident}).

\subsection{Hyperparameters}
\label{sec:appendix_experimental_design_hyperparameters}

\subsubsection{Network Architectures}

All models are implemented in the Deep Learning framework PyTorch \cite{paszke2019pytorch} and PyTorch Lightning \cite{Falcon_PyTorch_Lightning_2019}.
The trainings have been performed on a single NVIDIA TitanRTX GPU. 
Below, we give details about the specific architectures used.

\paragraph{VAEs} For all VAE architectures, we make use of the same network architecture to ensure a fair comparison.
The encoder is a convolutional architecture, applying several convolutions in sequence with stride 2 for decreasing the resolution.
Its output is two parameters per latent variable, which denote the mean and standard deviation for each latent variable.
Experiments with increased complexity of the predicted distribution did not show any improvements.
For the decoder, we found it beneficial to replace transposed convolutions with bilinear upsampling and using residual blocks.
The full architectures are outlined in \cref{tab:appendix_hyperparameters_architecture}.
Further, we make use of Batch Normalization \cite{ioffe2015batch} and the SiLU activation function \cite{hendrycks2016gelu, ramachandran2017searching} for non-linearity.
Together, the encoder and decoder have slightly less than 1 million parameters.

\paragraph{Autoencoders} For the standard autoencoder training, we found it to be beneficial to increase the complexity of the decoder by applying two residual blocks per resolution instead of one.
We also experimented with this increased complexity for the VAE approaches.
However, it did not show any noticeable improvement while requiring a smaller batch size and considerably longer training time.

\begin{table}[t!]
    \centering
    \caption{The architecture for both the encoder and decoder. Every residual block consists of two convolutions, each 64 channels, kernel 3 and stride 1 with BatchNorm+SiLU non-linearity. The upsampling represent bilinear upsampling of the features with a scaling factor of 2. The Tanh activation function on the output scales the output between -1 and 1.}
    \label{tab:appendix_hyperparameters_architecture}
    \vspace{0.8em}
    \resizebox{0.8\textwidth}{!}{
    \begin{tabular}{l@{\hskip 15pt}l@{\hskip 15pt}l@{\hskip 15pt}l@{\hskip 15pt}l@{\hskip 15pt}l@{\hskip 15pt}}
    \toprule
       & Layer &  Feature Dimension  & Kernel & Stride & Activation Function \\ 
      & &  \footnotesize{(H $\times$ W $\times$ C)} & & &  \\
        \midrule
    \multirow{11}{0.05\linewidth}{$f_{\textrm{enc}}$}    & Conv  & 32 $\times$ 32 $\times$ 64 & 3 & 2 & BatchNorm+SiLU \\ 
                                                        & Conv  & 32 $\times$ 32 $\times$ 64 & 3 & 1 & BatchNorm+SiLU \\ 
                                                        & Conv  & 16 $\times$ 16 $\times$ 64 & 3 & 2 & BatchNorm+SiLU \\ 
                                                        & Conv  & 16 $\times$ 16 $\times$ 64 & 3 & 1 & BatchNorm+SiLU \\ 
                                                        & Conv  & 8 $\times$ 8 $\times$ 64 & 3 & 2 & BatchNorm+SiLU \\ 
                                                        & Conv  & 8 $\times$ 8 $\times$ 64 & 3 & 1 & BatchNorm+SiLU \\ 
                                                        & Conv  & 4 $\times$ 4 $\times$ 64 & 3 & 2 & BatchNorm+SiLU \\ 
                                                        & Conv  & 4 $\times$ 4 $\times$ 64 & 3 & 1 & BatchNorm+SiLU \\ 
                                                        & Reshape  & 1 $\times$ 1 $\times$ 1024 & - & - & - \\
                                                        & Linear  & 1 $\times$ 1 $\times$ 256 & - & - & LayerNorm+SiLU \\
                                                        & Linear  & 1 $\times$ 1 $\times$ $2\cdot$num\_latents & - & - & - \\
    \midrule
    \multirow{14}{0.05\linewidth}{$f_{\textrm{dec}}$}    & Linear  & 1 $\times$ 1 $\times$ 256 & - & - & LayerNorm+SiLU \\ 
                                                        & Linear  & 1 $\times$ 1 $\times$ 1024 & - & - & - \\ 
                                                        & Reshape  & 4 $\times$ 4 $\times$ 64 & - & - & - \\
                                                        & Upsample  & 8 $\times$ 8 $\times$ 64 & - & - & - \\
                                                        & ResidualBlock  & 8 $\times$ 8 $\times$ 64 & 3 & 1 & - \\ 
                                                        & Upsample  & 16 $\times$ 16 $\times$ 64 & - & - & - \\
                                                        & ResidualBlock  & 16 $\times$ 16 $\times$ 64 & 3 & 1 & - \\ 
                                                        & Upsample  & 32 $\times$ 32 $\times$ 64 & - & - & - \\
                                                        & ResidualBlock  & 32 $\times$ 32 $\times$ 64 & 3 & 1 & - \\ 
                                                        & Upsample  & 64 $\times$ 64 $\times$ 64 & - & - & - \\
                                                        & ResidualBlock  & 64 $\times$ 64 $\times$ 64 & 3 & 1 & - \\ 
                                                        & Pre-Activation  & 64 $\times$ 64 $\times$ 64 & - & - & BatchNorm+SiLU \\ 
                                                        & Conv  & 64 $\times$ 64 $\times$ 64 & 1 & 1 & BatchNorm+SiLU \\ 
                                                        & Conv  & 64 $\times$ 64 $\times$ 3 & 1 & 1 & Tanh \\ 
        \bottomrule
    \end{tabular}
    }
\end{table}

\paragraph{Prior distribution} For \OurApproach{}, the prior distribution $p_{\phi}(\zpsi{i}^{t+1}|z^{t}, I_{i}^{t+1})$ is implemented using an autoregressive model.
We use a single autoregressive model that can model the distributions for all latent sets $\zpsi{i}$ by providing a mask on the intervention vector $I^{t+1}$.
The autoregressive model follow a MADE architecture \cite{germain2015made} where we assign 16 neurons per layer to each latent variable, and the input to these neurons are the features of all previous latents.
Additionally, to the latents, we use the latents of the previous layer as input.
The prior is 2 layers deep, and uses the SiLU activation function.
Finally, the prior predicts a Gaussian distribution per latent.
Similarly to the encoder distribution, one can make the prior more flexible by using normalizing flows.
However, in experiments, this showed to not provide any improvement compared to increased computational cost and parameters.
For the \iVAEAdapt{}, we use a two-layer MLP with a hidden dimensionality of 128.

\paragraph{Normalizing Flows} For the normalizing flow learned on top of the autoencoder, we found a simple affine autoregressive flow to work well.
More flexible coupling transformations such as a MixtureCDF layer \cite{ho2019flow} did not show to improve the results.
For the autoregressive network, we use the same MADE architecture as for the prior distribution.
Further, in between the coupling layer, we use common flow layers such as Activation Normalization and Invertible 1x1 Convolutions \cite{kingma2018glow}.

\paragraph{Target classifier} The target classifier uses a single-layer MLP with hidden dimensionality of 128, Layer Normalization \cite{ba2016layer} and a SiLU activation function.

\subsubsection{Training Hyperparameters}

\begin{table}[t!]
    \centering
    \caption{An overview of the hyperparameter used for \OurApproach{}-VAE, \iVAEAdapt{} and SlowVAE. This parameter configuration resulted in a training time of about 2 days on one NVIDIA TitanRTX GPU for the Temporal Causal3DIdent, and 12 hours for the Interventional Pong dataset.}
    \label{tab:appendix_hyperparameter_overview_vae}
    \footnotesize
    \begin{tabular}{ll}
        \toprule
        \textbf{Hyperparameter} & \textbf{Value} \\
        \midrule
        Batch size & 512 \\
        Optimizer & Adam \cite{kingma2015adam} \\
        Learning rate & 1e-3 \\
        Learning rate scheduler & Cosine Warmup (100 steps) \\
        KL divergence factor $\beta$ & 1.0 \\
        KL divergence factor $\psi_0$ ($\lambda$) & 0.01 \\
        Number of latents & 32 (Causal3DIdent), 16 (Pong) \\
        Number of epochs & 600 \\
        Target classifier weight & 2.0 \\
        Gumbel Softmax temperature & 1.0 \\
        \bottomrule
    \end{tabular}
\end{table}

\begin{table}[t!]
    \centering
    \caption{An overview of the hyperparameter used for the standard autoencoder. This parameter configuration resulted in a training time of about 2 days on one NVIDIA TitanRTX GPU for the Temporal Causal3DIdent, and 12 hours for the Interventional Pong dataset.}
    \label{tab:appendix_hyperparameter_overview_autoencoder}
    \footnotesize
    \begin{tabular}{ll}
        \toprule
        \textbf{Hyperparameter} & \textbf{Value} \\
        \midrule
        Batch size & 512 \\
        Optimizer & Adam \cite{kingma2015adam} \\
        Learning rate & 1e-3 \\
        Learning rate scheduler & Cosine Warmup (100 steps) \\
        Number of latents & 32 (Causal3DIdent), 16 (Pong) \\
        Gaussian noise std & 0.05 \\
        Number of epochs & 1000 \\
        \bottomrule
    \end{tabular}
\end{table}

\begin{table}[t!]
    \centering
    \caption{An overview of the hyperparameter used for \OurApproach{}-NF. This parameter configuration resulted in a training time of 6-10 hours on one NVIDIA TitanRTX GPU.}
    \label{tab:appendix_hyperparameter_overview_flow}
    \footnotesize
    \begin{tabular}{ll}
        \toprule
        \textbf{Hyperparameter} & \textbf{Value} \\
        \midrule
        Batch size & 1024 \\
        Optimizer & Adam \cite{kingma2015adam} \\
        Learning rate & 1e-3 \\
        Learning rate scheduler & Cosine Warmup (100 steps) \\
        KL divergence factor $\beta$ & 1.0 \\
        KL divergence factor $\psi_0$ ($\lambda$) & 0.01 (Causal3D), 0.1 (Pong) \\
        Number of latents & 32 (Causal3DIdent), 16 (Pong) \\
        Number of coupling layers & 4 (Causal3DIdent Teapot, Pong), 6 (Causal3DIdent 7-shapes) \\
        Number of epochs & 1000 \\
        Target classifier weight & 2.0 \\
        Gumbel Softmax temperature & 1.0 \\
        \bottomrule
    \end{tabular}
\end{table}

We provide an overview of the used hyperparameters in \cref{tab:appendix_hyperparameter_overview_vae} (VAE models), \cref{tab:appendix_hyperparameter_overview_autoencoder} (Autoencoder training), and \cref{tab:appendix_hyperparameter_overview_flow}.
In general, we use the Adam optimizer \cite{kingma2015adam} with a learning rate of 1e-3. 
We use a Cosine Warmup learning rate scheduler with 100 steps warmup. 
For the VAEs and autoencoder, we use the MSE loss as reconstruction loss, which corresponds to predicting a Gaussian mean per pixel with fixed standard deviation.
We found this to perform considerably better than discrete approaches such as a softmax over the 256 pixel values.
To balance the KL divergence and reconstruction loss, we have experimented with scaling the KL divergence like in the $\beta$-VAE \cite{higgins2017beta} in the range of $\beta\in\{0.25, 0.5, 1.0, 2.0, 4.0\}$.
However, the best results for all three models were achieved when using a value of $\beta=1$.
Further, we ran a grid search over the $\gamma$ hyperparameter for SlowVAE for the values $\gamma\in\{1, 2, 4, 8\}$, with the model achieving the best disentanglement results for $\gamma=1$.
Finally, we also experimented with a KL warmup, which slowly increases $\beta$ to its maximum value.
However, no noticeable improvement was found for all three models on the Temporal Causal3DIdent dataset, such that we did not include it for the final experiments.

The autoencoder is trained via the MSE reconstruction loss.
As mentioned in \cref{sec:method_normalizing_flow}, we add small Gaussian noise to the latents during training to prevent the latent space from collapsing into Dirac deltas.
We have used a noise standard deviation of $0.05$, which did not show to negatively impact the reconstruction loss.

In the normalizing flow, we use $4$ coupling layers for the Interventional Pong dataset and the Temporal Causal3DIdent Teapot dataset.
However, the more challenging dataset, Temporal Causal3DIdent 7-shapes, showed to benefit from using more coupling layers, hence using 6 here.
Note that one considerable benefit of the normalizing flow approach is that it trains much quicker than the VAE once the autoencoder has been trained.
Since one autoencoder can be used for multiple different settings, it reduced the overall computational cost and simplified the hyperparameter search.

%% file: sections/appendix_sections/3_additional_experiments.tex
\section{Additional Experiments and Results}
\label{sec:appendix_additional_experiments}

\subsection{Temporal Causal3DIdent}
\label{sec:appendix_additional_experiments_causal3d}

\subsubsection{Details of the Main Results}

\begin{table*}[t!]
    \centering
    \caption{Experimental results for the Temporal-Causal3DIdent Teapot dataset, including standard deviations over 3 seeds. See \cref{tab:experiments_results_teapot} for a detailed discussion on the table and metrics.}
    \label{tab:appendix_experiments_full_results_causal3d_teapot_indeptargets}
    \resizebox{\textwidth}{!}{%
    \begin{tabular}{lcccccccccCcccc}
        \toprule
        & \multicolumn{10}{c}{\textbf{Triplet evaluation distances} $\downarrow$} & \multicolumn{4}{c}{\textbf{Correlation metrics}}\\\cmidrule(r{4mm}){2-11}\cmidrule{12-15}
        & \texttt{pos\_x} & \texttt{pos\_y} & \texttt{pos\_z} & \texttt{rot\_}$\alpha$ & \texttt{rot\_}$\beta$ & \texttt{rot\_s} & \texttt{hue\_s} & \texttt{hue\_b} & \texttt{hue\_o} & Mean & $R^2$ diag $\uparrow$ & $R^2$ sep $\downarrow$ & Spearman diag $\uparrow$ & Spearman sep $\downarrow$\\
        \midrule
        \textbf{Oracle} & 0.02 & 0.02 & 0.02 & 0.02 & 0.03 & 0.01 & 0.02 & 0.01 & 0.02 & 0.02 & - & - & - & - \\
        \midrule
        \textbf{SlowVAE} & 0.13 & 0.10 & 0.12 & 0.50 & 0.59 & 0.22 & 0.64 & 0.21 & 0.17 & 0.30 & 0.65 & 0.20 & 0.62 & 0.27 \\
        (stds) & \footnotesize$\pm$0.002 & \footnotesize$\pm$0.004 & \footnotesize$\pm$0.000 & \footnotesize$\pm$0.009 & \footnotesize$\pm$0.009 & \footnotesize$\pm$0.005 & \footnotesize$\pm$0.004 & \footnotesize$\pm$0.006 & \footnotesize$\pm$0.003 & \footnotesize$\pm$0.003 & \footnotesize$\pm$0.007 & \footnotesize$\pm$0.007 & \footnotesize$\pm$0.005 & \footnotesize$\pm$0.012 \\[5pt]
        \textbf{\iVAEAdapt{}} & \highlight{0.04} & \highlight{0.03} & \highlight{0.04} & 0.25 & 0.31 & \highlight{0.03} & 0.58 & 0.02 & 0.05 & 0.15 & 0.78 & 0.21 & 0.77 & 0.17\\
        (stds) & \footnotesize$\pm$0.002 & \footnotesize$\pm$0.001 & \footnotesize$\pm$0.002 & \footnotesize$\pm$0.102 & \footnotesize$\pm$0.243 & \footnotesize$\pm$0.001 & \footnotesize$\pm$0.178 & \footnotesize$\pm$0.002 & \footnotesize$\pm$0.010 & \footnotesize$\pm$0.012 & \footnotesize$\pm$0.036 & \footnotesize$\pm$0.098 & \footnotesize$\pm$0.045 & \footnotesize$\pm$0.043 \\
        \midrule
        \textbf{\OurApproach-VAE} & 0.05 & 0.04 & 0.05 & 0.10 & 0.20 & \highlight{0.03} & 0.08 & 0.02 & 0.05 & 0.07 & 0.96 & 0.02 & 0.95 & \highlight{0.04} \\
        (stds) & \footnotesize$\pm$0.001 & \footnotesize$\pm$0.001 & \footnotesize$\pm$0.000 & \footnotesize$\pm$0.038 & \footnotesize$\pm$0.055 & \footnotesize$\pm$0.000 & \footnotesize$\pm$0.002 & \footnotesize$\pm$0.001 & \footnotesize$\pm$0.001 & \footnotesize$\pm$0.010 & \footnotesize$\pm$0.007 & \footnotesize$\pm$0.002 & \footnotesize$\pm$0.013 & \footnotesize$\pm$0.002 \\[5pt]
        \textbf{\OurApproach-NF} & \highlight{0.04} & \highlight{0.03} & \highlight{0.04} & \highlight{0.06} & \highlight{0.10} & \highlight{0.03} & \highlight{0.04} & \highlight{0.01} & \highlight{0.04} & \highlight{0.04} & \highlight{0.98} & \highlight{0.01} & \highlight{0.97} & 0.05 \\
        (stds) & \footnotesize$\pm$0.001 & \footnotesize$\pm$0.001 & \footnotesize$\pm$0.001 & \footnotesize$\pm$0.002 & \footnotesize$\pm$0.004 & \footnotesize$\pm$0.003 & \footnotesize$\pm$0.002 & \footnotesize$\pm$0.004 & \footnotesize$\pm$0.005 & \footnotesize$\pm$0.002 & \footnotesize$\pm$0.001 & \footnotesize$\pm$0.004 & \footnotesize$\pm$0.001 & \footnotesize$\pm$0.007 \\
        \bottomrule
    \end{tabular}%
    }
\end{table*}

\begin{table*}[t!]
    \centering
    \caption{Experimental results for the Temporal-Causal3DIdent 7 shapes dataset, including standard deviations over 3 seeds. See \cref{tab:experiments_results_all_shapes} for a detailed discussion on the table and metrics.}
    \label{tab:appendix_experiments_full_results_causal3d_7shapes}
    \resizebox{\textwidth}{!}{%
    \begin{tabular}{lccccccccccCcccc}
        \toprule
        & \multicolumn{11}{c}{\textbf{Triplet evaluation distances} $\downarrow$} & \multicolumn{4}{c}{\textbf{Correlation metrics}}\\\cmidrule(r{4mm}){2-12}\cmidrule{13-16}
        & \texttt{pos\_x} & \texttt{pos\_y} & \texttt{pos\_z} & \texttt{rot\_}$\alpha$ & \texttt{rot\_}$\beta$ & \texttt{rot\_s} & \texttt{hue\_s} & \texttt{hue\_b} & \texttt{hue\_o} & \texttt{obj\_s} & Mean & $R^2$ diag $\uparrow$ & $R^2$ sep $\downarrow$ & Spearman diag $\uparrow$ & Spearman sep $\downarrow$\\ 
        \midrule
        \textbf{Oracle} & 0.08 & 0.06 & 0.08 & 0.06 & 0.09 & 0.04 & 0.04 & 0.01 & 0.04 & 0.00 & 0.05 & - & - & - & - \\
        \midrule
        \textbf{SlowVAE} & 0.44 & 0.25 & 0.41 & 0.69 & 0.75 & 0.25 & 0.57 & 0.10 & 0.14 & 0.37 & 0.40 & 0.61 & 0.23 & 0.59 & 0.27 \\
        (stds) & \footnotesize$\pm$0.035 & \footnotesize$\pm$0.030 & \footnotesize$\pm$0.021 & \footnotesize$\pm$0.012 & \footnotesize$\pm$0.003 & \footnotesize$\pm$0.003 & \footnotesize$\pm$0.064 & \footnotesize$\pm$0.001 & \footnotesize$\pm$0.004 & \footnotesize$\pm$0.040 & \footnotesize$\pm$0.018 & \footnotesize$\pm$0.053 & \footnotesize$\pm$0.005 & \footnotesize$\pm$0.038 & \footnotesize$\pm$0.006 \\[5pt]
        \textbf{\iVAEAdapt{}} & 0.26 & 0.23 & 0.34 & 0.58 & 0.65 & 0.10 & 0.31 & 0.02 & 0.09 & 0.14 & 0.27 & 0.80 & 0.29 & 0.77 & 0.28\\
        (stds) & \footnotesize$\pm$0.013 & \footnotesize$\pm$0.038 & \footnotesize$\pm$0.015 & \footnotesize$\pm$0.051 & \footnotesize$\pm$0.043 & \footnotesize$\pm$0.011 & \footnotesize$\pm$0.003 & \footnotesize$\pm$0.000 & \footnotesize$\pm$0.001 & \footnotesize$\pm$0.039 & \footnotesize$\pm$0.002 & \footnotesize$\pm$0.002 & \footnotesize$\pm$0.013 & \footnotesize$\pm$0.001 & \footnotesize$\pm$0.019\\
        \midrule
        \textbf{\OurApproach-VAE} & 0.15 & 0.13 & 0.23 & 0.54 & 0.71 & 0.07 & 0.05 & 0.02 & 0.06 & 0.18 & 0.21 & 0.89 & 0.10 & 0.88 & 0.12\\
        (stds) & \footnotesize$\pm$0.002 & \footnotesize$\pm$0.000 & \footnotesize$\pm$0.001 & \footnotesize$\pm$0.009 & \footnotesize$\pm$0.009 & \footnotesize$\pm$0.002 & \footnotesize$\pm$0.002 & \footnotesize$\pm$0.000 & \footnotesize$\pm$0.002 & \footnotesize$\pm$0.050 & \footnotesize$\pm$0.004 & \footnotesize$\pm$0.001 & \footnotesize$\pm$0.007 & \footnotesize$\pm$0.002 & \footnotesize$\pm$0.006 \\[5pt]
        \textbf{\OurApproach-NF} & \highlight{0.12} & \highlight{0.08} & \highlight{0.11} & \highlight{0.09} & \highlight{0.14} & \highlight{0.05} & \highlight{0.05} & \highlight{0.02} & \highlight{0.06} & \highlight{0.00} & \highlight{0.07} & \highlight{0.98} & \highlight{0.04} & \highlight{0.97} & \highlight{0.08} \\
        (stds) & \footnotesize$\pm$0.001 & \footnotesize$\pm$0.000 & \footnotesize$\pm$0.001 & \footnotesize$\pm$0.000 & \footnotesize$\pm$0.004 & \footnotesize$\pm$0.001 & \footnotesize$\pm$0.001 & \footnotesize$\pm$0.004 & \footnotesize$\pm$0.002 & \footnotesize$\pm$0.000 & \footnotesize$\pm$0.001 & \footnotesize$\pm$0.000 & \footnotesize$\pm$0.003 & \footnotesize$\pm$0.000 & \footnotesize$\pm$0.007 \\
        \bottomrule
    \end{tabular}%
    }
\end{table*}

In this section, we provide additional results on the Temporal Causal3DIdent dataset. 
Firstly, we report the standard deviations over 3 seeds for the results in \cref{tab:experiments_results_teapot}, which can be found in \cref{tab:appendix_experiments_full_results_causal3d_teapot_indeptargets} and \cref{tab:appendix_experiments_full_results_causal3d_7shapes}.
Generally, \OurApproach{}-VAE and \OurApproach{}-NF showed to be stable across seeds and reach similar performance with different seeds. 

For the teapot experiments, the greatest difference is found in the modeling of the rotation angles, since the reconstruction error of a VAE can already be optimized well when modeling the teapot as a sphere.
Furthermore, the most common failure mode in the rotations is having a difference of 180 degrees, which increases the triplet distance considerably.
The reason is yet again the form of the teapot.
Rotating the teapot by 180 degrees in the $\beta$ angle switches the handle with the spout, which yet again gives a lower reconstruction error than other rotations.
\OurApproach{}-NF suffered much less from this problem since the disentanglement is performed independently of the reconstruction error, and the pretrained autoencoder precisely modeled the different rotations.

The \iVAEAdapt{}, on the other hand, showed a more instable behavior. 
Especially, the hue of the spotlight was very often entangled with different dimensions such as the hue of the object or background.
Similarly to \OurApproach{}-VAE, it struggled most with the rotation angles, which also had a high variance in the triplet evaluation.
Finally, the SlowVAE showed to be stable across seeds, but achieve lower disentanglement than the other models due to assuming independent causal factors, which was not the case here.

\begin{figure}
    \centering
    \ifarxiv
        \resizebox{0.5\textwidth}{!}{
        \footnotesize
        \begin{tabular}{cccc}
            \includegraphics[width=0.1\textwidth]{figures/causal3d_dataset/triplets/000019_img1.png} & 
            \includegraphics[width=0.1\textwidth]{figures/causal3d_dataset/triplets/000019_img2.png} & 
            \includegraphics[width=0.1\textwidth]{figures/causal3d_dataset/triplets/000019_gt.png} & 
            \includegraphics[width=0.1\textwidth]{figures/causal3d_dataset/triplets/000019_pred.png} \\
            Image 1 & Image 2 & Ground Truth & Prediction \\[7pt]
            \includegraphics[width=0.1\textwidth]{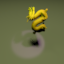} & 
            \includegraphics[width=0.1\textwidth]{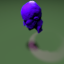} & 
            \includegraphics[width=0.1\textwidth]{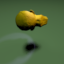} & 
            \includegraphics[width=0.1\textwidth]{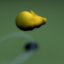} \\
            Image 1 & Image 2 & Ground Truth & Prediction \\[7pt]
            \includegraphics[width=0.1\textwidth]{figures/causal3d_dataset/triplets/000170_img1.png} & 
            \includegraphics[width=0.1\textwidth]{figures/causal3d_dataset/triplets/000170_img2.png} & 
            \includegraphics[width=0.1\textwidth]{figures/causal3d_dataset/triplets/000170_gt.png} & 
            \includegraphics[width=0.1\textwidth]{figures/causal3d_dataset/triplets/000170_pred.png} \\
            Image 1 & Image 2 & Ground Truth & Prediction \\[7pt]
            \includegraphics[width=0.1\textwidth]{figures/causal3d_dataset/triplets/000188_img1.png} & 
            \includegraphics[width=0.1\textwidth]{figures/causal3d_dataset/triplets/000188_img2.png} & 
            \includegraphics[width=0.1\textwidth]{figures/causal3d_dataset/triplets/000188_gt.png} & 
            \includegraphics[width=0.1\textwidth]{figures/causal3d_dataset/triplets/000188_pred.png} \\
            Image 1 & Image 2 & Ground Truth & Prediction \\
        \end{tabular}
        }
        \hspace{5mm}
        \resizebox{0.45\textwidth}{!}{
        \begin{tabular}{ccl}
            \toprule
            \multirow{2}{*}{\textbf{Row 1}} & Image 1 & rot\_s, obj\_s \\
            & Image 2 & pos\_o, rot\_o, hue\_o, hue\_s, hue\_b \\
            \midrule
            \multirow{2}{*}{\textbf{Row 2}} & Image 1 & rot\_o, rot\_s, hue\_o, hue\_s \\
            & Image 2 & pos\_o, rot\_o, hue\_b \\
            \midrule
            \multirow{2}{*}{\textbf{Row 3}} & Image 1 & pos\_o, rot\_s, hue\_b, obj\_s \\ %
            & Image 2 & rot\_o, hue\_o, hue\_s \\
            \midrule
            \multirow{2}{*}{\textbf{Row 4}} & Image 1 & rot\_s, hue\_o, obj\_s \\ %
            & Image 2 & pos\_o, rot\_o, hue\_b, hue\_s \\
            \bottomrule
        \end{tabular}
        }
    \else
        \footnotesize
        \begin{tabular}{cccc}
            \includegraphics[width=0.1\textwidth]{figures/causal3d_dataset/triplets/000019_img1.png} & 
            \includegraphics[width=0.1\textwidth]{figures/causal3d_dataset/triplets/000019_img2.png} & 
            \includegraphics[width=0.1\textwidth]{figures/causal3d_dataset/triplets/000019_gt.png} & 
            \includegraphics[width=0.1\textwidth]{figures/causal3d_dataset/triplets/000019_pred.png} \\
            Image 1 & Image 2 & Ground Truth & Prediction \\[7pt]
            \includegraphics[width=0.1\textwidth]{figures/causal3d_dataset/triplets/000125_img1.png} & 
            \includegraphics[width=0.1\textwidth]{figures/causal3d_dataset/triplets/000125_img2.png} & 
            \includegraphics[width=0.1\textwidth]{figures/causal3d_dataset/triplets/000125_gt.png} & 
            \includegraphics[width=0.1\textwidth]{figures/causal3d_dataset/triplets/000125_pred.png} \\
            Image 1 & Image 2 & Ground Truth & Prediction \\[7pt]
            \includegraphics[width=0.1\textwidth]{figures/causal3d_dataset/triplets/000170_img1.png} & 
            \includegraphics[width=0.1\textwidth]{figures/causal3d_dataset/triplets/000170_img2.png} & 
            \includegraphics[width=0.1\textwidth]{figures/causal3d_dataset/triplets/000170_gt.png} & 
            \includegraphics[width=0.1\textwidth]{figures/causal3d_dataset/triplets/000170_pred.png} \\
            Image 1 & Image 2 & Ground Truth & Prediction \\[7pt]
            \includegraphics[width=0.1\textwidth]{figures/causal3d_dataset/triplets/000188_img1.png} & 
            \includegraphics[width=0.1\textwidth]{figures/causal3d_dataset/triplets/000188_img2.png} & 
            \includegraphics[width=0.1\textwidth]{figures/causal3d_dataset/triplets/000188_gt.png} & 
            \includegraphics[width=0.1\textwidth]{figures/causal3d_dataset/triplets/000188_pred.png} \\
            Image 1 & Image 2 & Ground Truth & Prediction \\
        \end{tabular}
        \hspace{5mm}
        \begin{tabular}{ccl}
            \toprule
            \multirow{2}{*}{\textbf{Row 1}} & Image 1 & rot\_s, obj\_s \\
            & Image 2 & pos\_o, rot\_o, hue\_o, hue\_s, hue\_b \\
            \midrule
            \multirow{2}{*}{\textbf{Row 2}} & Image 1 & rot\_o, rot\_s, hue\_o, hue\_s \\
            & Image 2 & pos\_o, rot\_o, hue\_b \\
            \midrule
            \multirow{2}{*}{\textbf{Row 3}} & Image 1 & pos\_o, rot\_s, hue\_b, obj\_s \\ %
            & Image 2 & rot\_o, hue\_o, hue\_s \\
            \midrule
            \multirow{2}{*}{\textbf{Row 4}} & Image 1 & rot\_s, hue\_o, obj\_s \\ %
            & Image 2 & pos\_o, rot\_o, hue\_b, hue\_s \\
            \bottomrule
        \end{tabular}
    \fi
    \caption{Visualizations of triplet generations. Each row represents one example of the triplet evaluation. The table on the right shows which causal factors were combines from Image 1 and 2 respectively. The predictions were generated by \OurApproach{}-NF trained on all 7 shapes. Despite the blurriness of some predictions, the model still clearly identifies the correct causal factors.}
    \label{fig:appendix_experiments_triplet_visualizations}
\end{figure}

In the experiments on 7 shapes, the most common entanglement was the object shape with the rotation angle.
For modeling a change in the object shape independent of the rotation angle, the models need to learn the default rotation angles for each object.
Otherwise, one cannot align the different shapes correctly.
Furthermore, since having multiple object shapes considerably increased the modeling complexity, the VAE models often had blurry predictions on which the rotation could not be determined. 
Thus, \OurApproach{}-VAE was able to disentangle most factors well, except the rotation and the object shape.
However, \OurApproach{}-NF did not have this difficulty due to relying on an autoencoder, which was trained independently of any latent space prior regularization.
This is why \OurApproach{}-NF yet disentangled the different causal factors well, even the rotation.
We show some examples of the triplet generation of \OurApproach{}-NF on this dataset in \cref{fig:appendix_experiments_triplet_visualizations}.

\begin{table*}[t!]
    \centering
    \caption{Experimental results for the Temporal-Causal3DIdent 5 shapes dataset with additional testing on 2 unknown shapes (Cow and Head), including standard deviations over 3 seeds for \OurApproach{}-NF. See \cref{tab:experiments_results_causal3d_generalization} for a detailed discussion on the table and metrics. Further, we provide \iVAEAdapt{} as a baseline being trained on the same setting.}
    \label{tab:appendix_experiments_full_results_causal3d_generalization}
    \resizebox{\textwidth}{!}{%
    \begin{tabular}{lccccccccccCcccc}
        \toprule
        & \multicolumn{11}{c}{\textbf{Triplet evaluation distances} $\downarrow$} & \multicolumn{4}{c}{\textbf{Correlation metrics}}\\\cmidrule(r{4mm}){2-12}\cmidrule{13-16}
        & \texttt{pos\_x} & \texttt{pos\_y} & \texttt{pos\_z} & \texttt{rot\_}$\alpha$ & \texttt{rot\_}$\beta$ & \texttt{rot\_s} & \texttt{hue\_s} & \texttt{hue\_b} & \texttt{hue\_o} & \texttt{obj\_s} & Mean & $R^2$ diag $\uparrow$ & $R^2$ sep $\downarrow$ & Spearman diag $\uparrow$ & Spearman sep $\downarrow$\\ 
        \midrule
        \textbf{Oracle} & 0.08 & 0.06 & 0.08 & 0.06 & 0.09 & 0.04 & 0.04 & 0.01 & 0.04 & 0.00 & 0.05 & - & - & - & - \\
        \midrule
        \textbf{\OurApproach{}-NF} & & & & & & & & & & & & & \\
        5 seen shapes & 0.14 & 0.10 & 0.14 & 0.12 & 0.16 & 0.06 & 0.06 & 0.02 & 0.07 & 0.01 & 0.09 & 0.98 & 0.05 & 0.97 & 0.10 \\
        (stds) & \footnotesize$\pm$0.003 & \footnotesize$\pm$0.004 & \footnotesize$\pm$0.006 & \footnotesize$\pm$0.010 & \footnotesize$\pm$0.012 & \footnotesize$\pm$0.002 & \footnotesize$\pm$0.001 & \footnotesize$\pm$0.002 & \footnotesize$\pm$0.002 & \footnotesize$\pm$0.000 & \footnotesize$\pm$0.004 & \footnotesize$\pm$0.001 & \footnotesize$\pm$0.002 & \footnotesize$\pm$0.000 & \footnotesize$\pm$0.003 \\[5pt]
        2 unseen shapes & 0.32 & 0.26 & 0.35 & 0.36 & 0.52 & 0.12 & 0.10 & 0.03 & 0.11 & 0.10 & 0.23 & 0.94 & 0.15 & 0.93 & 0.19 \\
        (stds) & \footnotesize$\pm$0.026 & \footnotesize$\pm$0.027 & \footnotesize$\pm$0.021 & \footnotesize$\pm$0.037 & \footnotesize$\pm$0.048 & \footnotesize$\pm$0.007 & \footnotesize$\pm$0.009 & \footnotesize$\pm$0.005 & \footnotesize$\pm$0.008 & \footnotesize$\pm$0.033 & \footnotesize$\pm$0.018 & \footnotesize$\pm$0.003 & \footnotesize$\pm$0.012 & \footnotesize$\pm$0.005 & \footnotesize$\pm$0.012 \\[5pt]
        \textbf{\iVAEAdapt{}} & & & & & & & & & & & & & \\
        5 seen shapes & 0.26 & 0.18 & 0.29 & 0.66 & 0.68 & 0.13 & 0.25 & 0.03 & 0.08 & 0.15 & 0.27 & 0.80 & 0.21 & 0.81 & 0.25 \\
        2 unseen shapes & 0.37 & 0.28 & 0.70 & 0.95 & 0.94 & 0.15 & 0.24 & 0.03 & 0.09 & 0.86 & 0.46 & 0.69 & 0.19 & 0.68 & 0.25 \\
        \midrule
        \textbf{\OurApproach{}-NF} & & & & & & & & & & & & & \\
        - Cow shape & 0.32 & 0.25 & 0.36 & 0.38 & 0.53 & 0.10 & 0.09 & 0.03 & 0.10 & 0.09 & 0.22 & 0.93 & 0.14 & 0.92 & 0.22 \\
        (stds) & \footnotesize$\pm$0.030 & \footnotesize$\pm$0.024 & \footnotesize$\pm$0.017 & \footnotesize$\pm$0.035 & \footnotesize$\pm$0.049 & \footnotesize$\pm$0.002 & \footnotesize$\pm$0.006 & \footnotesize$\pm$0.004 & \footnotesize$\pm$0.004 & \footnotesize$\pm$0.013 & \footnotesize$\pm$0.015 & \footnotesize$\pm$0.004 & \footnotesize$\pm$0.010 & \footnotesize$\pm$0.004 & \footnotesize$\pm$0.009 \\[5pt]
        - Head shape & 0.32 & 0.27 & 0.35 & 0.33 & 0.51 & 0.13 & 0.11 & 0.04 & 0.12 & 0.12 & 0.23 & 0.94 & 0.15 & 0.93 & 0.16 \\
        (stds) & \footnotesize$\pm$0.023 & \footnotesize$\pm$0.029 & \footnotesize$\pm$0.026 & \footnotesize$\pm$0.038 & \footnotesize$\pm$0.047 & \footnotesize$\pm$0.012 & \footnotesize$\pm$0.011 & \footnotesize$\pm$0.005 & \footnotesize$\pm$0.012 & \footnotesize$\pm$0.052 & \footnotesize$\pm$0.021 & \footnotesize$\pm$0.001 & \footnotesize$\pm$0.014 & \footnotesize$\pm$0.006 & \footnotesize$\pm$0.015\\[5pt]
        \textbf{\iVAEAdapt{}} & & & & & & & & & & & & & \\
        - Cow shape & 0.46 & 0.27 & 0.98 & 0.97 & 0.96 & 0.14 & 0.25 & 0.03 & 0.09 & 0.90 & 0.51 & 0.69 & 0.17 & 0.68 & 0.26 \\
        - Head shape & 0.27 & 0.29 & 0.42 & 0.95 & 0.94 & 0.16 & 0.24 & 0.03 & 0.10 & 0.83 & 0.42 & 0.69 & 0.20 & 0.69 & 0.25 \\
        \bottomrule
    \end{tabular}%
    }
\end{table*}

In correspondence to the generalization experiments on the Temporal-Causal3DIdent dataset (\cref{tab:experiments_results_causal3d_generalization}), we show a detailed version of the results in \cref{tab:appendix_experiments_full_results_causal3d_generalization}.
Overall, the generalization performance shows to be stable across the 2 unseen shapes, Cow \cite{keenan2021cow} and Head \cite{rusinkiewicz2021head}.
The most difficult causal factors to generalize are the position and rotation, since both of them are heavily dependent on the object shape and hence also entangled in the autoencoder's representation.
Nonetheless, the performance well above random holds promise for future work on exploring these generalization capabilities.
Furthermore, we provide a baseline in \cref{tab:experiments_results_causal3d_generalization} by training an \iVAEAdapt{} on the same setting.
To reduce computational cost and previous experiments showing stable performance across seeds, we run the \iVAEAdapt{} for a single seed.
As one would expect, \iVAEAdapt{} heavily struggles with generalizing to unseen shapes since it's en- and decoder have not been trained on these shapes and the new shapes require non-trivial extrapolation.
Hence, the model generates arbitrary shapes for the new objects, from which no rotations can be inferred.
While the factors independent of the object shape obtain similar performance as for the known shapes, the model's overall disentanglement performance degrades considerably.
This underlines the promise of \OurApproach{}-NF to generalize to unseen instances of causal factors.

\subsubsection{Correlation Matrices}

\begin{figure}[t!]
    \centering
    \footnotesize
    \begin{tabular}{cc}
        \includegraphics[width=0.32\textwidth]{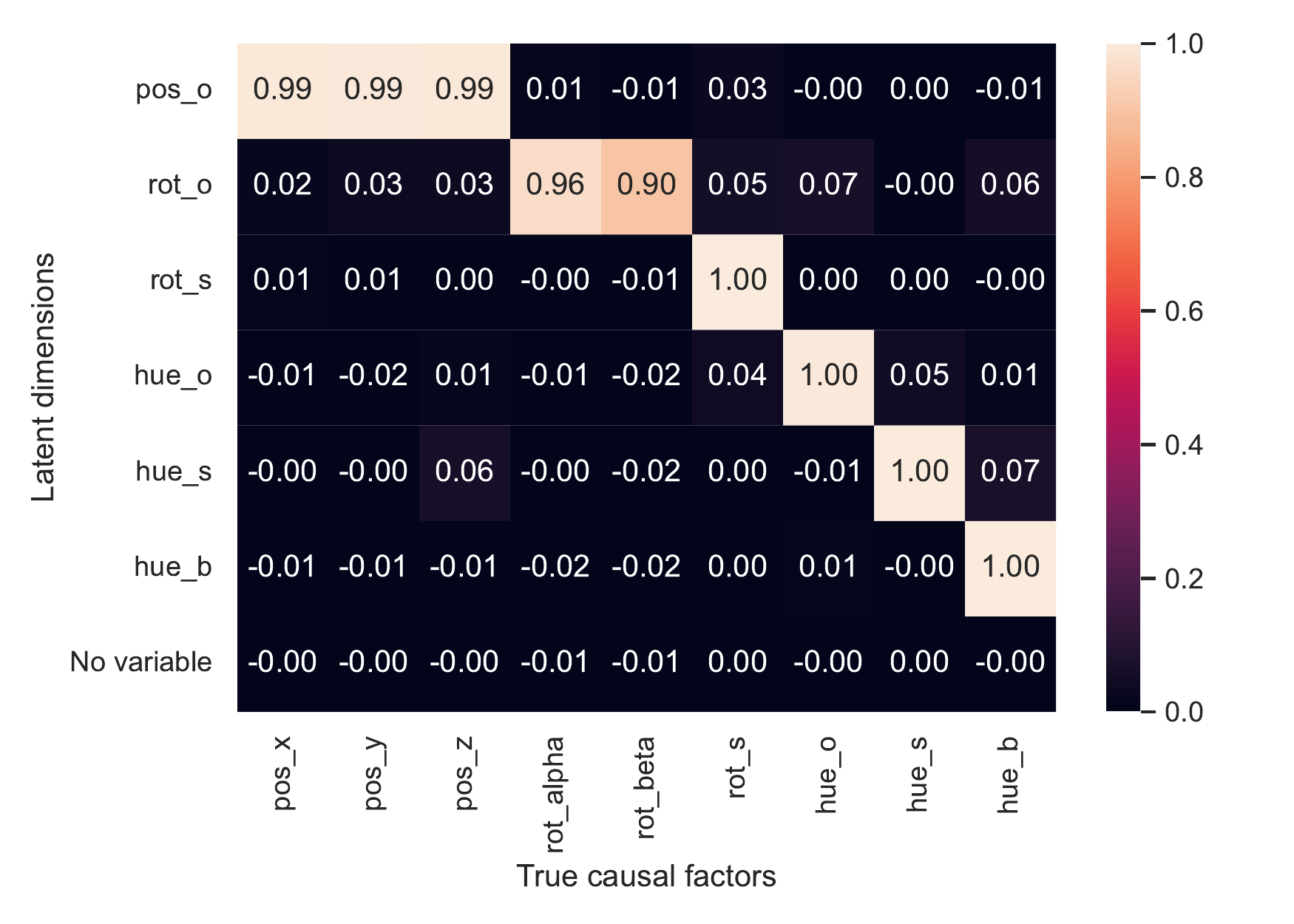} & 
        \includegraphics[width=0.32\textwidth]{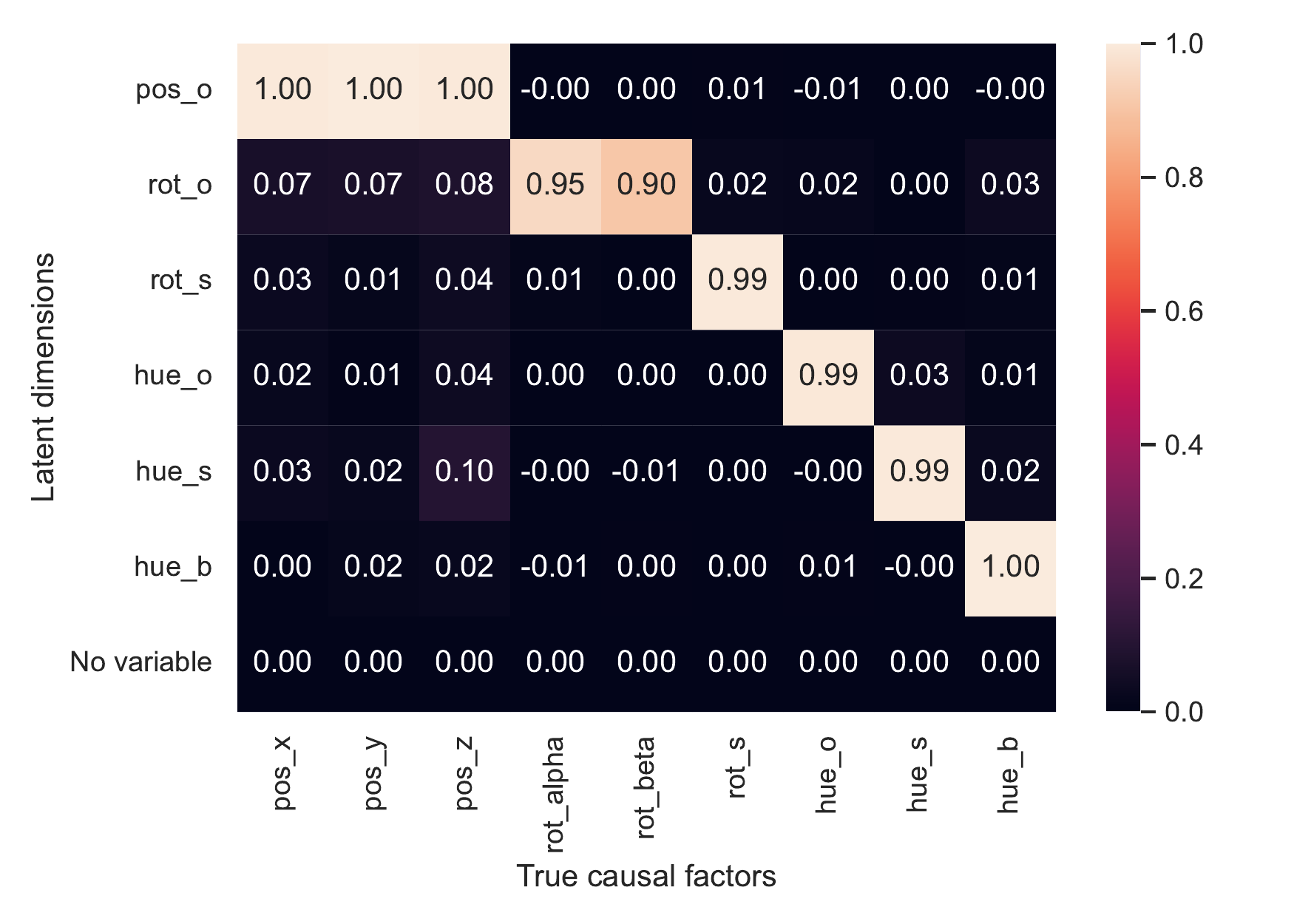} \\
        (a) \OurApproach{}-NF, $R^2$ correlation matrix & (b) \OurApproach{}-NF, Spearman correlation matrix \\[8pt]
        \includegraphics[width=0.32\textwidth]{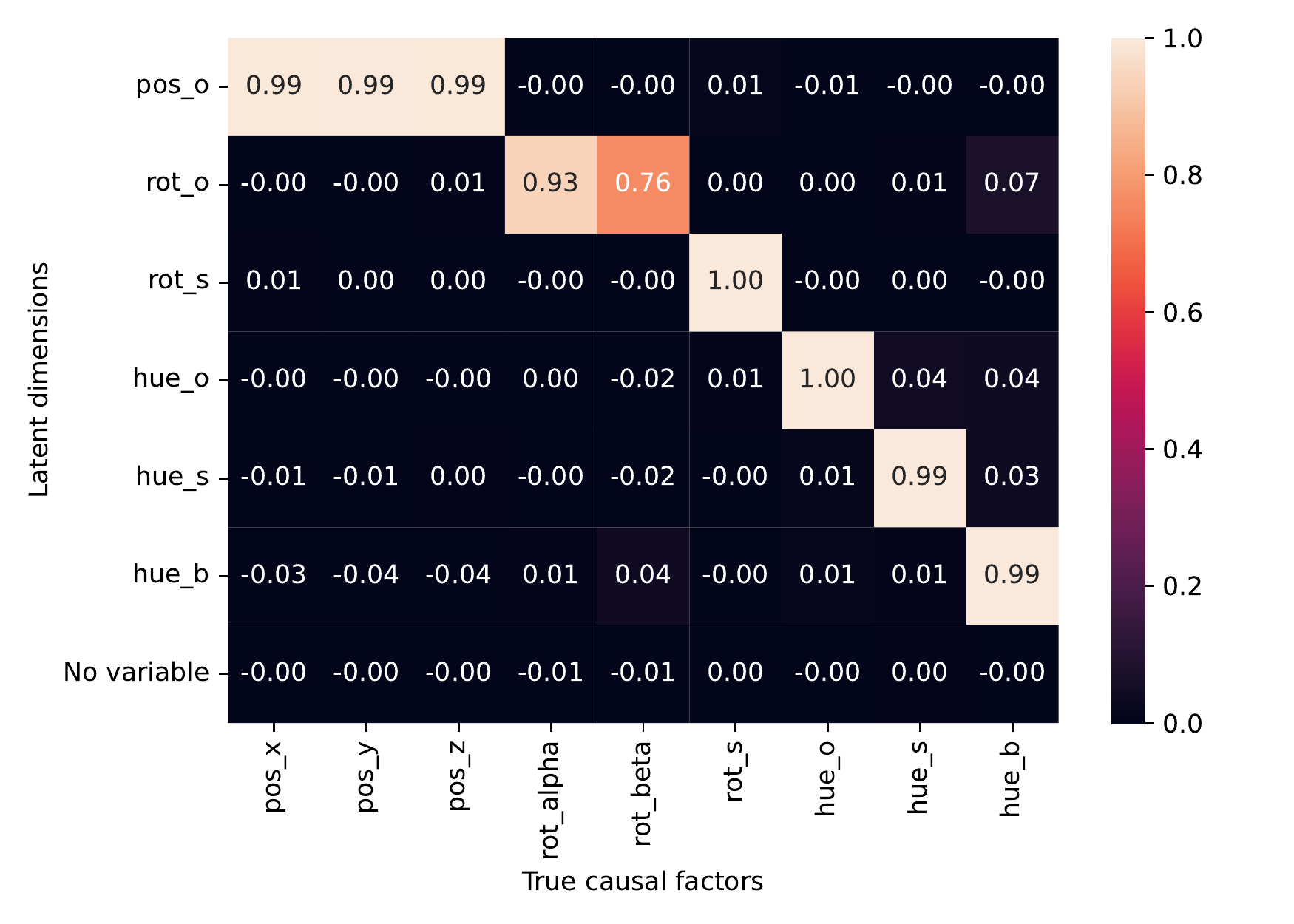} & 
        \includegraphics[width=0.32\textwidth]{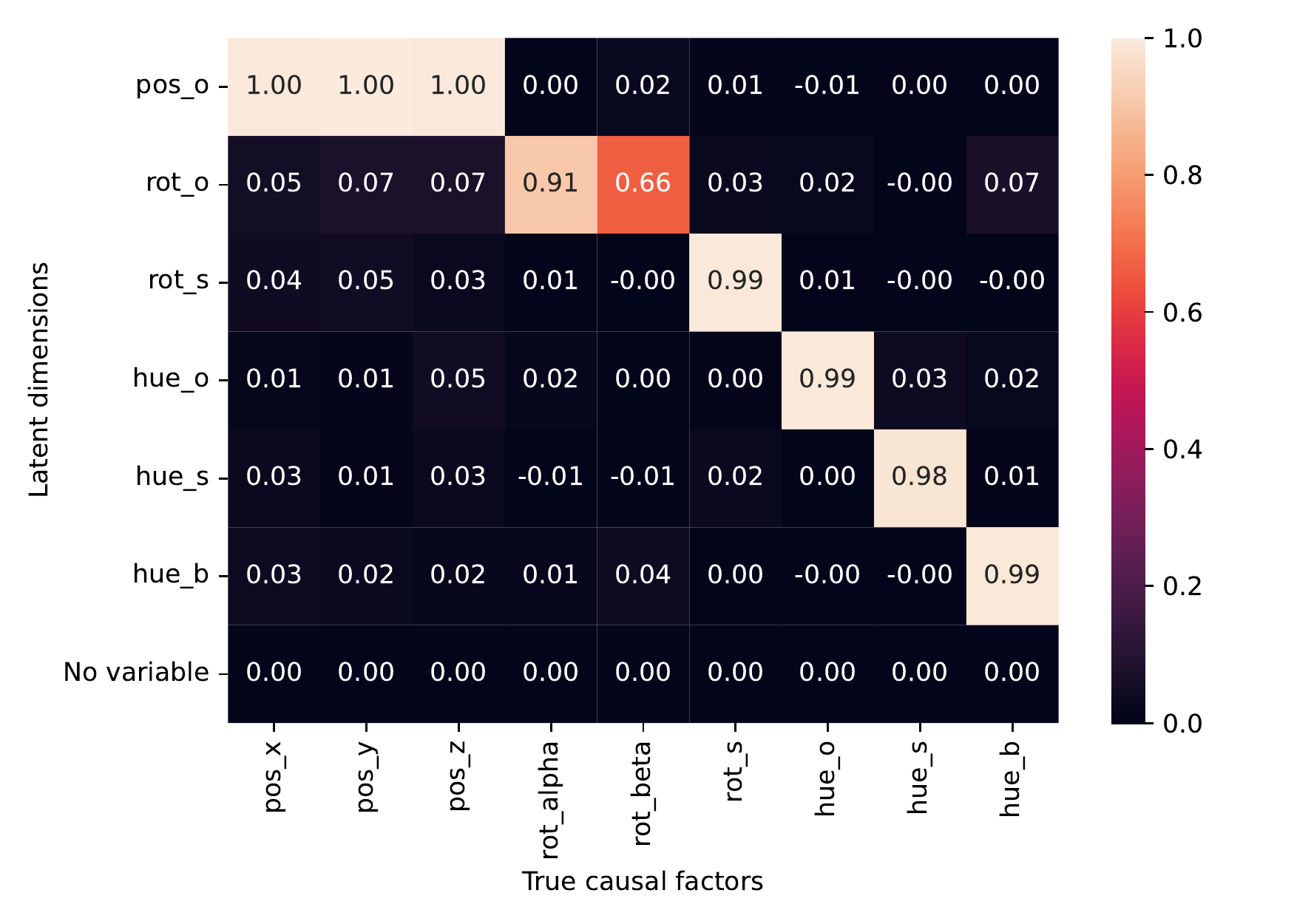} \\
        (c) \OurApproach{}-VAE, $R^2$ correlation matrix & (d) \OurApproach{}-VAE, Spearman correlation matrix \\[8pt]
        \includegraphics[width=0.30\textwidth]{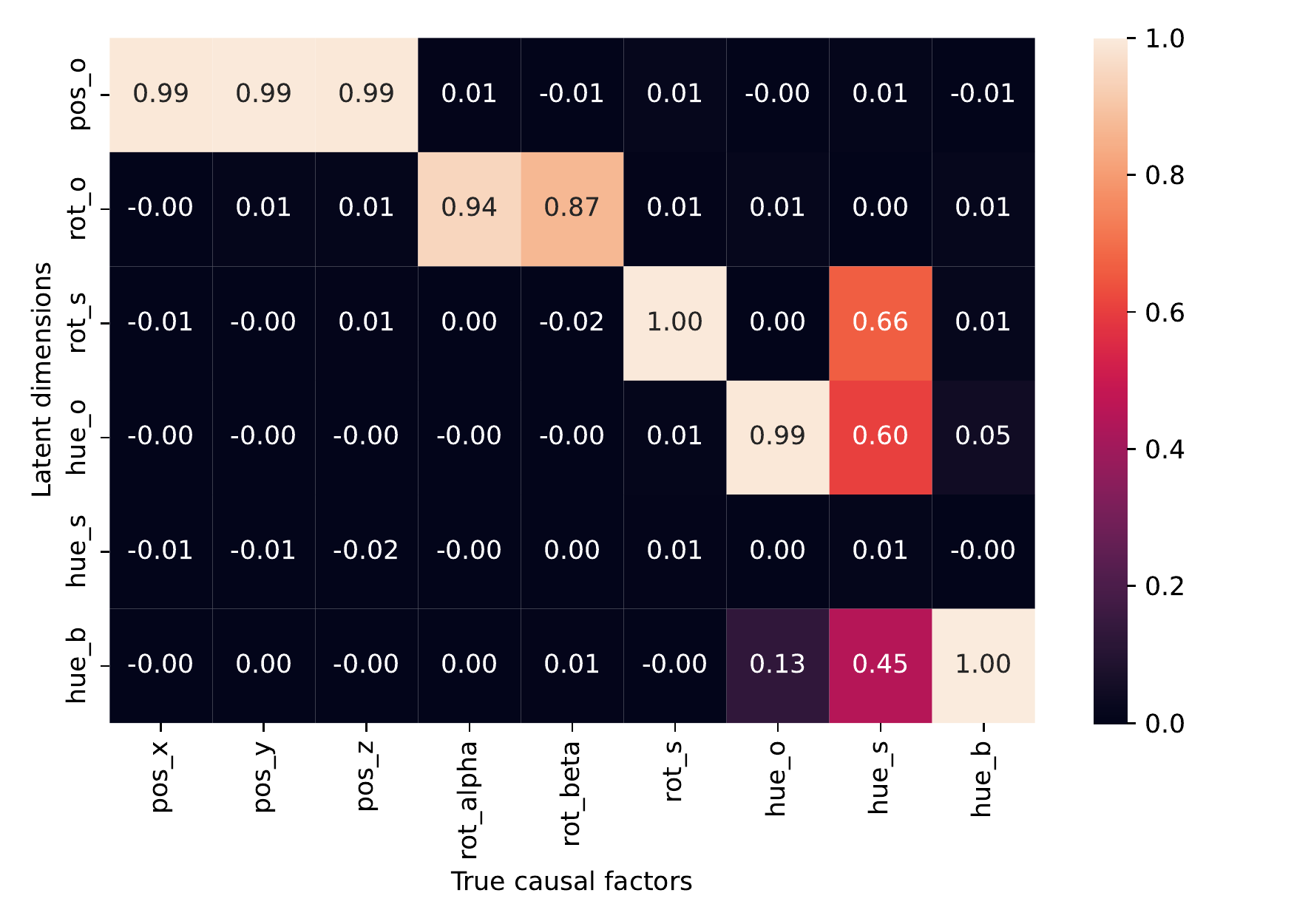} & 
        \includegraphics[width=0.30\textwidth]{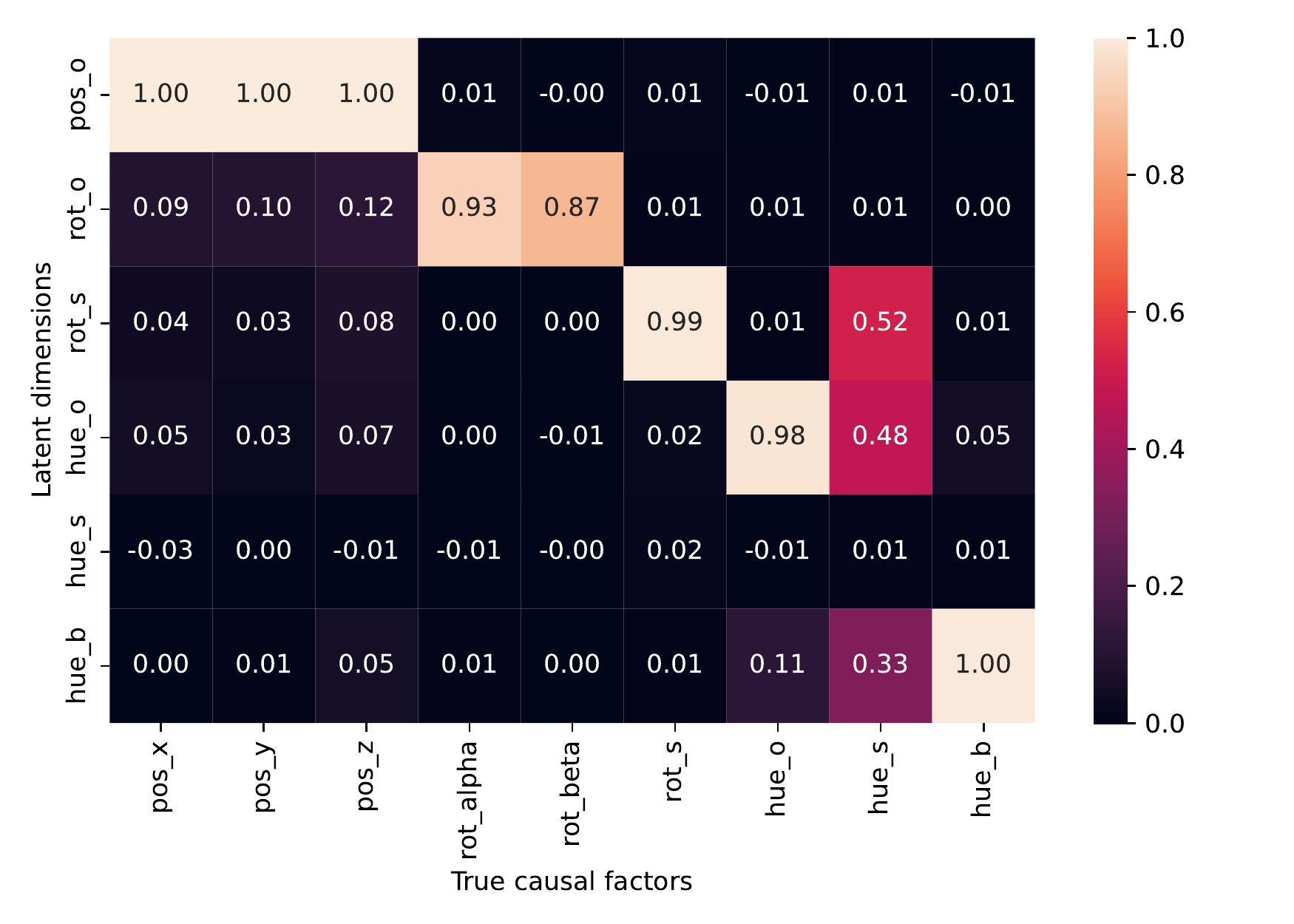} \\
        (e) \iVAEAdapt{}, $R^2$ correlation matrix & (f) \iVAEAdapt{}, Spearman correlation matrix \\[8pt]
        \includegraphics[width=0.30\textwidth]{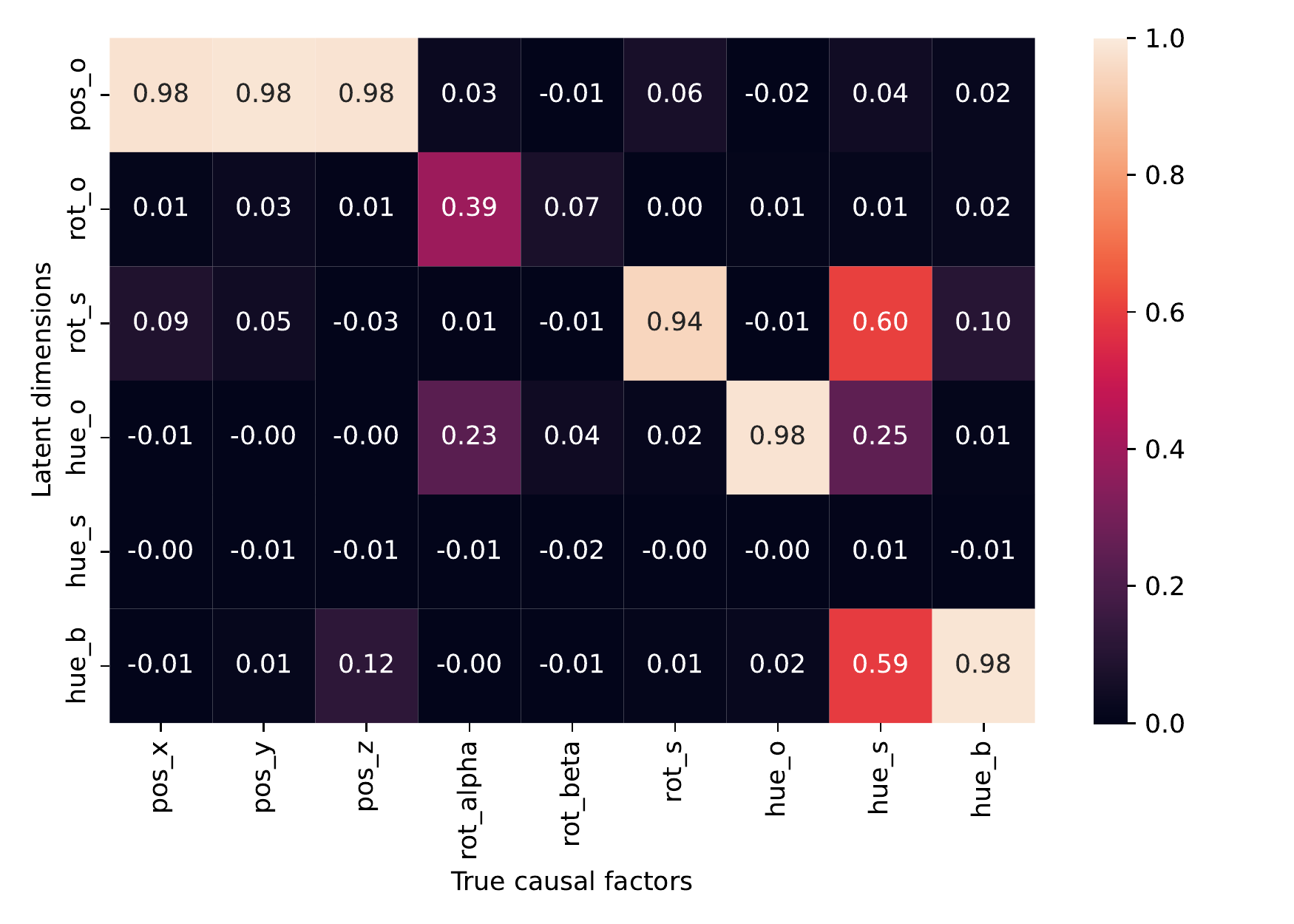} &  
        \includegraphics[width=0.30\textwidth]{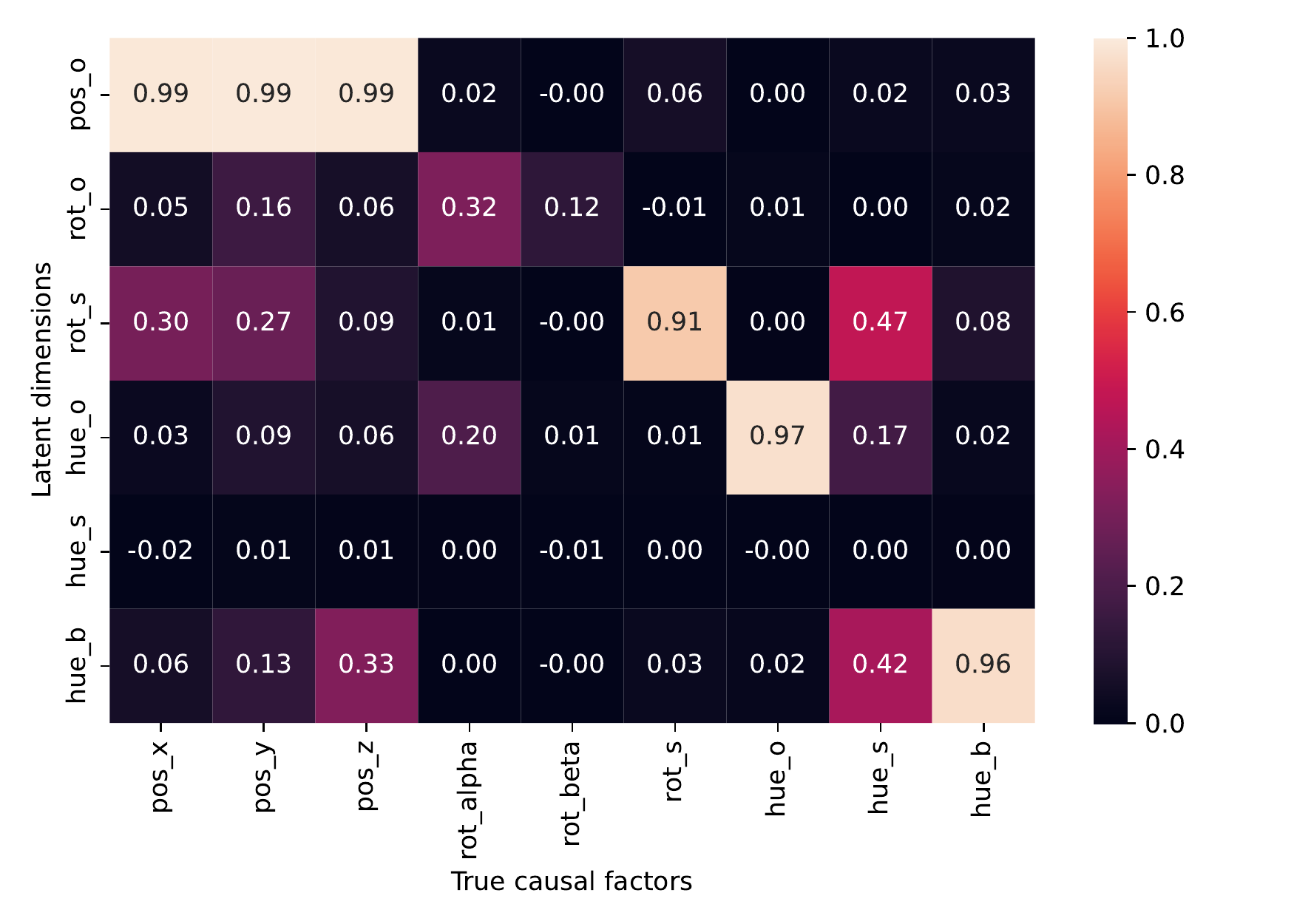} \\
        (e) SlowVAE, $R^2$ correlation matrix & (f) SlowVAE, Spearman correlation matrix \\
    \end{tabular}
    \caption{Correlation matrices for the experiments on the Temporal-Causal3DIdent Teapot dataset. The y-axis shows the sets of latent dimensions that were assigned to a certain causal factor. The set $\zpsi{0}$ is represented by 'no variable' in the plots of \OurApproach{}. The x-axis shows the ground truth causal factors with all dimensions, \ie{}, pos\_o represented by pos\_x, pos\_y, pos\_z. The heatmap is the correlation matrix between those factors ($R^2$ left, Spearman right).}
    \label{fig:appendix_additional_experiments_correlation_matrices_causal3d_teapot}
\end{figure}
\begin{figure}[t!]
    \centering
    \footnotesize
    \begin{tabular}{cc}
        \includegraphics[width=0.32\textwidth]{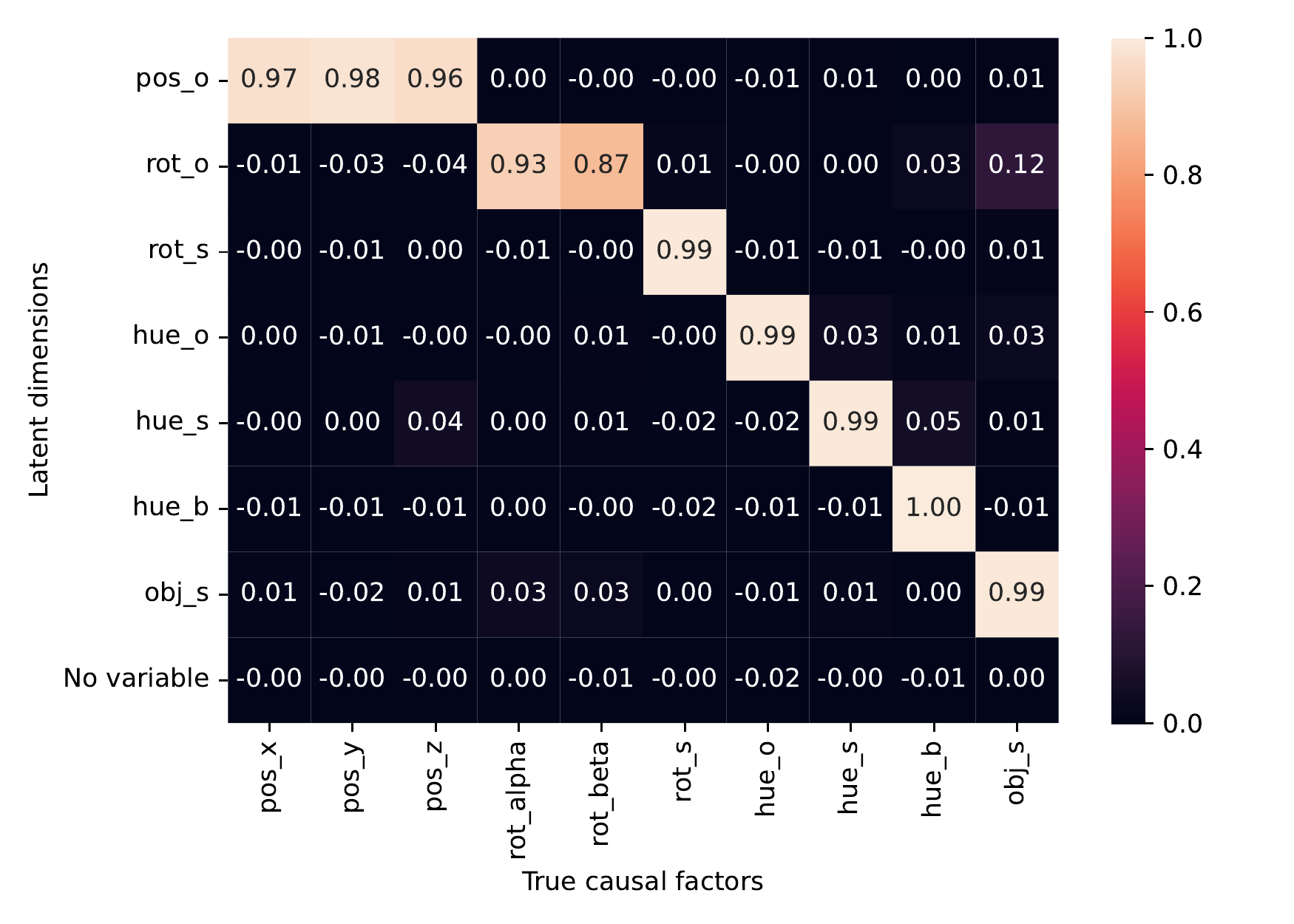} & 
        \includegraphics[width=0.32\textwidth]{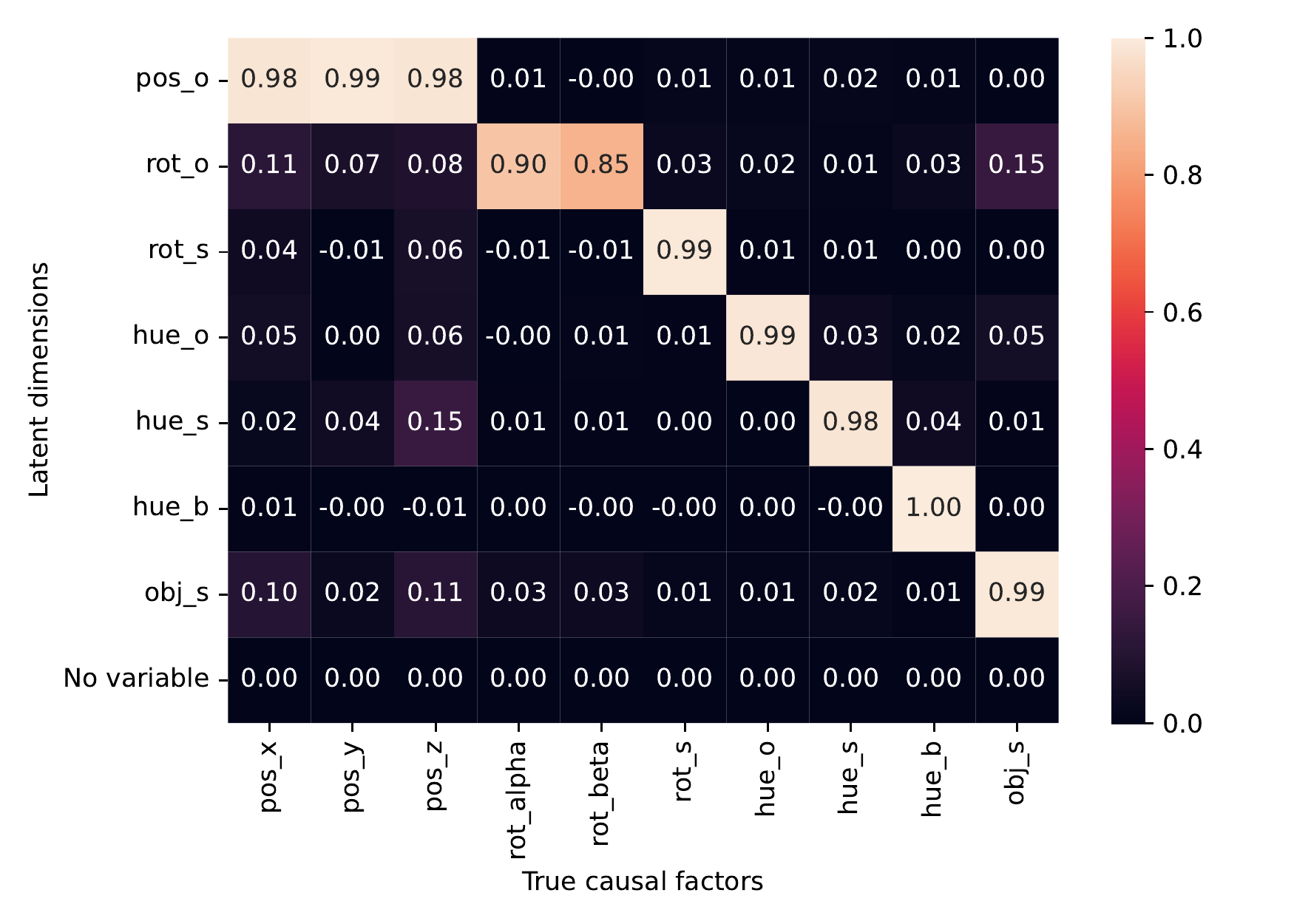} \\
        (a) \OurApproach{}-NF, $R^2$ correlation matrix & (b) \OurApproach{}-NF, Spearman correlation matrix \\[8pt]
        \includegraphics[width=0.32\textwidth]{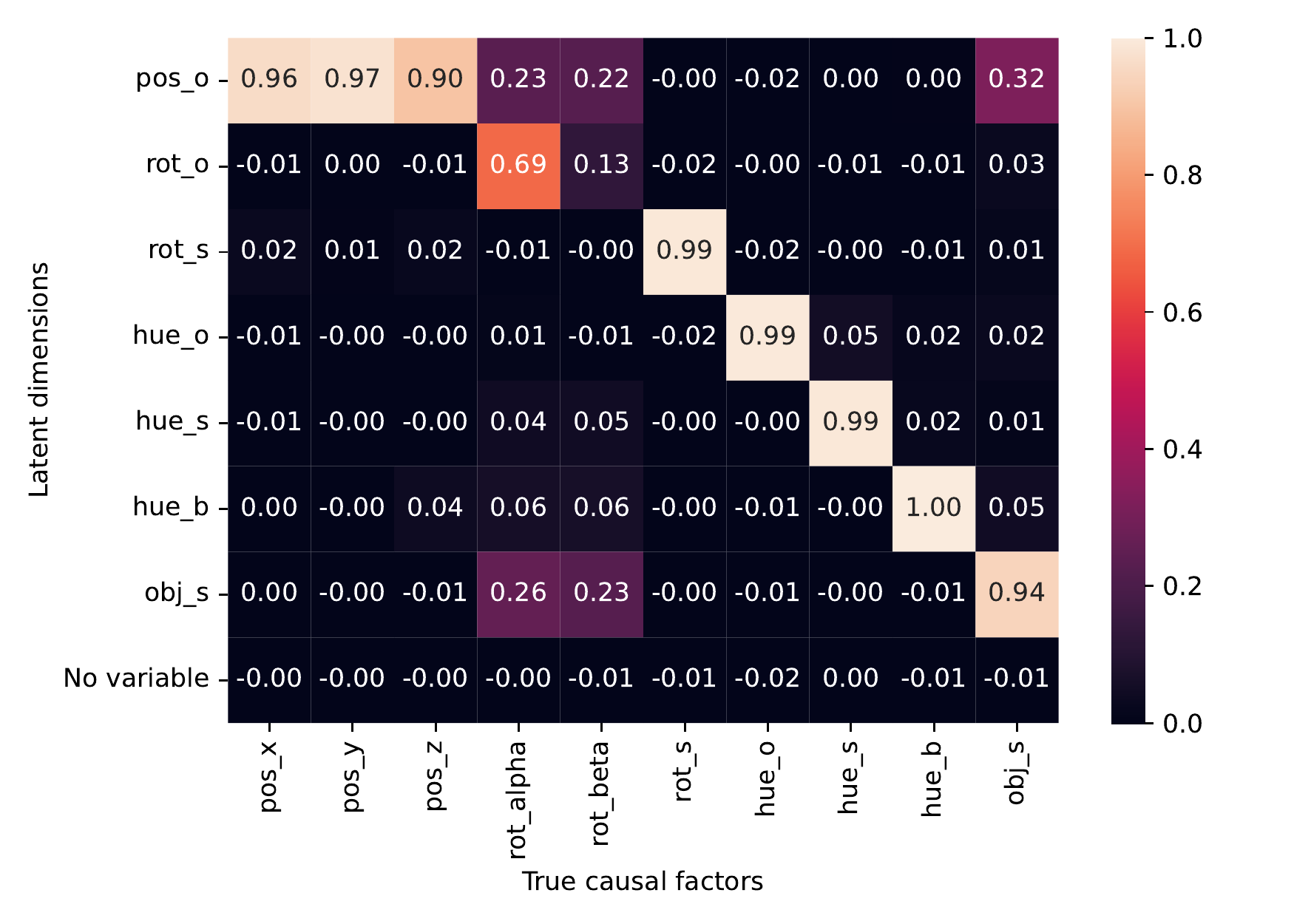} & 
        \includegraphics[width=0.32\textwidth]{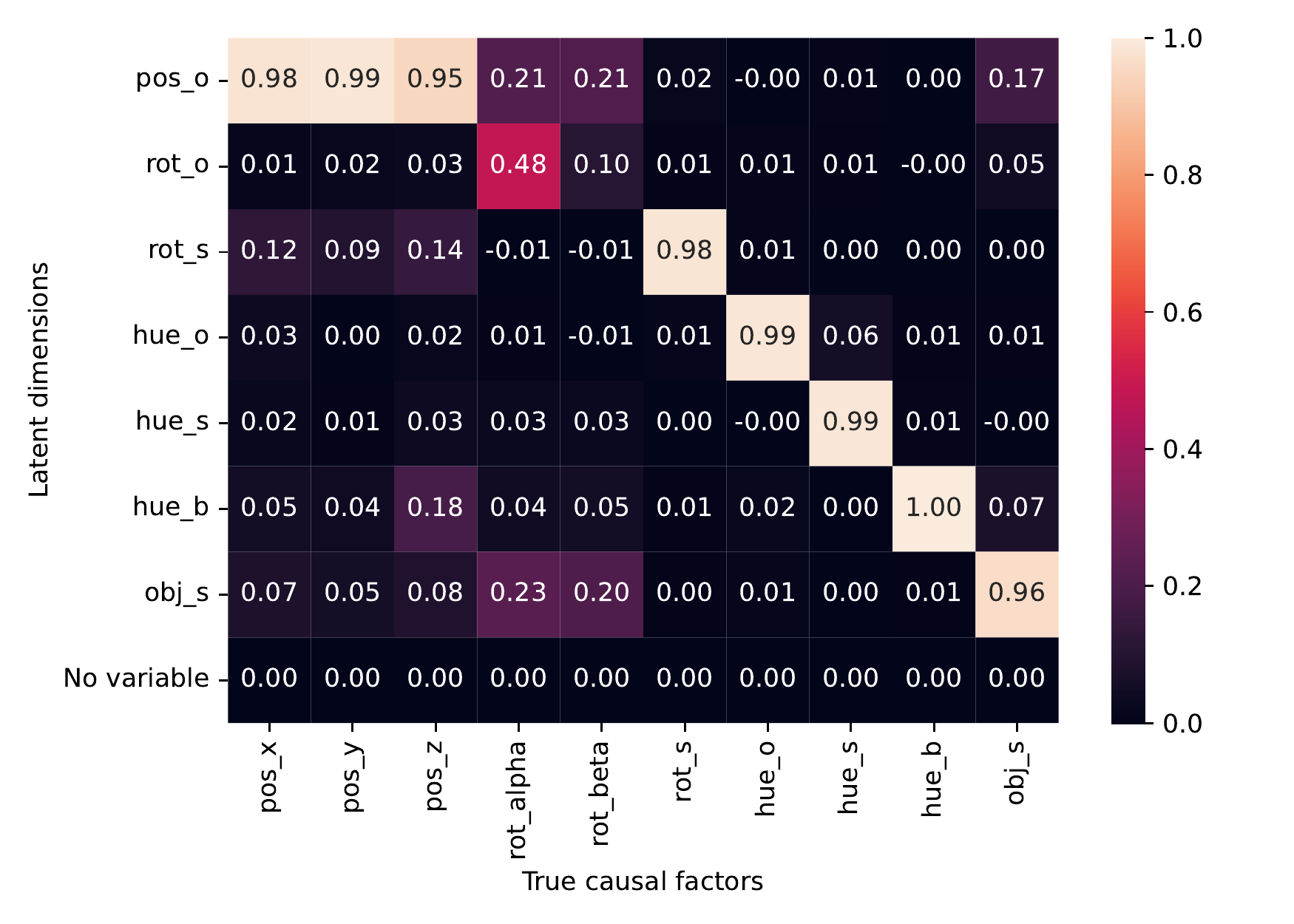} \\
        (c) \OurApproach{}-VAE, $R^2$ correlation matrix & (d) \OurApproach{}-VAE, Spearman correlation matrix \\[8pt]
        \includegraphics[width=0.30\textwidth]{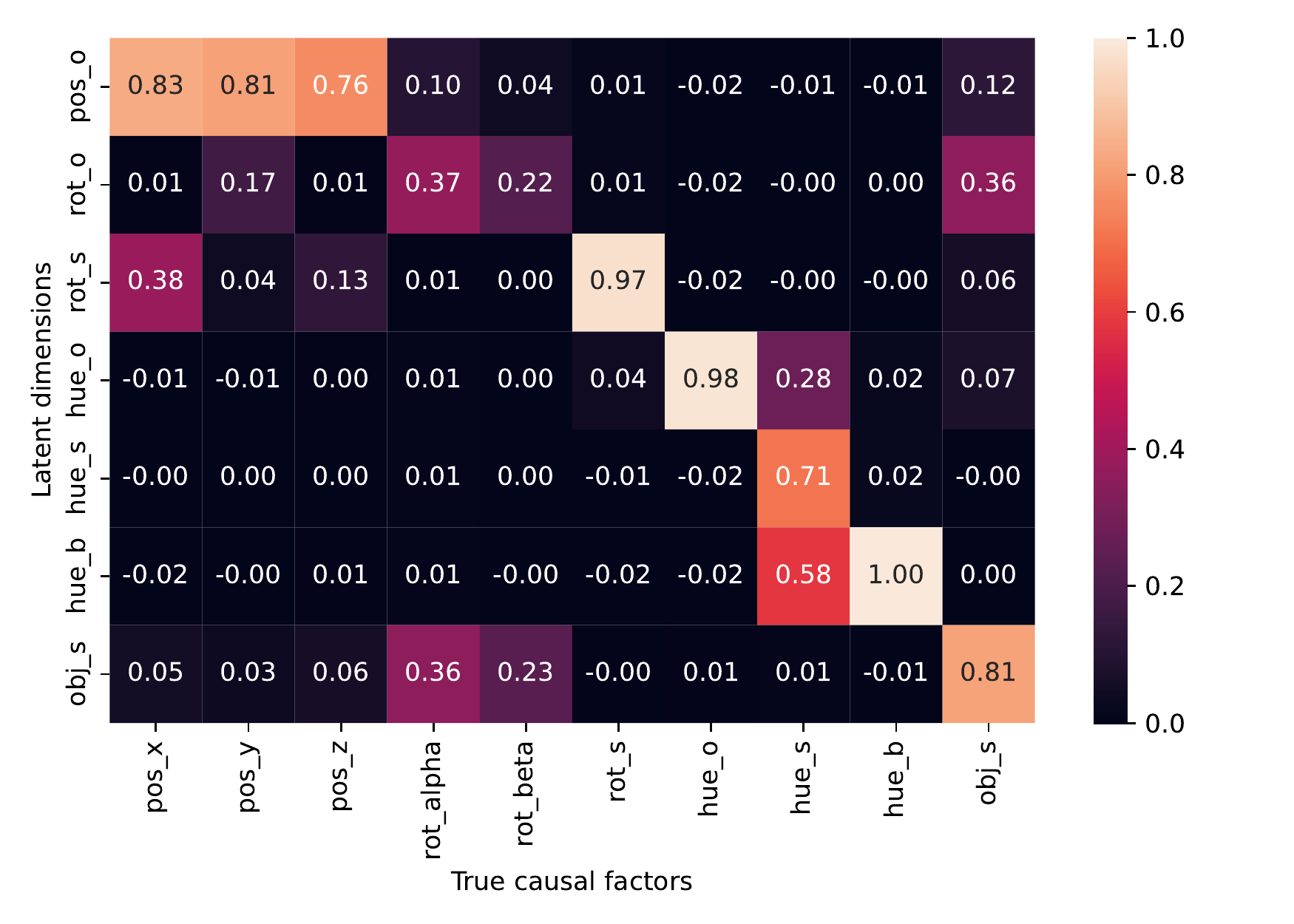} & 
        \includegraphics[width=0.30\textwidth]{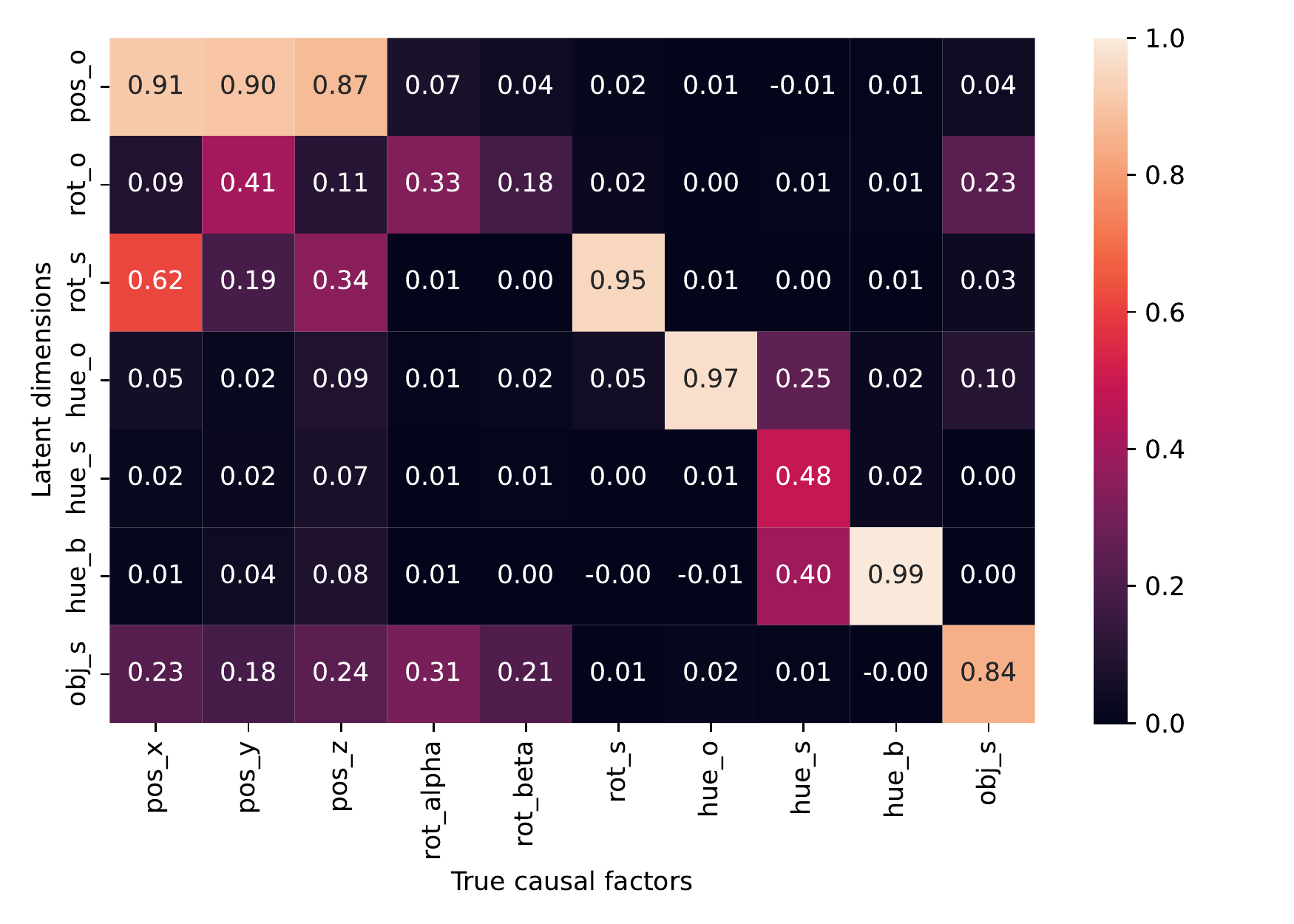} \\
        (e) \iVAEAdapt{}, $R^2$ correlation matrix & (f) \iVAEAdapt{}, Spearman correlation matrix \\[8pt]
        \includegraphics[width=0.30\textwidth]{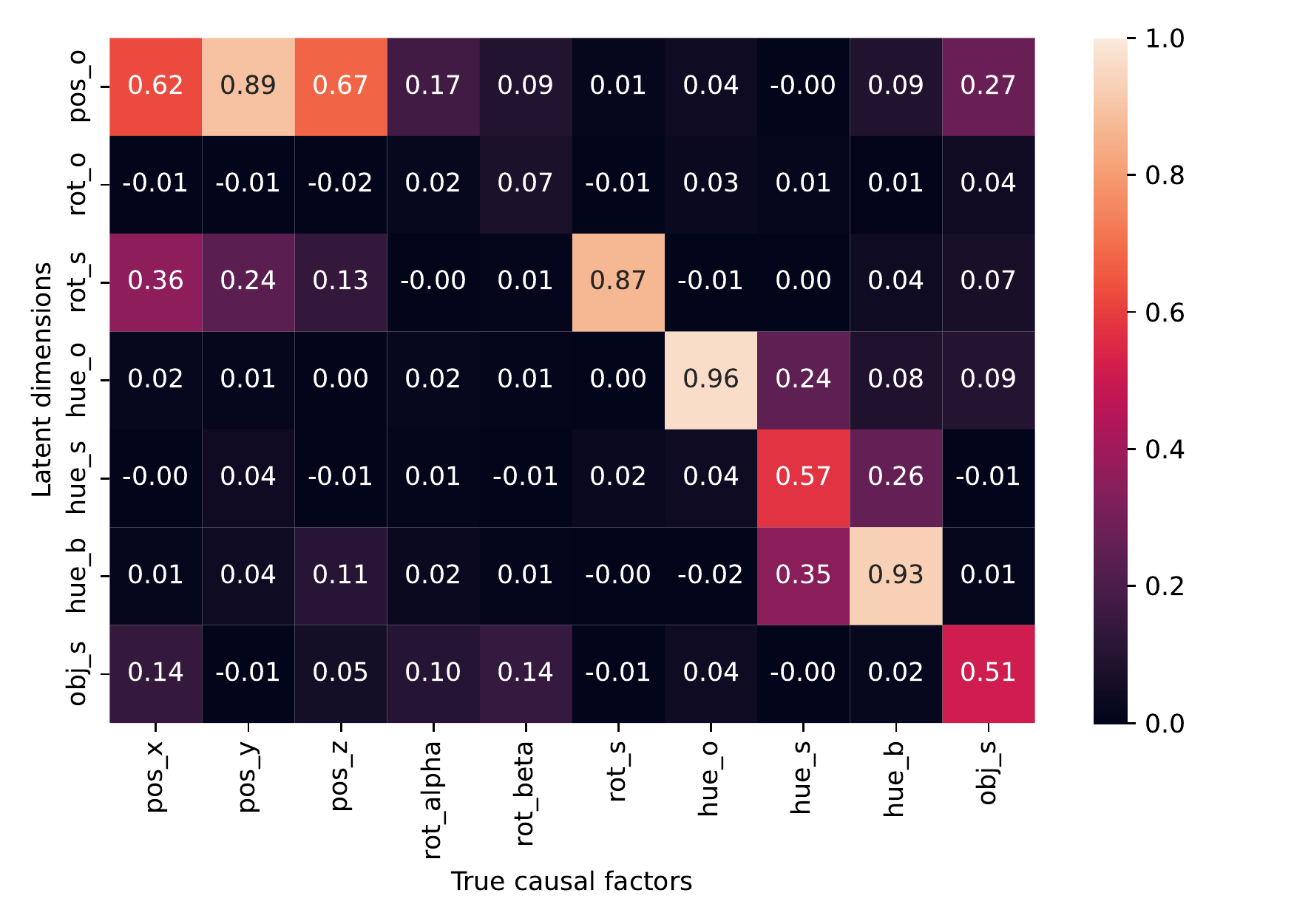} &  
        \includegraphics[width=0.30\textwidth]{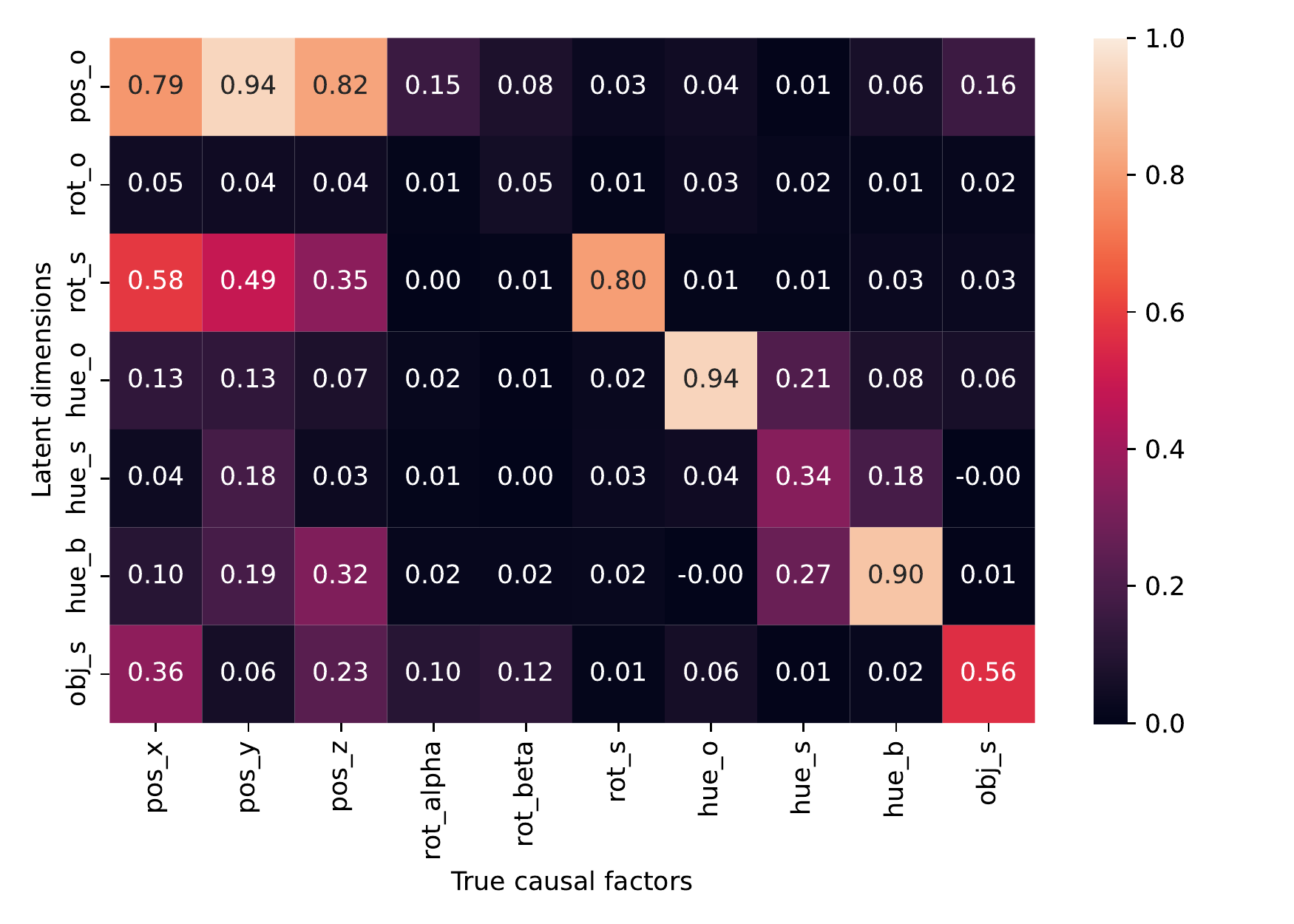} \\
        (e) SlowVAE, $R^2$ correlation matrix & (f) SlowVAE, Spearman correlation matrix \\
    \end{tabular}
    \caption{Correlation matrices for the experiments on the Temporal-Causal3DIdent 7-shapes dataset. The y-axis shows the sets of latent dimensions that were assigned to a certain causal factor. The set $\zpsi{0}$ is represented by 'no variable' in the plots of \OurApproach{}. The x-axis shows the ground truth causal factors with all dimensions, \ie{}, pos\_o represented by pos\_x, pos\_y, pos\_z. The heatmap is the correlation matrix between those factors ($R^2$ left, Spearman right).}
    \label{fig:appendix_additional_experiments_correlation_matrices_causal3d_all7}
\end{figure}

The results of \cref{tab:experiments_results_all_shapes} summarized the correlation matrices by reporting the mean on the diagonal and the maximum for any other causal factor.
In this section, we additionally show examples of full correlation matrices for all models on the Temporal Causal3DIdent dataset.
\cref{fig:appendix_additional_experiments_correlation_matrices_causal3d_teapot} shows the results for the Temporal Causal3dIdent Teapot dataset.
The x-axis shows all dimensions of all causal factors, which includes the $x$, $y$ and $z$ dimensions of the position causal factor.
The optimal model would have 1s between pos\_o and pos\_x, pos\_y, pos\_z.
The figure shows the difficulty with the rotation for all models.
Further, in the \iVAEAdapt{}, one can see the difficulty the model had with the hue of the spotlight.

The results for the 7-shapes datasets are shown in \cref{fig:appendix_additional_experiments_correlation_matrices_causal3d_all7}.
As one could already see from the results in \cref{tab:experiments_results_all_shapes}, only \OurApproach{}-NF is able to disentangle all causal factors well while the other models have especially difficulties with the rotation.

\subsubsection{Confounded Interventions}
\label{sec:appendix_additional_experiments_causal3d_confounded}

\begin{table*}[t!]
    \centering
    \caption{Experimental results for the Temporal-Causal3DIdent Teapot dataset with a limited intervention set, including standard deviations over 3 seeds. \OurApproach{} performs very similar as in the setting of independent interventions (\cref{tab:appendix_experiments_full_results_causal3d_teapot_indeptargets}), showing that it can handle such intervention sets as well.}
    \label{tab:appendix_experiments_full_results_causal3d_teapot_grouptargets}
    \resizebox{\textwidth}{!}{%
    \begin{tabular}{lcccccccccCcccc}
        \toprule
        & \multicolumn{10}{c}{\textbf{Triplet evaluation distances} $\downarrow$} & \multicolumn{4}{c}{\textbf{Correlation metrics}}\\\cmidrule(r{4mm}){2-11}\cmidrule{12-15}
        & \texttt{pos\_x} & \texttt{pos\_y} & \texttt{pos\_z} & \texttt{rot\_}$\alpha$ & \texttt{rot\_}$\beta$ & \texttt{rot\_s} & \texttt{hue\_s} & \texttt{hue\_b} & \texttt{hue\_o} & Mean & $R^2$ diag $\uparrow$ & $R^2$ sep $\downarrow$ & Spearman diag $\uparrow$ & Spearman sep $\downarrow$\\
        \midrule
        \textbf{Oracle} & 0.02 & 0.02 & 0.02 & 0.02 & 0.03 & 0.01 & 0.02 & 0.01 & 0.02 & 0.02 & - & - & - & - \\
        \midrule
        \textbf{SlowVAE} & 0.10 & 0.08 & 0.10 & 0.39 & 0.56 & 0.14 & 0.61 & 0.07 & 0.10 & 0.24 & 0.66 & 0.27 & 0.65 & 0.23\\
        (stds) & \footnotesize$\pm$0.007 & \footnotesize$\pm$0.003 & \footnotesize$\pm$0.001 & \footnotesize$\pm$0.078 & \footnotesize$\pm$0.068 & \footnotesize$\pm$0.000 & \footnotesize$\pm$0.002 & \footnotesize$\pm$0.003 & \footnotesize$\pm$0.003 & \footnotesize$\pm$0.016 & \footnotesize$\pm$0.016 & \footnotesize$\pm$0.014 & \footnotesize$\pm$0.012 & \footnotesize$\pm$0.008 \\[5pt]
        \textbf{\iVAEAdapt{}} & 0.10 & 0.07 & 0.09 & 0.23 & 0.37 & 0.06 & 0.31 & 0.02 & 0.06 & 0.15 & 0.81 & 0.18 & 0.80 & 0.18\\
        (stds) & \footnotesize$\pm$0.002 & \footnotesize$\pm$0.001 & \footnotesize$\pm$0.001 & \footnotesize$\pm$0.124 & \footnotesize$\pm$0.213 & \footnotesize$\pm$0.001 & \footnotesize$\pm$0.268 & \footnotesize$\pm$0.001 & \footnotesize$\pm$0.014 & \footnotesize$\pm$0.006 & \footnotesize$\pm$0.045 & \footnotesize$\pm$0.111 & \footnotesize$\pm$0.042 & \footnotesize$\pm$0.068 \\
        \midrule
        \textbf{\OurApproach-VAE} & \highlight{0.05} & \highlight{0.03} & 0.05 & 0.09 & 0.24 & \highlight{0.03} & 0.05 & \highlight{0.01} & 0.05 & 0.07 & 0.97 & \highlight{0.03} & 0.96 & \highlight{0.04} \\
        (stds) & \footnotesize$\pm$0.002 & \footnotesize$\pm$0.001 & \footnotesize$\pm$0.002 & \footnotesize$\pm$0.026 & \footnotesize$\pm$0.091 & \footnotesize$\pm$0.000 & \footnotesize$\pm$0.005 & \footnotesize$\pm$0.000 & \footnotesize$\pm$0.002 & \footnotesize$\pm$0.014 & \footnotesize$\pm$0.001 & \footnotesize$\pm$0.002 & \footnotesize$\pm$0.006 & \footnotesize$\pm$0.007 \\[5pt]
        \textbf{\OurApproach-NF} & \highlight{0.05} & \highlight{0.03} & \highlight{0.04} & \highlight{0.04} & \highlight{0.07} & \highlight{0.03} & \highlight{0.04} & 0.02 & \highlight{0.04} & \highlight{0.04} & \highlight{0.98} & 0.05 & \highlight{0.98} & 0.08 \\
        (stds) & \footnotesize$\pm$0.002 & \footnotesize$\pm$0.000 & \footnotesize$\pm$0.001 & \footnotesize$\pm$0.001 & \footnotesize$\pm$0.003 & \footnotesize$\pm$0.001 & \footnotesize$\pm$0.001 & \footnotesize$\pm$0.000 & \footnotesize$\pm$0.001 & \footnotesize$\pm$0.001 & \footnotesize$\pm$0.002 & \footnotesize$\pm$0.015 & \footnotesize$\pm$0.002 & \footnotesize$\pm$0.023 \\
        \bottomrule
    \end{tabular}%
    }
\end{table*}

The previous experiments on the Temporal Causal3DIdent dataset were performed on interventions that were independently sampled to show that \OurApproach{} can handle both single-target and joint interventions.
In this section, we repeat the experiments on the Teapot dataset, but with a different intervention settings.
Instead of sampling each intervention target independently, we consider 6 possible intervention sets:
\begin{enumerate}
    \item The observational regime, \ie{}, no causal variable being intervened upon
    \item A single target intervention on the object position (pos\_o)
    \item A single target intervention on the object rotation (rot\_o)
    \item A joint intervention on the rotation and hue of the spotlight (rot\_s, hue\_s)
    \item A joint intervention on the hue of the object and spotlight (hue\_o, hue\_s)
    \item A joint intervention on the hue of the object and background (hue\_o, hue\_b)
\end{enumerate}
Note that this set of interventions fulfills the intervention conditions of \cref{theo:method_intv_over_time} for all causal variables.
At each time step, we perform one of these interventions, and use the same temporal dependencies and dynamics as before.

The results over 3 seeds for these experiments are summarized in \cref{tab:appendix_experiments_full_results_causal3d_teapot_grouptargets}.
All models achieved very similar results as for the setting with independent intervention targets.
This verifies that \OurApproach{} can also disentangle the different causal factors in a setting with limited diversity in the observed intervention settings.

\subsubsection{Learning the Causal Graph}
\label{sec:appendix_additional_experiments_causal_graph}

\begin{figure*}[t!]
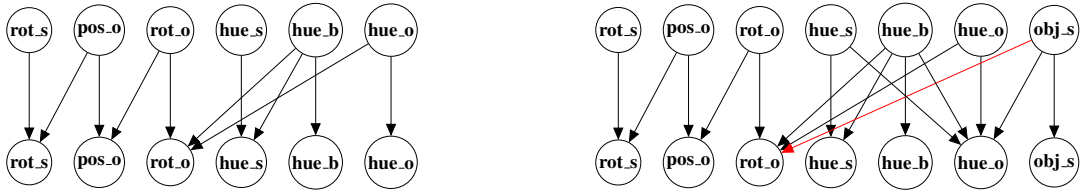

    \centering
    \begin{subfigure}{0.45\textwidth}
        \centering
        \resizebox{0.725\textwidth}{!}{
            \tikz{ %
        		\node[latent] (pos) {\textbf{pos\_o}} ; %
        		\node[latent, right=of pos, xshift=-.6cm] (rot) {\textbf{rot\_o}} ; %
        		\node[latent, left=of pos, xshift=.6cm] (rots) {\textbf{rot\_s}} ; %
        		\node[latent, right=of rot, xshift=-.6cm] (hues) {\textbf{hue\_s}} ; %
        		\node[latent, right=of hues, xshift=-.6cm] (hueb) {\textbf{hue\_b}} ; %
        		\node[latent, right=of hueb, xshift=-.6cm] (hueo) {\textbf{hue\_o}} ; %
        		
        		\node[latent, below=of pos, yshift=-.5cm] (pos1) {\textbf{pos\_o}} ; %
        		\node[latent, right=of pos1, xshift=-.6cm] (rot1) {\textbf{rot\_o}} ; %
        		\node[latent, left=of pos1, xshift=.6cm] (rots1) {\textbf{rot\_s}} ; %
        		\node[latent, right=of rot1, xshift=-.6cm] (hues1) {\textbf{hue\_s}} ; %
        		\node[latent, right=of hues1, xshift=-.6cm] (hueb1) {\textbf{hue\_b}} ; %
        		\node[latent, right=of hueb1, xshift=-.6cm] (hueo1) {\textbf{hue\_o}} ; %
        		
        		\edge{pos}{pos1} ;
        		\edge{rot}{rot1} ;
        		\edge{rots}{rots1} ;
        		\edge{hues}{hues1} ;
        		\edge{hueb}{hueb1} ;
        		\edge{hueo}{hueo1} ;
        		
        		\edge{rot}{pos1} ;
        		\edge{hueb}{rot1} ;
        		\edge{hueo}{rot1} ;
        		\edge{hueb}{hues1} ;
        		\edge{pos}{rots1} ;
        	}
        }
    \end{subfigure}
    \hspace{12pt}
    \begin{subfigure}{0.45\textwidth}
        \centering
        \resizebox{0.85\textwidth}{!}{
            \tikz{ %
        		\node[latent] (pos) {\textbf{pos\_o}} ; %
        		\node[latent, right=of pos, xshift=-.6cm] (rot) {\textbf{rot\_o}} ; %
        		\node[latent, left=of pos, xshift=.6cm] (rots) {\textbf{rot\_s}} ; %
        		\node[latent, right=of rot, xshift=-.6cm] (hues) {\textbf{hue\_s}} ; %
        		\node[latent, right=of hues, xshift=-.6cm] (hueb) {\textbf{hue\_b}} ; %
        		\node[latent, right=of hueb, xshift=-.6cm] (hueo) {\textbf{hue\_o}} ; %
        		\node[latent, right=of hueo, xshift=-.6cm] (objs) {\textbf{obj\_s}} ; %
        		
        		\node[latent, below=of pos, yshift=-.5cm] (pos1) {\textbf{pos\_o}} ; %
        		\node[latent, right=of pos1, xshift=-.6cm] (rot1) {\textbf{rot\_o}} ; %
        		\node[latent, left=of pos1, xshift=.6cm] (rots1) {\textbf{rot\_s}} ; %
        		\node[latent, right=of rot1, xshift=-.6cm] (hues1) {\textbf{hue\_s}} ; %
        		\node[latent, right=of hues1, xshift=-.6cm] (hueb1) {\textbf{hue\_b}} ; %
        		\node[latent, right=of hueb1, xshift=-.6cm] (hueo1) {\textbf{hue\_o}} ; %
        		\node[latent, right=of hueo1, xshift=-.6cm] (objs1) {\textbf{obj\_s}} ; %
        		
        		\edge{pos}{pos1} ;
        		\edge{rot}{rot1} ;
        		\edge{rots}{rots1} ;
        		\edge{hues}{hues1} ;
        		\edge{hueb}{hueb1} ;
        		\edge{hueo}{hueo1} ;
        		\edge{objs}{objs1} ;
        		
        		\edge{rot}{pos1} ;
        		\edge{hueb}{rot1} ;
        		\edge{hueo}{rot1} ;
        		\edge{hueb}{hues1} ;
        		\edge{objs}{hueo1} ;
        		\edge[red]{objs}{rot1} ;
        		\edge{hues}{hueo1} ;
        		\edge{hueb}{hueo1} ;
        		\edge{pos}{rots1} ;
        	}
        }
    \end{subfigure}
    \vspace{-2mm}
    \caption{The learned graphs of \OurApproach{}-NF from the Temporal-Causal3DIdent dataset. False positive edges are colored red. \textbf{Left}: Temporal-Causal3DIdent Teapot. The learned causal graph is identical to the ground truth graph. Since we have a single object shape, we do not include obj\_s as a causal variable in the graph. Additionally, the parent set of hue\_o is reduced due to the constant object shape. \textbf{Right}: Temporal-Causal3DIdent 7-shapes. The edge obj\_s$\to$rot\_o is predicted due to disentangling these two variables being the hardest task of the dataset, and small correlations remain.}
    \label{fig:appendix_additional_experiments_causal3d_graph}
\end{figure*}

The transition prior $p_{\phi}$ of \OurApproach{} takes as input all latent variables from the previous time-step, $z^t$, in order to model a distribution over $z^{t+1}$.
Once the model has converged, we can try to learn a sparser prior, which removes unnecessary edges between sets of latent variables that have been assigned to different causal factors.
More specifically, we are trying to find the causal graph between $z_{\Psi_0}^t, z_{\Psi_1}^t, ...,z_{\Psi_K}^t$ and $z_{\Psi_0}^{t+1}, z_{\Psi_1}^{t+1}, ...,z_{\Psi_K}^{t+1}$, where the directions of the directions of the edges are determined by $t\to t+1$, and no edges within a time step exist.
We find this graph by using ENCO \cite{lippe2022enco}, a continuous-optimization causal discovery method which supports the usage of arbitrary neural networks.
The causal graph is parameterized by a weight matrix $\bm{\gamma}\in\mathbb{R}^{K+1\times K+1}$, where $\sigmoid(\gamma_{ij})$ represents the probability of having the edge $z_{\Psi_i}^t\to z_{\Psi_j}^{t+1}$ in the causal graph.
To estimate the conditional likelihoods under different causal graphs, we learn a new autoregressive prior $p_{\varepsilon}(z^{t+1}_{\Psi_i}|z^T\cdot M, I_i^{t+1})$ where $M$ is mask on the latents at time-step $t$ according to the causal graph: $M_{j}=\hat{M}_{\psi(j)}, \hat{M}_{k}\sim\text{Bernoulli}(\gamma_{ki})$.
In other words, we sample a causal graph, and then mask a set of latents $z^t_{\Psi_j}$ if the causal variable $C_j$ does not have an edge to $C_i$: $z_{\Psi_j}^t\not\to z_{\Psi_i}^{t+1}$.
With this prior, we can then determine the gradients for $\bm{\gamma}$ on a held-out test set following ENCO's gradient updates.
We use a sparsity regularizer of $\lambda=0.05$ and train for 100 epochs.

We perform the experiments on two learned \OurApproach{}-NF, one for Temporal-Causal3DIdent Teapot, and one for Temporal-Causal3DIdent 7-shapes.
The learned graphs are shown in \cref{fig:appendix_additional_experiments_causal3d_graph}.
For the teapot experiments, the graph is identical to the ground truth, showing that \OurApproach{} has indeed learned the causal variables.
For the 7 shapes experiment, the model has one false positive prediction, the edge obj\_s$\to$rot\_o.
This fits our previous discussion on the results of the 7 shape experiments (\cref{tab:appendix_experiments_full_results_causal3d_7shapes}) since the object shape and rotation was the most difficult to disentangle for all models.
Hence, there remains some correlation between the object shape and rotation in the \OurApproach{}-NF model.

\subsubsection{Interventions on a Subset of Variables}
\label{sec:appendix_additional_experiments_causal3d_novars}

\begin{figure}[t!]
    \centering
    \footnotesize
    \begin{tabular}{cc}
        \includegraphics[width=0.4\textwidth]{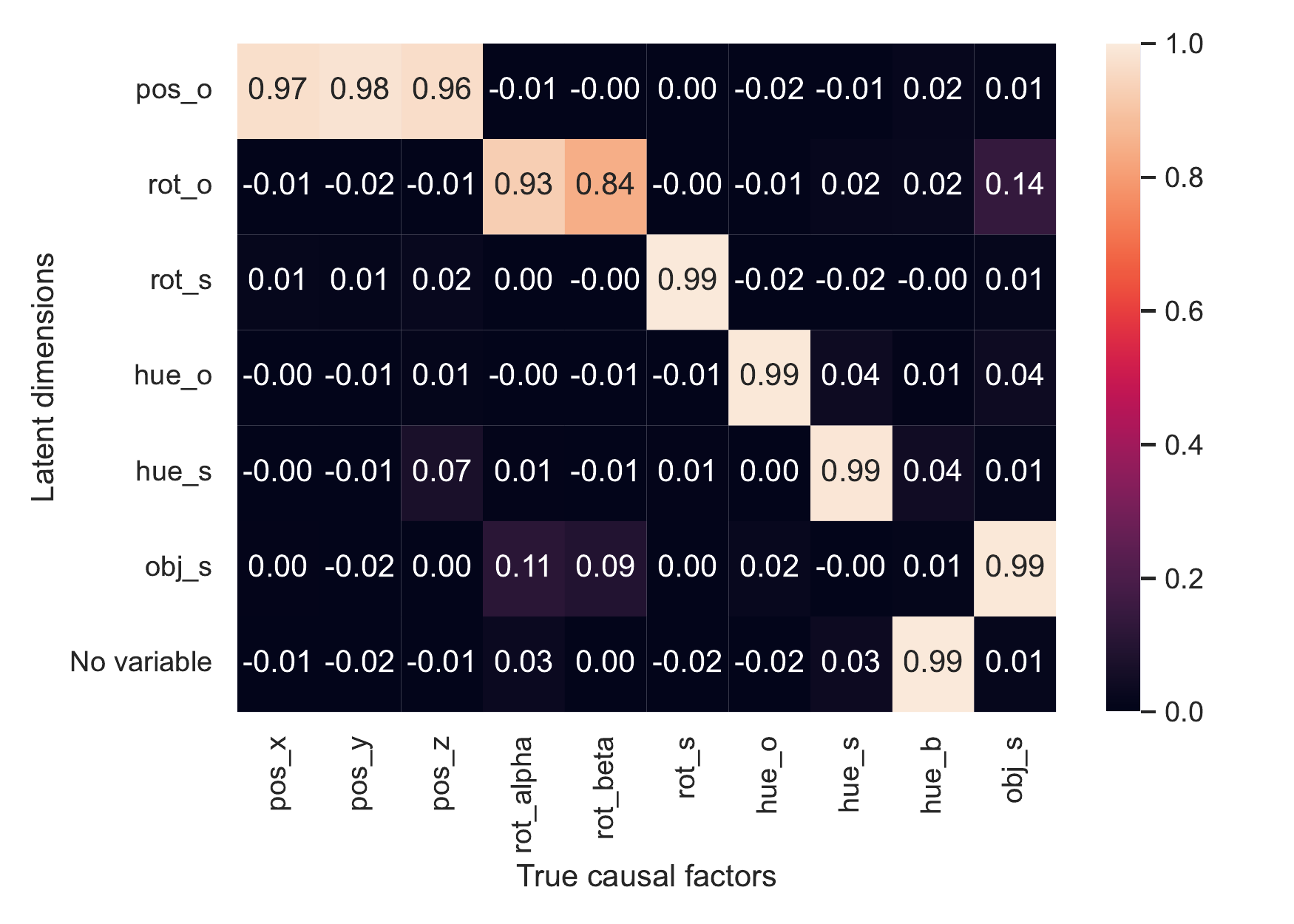} & 
        \includegraphics[width=0.4\textwidth]{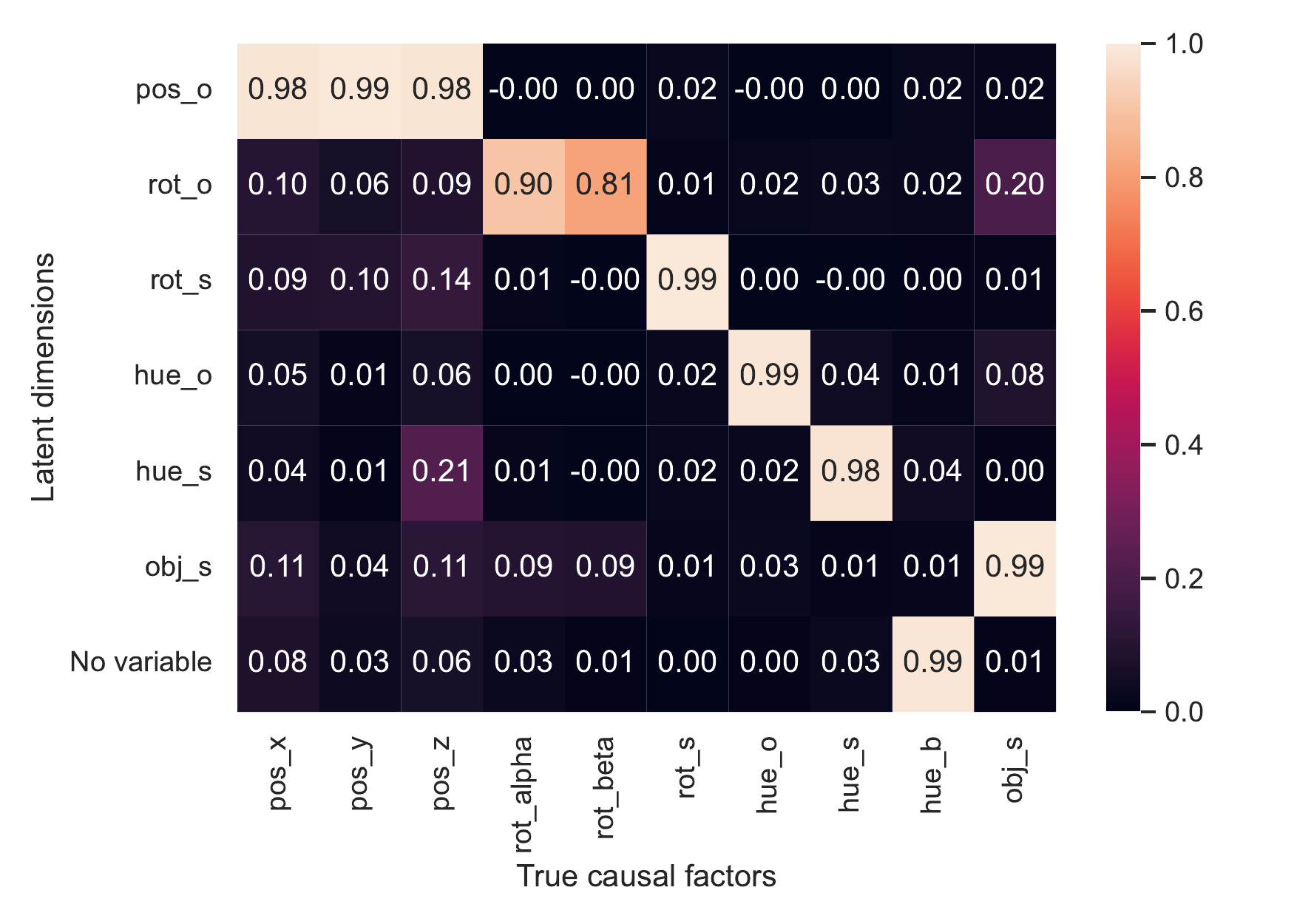} \\
        (a) \OurApproach{}-NF, $R^2$ correlation matrix & (b) \OurApproach{}-NF, Spearman correlation matrix \\
    \end{tabular}
    \caption{Correlation matrices for the experiments on the Temporal-Causal3DIdent 7-shapes dataset where no intervention targets for the causal variable hue-background are shown (\textbf{left}: $R^2$ correlation, \textbf{right}: Spearman correlation). The set $\zpsi{0}$ is represented by 'no variable' in the plots of \OurApproach{}. As intended, \OurApproach{} learns to map the background hue to $\zpsi{0}$, i.e. the latents that represent all information that does not belong to the causal variables for which we are given intervention targets. The remaining causal variables are disentangled as well as when interventions were provided for all variables (compare to \cref{fig:appendix_additional_experiments_correlation_matrices_causal3d_all7}).}
    \label{fig:appendix_additional_experiments_correlation_matrices_causal3d_novars_all7}
\end{figure}
\begin{figure}[t!]
    \centering
    \footnotesize
    \begin{tabular}{cc}
        \includegraphics[width=0.4\textwidth]{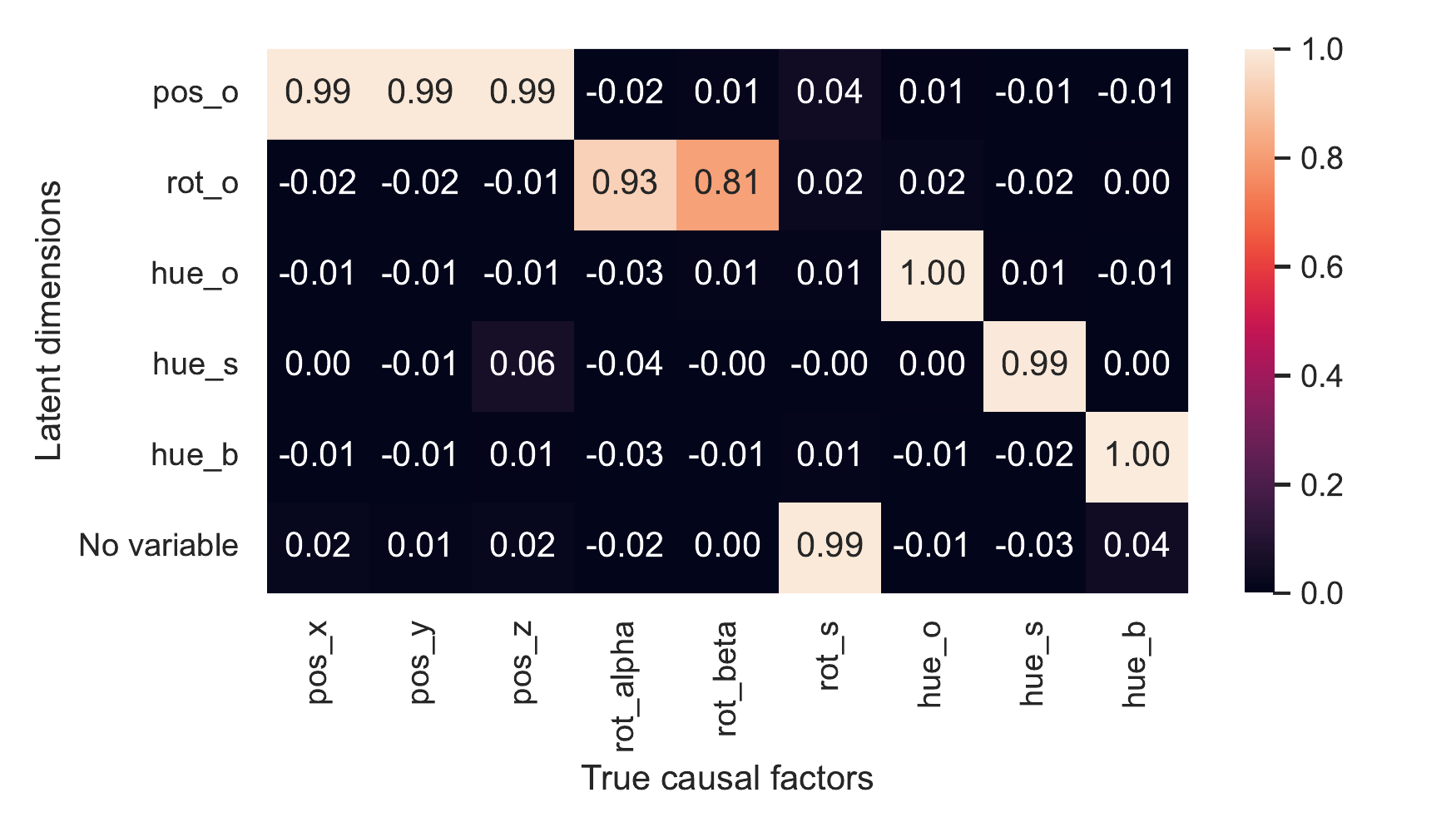} & 
        \includegraphics[width=0.4\textwidth]{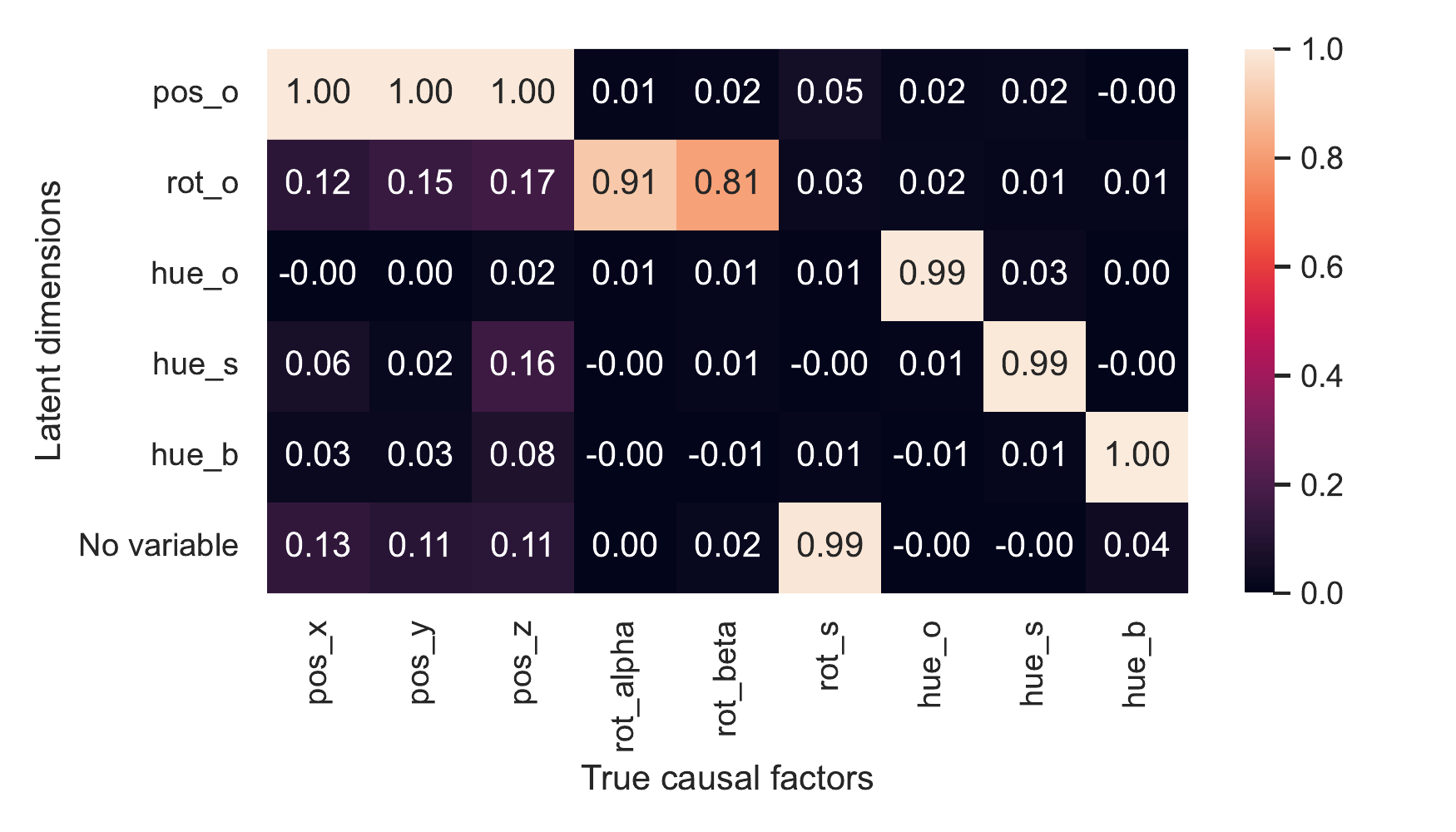} \\
        (a) Excludes rot\_s, $R^2$ correlation matrix & (b) Excludes rot\_s, Spearman correlation matrix \\
        \includegraphics[width=0.4\textwidth]{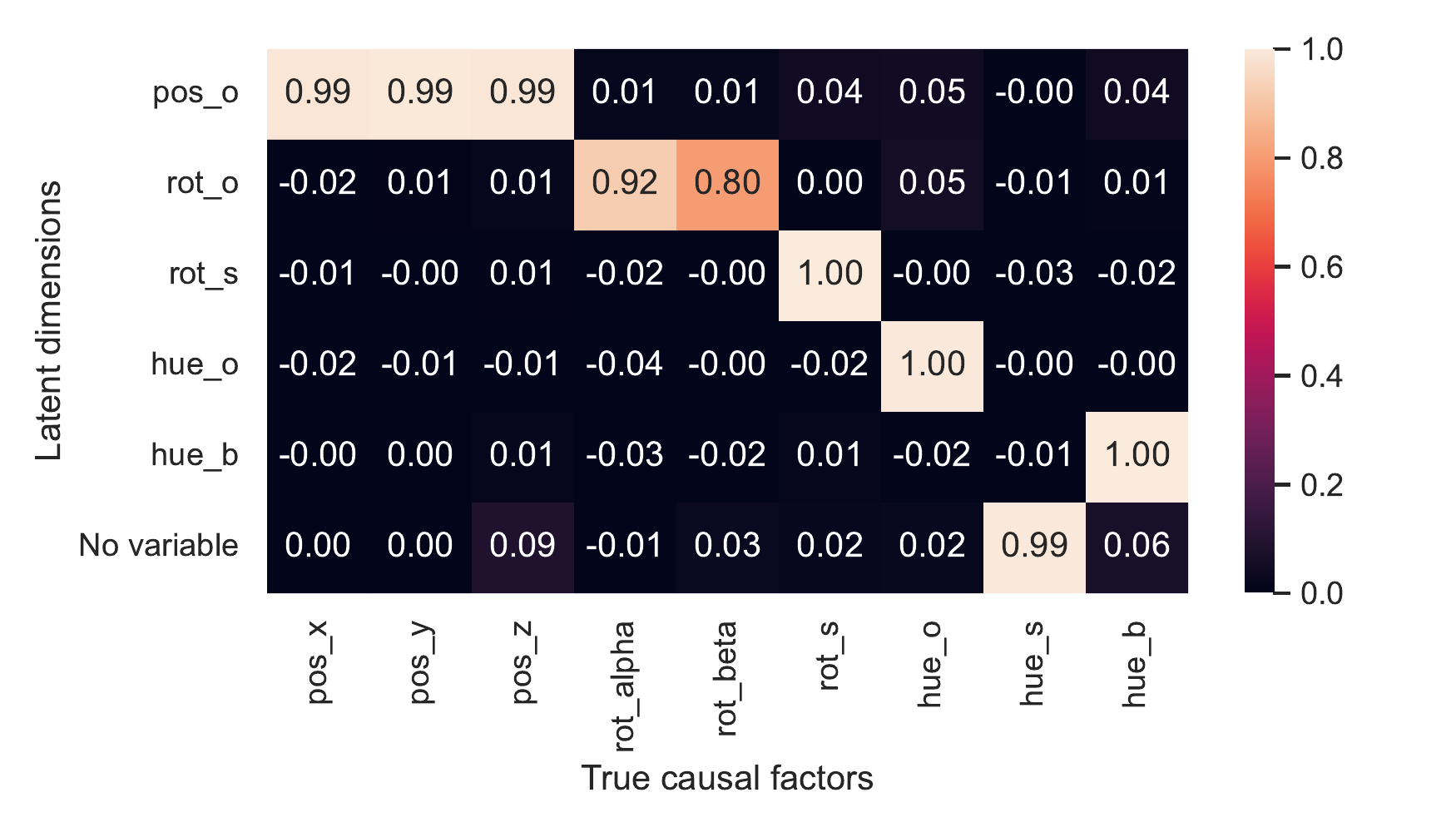} & 
        \includegraphics[width=0.4\textwidth]{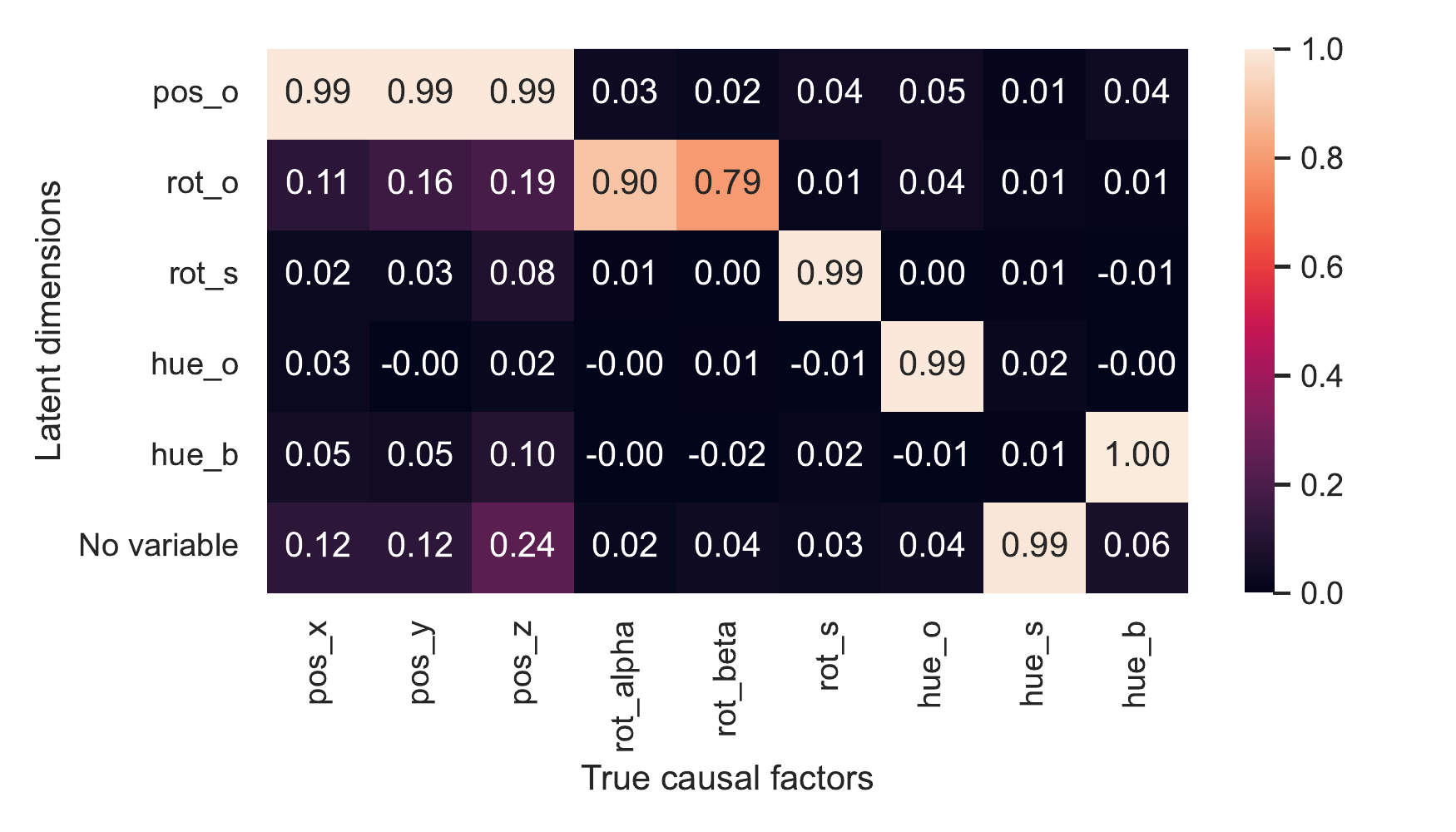} \\
        (c) Excludes hue\_s, $R^2$ correlation matrix & (d) Excludes hue\_s, Spearman correlation matrix \\
    \end{tabular}
    \caption{Correlation matrices for the experiments on the Temporal-Causal3DIdent Teapot dataset where no intervention targets for the causal variable rotation-spotlight (a,b) and hue-spotlight (c,d) are shown (\textbf{left}: $R^2$ correlation, \textbf{right}: Spearman correlation). The set $\zpsi{0}$ is represented by 'no variable' in the plots of \OurApproach{}. \OurApproach{} assigns the variables without intervention targets to $\zpsi{0}$, and the remaining causal variables only show slightly worse disentanglement than for the full intervention set (compare to \cref{fig:appendix_additional_experiments_correlation_matrices_causal3d_teapot}).}
    \label{fig:appendix_additional_experiments_correlation_matrices_causal3d_novars_teapot}
\end{figure}

One property of \OurApproach{} is that when interventions are only provided for a subset of variables, \OurApproach{} is yet able to disentangle these variables, while all remaining information is grouped into the latents $z_{\Psi_0}$.
We have used this to disentangle the score in the Interventional Pong experiments (see \cref{sec:experiments_interventional_pong}), and show here that it can also be applied to the more complex Temporal Causal3DIdent dataset.
Specifically, we reuse the datasets that have been used for the experiments in \cref{sec:experiments_causal3dident}, and remove the intervention targets for a subset of variables.
This keeps all temporal relations intact, while providing an additional challenge of modeling a multimodal distribution for the excluded variables.

As a first setting, we train \OurApproach{}-NF on the Temporal Causal3DIdent 7-shapes dataset, where the interventions on the hue of the background have been omitted.
The correlation matrices in \cref{fig:appendix_additional_experiments_correlation_matrices_causal3d_novars_all7} show that \OurApproach{} learned to disentangle all causal variables in a similar accuracy as in the full experiment (see \cref{fig:appendix_additional_experiments_correlation_matrices_causal3d_all7}), while the background hue is assigned to the latents of $\zpsi{0}$.
Additionally, we show in \cref{fig:appendix_additional_experiments_correlation_matrices_causal3d_teapot} results on the Temporal Causal3DIdent Teapot dataset, in which we exclude intervention targets for the spotlight rotation (a,b) and spotlight hue (c,d) respectively.
Both models assign the correct information to the 'no-variable' slot.
The disentanglement for all other variables shows to still be very good, but slightly worse than the results on the full intervention set, which comes from additional entanglement in $\zpsi{0}$.
Still, the experiments show that \OurApproach{} can also handle variables without interventions in complex 3D rendered scenes.

\subsubsection{Visualizing the Learned Assignment Function $\psi$}
\label{sec:appendix_additional_experiments_causal3d_assignment_map}

\begin{figure}[t!]
    \centering
    \footnotesize
    \includegraphics[width=\textwidth]{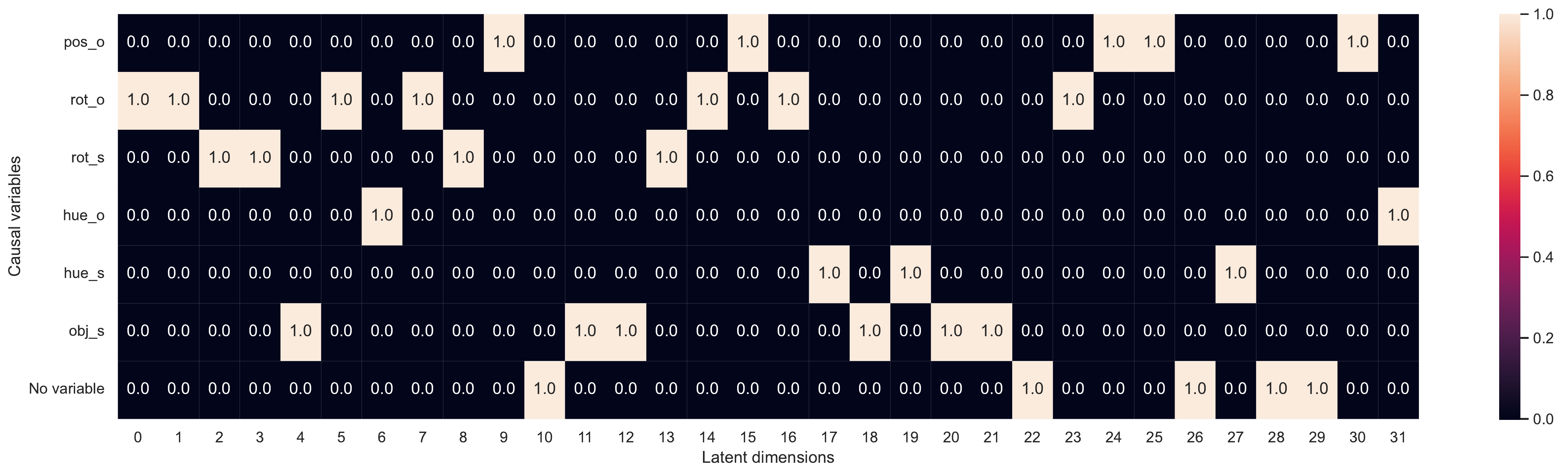}
    \caption{Learned latent-to-causal variable assignment map $\psi$ for an \OurApproach{}-NF trained on the Temporal Causal3DIdent 7-shapes datasets with no interventions on the hue of the background. The map is learned via a Gumbel-Softmax, meaning that we have a softmax distribution over the possible causal variables for each latent variable. Thus, the figure shows the probability that latent dimension $i$ (x-axis) is assigned to a specific causal variable (y-axis), where the distribution usually gets strongly peaked as training progresses. The model has 32 dimensions, of which 5 are assigned to pos\_o, 7 to rot\_o, 4 to rot\_s, 2 to hue\_o, 3 to hue\_s, 6 to obj\_s, and 5 to 'no variable', i.e. $\zpsi{0}$. In general, we find that the most complex causal variable, here rot\_o and obj\_s, are assigned the most dimensions, while circular values have always at least 2 dimensions (the model commonly encodes them as two or more shifted sine-waves).}
    \label{fig:appendix_additional_experiments_target_assignment}
\end{figure}

\OurApproach{} learns an assignment between latent and causal variable, $\psi$. 
To get an intuition of what a learned function of such looks like, we visualize the parameters of $\psi$ of a trained \OurApproach{}-NF in \cref{fig:appendix_additional_experiments_target_assignment}.
The model is trained on the Temporal Causal3DIdent 7-shapes dataset, where no interventions have been shown for the background hue (i.e., the same model as in \cref{fig:appendix_additional_experiments_correlation_matrices_causal3d_novars_all7}).
Overall, the probabilities of a latent variable belonging to a certain causal variable are very peaked after training, with the second-highest value per latent variable to be between $1e$-$6$ and $1e$-$9$.
This shows that the soft relaxation of $\psi$ becomes discrete as training progresses.
Further, we find that the most latent variables are assigned to the intuitively most complex causal variables.
In the Temporal Causal3DIdent 7-shapes dataset, this corresponds to the rotation of the object and the shape of the object, assigned 7 and 6 latent respectively.
The property of learning a flexible latent space size per causal variable is thereby important, since when the causal variables are not known, it would be difficult to manually estimate the number of latent variables per causal variable.
Moreover, an interesting property we observe across all models is that circular values like the hues are always encoded in at least two latent dimensions.
On closer inspection, the model encodes the angles as two or more shifted sine-waves, verifying our motivation of using multidimensional latent representation even for single-dimensional causal variables.

\subsection{Interventional Pong}
\label{sec:appendix_additional_experiments_pong}

\begin{table*}[t!]
    \centering
    \caption{Experimental results for the Interventional Pong dataset, extension of \cref{tab:experiments_results_pong}. The six causal factors are the ball x and y position, the velocity direction, the left paddle y position, the right paddle y position, and the player's score. In the correlation metrics, \textit{diag} refers to the average score on its diagonal (optimal 1), and \textit{sep} for the average of the maximum correlation per causal variable besides itself (optimal 0).}
    \label{tab:appendix_experiments_full_results_pong}
    \resizebox{\textwidth}{!}{%
    \begin{tabular}{lccccccCcccc}
        \toprule
        & \multicolumn{7}{c}{\textbf{Triplet evaluation distances} $\downarrow$} & \multicolumn{4}{c}{\textbf{Correlation metrics}}\\\cmidrule(r{4mm}){2-8}\cmidrule{9-12}
        & \texttt{ball\_x} & \texttt{ball\_y} & \texttt{ball\_vel\_dir} & \texttt{pleft\_y} & \texttt{pright\_y} & \texttt{score} & Mean & $R^2$ diag $\uparrow$ & $R^2$ sep $\downarrow$ & Spearman diag $\uparrow$ & Spearman sep $\downarrow$\\
        \midrule
        \textbf{Oracle} & 0.01 & 0.01 & 0.01 & 0.01 & 0.01 & 0.00 & 0.01 & - & - & - & - \\
        \midrule
        \textbf{SlowVAE} & 0.06 & 0.21 & 0.99 & 0.48 & 0.07 & 0.23 & 0.34 & 0.61 & 0.17 & 0.66 & 0.23\\
        (stds) & \footnotesize$\pm$0.006 & \footnotesize$\pm$0.012 & \footnotesize$\pm$0.002 & \footnotesize$\pm$0.121 & \footnotesize$\pm$0.089 & \footnotesize$\pm$0.001 & \footnotesize$\pm$0.001 & \footnotesize$\pm$0.004 & \footnotesize$\pm$0.012 & \footnotesize$\pm$0.007 & \footnotesize$\pm$0.013\\[5pt]
        \textbf{\iVAEAdapt{}} & 0.03 & 0.21 & 0.05 & \highlight{0.03} & \highlight{0.03} & 0.20 & 0.09 & 0.91 & 0.04 & 0.92 & 0.06\\
        (stds) & \footnotesize$\pm$0.008 & \footnotesize$\pm$0.255 & \footnotesize$\pm$0.005 & \footnotesize$\pm$0.011 & \footnotesize$\pm$0.009 & \footnotesize$\pm$0.015 & \footnotesize$\pm$0.039 & \footnotesize$\pm$0.135 & \footnotesize$\pm$0.056 & \footnotesize$\pm$0.131 & \footnotesize$\pm$0.058 \\
        \midrule
        \textbf{\OurApproach-VAE} & \highlight{0.02} & 0.03 & 0.05 & 0.04 & \highlight{0.03} & \highlight{0.00} & 0.03 & 0.99 & \highlight{0.01} & 0.99 & \highlight{0.05} \\
        (stds) & \footnotesize$\pm$0.001 & \footnotesize$\pm$0.004 & \footnotesize$\pm$0.009 & \footnotesize$\pm$0.006 & \footnotesize$\pm$0.002 & \footnotesize$\pm$0.001 & \footnotesize$\pm$0.001 & \footnotesize$\pm$0.001 & \footnotesize$\pm$0.000 & \footnotesize$\pm$0.001 & \footnotesize$\pm$0.010 \\[5pt]
        \textbf{\OurApproach-NF} & \highlight{0.02} & \highlight{0.02} & \highlight{0.02} & \highlight{0.03} & \highlight{0.03} & \highlight{0.00} & \highlight{0.02} & \highlight{1.00} & 0.04 & \highlight{1.00} & 0.10 \\
        (stds) & \footnotesize$\pm$0.002 & \footnotesize$\pm$0.002 & \footnotesize$\pm$0.001 & \footnotesize$\pm$0.002 & \footnotesize$\pm$0.001 & \footnotesize$\pm$0.001 & \footnotesize$\pm$0.001 & \footnotesize$\pm$0.001 & \footnotesize$\pm$0.011 & \footnotesize$\pm$0.001 & \footnotesize$\pm$0.023 \\
        \bottomrule
    \end{tabular}%
    }
\end{table*}
\begin{figure}[t!]
    \centering
    \footnotesize
    \begin{tabular}{cc}
        \includegraphics[width=0.33\textwidth]{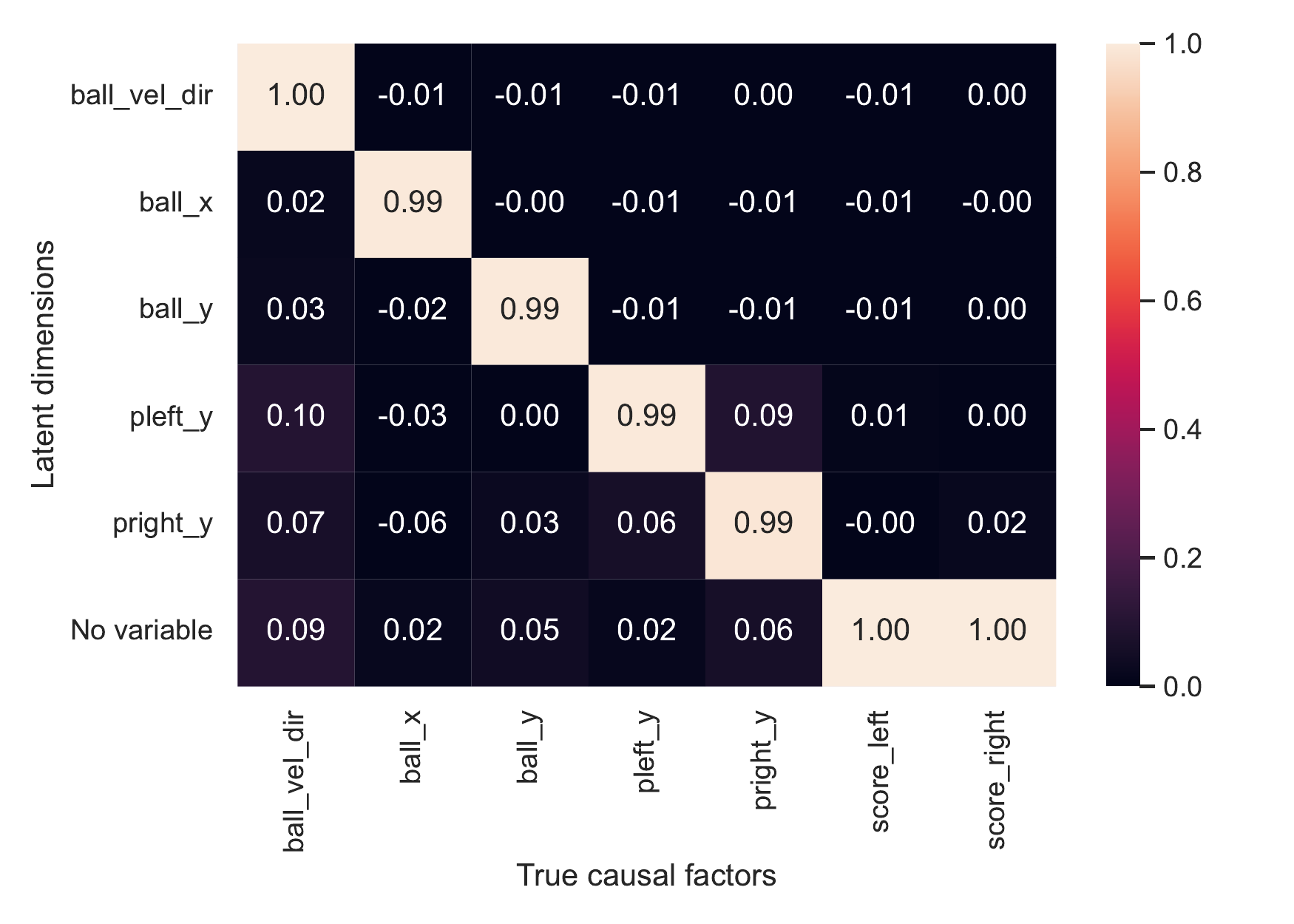} & 
        \includegraphics[width=0.33\textwidth]{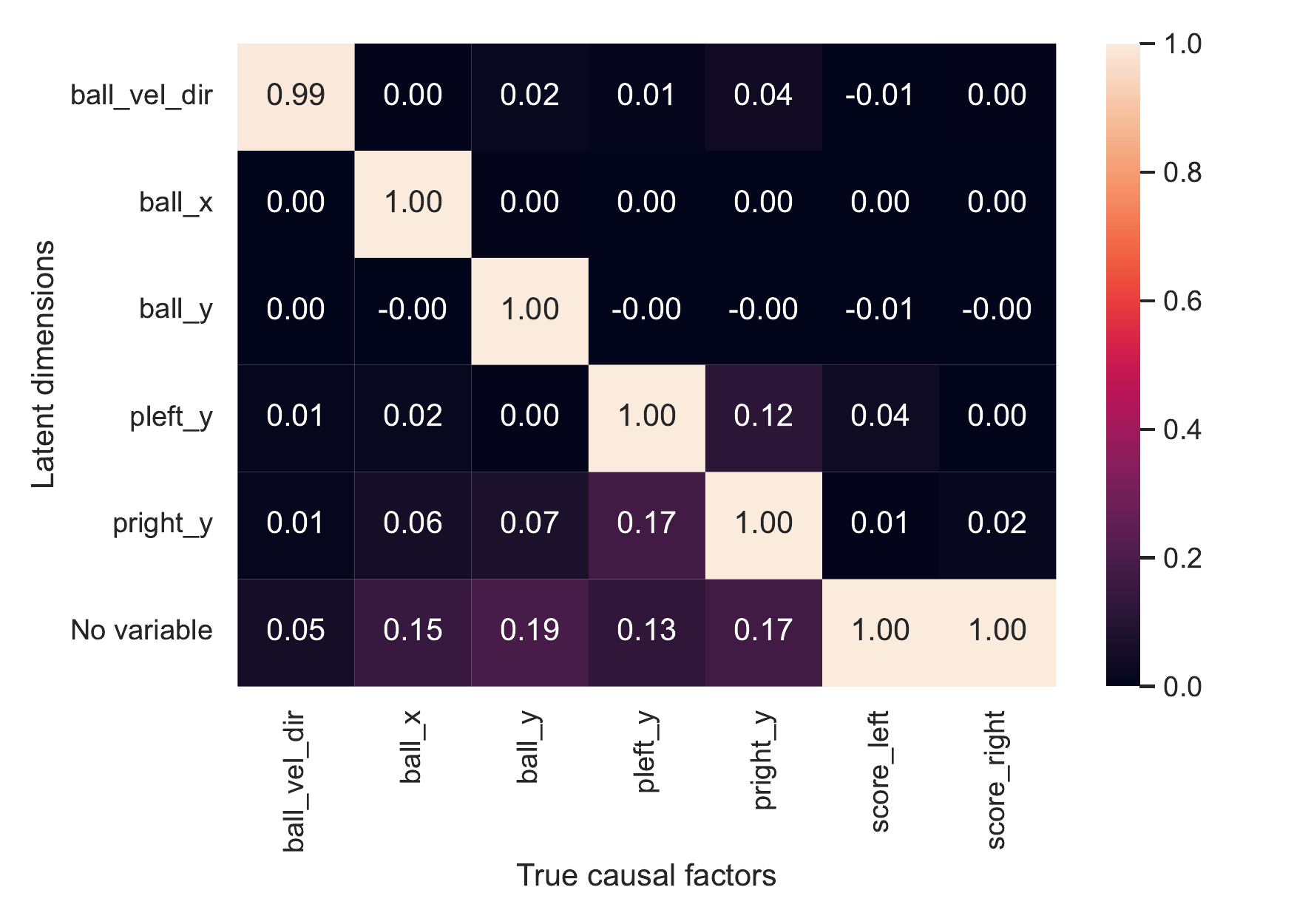} \\
        (a) \OurApproach{}-NF, $R^2$ correlation matrix & (b) \OurApproach{}-NF, Spearman correlation matrix \\[8pt]
        \includegraphics[width=0.33\textwidth]{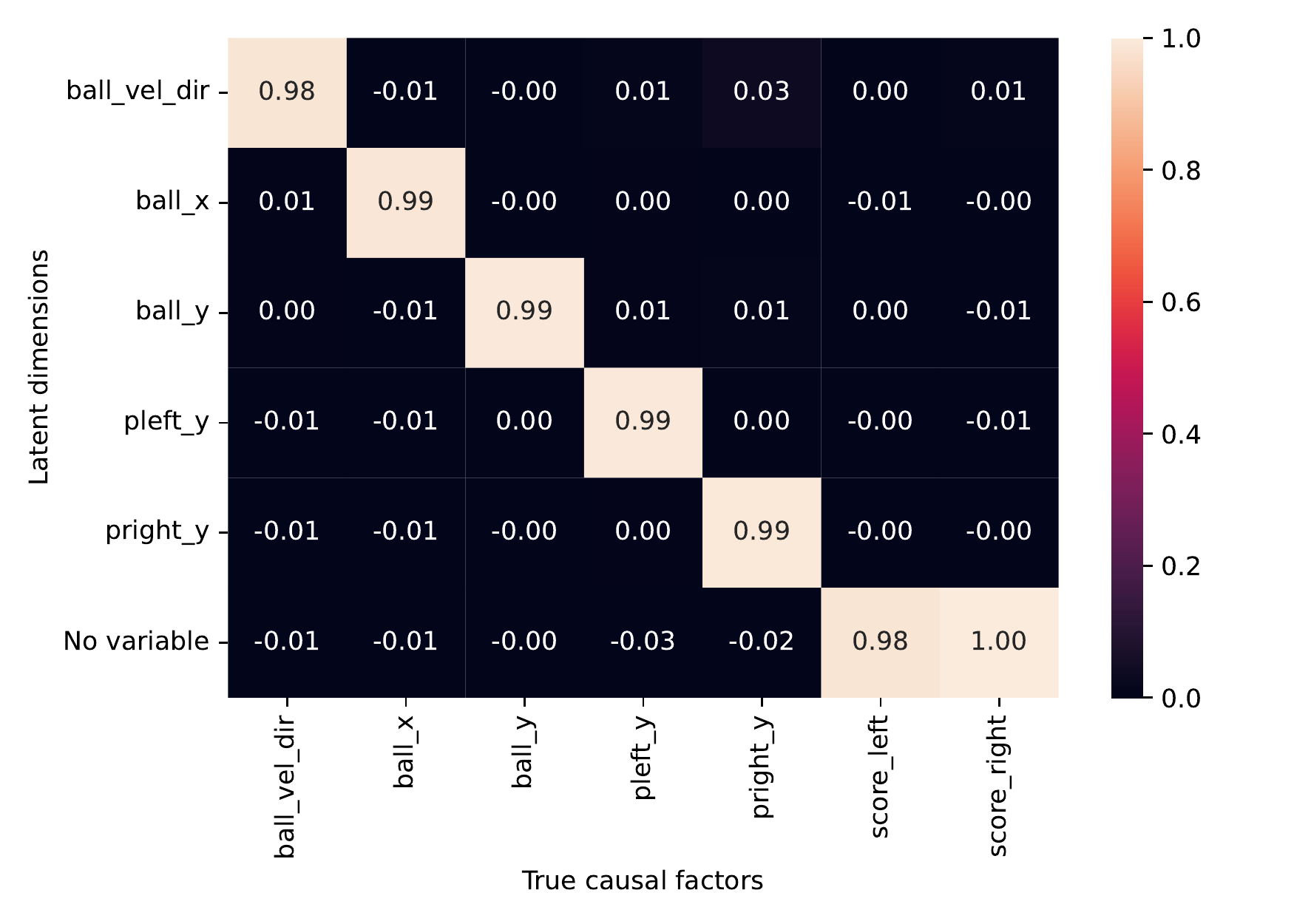} & 
        \includegraphics[width=0.33\textwidth]{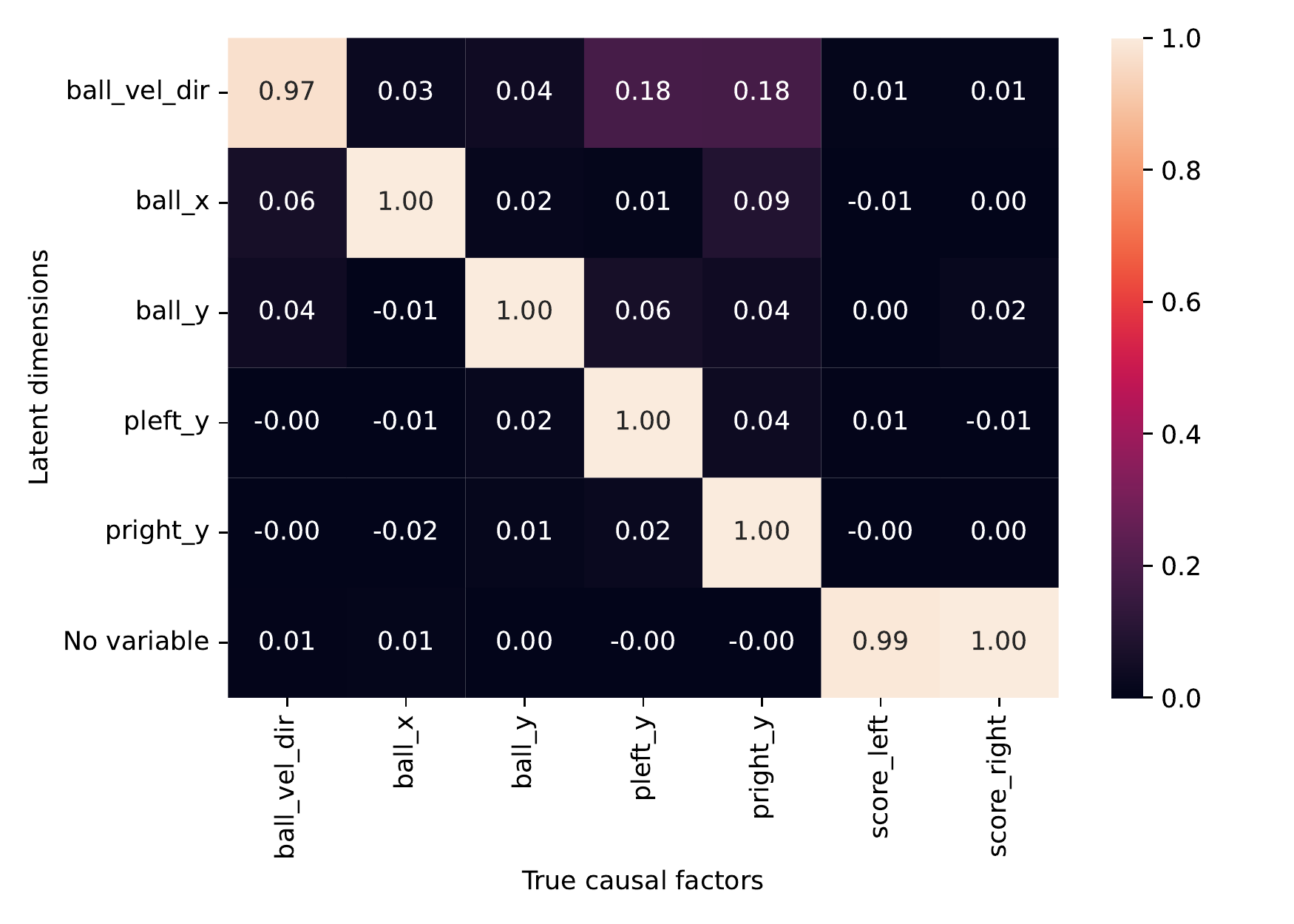} \\
        (c) \OurApproach{}-VAE, $R^2$ correlation matrix & (d) \OurApproach{}-VAE, Spearman correlation matrix \\[8pt]
        \includegraphics[width=0.33\textwidth]{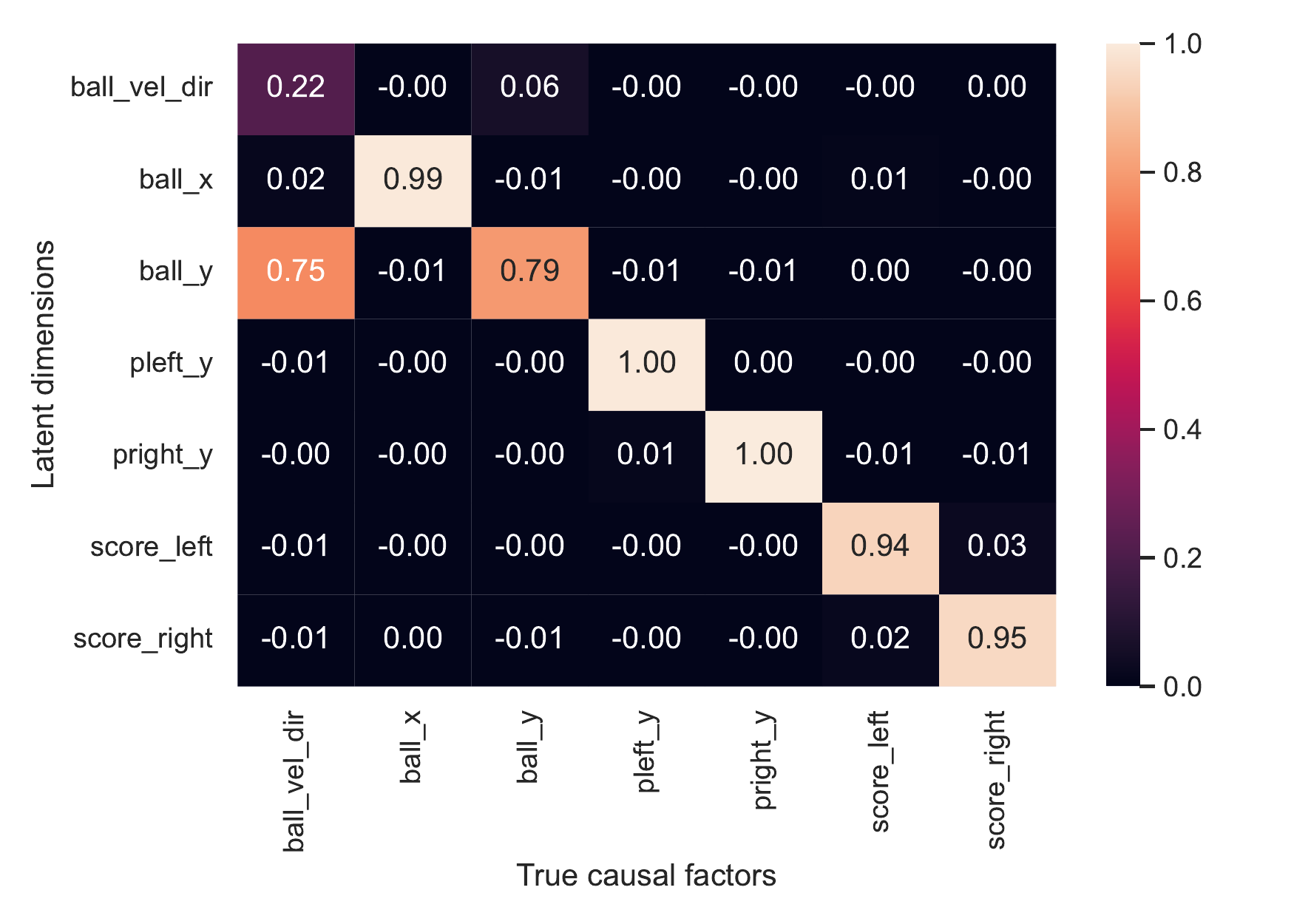} & 
        \includegraphics[width=0.33\textwidth]{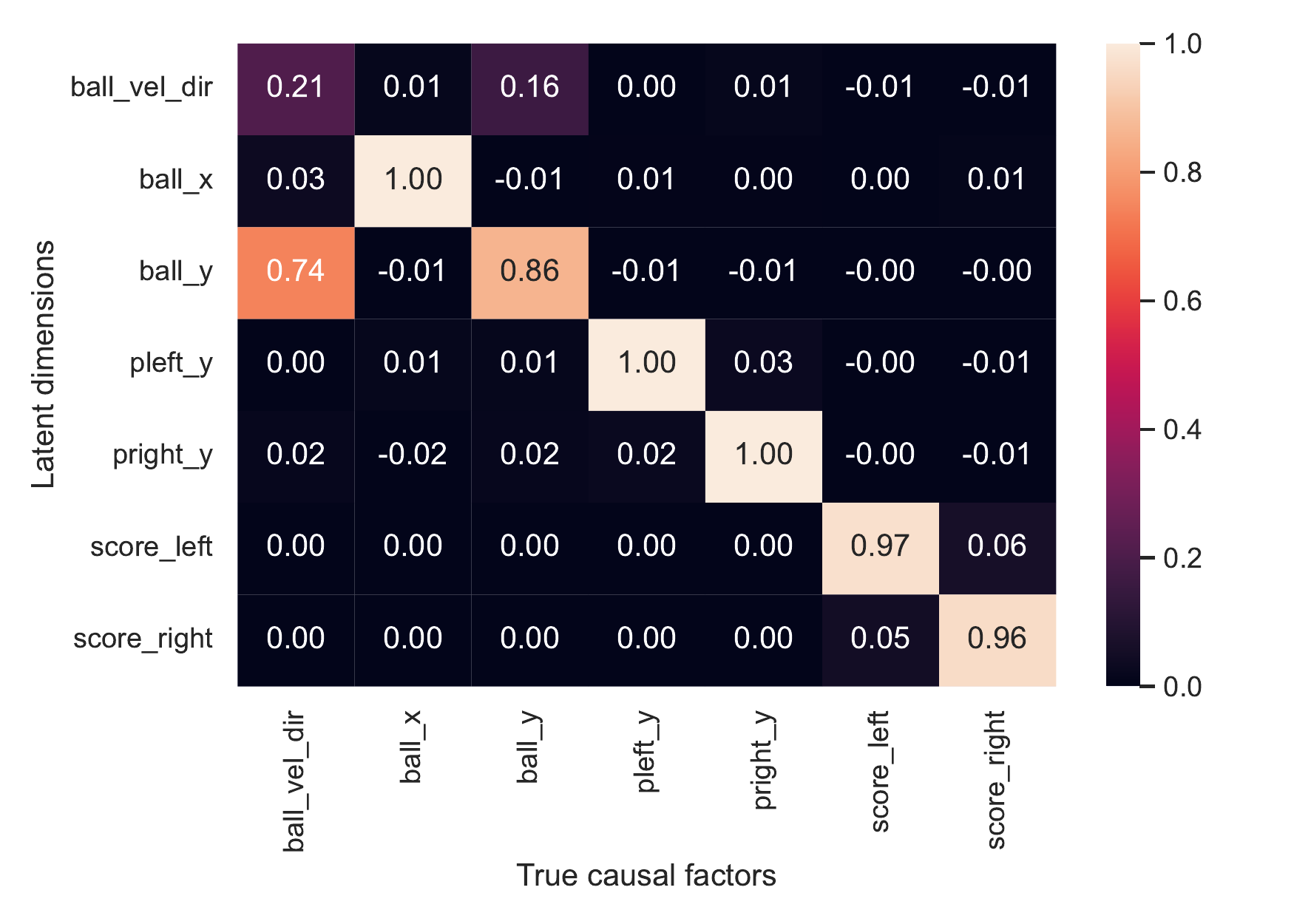} \\
        (e) \iVAEAdapt{}, $R^2$ correlation matrix & (f) \iVAEAdapt{}, Spearman correlation matrix \\[8pt]
        \includegraphics[width=0.33\textwidth]{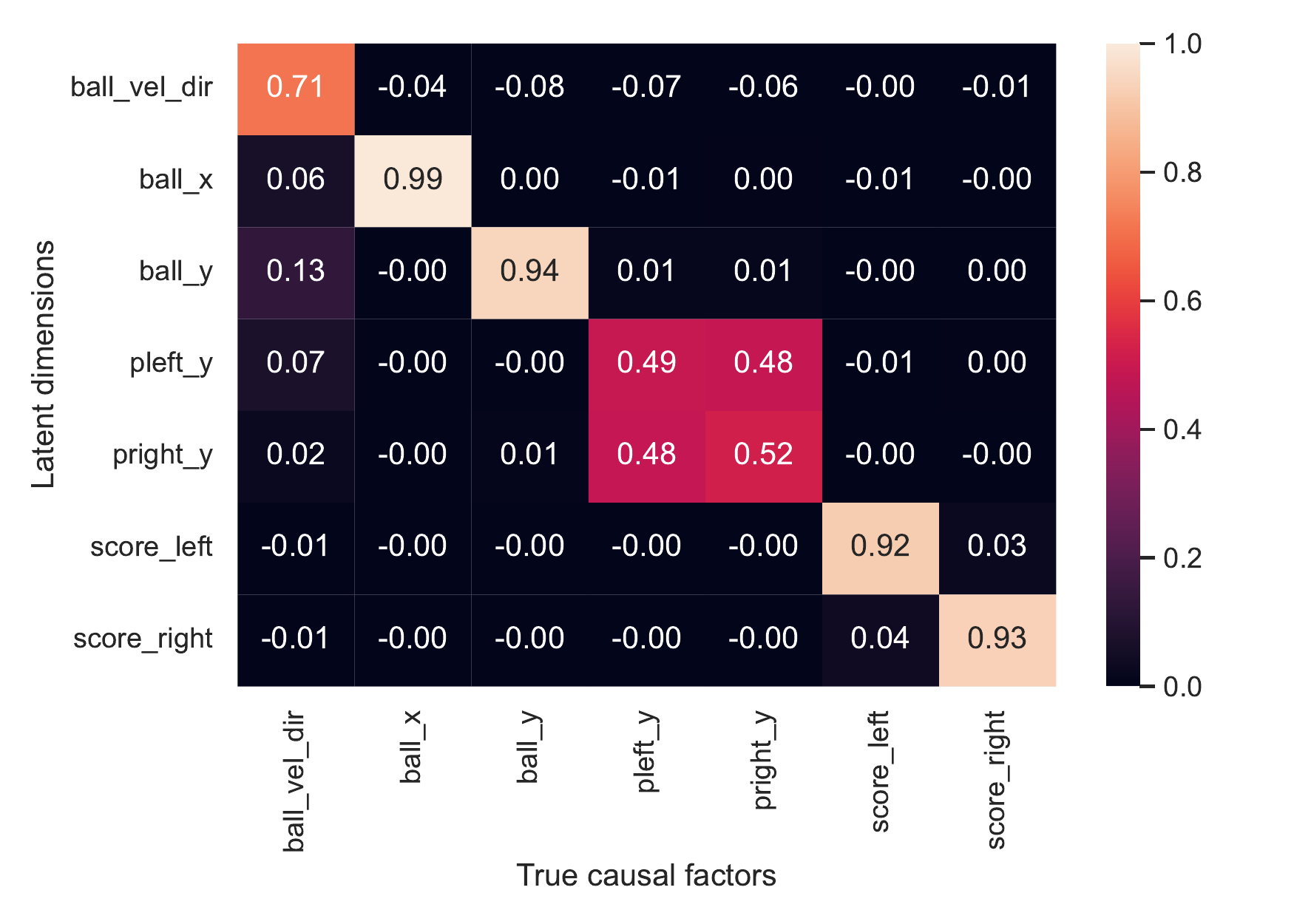} &  
        \includegraphics[width=0.33\textwidth]{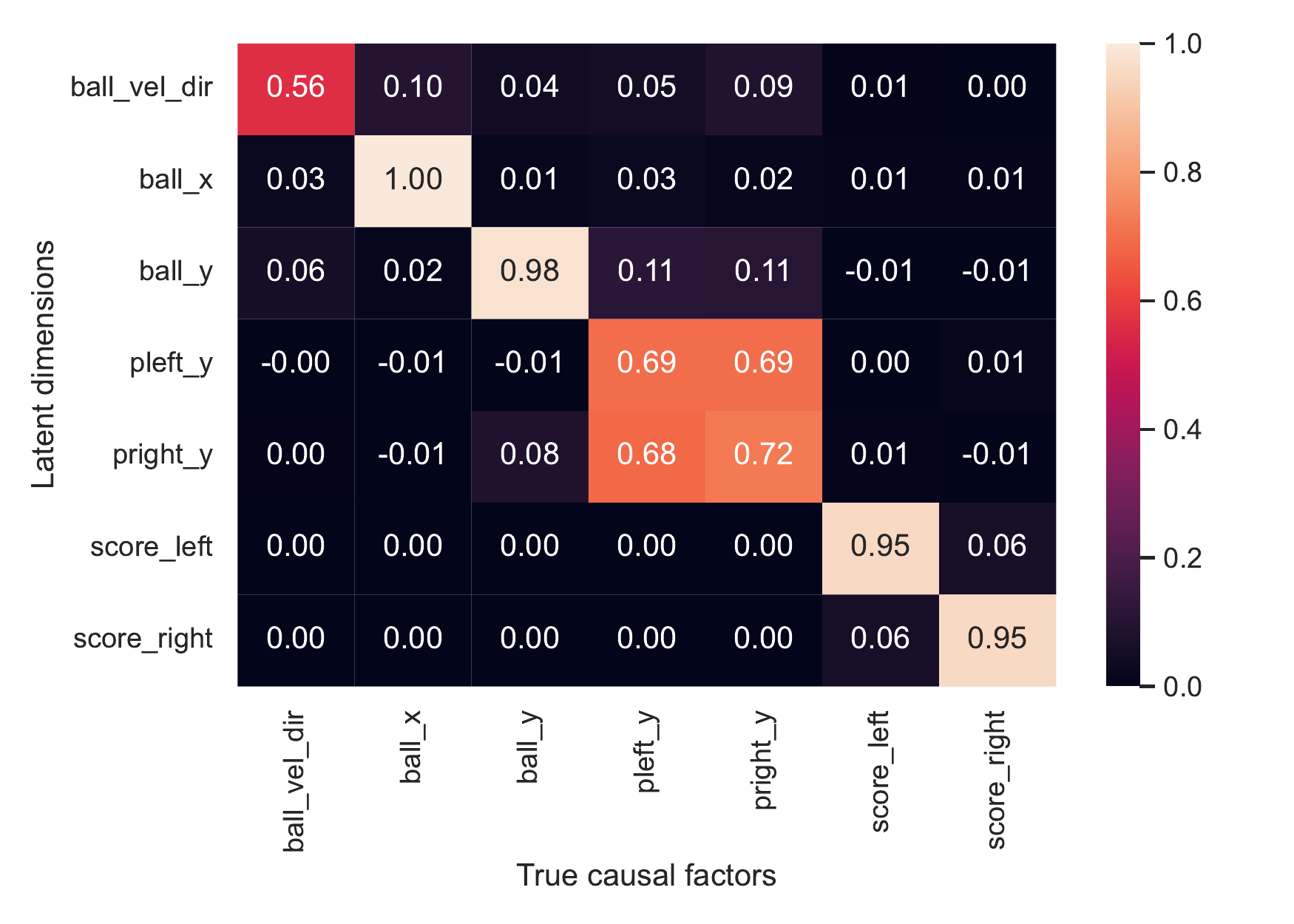} \\
        (e) SlowVAE, $R^2$ correlation matrix & (f) SlowVAE, Spearman correlation matrix \\
    \end{tabular}
    \caption{Correlation matrices for the experiments on the Interventional Pong dataset. The y-axis shows the sets of latent dimensions that were assigned to a certain causal factor. The set $\zpsi{0}$ is represented by 'no variable' in the plots of \OurApproach{}, where no interventions on score\_left and score\_right have been provided. The x-axis shows the ground truth causal factors. The heatmap is the correlation matrix between those factors ($R^2$ left, Spearman right).}
    \label{fig:appendix_additional_experiments_correlation_matrices_pong}
\end{figure}

The full experimental results for the interventional Pong dataset can be found in \cref{tab:appendix_experiments_full_results_pong}, and we plot the correlation matrices for all models in \cref{fig:appendix_additional_experiments_correlation_matrices_pong}.
The triplet evaluation distances show that SlowVAE had difficulties with disentangling all the factors.
This is since they are highly correlated, while SlowVAE assumes independence in the factors.
Both the \iVAEAdapt{} and SlowVAE had troubles with disentangling the y position of the ball (ball\_y), which was often entangled together with the ball velocity direction (ball\_vel\_dir) in the models. 
Further, the player's score was only consistently recovered by \OurApproach{}.
Note that for the triplet evaluation, since there are no interventions given on the player's score, we use the variables $\zpsi{0}$ of \OurApproach{} of that image of which the score it used for the third image.
\OurApproach{}-NF slightly improves upon \OurApproach{}-VAE, mostly due to better modeling of ball\_vel\_dir.
This is because it only affects a single channel and hence has a lower effect in the reconstruction error.
Nonetheless, the experiments show that for simpler datasets, \OurApproach{}-VAE can disentangle the causal factors as well as the normalizing flow.
\newpage

\subsection{Ball-in-Boxes Example}
\label{sec:appendix_additional_experiments_ball_in_box}

\begin{figure}
    \centering
    \footnotesize
    \begin{tabular}{cccccccccc}
        \includegraphics[width=0.1\textwidth]{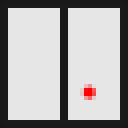} & 
        \includegraphics[width=0.1\textwidth]{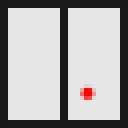} & 
        \includegraphics[width=0.1\textwidth]{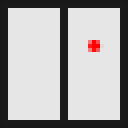} & 
        \includegraphics[width=0.1\textwidth]{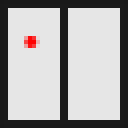} & 
        \includegraphics[width=0.1\textwidth]{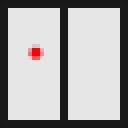} & 
        \includegraphics[width=0.1\textwidth]{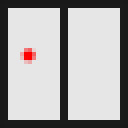} & 
        \includegraphics[width=0.1\textwidth]{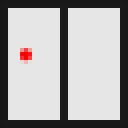} \\
         (1) - & (2) - & (3) ball\_y & (4) ball\_x & (5) ball\_y & (6) - & (7) - \\[7pt]
        \includegraphics[width=0.1\textwidth]{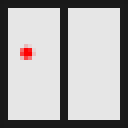} & 
        \includegraphics[width=0.1\textwidth]{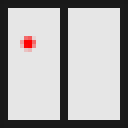} & 
        \includegraphics[width=0.1\textwidth]{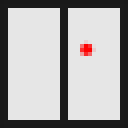} & 
        \includegraphics[width=0.1\textwidth]{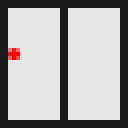} & 
        \includegraphics[width=0.1\textwidth]{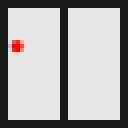} & 
        \includegraphics[width=0.1\textwidth]{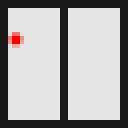} & 
        \includegraphics[width=0.1\textwidth]{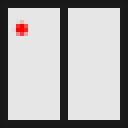}
        \\
        (8) - & (9) - & (10) ball\_x & (11) ball\_x & (12) - & (13) - & (14) - \\
    \end{tabular}
    \caption{An example sequence with 14 frames in the Ball-in-Boxes dataset. The causal variables denoted below each image indicate the variables which were intervened on at this time step, where an intervention on ball\_x changes the box identifier (left or right), and ball\_y the y position of the ball. }
    \label{fig:appendix_box_sequence_example}
\end{figure}

Throughout the paper, we have referred to the example of a ball in a 2D space, which can be in one of two boxes. 
It can move freely within a box, but cannot switch between them by itself.
Only under interventions on its $x$ position, we switch the box. 
However, this intervention does not influence the internal position of the ball in the box.
Under our theoretical setup of \OurApproach{}, the box identifier represents the minimal causal variable of $x$, while the internal position should be modeled in $\zpsi{0}$.

In this section, we give empirical evidence that this is indeed also the case in practice.
For simplicity, we implement it by generating frames of $32\times 32$ resolution similar to our Pong environment (see \cref{fig:appendix_box_sequence_example}). 
However, the dynamics are replaced by the ball-in-boxes example: we consider two causal variables, ball\_x and ball\_y.
Both ball\_x and ball\_y follow a truncated Gaussian distribution, \ie{} move randomly with Gaussian noise over time in the boundaries of the box the ball is in.
Under interventions on ball\_y, we uniformly resample the position of the ball.
Under interventions on ball\_x, we switch the box without influencing the internal ball position.
We generate the dataset by using a sequence of $100,000$ frames, and sample the intervention targets from $I^t_i\sim \text{Bernoulli}(0.2)$.
On this dataset, we apply \OurApproach{}-VAE with the same experimental setup as for the Pong dataset, including the small regularizer $\lambda=0.01$, to show that little change in the KL divergence already encourages the disentanglement to $\zpsi{0}$.

\begin{table*}[t!]
    \centering
    \caption{Experimental results for the Ball-in-Boxes dataset. The three evaluation causal factors are the box identifier ball\_b, the internal ball x position ball\_xin, and the ball y position. In the correlation metrics, \textit{diag} refers to the average score on its diagonal (optimal 1), and \textit{sep} for the average of the maximum correlation per causal variable besides itself (optimal 0). For the correlation metrics, we consider $\zpsi{0}$ to be the latent variables in which ball\_xin should be modeled, \ie{} have a high correlation with. From the results, one can clearly see that \OurApproach{} has indeed learned to disentangle ball\_b and ball\_xin, showing that \OurApproach{} learns the minimal causal variables.}
    \label{tab:appendix_experiments_ball_in_box}
    \resizebox{0.8\textwidth}{!}{%
    \begin{tabular}{lcccCcccc}
        \toprule
        & \multicolumn{4}{c}{\textbf{Triplet evaluation distances} $\downarrow$} & \multicolumn{4}{c}{\textbf{Correlation metrics}}\\\cmidrule(r{4mm}){2-5}\cmidrule{6-9}
        & \texttt{ball\_b} & \texttt{ball\_y} & \texttt{ball\_xin} & Mean & $R^2$ diag $\uparrow$ & $R^2$ sep $\downarrow$ & Spearman diag $\uparrow$ & Spearman sep $\downarrow$\\
        \midrule
        \textbf{Oracle} & 0.00 & 0.01 & 0.01 & 0.01 & - & - & - & - \\
        \midrule
        \textbf{\OurApproach-VAE} & 0.00 & 0.01 & 0.02 & 0.01 & 0.99 & 0.00 & 1.00 & 0.02 \\
        (stds) & \footnotesize$\pm$0.001 & \footnotesize$\pm$0.003 & \footnotesize$\pm$0.005 & \footnotesize$\pm$0.002 & \footnotesize$\pm$0.001 & \footnotesize$\pm$0.001 & \footnotesize$\pm$0.001 & \footnotesize$\pm$0.002 \\[5pt]
        \bottomrule
    \end{tabular}%
    }
\end{table*}

\begin{figure}[t!]
    \centering
    \footnotesize
    \begin{tabular}{cc}
        \includegraphics[width=0.32\textwidth]{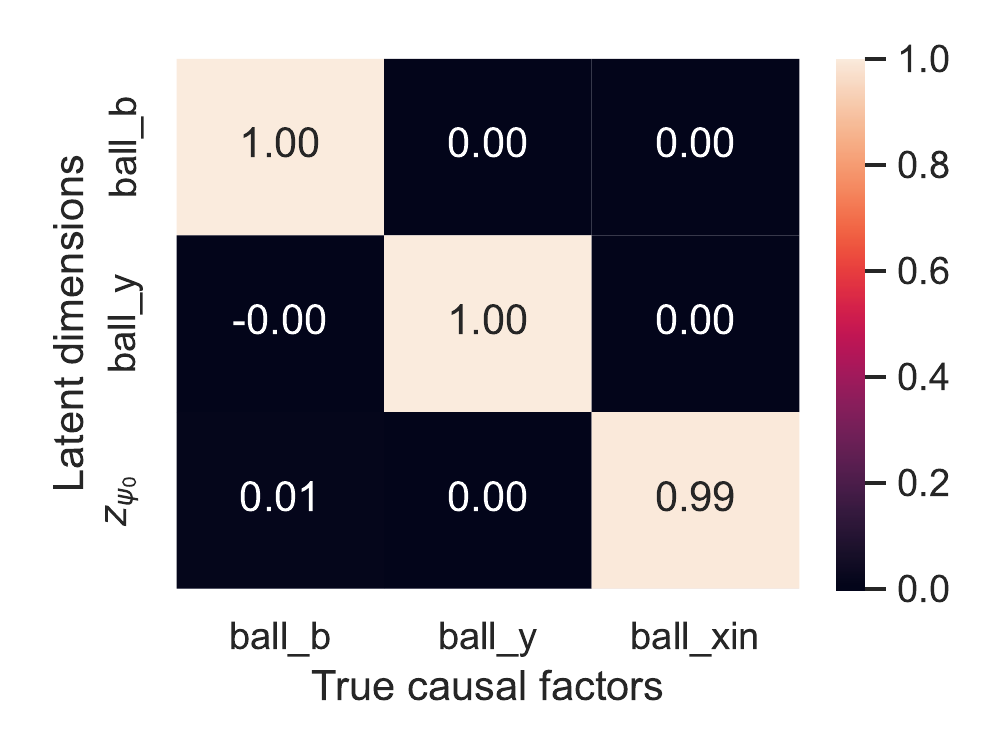} & 
        \includegraphics[width=0.32\textwidth]{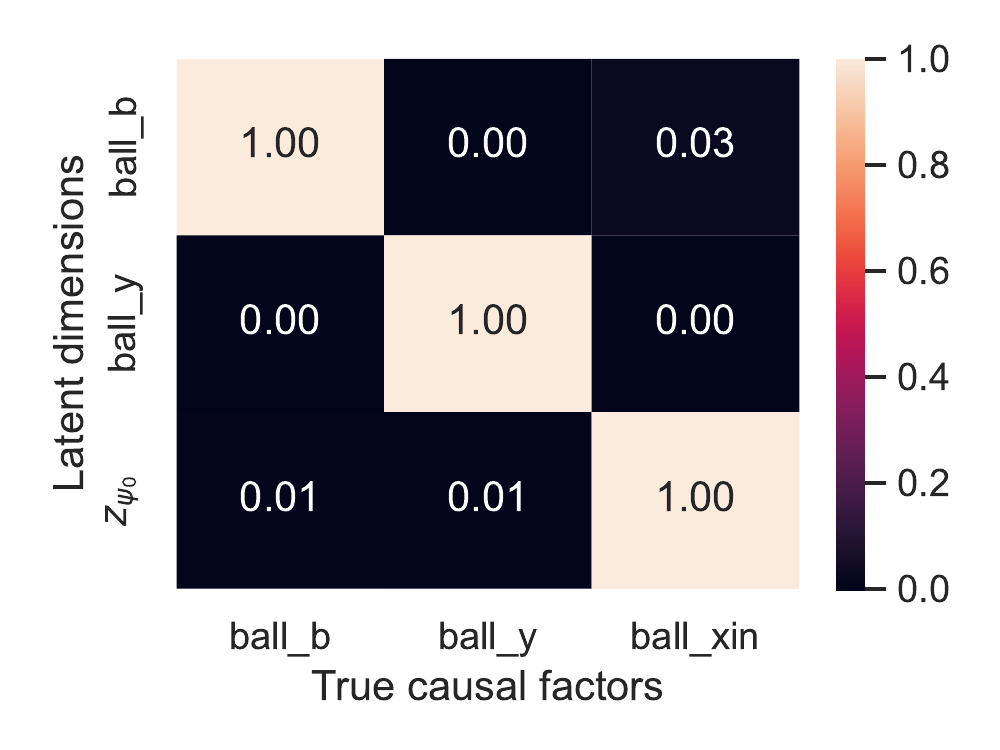} \\
        (a) \OurApproach{}-VAE, $R^2$ correlation matrix & (b) \OurApproach{}-VAE, Spearman correlation matrix \\
    \end{tabular}
    \caption{The learned correlation matrices of \OurApproach{}-VAE on the Ball-in-Boxes dataset. The y-axis shows the sets of latent dimensions which are assigned to causal variable 1 (ball\_b/ball\_x), causal variable 2 (ball\_y), and to none of them ($\zpsi{0}$). The x-axis shows the true causal factors. Each element of the matrix therefore describes the correlation between a set of latent factors and a ground truth causal factor. \OurApproach{} disentangles the causal factors as intended, and assigning the internal box position ball\_xin to $\zpsi{0}$ since it is independent of the provided interventions.}
    \label{fig:appendix_box_correlation_matrices}
\end{figure}

For evaluation purposes, we perform the correlation and triplet evaluation by predicting both the box identifier ball\_b and the box internal position ball\_xin instead of just ball\_x.
Note that this choice does not influence the training of \OurApproach{} in any way, but simply allows us to identify whether the split between ball\_b and ball\_xin has been learned or not.
The results are shown in \cref{tab:appendix_experiments_ball_in_box}.
One can clearly see that \OurApproach{} learned to disentangle the internal ball position ball\_xin and ball\_b, since $\zpsi{0}$ has a $R^2$ correlation score of almost one to ball\_xin, and close to zero for ball\_b.
We show a visualization of the $R^2$ and Spearman correlation matrix in \cref{fig:appendix_box_correlation_matrices}.
In conclusion, this experiment validates that \OurApproach{} indeed learns the minimal causal variables, \ie{} the information that strictly depends on the individual interventions.